\documentclass[10pt,twocolumn,letterpaper]{article}
\PassOptionsToPackage{table}{xcolor}
\PassOptionsToPackage{pagebackref,breaklinks,colorlinks,allcolors=cvprblue}{hyperref}

\usepackage[pagenumbers]{cvpr}

\definecolor{cvprblue}{rgb}{0.21,0.49,0.74}
\usepackage{booktabs}  % for \toprule, \midrule, \bottomrule
\usepackage{pifont}    % for \xmark and \cmark
\newcommand{\cmark}{\ding{51}}  % checkmark
\newcommand{\xmark}{\ding{55}}  % cross mark
\usepackage{orcidlink}
\usepackage{array}
\usepackage{url}
\usepackage{graphicx}
\usepackage{pgffor} 
\usepackage{multirow}
\usepackage{titletoc}

\usepackage{wrapfig}
\usepackage[table]{xcolor}
\usepackage{epigraph}
\usepackage{listings}
\usepackage{xcolor}
\usepackage{graphicx}
\usepackage{tikz}
\usepackage{amsmath, amssymb}
\usepackage{algorithm}
\usepackage{algorithmic}
\usepackage{array}
\usepackage{float}
\usepackage{arydshln}
\usepackage{adjustbox}
\usepackage{array}
\usepackage{booktabs}
\usepackage{multirow}
\usepackage{makecell}
\usepackage{colortbl}
\usepackage{xcolor}
\usepackage{makecell}
\usepackage{tikz}
\usepackage{xcolor}
\usepackage{placeins}
\usepackage{longtable}
\usepackage{graphicx}
\usepackage{array}
\usepackage{arydshln}
\usepackage{array}
\usepackage{url}
\usepackage{tabularx}
\usepackage{booktabs}
\usepackage{tikz}
\usepackage{makecell}
\usepackage{lscape}
\usepackage{subcaption}
\usepackage{enumitem}
\usepackage{textcomp}
\usepackage{float}
\usepackage[T1]{fontenc}
\usepackage{listings}
\usepackage{xcolor}
\usepackage{amsmath}
\usepackage{amssymb}
\usepackage{amsthm}
\usepackage{tikz}
\usepackage[most]{tcolorbox}
\usetikzlibrary{positioning,calc,fit}
\newtheorem{theorem}{Theorem}
\newtheorem{assumption}{Assumption}
\newtheorem{definition}{Definition}

\definecolor{boxbg}{RGB}{245,245,245}
\definecolor{boxborder}{RGB}{80,80,80}

\tcbset{
    colback=boxbg,
    colframe=boxborder,
    arc=3mm,
    boxrule=0.8pt,
    left=3mm,
    right=3mm,
    top=2mm,
    bottom=2mm
}
\usepackage{etoc}
\usepackage{wrapfig}
\etocsetlevel{author}{6}
\etocsetlevel{title}{6}
\providecommand{\authcount}[1]{}
\providecommand{\emailauthor}[2]{}
\providecommand{\urlauthor}[2]{}
\usepackage[numbers]{natbib}

\definecolor{grey}{rgb}{0.5,0.5,0.5}

\usepackage[pagebackref,breaklinks,colorlinks,allcolors=cvprblue]{hyperref}

\def\confName{CVPR}
\def\confYear{2027}

\title{Seeing Through Preferences: Recovering Hidden Local Preference Neighborhoods for Text-to-Image Alignment}
\author{
Pulkit Bansal \quad Vivek Srivastava \quad Shirish Karande\\
TCS Research, India\\
{\tt\small
pulkitbansal996@gmail.com, srivastava.vivek2@tcs.com, shirish.karande@tcs.com
}
}

\begin{document}
\maketitle
\raggedbottom
\setlength{\dbltextfloatsep}{10pt plus 2pt minus 2pt}
\begin{abstract}
Aligning text-to-image (T2I) diffusion models with human preferences typically relies on large-scale pairwise annotations, which are expensive to obtain and provide only sparse supervision. Moreover, not all preference comparisons contribute equally to learning: comparisons between semantically distant images often reveal little about the fine-grained perceptual factors underlying human judgments. We argue that informative supervision can arise from \textbf{local preference neighborhoods} surrounding human-preferred images, where semantically aligned alternatives expose meaningful perceptual and structural differences. Motivated by this observation, we propose \textsc{\textbf{PrefLens}}, a framework that recovers local preference neighborhoods from sparse human feedback. Specifically, \textsc{PrefLens} generates controlled perceptual variations around preferred images that largely retain semantic identity and selects compact sets of informative neighbors in a reward-defined preference space induced by pretrained reward models. By transforming each human preference annotation into a structured neighborhood of informative comparisons, \textsc{PrefLens} enriches supervision without requiring additional human labeling. Our experiments and diagnostic evaluations show that the recovered neighborhoods capture human-relevant preference characteristics and provide effective supervision for preference alignment across diverse settings.
\end{abstract}
\begin{figure}[t]
    \centering
    
    \includegraphics[width=0.85\linewidth]{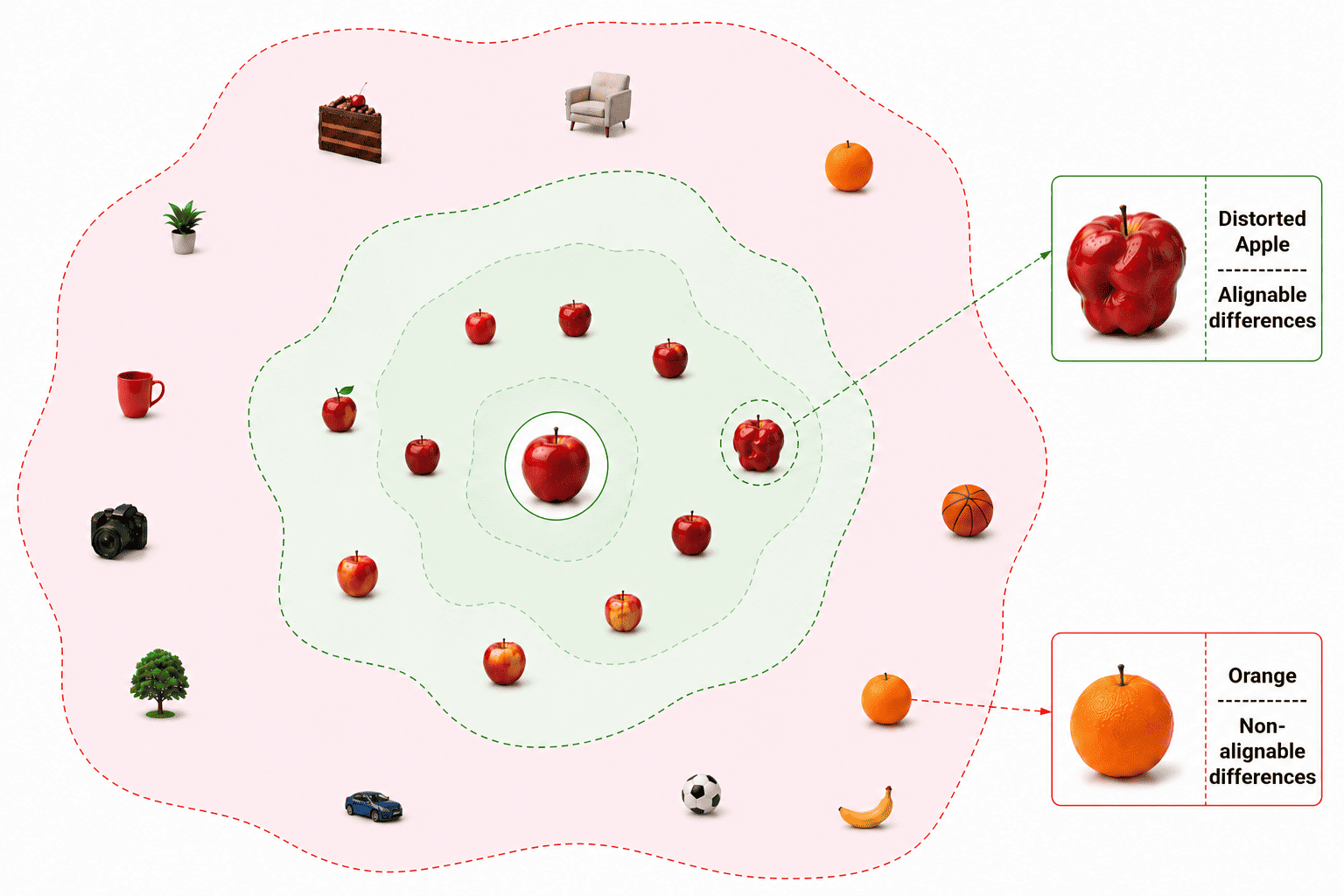}
        
\caption{
\textbf{Illustration of local preference neighborhoods.} The \textcolor{green!60!black}{green region} denotes a \emph{local preference neighborhood} surrounding a preferred image, where semantically aligned alternatives expose localized, \emph{alignable differences}. Samples in the \textcolor{red!70!black}{red region} are semantically distant and often lead to \emph{non-alignable differences}.
}

    \label{fig:teaser}
\end{figure}

\section{Introduction}

Recent advances in T2I diffusion models~\citep{saharia2022photorealistic, betker2023improving, nichol2021glide, Rombach_2022_CVPR, podell2023sdxl} have enabled remarkable progress in generating images. However, aligning these models with nuanced human preferences remains a fundamental challenge, since many desirable attributes are difficult to specify through prompts alone. A common strategy for aligning generative models is to collect pairwise human preference annotations and train preference models or reward predictors that guide generation toward preferred outputs. Recent efforts such as Pick-a-Pic~\cite{kirstain2023pick}, ImageReward~\cite{xu2023imagereward}, and subsequent preference modeling approaches~\cite{wallace2024diffusion,zhu2025dspo,bai2025towards,lee2025calibrated,jang2025multi,liang2024rich} demonstrate that preference learning can substantially influence the behavior of T2I models. More importantly, recent evidence suggests that existing datasets provide only an incomplete characterization of human preferences. For instance, a recent study~\cite{singh2024effective}
shows that carefully selected subsets comprising only a small fraction ($\approx$5\%) of the available preference data can outperform training on the full dataset, indicating the presence of substantial redundancy and non-informative comparisons in existing preference corpora. Such redundancy may bias optimization toward over-represented preference patterns while leaving large regions of the preference landscape underexplored.

A natural response to these limitations is to acquire additional preference supervision by collecting more human annotations, but this remains costly. Another possibility is to associate each preferred image with multiple candidate negatives and obtain rankings over the resulting candidate set~\cite{bai2025towards,wu2025ranking}. While such strategies can potentially provide richer supervision, they require evaluating large numbers of candidate images, which is computationally expensive and may inherit biases and inconsistencies from VLM-based judges~\cite{lee2026mm,laskar2025judging,zou2026vision}. Moreover, although inverse-constitution-based approaches~\cite{findeis2025inverse,kingslin2026democratic} can improve the robustness of such judges, applying these strategies in the image domain remains challenging due to
% both the limited reliability of multimodal reasoning for fine-grained visual judgments and
the cost of repeated multimodal evaluation.
% Simply increasing the number of generated negatives is also insufficient, as large candidate pools may contain substantial redundancy and repeatedly expose models to highly similar preference patterns while incurring additional ranking costs.

We instead argue that a key limitation of existing preference datasets lies in how human judgments are represented. Current datasets typically encode each human annotation as a single pairwise comparison between a preferred and a rejected image.
% , reducing a potentially rich human evaluation to a binary ordering relation.
Consequently, each annotation provides only a sparse observation of the broader preference structure surrounding the preferred image. We hypothesize that each preference pair can provide a starting point for recovering a richer local preference neighborhood, in which semantically aligned alternatives reveal how human judgments vary under controlled perceptual changes. Recovering such neighborhoods can therefore transform each human annotation into substantially richer and more informative supervision for preference alignment.

% ---- Helper commands to draw an image with a "Win"/"Lose" tag on top-left ----
\newcommand{\winimg}[2][0.44\linewidth]{%
    \begin{tikzpicture}[remember picture]
        \node[anchor=south west, inner sep=0] (img) {\includegraphics[width=#1]{#2}};
        \node[anchor=north west, fill=green!60!black, text=white,
              font=\bfseries\footnotesize, inner sep=3pt,
              xshift=-3pt, yshift=3pt] at (img.north west) {Win};
    \end{tikzpicture}%
}

\newcommand{\loseimg}[2][0.44\linewidth]{%
    \begin{tikzpicture}[remember picture]
        \node[anchor=south west, inner sep=0] (img) {\includegraphics[width=#1]{#2}};
        \node[anchor=north west, fill=red!70!black, text=white,
              font=\bfseries\footnotesize, inner sep=3pt,
              xshift=-3pt, yshift=3pt] at (img.north west) {Lose};
    \end{tikzpicture}%
}

% ---- Helper for one "column" block: prompt + win/lose pair --
\newcommand{\promptpair}[3]{%
    \begin{minipage}{0.3\textwidth}
        \centering
        {\scriptsize
        \linespread{0.85}\selectfont
        \parbox{\linewidth}{\centering\textbf{#1}}
        }\\[2pt]
        \winimg{#2}\hfill\loseimg{#3}
    \end{minipage}%
}
% ---- Helper for one "column" block: prompt + 

\begin{figure*}[t]
    \centering
    
    % ===================== ROW 1 =====================
    \promptpair{A 3D model of a scientist, dramatic chiaroscuro lighting, intricate textures, .....}{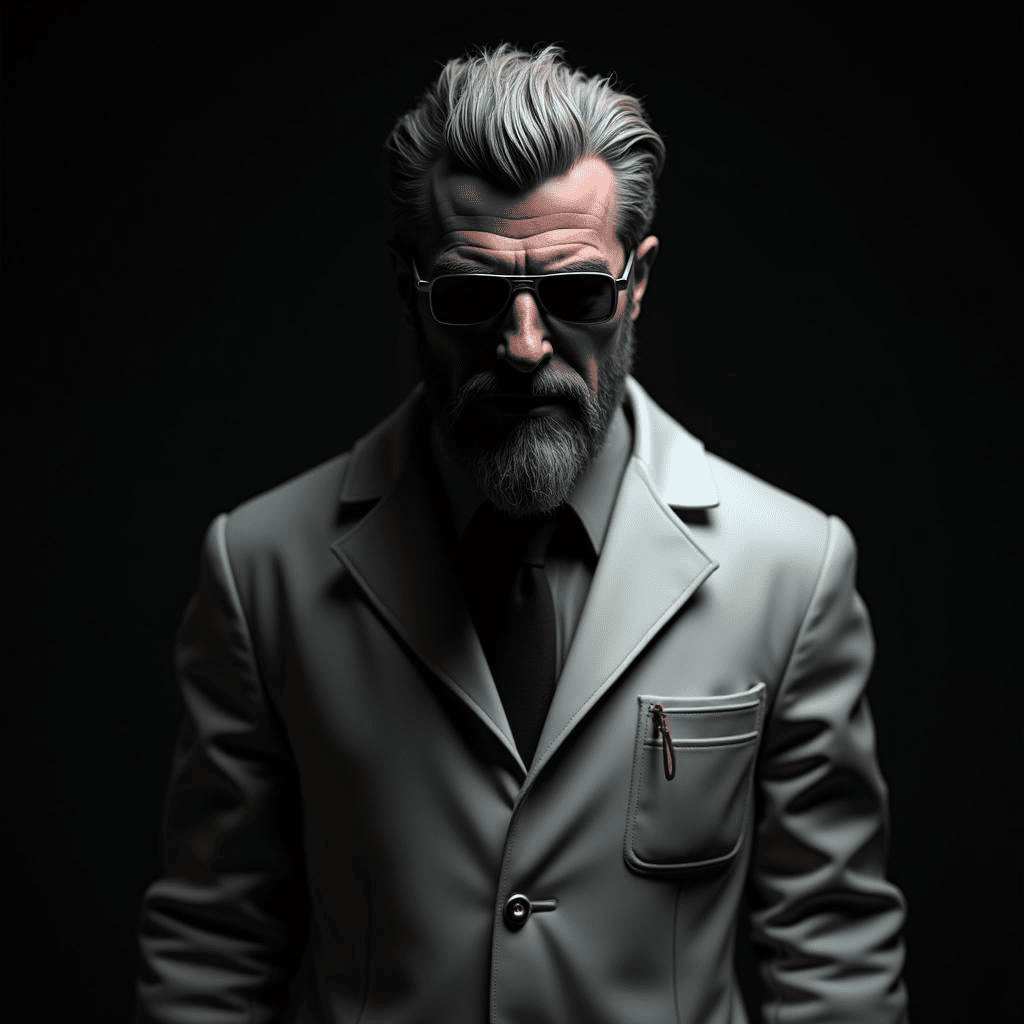}{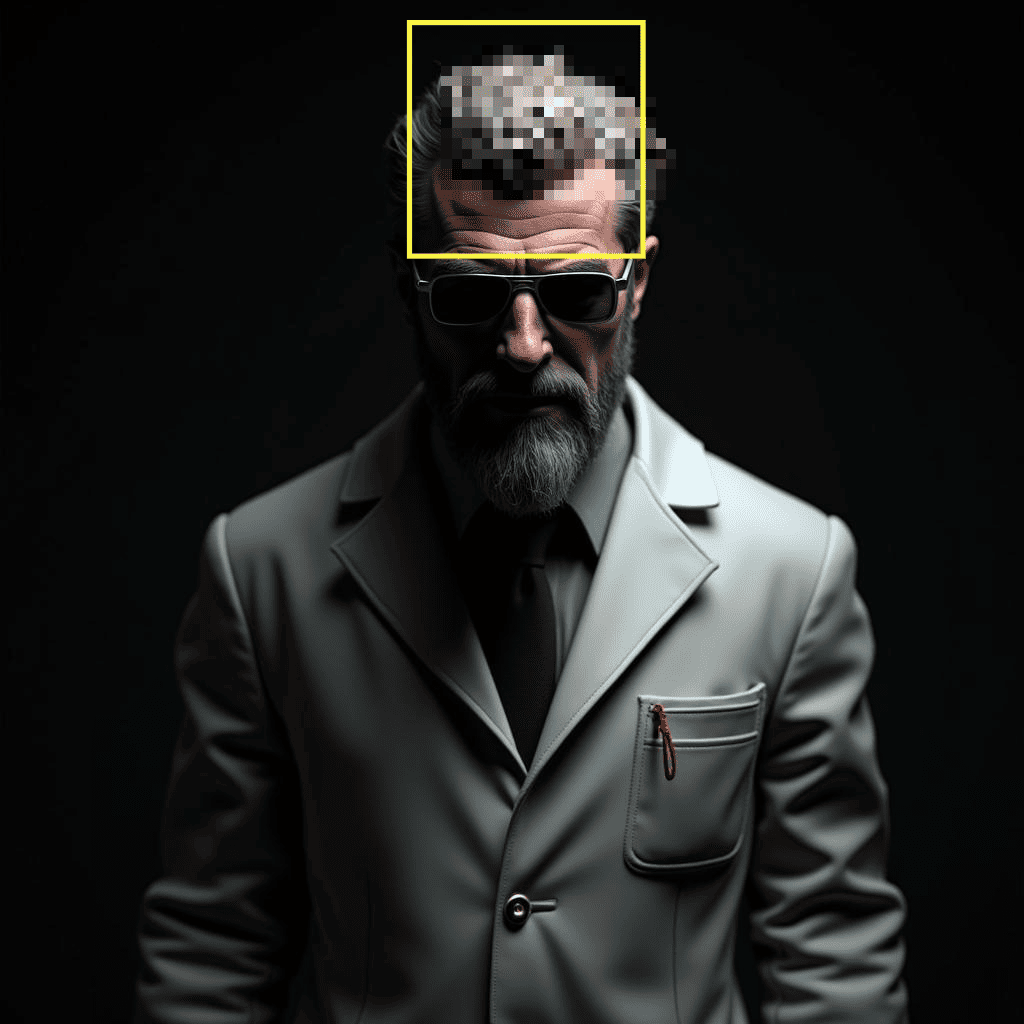}
    \hfill
    \promptpair{Kommissar Tarsis, anime style, pale skin, navy hat, imperial guard, sharp eyes, .....}{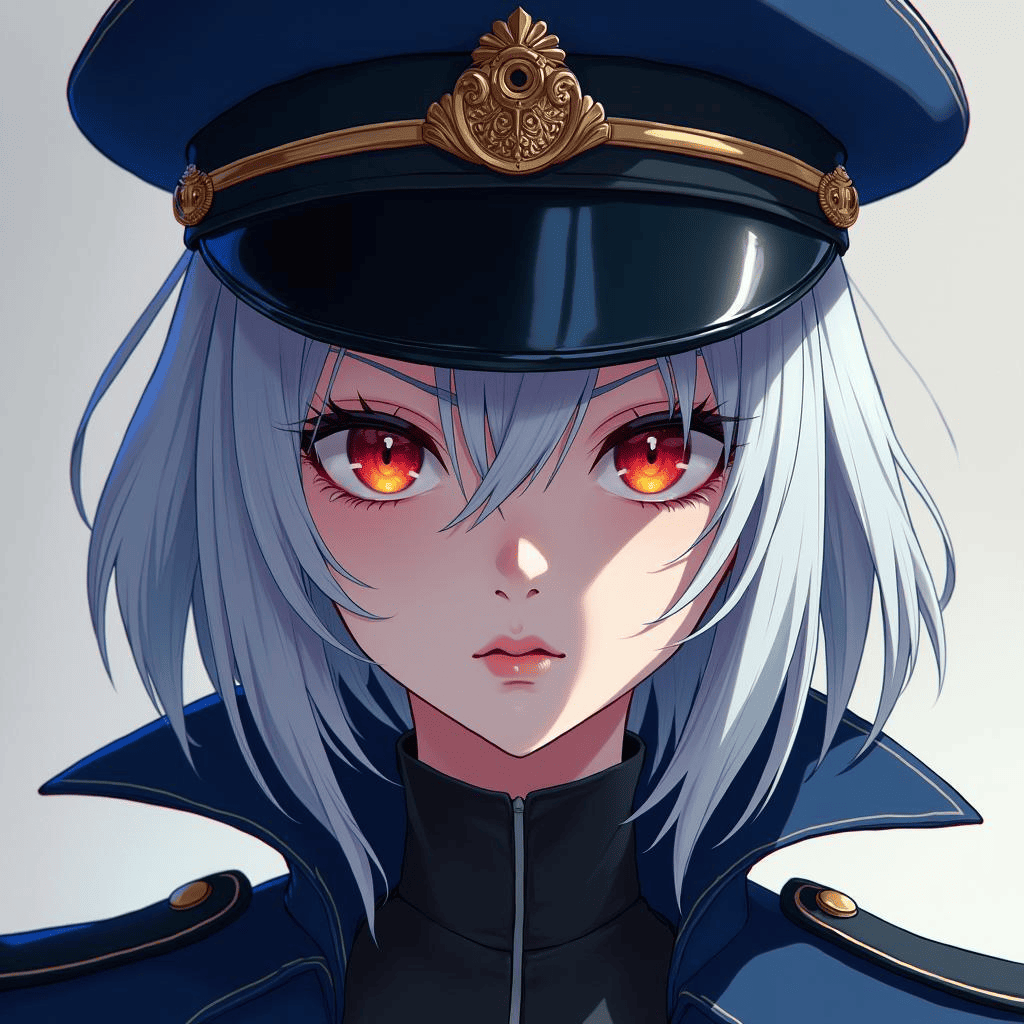}{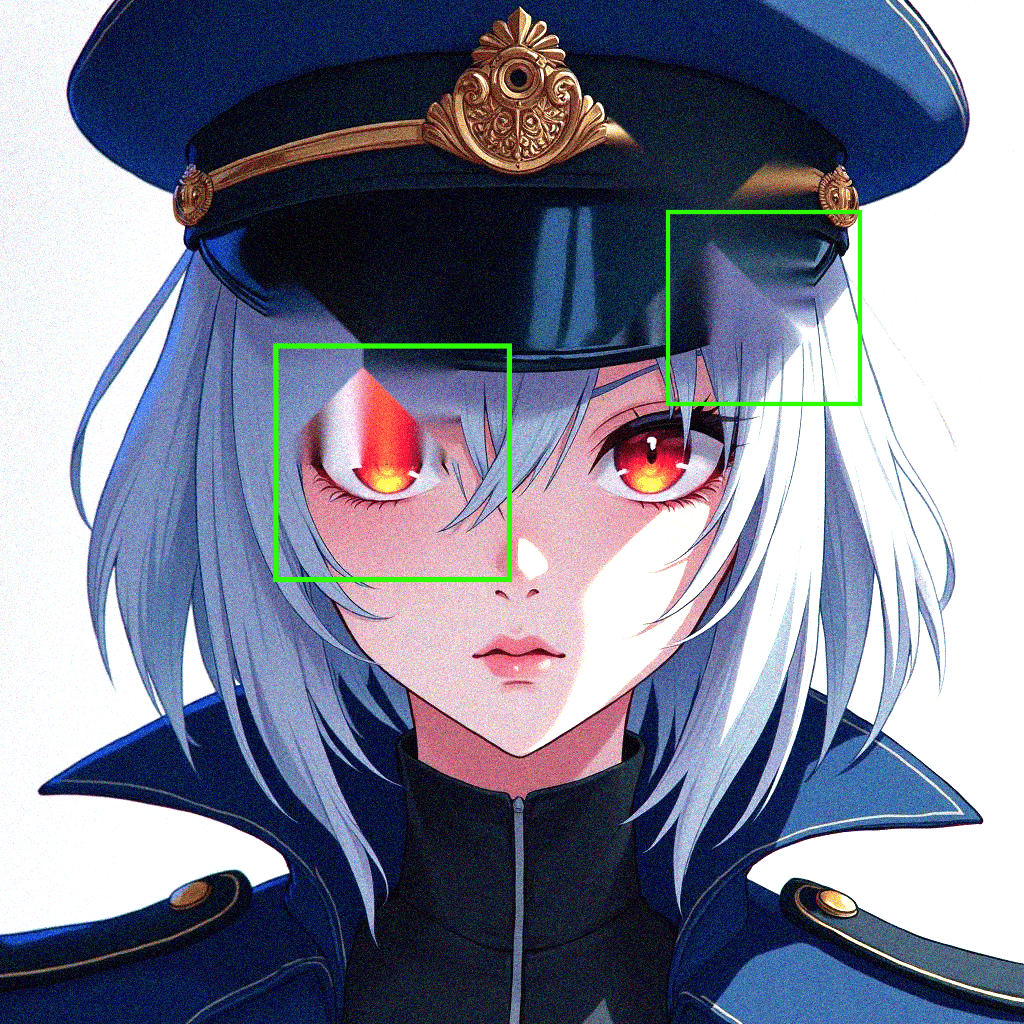}
    \hfill
    \promptpair{A yellow wall with Hello, world. in bold, cinematic letters .....}{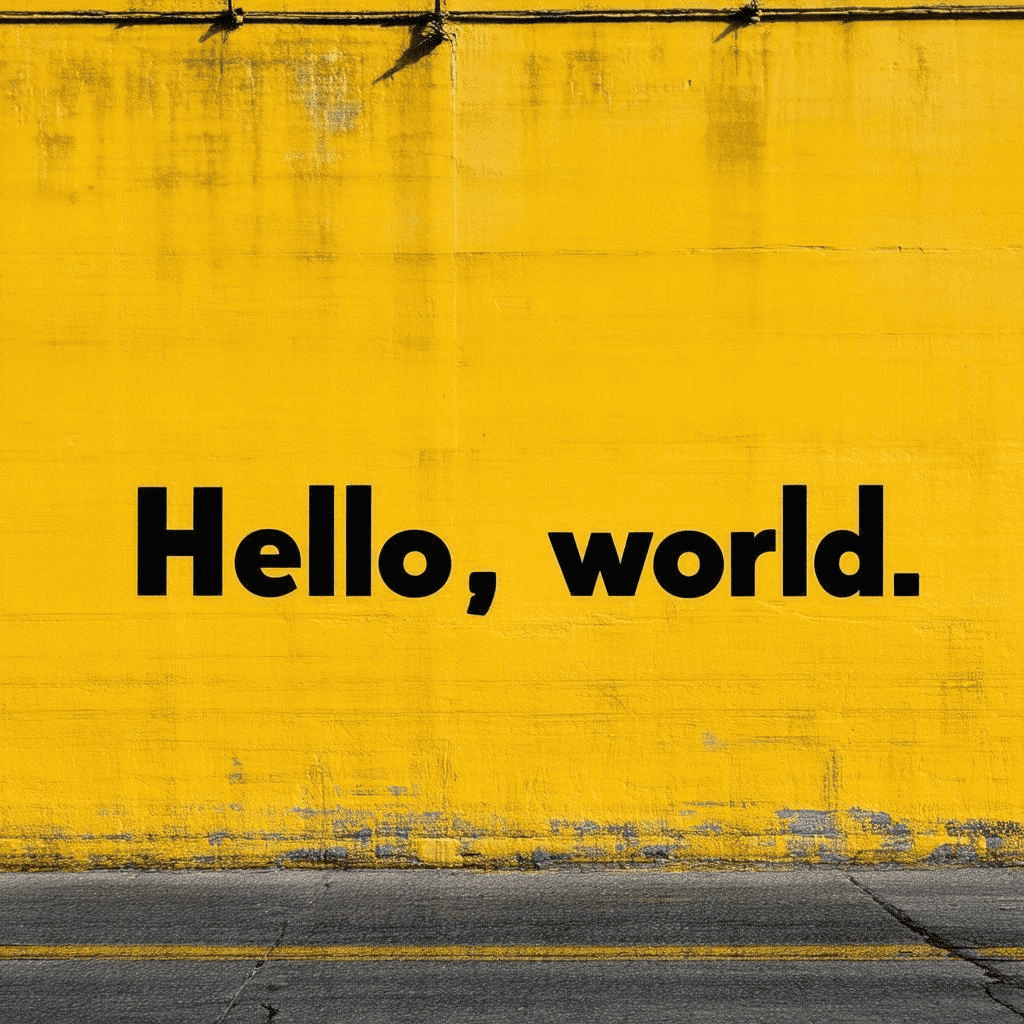}{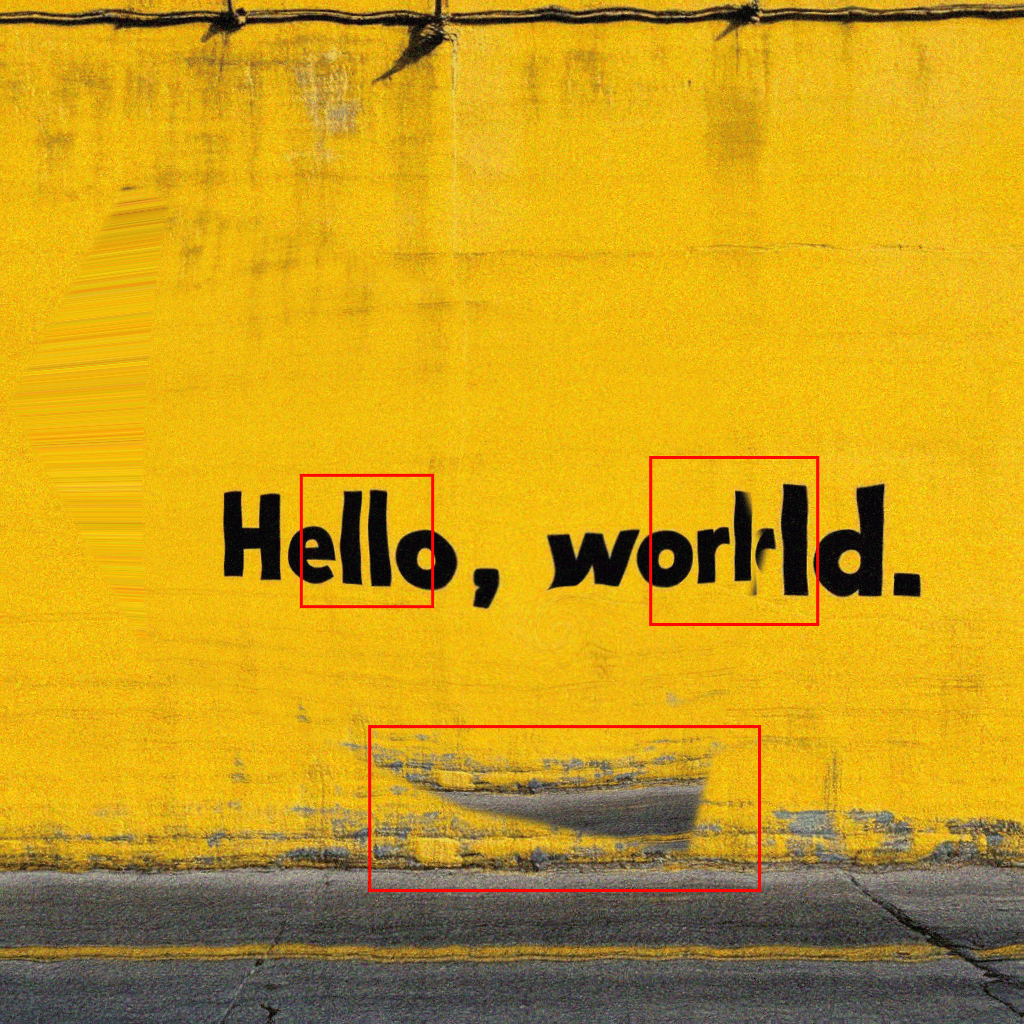}

    % \vspace{17pt}
    
    % \promptpair{Cinematic moody steampunk baroque mad max cars, rocket league, .....}{Images/output_images/win_9675.png}{Images/output_images/lose_9675.png}
    % \hfill
    % % ===================== ROW 3 =====================
    % \promptpair{An intricate neonpunk phonograph, crafted from layered cardboard .....}{Images/output_images/win_44539.png}{Images/output_images/lose_44539.png}
    % \hfill
    % \promptpair{A fantasy poster of three ancient trees, glowing with a mystical aura .....}{Images/output_images/win_7749.png}{Images/output_images/lose_7749.png}

    \caption{\textbf{Examples of preference pairs constructed with \textsc{PrefLens} across different prompts.} Bounding boxes highlight alignable perceptual differences relevant to human preference judgments, while largely retaining the overall semantic content.}
    \label{fig:dataset_examples}
\end{figure*}

Recovering such a \emph{local preference neighborhood} is itself challenging, as a preferred image may admit a large number of plausible alternatives. However, not all alternatives provide equally informative supervision. Cognitive theories of similarity offer a possible principle for identifying informative comparisons. In particular, Structure-Mapping Theory (SMT) posits that humans compare examples by first aligning their common structure and subsequently attending to \emph{\textbf{alignable differences}} between corresponding components~\cite{sagi2012difference,gentner1983structure,gentner1997structure}. As illustrated in Figure~\ref{fig:teaser}, the broader preference space surrounding a preferred image may contain both semantically distant and semantically aligned alternatives. SMT suggests that these comparisons contribute differently to learning: semantically distant alternatives primarily expose non-alignable differences, whereas semantically aligned alternatives preserve common structure and highlight fine-grained perceptual differences. Consequently, an informative \emph{local preference neighborhood} should consist of semantically aligned alternatives that retain common structure while exposing perceptual differences relevant to human judgments.

The notion of local preference neighborhoods naturally suggests three requirements for effective preference expansion. First, neighborhood construction should be \textbf{preferred-image anchored}, ensuring that generated comparisons remain centered around the preferred reference image so that common semantic structure is retained and meaningful alignable differences can emerge. Second, effective neighborhood exploration requires \textbf{fine-grained control over perceptual variations}, enabling systematic manipulation of localized visual attributes while largely retaining semantic identity. Finally, effective neighborhood recovery should maintain \textbf{diversity within the local neighborhood}. Since local preference neighborhoods may be highly dense, naively incorporating all neighboring samples can introduce substantial redundancy, which may negatively affect preference optimization~\cite{singh2024effective,surana2026mass}. Consequently, effective preference expansion should select compact subsets of neighboring samples that retain the information while reducing redundancy.

\setlength{\dashlinedash}{1pt}
\setlength{\dashlinegap}{1.5pt}
\setlength{\arrayrulewidth}{1.2pt}

\begin{table}[t]
\centering
\caption{
Comparison of preference dataset construction strategies across the proposed design principles.
}
\label{tab:expansion_comparison}
\small
\resizebox{\linewidth}{!}{%
\setlength{\tabcolsep}{8pt}
\renewcommand{\arraystretch}{1.25}

\begin{tabular}{
>{\centering\arraybackslash}m{4.2cm}
ccc}

\hline
\textbf{Method} 
& \makecell{\textbf{Anchor}\\\textbf{Based}} 
& \makecell{\textbf{Fine-Grained}\\\textbf{Visual Control}} 
& \makecell{\textbf{Preference}\\\textbf{Diversity}} \\
\hline

Human Preference Collection\\
~\citep{kirstain2023pick,xu2023imagereward,bai2025towards}
& \xmark 
& \xmark 
& Partial \\
\hdashline

Model Sampling + Ranking\\
~\cite{karthik2025scalable,croitoru2025curriculum,gu2024diffusion,liang2024step}
& \xmark 
& Partial 
& Partial \\
\hdashline

Self-Generated Supervision\\
~\cite{yuan2024self,zhang2024seppo}
& \xmark
& \xmark
& Partial \\
\hdashline

Multi-Dimensional Preferences\\
~\cite{lee2025calibrated,jang2025multi,liang2024rich}
& \xmark 
& Partial 
& Partial \\
\hdashline

Prompt-side Expansion\\
~\cite{li2026preference,xian2025free}
& Partial 
& \xmark 
& \cmark \\
\hdashline

% Image-side Expansion\\
% ~\cite{bansal2026preference}
% & \xmark 
% & \xmark 
% & \cmark \\
% \hdashline

\rowcolor{black!5}
\textbf{\textsc{PrefLens}}
& \cmark 
& \cmark 
& \cmark \\

\hline
\end{tabular}
}
\end{table}

Table~\ref{tab:expansion_comparison} compares existing preference dataset construction strategies along the three proposed design principles. While human-annotated datasets provide reliable supervision, they do not explicitly recover local neighborhoods around preferred images. Sampling-based, self-generated, and multi-dimensional preference approaches enrich supervision but offer limited control over anchor-centered perceptual variations~\cite{gu2024diffusion,yuan2024self,lee2025calibrated}. Prompt-side expansion methods diversify textual descriptions without directly controlling visual attributes~\cite{li2026preference,xian2025free}. Moreover, expanding candidate sets without considering their distribution can introduce redundant supervision~\cite{singh2024effective}. 

% Table~\ref{tab:expansion_comparison} summarizes existing approaches through the proposed design principles. Human preference collection provides high-quality supervision but does not explicitly recover local neighborhoods around preferred examples. Sampling-based methods generate additional candidates from the same prompts~\cite{karthik2025scalable,croitoru2025curriculum,gu2024diffusion}, preserving semantic context but without anchoring variations to the preferred image or explicitly controlling localized perceptual changes. SPO~\cite{liang2024step} constructs finer-grained preference pairs from candidates sharing a noisy latent, while~\cite{yuan2024self} and~\cite{zhang2024seppo} derive supervision from current or previous model generations, making the resulting variations model-dependent and less explicitly controlled. Prompt-side expansion methods~\cite{li2026preference,xian2025free} diversify textual descriptions but provide limited control over localized visual attributes. Moreover, existing expansion pipelines rarely account explicitly for the distribution of generated candidates, potentially resulting in redundant supervision concentrated in dense regions~\cite{singh2024effective}.

Motivated by these observations, we propose \textsc{PrefLens}, a framework for recovering \emph{local preference neighborhoods} around human-preferred images. \textsc{PrefLens} treats the preferred image as an anchor and systematically recovers controlled perceptual variations that largely retain its semantic identity. These variations expose diverse alignable differences and provide a richer characterization of the local preference structure surrounding the preferred example (Figure~\ref{fig:dataset_examples}). To obtain informative supervision, \textsc{PrefLens} subsequently selects diverse samples from the recovered neighborhood in a reward-defined preference space, reducing redundant comparisons while encouraging broad coverage of the local preference neighborhood.

% The resulting structured preference dataset can be readily used with preference optimization objectives such as Direct Preference Optimization, enabling diffusion models to better capture the nuanced perceptual preferences reflected in human judgments.

\noindent\textbf{Contributions.}
\begin{itemize}[noitemsep,nolistsep,leftmargin=*]

\item \textbf{Local preference neighborhoods for T2I alignment.}
We introduce the notion of \emph{local preference neighborhoods} for T2I alignment, where controlled perceptual variations expose diverse alignable differences around human-preferred images.

\item \textbf{\textsc{PrefLens} framework.}
We propose \textsc{PrefLens}, a framework that recovers local preference neighborhoods through controlled perceptual perturbations and constructs compact sets of informative neighbors via reward-guided preference-space selection.

\item \textbf{Improved data-efficient alignment.}
Our experiments show that \textsc{PrefLens} improves preference alignment under limited human supervision across diverse settings.

\end{itemize}

\section{Related Work}

\subsection{Preference Alignment for Text-to-Image Models.} Recent text-to-image (T2I) diffusion models have achieved remarkable generative performance through large-scale multimodal training~\cite{saharia2022photorealistic, rombach2022high, betker2023improving, podell2023sdxl}. Despite these advances, aligning model outputs with human preferences remains challenging. Early approaches adapted Reinforcement Learning from Human Feedback (RLHF)~\cite{christiano2017deep, ouyang2022training, black2023training, fan2023dpok, liang2024rich} to generative models. While effective, these methods introduce significant computational overhead and can suffer from reward hacking.

To simplify alignment, several works have proposed reward-free preference optimization methods that directly learn from pairwise comparisons. Examples include Diffusion-DPO~\cite{wallace2024diffusion}, DSPO~\cite{zhu2025dspo}, and related variants~\cite{li2024aligning, lu2025inpo, lusmoothed, gu2024diffusion, yangdense} that formulate alignment as supervised learning over preference likelihoods. Although these approaches substantially reduce training complexity, they rely on preference datasets that provide only sparse observations of human judgments. Consequently, recovering richer preference structure from limited annotations remains a key challenge for scalable alignment.

\subsection{Preference Datasets and Synthetic Preference Generation.} Large-scale human preference datasets such as \textit{Pick-a-Pic}~\cite{kirstain2023pick}, \textit{ImageReward}~\cite{xu2023imagereward}, and \textit{HPD}~\cite{wu2023human} constitute the primary source of supervision for T2I preference alignment. These datasets provide pairwise comparisons between generated images and have enabled both reward modeling and preference optimization.

To reduce annotation costs and enrich preference supervision, recent work has explored automated preference generation pipelines. One class of approaches relies on pretrained models or vision-language models to simulate preference feedback and generate synthetic preference pairs~\cite{sun2024dreamsync,wu2024multimodal}. Another line of work constructs large candidate pools through model sampling and subsequently ranks candidates using automated evaluators~\cite{gu2024diffusion,karthik2025scalable}. While such approaches improve scalability, they typically provide limited control over the local preference structure surrounding human-preferred examples. In contrast, our work recovers local preference neighborhoods around preferred images and systematically selects informative neighbors that expose localized, alignable differences.

\subsection{Structured Preference Supervision and Dataset Expansion.}
Beyond the scale of preference datasets, recent work has highlighted the importance of constructing informative preference supervision for alignment. In representation learning and metric learning, techniques such as hard negative mining focus training on examples that reveal meaningful decision boundaries~\cite{schroff2015facenet}, and similar principles have been adopted in preference alignment to improve optimization stability. Several approaches have explored enriching preference supervision through prompt-side augmentation~\cite{xian2025free, li2026preference}, model sampling and ranking pipelines~\cite{gu2024diffusion, karthik2025scalable}, and multi-dimensional preference modeling using multiple reward signals~\cite{lee2025calibrated, jang2025multi}. While these approaches improve the quantity or expressiveness of preference supervision, they typically provide limited control over the localized perceptual factors that vary across comparisons.

Controlled image perturbations have also been explored as useful training signals, e.g., AugMix~\cite{hendrycks2019augmix} for robustness. In preference alignment, Diffusion-NPO~\cite{wang2025diffusion} primarily introduces a negative preference optimization objective and, alongside this contribution, studies controlled corruption of negative samples, showing that mild degradations can improve alignment. Compared with these prior directions, \textsc{PrefLens} combines anchor-centered controlled perturbations with reward-guided neighborhood organization and diversity-aware selection to construct compact local preference neighborhoods for optimization.

\section{Our Approach}

\subsection{Problem Formulation}
\begin{figure}[!tbh]
    \centering
    \includegraphics[width=\linewidth]{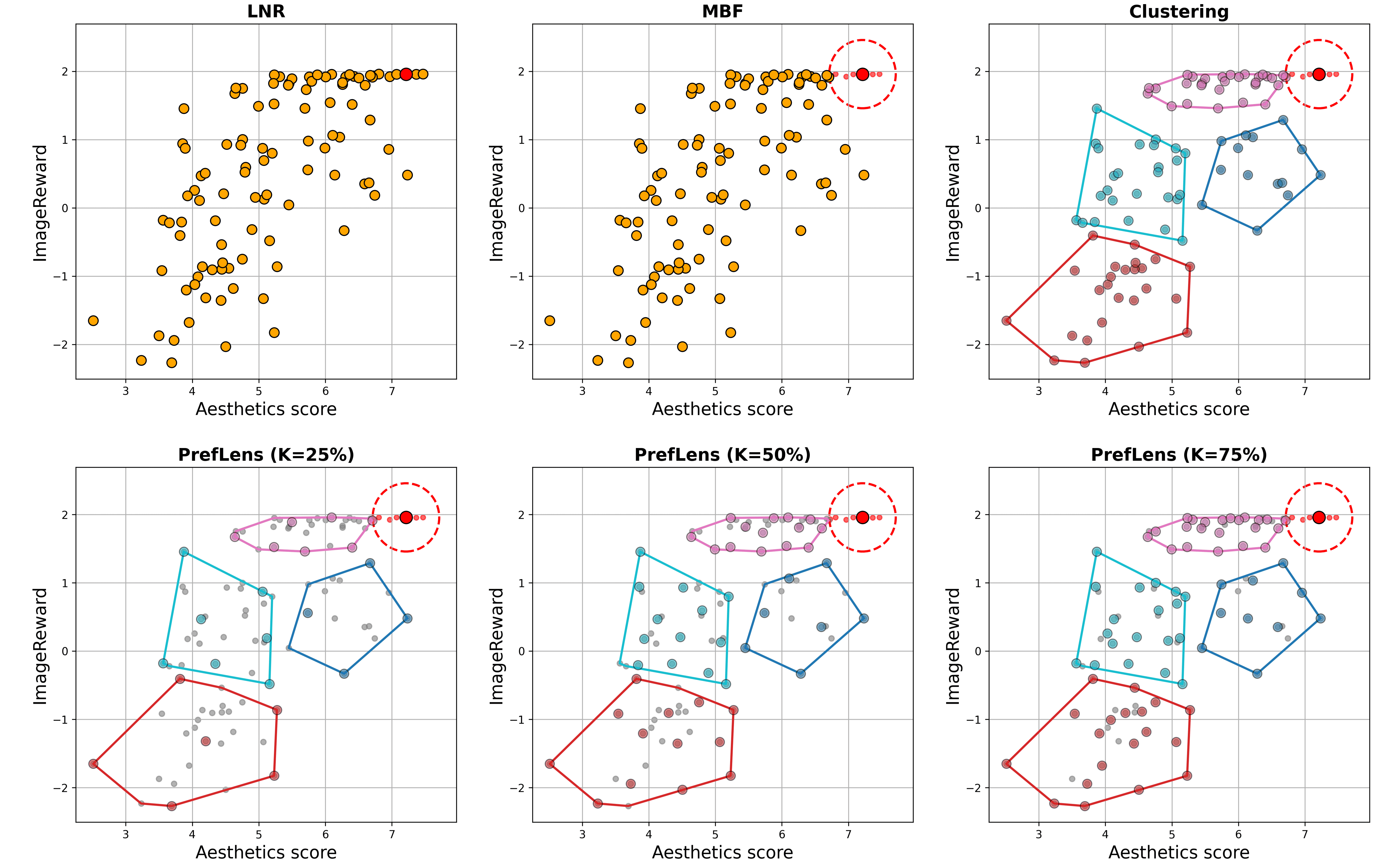}
    
    \caption{Visualization of LNR in the proposed reward-defined preference space (Aesthetics score and ImageReward). Margin-Based Filtering (MBF) removes candidates that are too close to the anchor in reward space (\textcolor{red}{\textbullet}), after which the remaining candidates are partitioned into distinct variation modes. Finally, \textsc{PrefLens} selects diverse and informative neighbors under different sampling budgets ($K=25\%, 50\%, 75\%$).}

    \label{fig:cluster_visualization_main}
\end{figure}

Let $\mathcal{D}=\{(p_i,x_i^+,x_i^-)\}_{i=1}^{|\mathcal D|}$ denote a human preference dataset, where $p_i$ is a text prompt, $x_i^+$ is the preferred image, and $x_i^-$ is the rejected image. For each preferred image $x^+$, our objective is to recover a local preference neighborhood
\[
\mathcal{N}(x^+)=\{x^-_1,\dots,x^-_{|\mathcal{N}(x^+)|}\}
\]
consisting of informative negatives associated with the anchor image. Each neighbor $x^-_i$ represents a controlled perceptual variation of the preferred image, enabling each human preference annotation to be transformed into a structured set of informative comparisons (Figure~\ref{fig:cluster_visualization_main}).

\subsection{Recovering Local Preference Neighborhoods (LNR)}
Given an anchor image $x^+$, we recover its local preference neighborhood by generating controlled perceptual variations around the anchor using a family of perturbation operators. These operators systematically modify visual attributes while largely retaining the semantic identity of the anchor, thereby producing neighbors that expose diverse \emph{alignable differences} relevant to preference alignment.

Let $\mathcal{T}=\{T_1,\dots,T_M\}$ denote a family of perturbation operators that introduce controlled variations in visual fidelity, structure, and appearance.

A candidate negative is generated by applying a composition of randomly sampled perturbation operators from $\mathcal{T}$:
\[
\tilde{x}^{-}=T_{j_k}\circ T_{j_{k-1}}\circ\dots\circ T_{j_1}(x^+),
\qquad T_{j_t}\in\mathcal{T}.
\]
By repeatedly sampling perturbation sequences and applying the corresponding compositions to the anchor image, we obtain a candidate pool
\[
\mathcal{C}=\{\tilde{x}_1^-,\dots,\tilde{x}_N^-\}.
\]
Sequential compositions of perturbation operators enable exploration of diverse regions of the local preference neighborhood while largely retaining semantic identity. Additional neighborhood visualizations and preference-pair
examples along with a theoretical analysis are provided in
Appendices~\ref{sec:neighborhood_visualization}, \ref{sec:dataset_examples}, and~\ref{sec:theory_appendix}, respectively.

\subsection{Reward-Guided Informative Neighborhood Selection}
In this step, we seek to obtain a compact set of informative neighbors while retaining the structure of the recovered local neighborhood. To induce an ordering over candidate neighbors, we require a notion of distance defined on the neighborhood. To this end, we map the recovered neighborhood from the image space $\mathcal X$ into a reward-defined preference space $\mathcal Q\subset\mathbb R^d$ using pretrained reward models.

We define a mapping
$\mathbf q:\mathcal X\rightarrow\mathcal Q$ as
\[
\mathbf q(x)=[r_1(x),r_2(x),\dots,r_d(x)],
\]
where $r_i(x)$ denotes the score assigned by the $i$-th reward model. We further equip $\mathcal Q$ with the Euclidean metric
\[
d(\mathbf q_i,\mathbf q_j)=\|\mathbf q_i-\mathbf q_j\|_2,
\]
so that $(\mathcal Q,d)$ forms a metric space. The induced metric provides a reward-guided notion of proximity for filtering and selecting candidate neighbors. 
% Let
% \[
% \mathcal N_{\mathbf q}
% =
% \{\mathbf q(x):x\in\mathcal C\cup\{x^+\}\}
% \]
% denote the recovered neighborhood represented in the preference metric space $(\mathcal Q,d)$.

To remove candidates that remain too close to the anchor in the reward-defined preference space, we retain the candidate set
\[
\mathcal C^{\epsilon}
=
\left\{
x\in\mathcal C:
d(\mathbf q(x),\mathbf q(x^+))\ge\epsilon
\right\}.
\]

The margin constraint removes candidates exhibiting only negligible deviations from the anchor while retaining more informative variations. Similar \textbf{margin-based filtering (MBF)} is widely used in metric learning and hard negative mining to focus optimization on informative examples~\cite{schroff2015facenet}. Conceptually, this filtering is also motivated by the notion of \emph{just noticeable differences} in perceptual psychology, where sufficiently small variations can be difficult to distinguish reliably~\cite{wu2019survey}.

To identify informative variation modes, we partition the retained candidates according to their representations in the reward-defined preference space:
\[
\mathcal C^{\epsilon}
=
C_1\cup C_2\cup\cdots\cup C_L,
\quad
C_i\cap C_j=\emptyset,\; i\neq j.
\]
Each cluster approximates a distinct variation mode within the recovered neighborhood.

To preserve the relative density of different variation modes, we allocate a sampling budget proportional to cluster size. Given a sampling ratio $K\in(0,1]$, the budget assigned to cluster $C_\ell$ is
\[
k_\ell=\lfloor K|C_\ell|\rfloor.
\]
For each cluster $C_\ell$, we select a subset of informative neighbors by solving
\[
S_\ell^*
=
\arg\max_{S\subset C_\ell,\; |S|=k_\ell}
\;
\min_{\substack{x_i,x_j\in S\\i\neq j}}
d(\mathbf q(x_i),\mathbf q(x_j)).
\]

The max--min objective promotes diversity within each variation mode by favoring mutually distant samples in the reward-defined preference space and avoiding redundant selections from the same region. This objective corresponds to the classical \emph{max--min dispersion} problem\footnote{The \emph{max--min dispersion} problem is NP-hard~\cite{ravi1994heuristic}. However, the candidate neighborhoods considered in our setting are moderate in size, making approximate solutions computationally practical.}~\cite{erkut1990discrete,ravi1991facility}. In practice, we approximate this objective using a graph-based selection procedure, where candidates are represented as nodes and edges connect sufficiently distant samples in the preference space.
The final neighborhood is obtained as,
\[
\mathcal N(x^+)=\bigcup_{\ell=1}^{L}S_\ell^*,
\qquad
|\mathcal N(x^+)|=\sum_{\ell=1}^{L}k_\ell.
\]
Overall, the proposed procedure recovers compact local preference neighborhoods that expose diverse alignable differences while largely retaining semantic content.
% The dataset $\widetilde{\mathcal D}$ can subsequently be used with preference optimization algorithms to provide more informative supervision.
\section{Experiments}

% \subsection{Experimental Setup}

\subsection{Training Data and Models}
We use the \texttt{open\allowbreak-image-\allowbreak preferences\allowbreak-v1-\allowbreak binarized} dataset~\cite{huggingfaceDataisbettertogetheropenimagepreferencesv1binarizedDatasets}, which contains 7,459 human-annotated image preference pairs. We refer to this dataset as \textbf{\textit{OpenPref}} throughout the paper. In \textit{OpenPref}, images are generated using \texttt{FLUX.1-dev}~\cite{huggingfaceBlackforestlabsFLUX1devHugging} and \texttt{SD-3.5-\allowbreak large}~\cite{esser2024scaling}. We apply \textsc{PrefLens} to \textit{OpenPref} to recover local preference neighborhoods around human-preferred images, which are subsequently used to construct expanded preference pairs for alignment. To represent images in preference space, we employ two reward models, \textit{ImageReward}~\cite{xu2023imagereward} and \textit{Aesthetic Score}~\cite{schuhmann2022laion}, which provide complementary preference signals for characterizing the recovered neighborhoods. For preference alignment, we fine-tune two widely used diffusion models: \textit{SD v1.5}~\cite{Rombach_2022_CVPR} and \textit{SDXL-1.0-base}~\cite{podell2023sdxl}.

\subsection{Baselines}
We evaluate \textsc{PrefLens} with \emph{Diff.-DPO} and \emph{DSPO}~\cite{wallace2024diffusion,zhu2025dspo}, and compare against alignment-method, preference-dataset, and neighborhood-construction baselines.

\textit{\textbf{Alignment-method baselines.}}
We first report \emph{SFT}, where the model is fine-tuned using only the
preferred samples from \textit{OpenPref}, providing a reference before preference-based optimization. For SD v1.5, we compare against
\emph{Diff.-DPO}~\cite{wallace2024diffusion}, \emph{SPIN-Diffusion}~\cite{yuan2024self}, \emph{SePPO}~\cite{zhang2024seppo}, and \emph{SPO}~\cite{liang2024step}.
For SDXL-Base, we compare against \emph{Diff.-DPO} and \emph{SPO}~\cite{liang2024step}. Together, these baselines span diverse preference-alignment paradigms,
including learning from human preference pairs, self-play, and
model-generated preference construction strategies. 
% Except for SFT, all alignment-method baselines are evaluated using their publicly released checkpoints.

\textit{\textbf{Human-preference-dataset baselines.}}
We compare different sources of human preference supervision while keeping the underlying alignment algorithm fixed. We train using \emph{HPDv2}, \emph{Pick-a-Pic-v2}, and \emph{OpenPref}. This comparison evaluates whether the preference dataset constructed by \textsc{PrefLens} provides more effective supervision than existing human-annotated preference datasets.

\textit{\textbf{Neighborhood-construction baselines.}}
We further compare against alternative strategies for constructing preference
pairs from the same preferred anchors: \textbf{(i)} \emph{LNR}, where local
preference neighborhoods are generated using the proposed perturbation strategy
and all recovered neighbors are retained for training; \textbf{(ii)}
\emph{LNR+RS}, where the same candidate neighborhoods are constructed but
negatives are selected uniformly at random instead of using the proposed
informative-neighbor selection; \textbf{(iii)} \emph{AugMix}, where AugMix
transformations are used to construct a single negative image for each preferred
image; and \textbf{(iv)} \emph{AugMix-MN}, where multiple AugMix-generated
negatives are associated with each preferred image. 
% Finally, \textsc{PrefLens} uses the complete proposed pipeline, comprising local neighborhood recovery followed by preference-space filtering and informative, diversity-aware neighbor selection.

\subsection{Evaluation Protocol}
We evaluate T2I generation using 500 prompts from each of three
widely used benchmarks: Pick-a-Pic-v2~\cite{kirstain2023pick},
PartiPrompts~\cite{yu2022scaling}, and GenAIBench~\cite{li2024genai}.

\textbf{\textit{Automatic metrics.}}
We report win rates (\%) over the corresponding base model and organize the automatic evaluation metrics into two categories:
\emph{image quality} and \emph{human preference}. For image quality, we
report Aesthetic Score~\cite{schuhmann2022laion} using both ViT-L/14 and
ViT-B/16 predictors. For human preference, we report
ImageReward~\cite{xu2023imagereward}, HPSv2~\cite{wu2023human}, and PickScore~\cite{kirstain2023pick}. Since Aesthetic Score (ViT-L/14) and ImageReward are used in
preference-space construction, we additionally report Aesthetic Score (ViT-B/16), HPSv2, and PickScore as held-out metrics to assess potential reward overfitting and evaluation circularity.

\textbf{\textit{Human evaluation.}}
We further conduct human evaluation using
three annotators, who rank images generated by the base model,
\emph{Diff.-DPO with OpenPref}, and
\emph{Diff.-DPO with \textsc{PrefLens}} across 100 samples from each
benchmark. We report both \textit{Top-1 preference} percentages and
\textit{Borda ranking}~\cite{emerson2013original} scores.
Inter-annotator agreement, measured using Kendall's $\tau$~\cite{mcleod2005kendall}, is 0.71.

\textbf{\textit{VLM-as-a-Judge.}}
We perform VLM-as-a-Judge evaluation using
\texttt{Qwen3-VL-8B-Instruct}~\cite{bai2025qwen3} over all 500 samples from each benchmark.

\textbf{\textit{Compositional fidelity.}} To evaluate compositional
fidelity, we report average
\textit{TIFA} scores~\cite{hu2023tifa} on 1,000 samples from
T2I-CompBench++~\cite{huang2025t2i}, using \texttt{Salesforce/blip-vqa-capfilt-large}~\cite{li2022blip} and \texttt{microsoft/git-large}~\cite{wang2022git} as VQA
backbones, with evaluation questions generated by
\texttt{GPT-4o}~\cite{hurst2024gpt}. All evaluation instructions are
provided in Appendix~\ref{sec:evaluation_instructions}.

\subsection{Implementation Details}

For each human-annotated preference pair, we recover a local preference neighborhood by generating $N=100$ candidates through controlled visual perturbations applied to the preferred image. Each candidate is produced by composing a sequence of $3$--$11$ perturbations randomly sampled from a set of transformations. We organize the perturbations into three groups according to the type of visual variation they induce: \textbf{\textit{(i) Photometric}}, consisting of Gaussian noise, salt-and-pepper noise, posterization, channel manipulation, and color jitter; \textbf{\textit{(ii) Geometric}}, consisting of shear, elastic deformation, swirl, sine-wave warping, twist deformation, and radial zoom; and \textbf{\textit{(iii) Fidelity Reduction}}, consisting of Gaussian blur, pixelation, JPEG compression, and inpainting-based region removal. Further details on the perturbations are provided in Appendix~\ref{sec:perturbations_desc}.

By default, we use the raw reward scores to construct the preference space and set the margin threshold to $\epsilon=0.5$. The preference representations are partitioned using $k$-means clustering. Training configurations, hyperparameter sensitivity,
and convergence analyses are provided in
Appendices~\ref{sec:hyper_data} and~\ref{sec:optimization_analysis}.

\subsection{Ablation Studies}
\textbf{\textit{Perturbation families.}}
We independently evaluate Group 1 (Photometric), Group 2 (Geometric), and Group 3 (Fidelity Reduction), and compare them with their combined use, \textbf{Group 1+2+3}.

\textbf{\textit{Perturbation strength.}}
By default, each perturbation samples its strength from
$[\lambda_{\min},\lambda_{\max}]$. We compare this range-based strategy with two fixed-strength variants: \textbf{Min} and \textbf{Max}, which always uses $\lambda_{\min}$ and $\lambda_{\max}$, respectively.

\textbf{\textit{Margin threshold.}}
We analyze the sensitivity of neighborhood recovery to the margin threshold
by evaluating $\epsilon\in\{0.1,0.5,1.0\}$.

\textbf{\textit{Neighborhood selection.}}
We compare three variants: \textbf{(i)} \textsc{PrefLens-RS}, which randomly samples negatives within each cluster; \textbf{(ii)} \textsc{PrefLens-US}, which samples an equal number from each cluster; and \textbf{(iii)}
\textsc{PrefLens}, which uses the proposed sampling strategy.
For \textsc{PrefLens-RS} and \textsc{PrefLens-US}, we use the same fixed number of negatives to isolate the effect of the sampling strategy, while for \textsc{PrefLens} we additionally evaluate budgets of $K=25\%$, $50\%$, and $75\%$.

\textbf{\textit{Preference-space normalization.}} We compare the default raw-score preference space with a $z$-score normalized variant using training-set statistics to assess sensitivity to reward-model scale.

\subsection{Preference-Neighborhood Diagnostics}
We evaluate the recovered preference neighborhoods along two complementary dimensions.

\textbf{\textit{Preference reliability and semantic proximity.}} We measure whether the preferred anchor is consistently favored over the recovered negative using Aesthetic Score, ImageReward, and two VLM judges (\texttt{LLaVA-v1.5-7B} and \texttt{Qwen3-VL-4B-Instruct}) over the full recovered dataset; human consistency is additionally evaluated on 500 pairs. We also measure anchor--neighbor semantic similarity using CLIP~\cite{radford2021learning} and DINO-v2~\cite{oquab2023dinov2}.

\textbf{\textit{Human-relevant preference structure.}}
To extract interpretable preference principles from each dataset, we apply Inverse Constitutional AI (ICAI)~\cite{findeis2025inverse} independently to HPDv2, Pick-a-Pic-v2, OpenPref, and \textsc{PrefLens}, and group the recovered principles into six categories. 
% We further ask two annotators to categorize 500 recovered pairs as \emph{easy}, \emph{medium}, or \emph{hard} based on the perceived difficulty of the preference comparison.
\section{Results}

\paragraph{\textbf{Q1: Do synthetic preference pairs capture human-relevant preference structure?}}
Applying ICAI to HPDv2, Pick-a-Pic-v2, OpenPref, and \textsc{PrefLens} yields 12, 56, 49, and 86 preference principles, respectively. We organize these principles into six preference categories, with the complete category-wise distributions for all datasets provided in Appendix~\ref{sec:additional_quantitative} (Table~\ref{tab:icai_principle_distribution}). As shown in Figure~\ref{fig:icai_constitution_comparison}, the principles recovered from \textsc{PrefLens} exhibit qualitative correspondence with those extracted from OpenPref, suggesting that the synthetic pairs capture preference characteristics present in human annotations rather than being determined solely by the applied perturbations.

% The perceptual-difficulty analysis provides complementary evidence. The two annotators assign easy:medium:hard ratios of 486:12:2 and 361:120:19, showing that perceived difficulty varies across annotators. Together with the principle-level correspondence, these results suggest that \textsc{PrefLens} produces human-relevant preference comparisons with varying perceptual difficulty rather than a homogeneous collection of clean-versus-perturbed examples.

\begin{figure}[t]
    \centering
        \includegraphics[width=\linewidth]{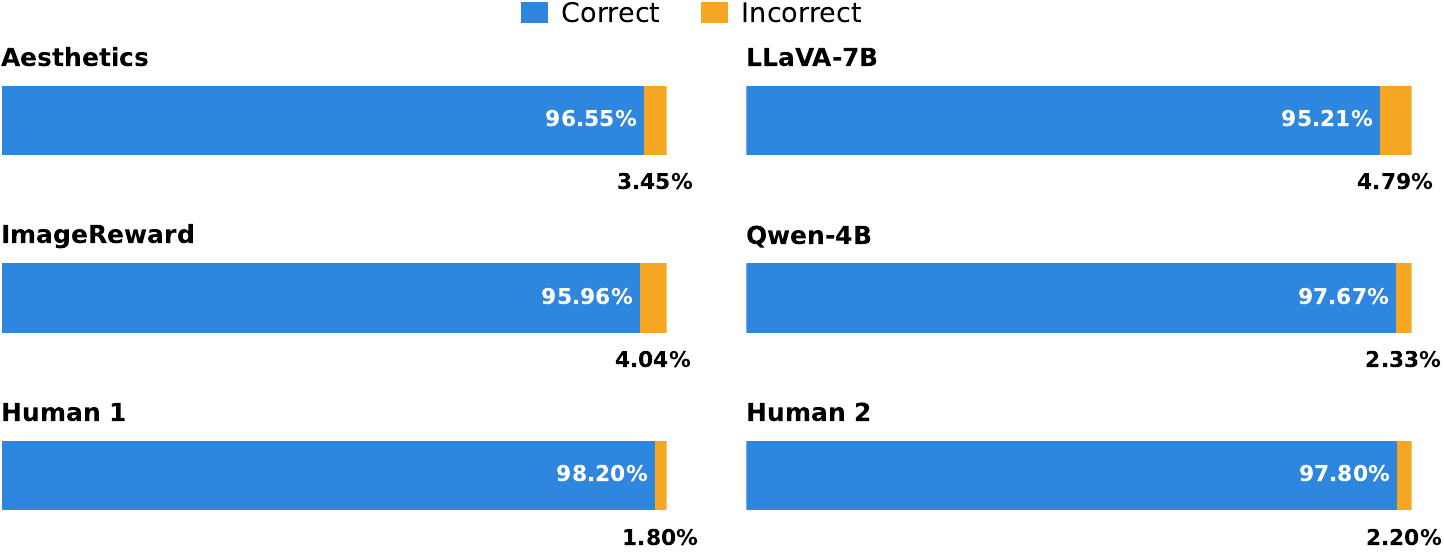}
    \caption{\textbf{Consistency analysis of recovered preference pairs.} Preference correctness (\%) of recovered local preference neighborhoods measured using multiple evaluators.}
    \label{fig:pref_consistency}
\end{figure}

\paragraph{\textbf{Q2: Are the recovered preference neighborhoods reliable and locally consistent?}}
To assess whether the synthetic preference pairs recovered by \textsc{PrefLens} provide reliable supervision, we evaluate their preference consistency across multiple evaluators. For each recovered pair $(x^+,x^-)$, we test whether the preferred anchor $x^+$ is consistently judged more favorable than the corresponding synthetic negative $x^-$. As shown in Fig.~\ref{fig:pref_consistency}, the recovered pairs exhibit high agreement across reward models, VLMs, and human evaluators, indicating that the generated comparisons largely preserve the intended preference ordering.
\begin{table}[!htb]
\centering
\caption{
Win-rate (\%) comparison on Pick-a-Pic-v2 for (a) SD v1.5 and (b) SDXL-Base.
Best results are highlighted in \textbf{boldface},
and second-best results are \underline{underlined}.
Results on PartiPrompts and GenAIBench are provided in Appendix~\ref{sec:extended_benchmark_results}.
}
\label{tab:results_win_rate}
% ===================== SUBTABLE (a) =====================
\begin{subtable}{\linewidth}
\centering
\resizebox{\linewidth}{!}{
\setlength{\tabcolsep}{6pt}
\begin{tabular}{l p{1.8cm} p{1.8cm} p{1.8cm} p{1.8cm} p{1.8cm} p{1.5cm}}
\toprule

\multirow{3}{*}{\textbf{Method}} &
\multicolumn{2}{c}{\textbf{Image Quality}} &
\multicolumn{3}{c}{\textbf{Human Preference}} &
\multirow{2}{*}{\centering\arraybackslash\textbf{Average}} \\
\cmidrule(lr){2-3}
\cmidrule(lr){4-6}
&
\centering\textbf{Aesthetics} &
\centering\textbf{Aesthetics} &
\centering\textbf{ImageReward} &
\centering\textbf{HPS v2} &
\centering\textbf{PickScore} & \\
&
\centering\textbf{(ViT-14)} &
\centering\textbf{(ViT-16)} &
\centering\textbf{} &
\centering\textbf{} &
\centering\textbf{} & \\
\midrule

% -------- PASTE FIRST TABLE ROWS HERE --------
SFT       & \centering 46.8 & \centering 47.8 & \centering 49.2 & \centering 50.6 & \centering 49.8 & \centering\arraybackslash 48.84 \\
SPIN      & \centering 64.2 & \centering 65.2 & \centering 62.6 & \centering 71.2 & \centering 66.8 & \centering\arraybackslash 66.00 \\
SePPO     & \centering 65.6 & \centering 62.8 & \centering 63.2 & \centering 76.2 & \centering 67.4 & \centering\arraybackslash 67.04 \\
SPO       & \centering 61.2 & \centering 62.0 & \centering 56.6 & \centering 72.8 & \centering 66.4 & \centering\arraybackslash 63.80 \\
Diff.-DPO & \centering 50.2 & \centering 52.2 & \centering 56.4 & \centering 61.6 & \centering 64.2 & \centering\arraybackslash 56.92 \\

\\[-1em] \cdashline{1-7}[0.4pt/1.5pt] \\[-1em]

\textbf{Diffusion-DPO} & & & & & & \\

\hspace{5mm}w/ HPDv2
& \centering 50.6 & \centering 49.4 & \centering 54.4
& \centering 60.0 & \centering 53.0
& \centering\arraybackslash 53.5 \\

\hspace{5mm}w/ Pick-a-Pic-v2
& \centering 50.6 & \centering 51.6 & \centering 50.6
& \centering 55.6 & \centering 51.8
& \centering\arraybackslash 52.0 \\

\hspace{5mm}w/ OpenPref
& \centering 47.2 & \centering 49.4 & \centering 49.8
& \centering 53.6 & \centering 50.4
& \centering\arraybackslash 50.1 \\

\hspace{5mm}w/ LNR
& \centering 58.6 & \centering 61.0 & \centering 58.8
& \centering 66.6 & \centering 61.8
& \centering\arraybackslash 61.4 \\

\hspace{5mm}w/ LNR + RS
& \centering 62.2 & \centering 54.0 & \centering 60.4
& \centering 68.0 & \centering 63.2
& \centering\arraybackslash 61.6 \\

\hspace{5mm}w/ AugMix
& \centering 42.0 & \centering 55.0 & \centering 49.0
& \centering 57.8 & \centering 37.0
& \centering\arraybackslash 48.2 \\

\hspace{5mm}w/ AugMix-MN
& \centering 42.0 & \centering 52.6 & \centering 53.2
& \centering 63.4 & \centering 37.6
& \centering\arraybackslash 49.8 \\

\hspace{5mm}w/ \textsc{PrefLens}
& \centering \textbf{68.8}
& \centering \underline{67.0}
& \centering \underline{66.6}
& \centering 72.4
& \centering \textbf{71.0}
& \centering\arraybackslash \underline{69.16} \\

\\[-1em] \cdashline{1-7}[0.4pt/1.5pt] \\[-1em]

\textbf{DSPO} & & & & & & \\

\hspace{5mm}w/ HPDv2
& \centering 56.8 & \centering 53.8 & \centering 61.0
& \centering 71.6 & \centering 63.0
& \centering\arraybackslash 61.2 \\

\hspace{5mm}w/ Pick-a-Pic-v2
& \centering 59.6 & \centering 58.2 & \centering 62.8
& \centering 74.4 & \centering 62.0
& \centering\arraybackslash 63.4 \\

\hspace{5mm}w/ OpenPref
& \centering 64.2 & \centering 63.2 & \centering 59.8
& \centering 74.8 & \centering 65.8
& \centering\arraybackslash 65.6 \\

\hspace{5mm}w/ LNR
& \centering 64.8 & \centering 65.0 & \centering 63.8
& \centering \underline{77.6}
& \centering 68.0
& \centering\arraybackslash 67.8 \\

\hspace{5mm}w/ LNR + RS
& \centering 62.0 & \centering 59.2 & \centering 62.4
& \centering 74.8 & \centering 62.2
& \centering\arraybackslash 64.1 \\

\hspace{5mm}w/ AugMix
& \centering 62.0 & \centering 66.0 & \centering 60.0
& \centering 75.8 & \centering 62.6
& \centering\arraybackslash 65.3 \\

\hspace{5mm}w/ AugMix-MN
& \centering 60.0 & \centering 66.2 & \centering 61.8
& \centering 75.4 & \centering 63.0
& \centering\arraybackslash 65.3 \\

\hspace{5mm}w/ \textsc{PrefLens}
& \centering \underline{68.0}
& \centering \textbf{71.0}
& \centering \textbf{66.8}
& \centering \textbf{80.6}
& \centering \underline{70.4}
& \centering\arraybackslash \textbf{71.4} \\

\bottomrule
\end{tabular}
}
\caption{SD v1.5}
\label{tab:results_win_rate_a}
\end{subtable}

\vspace{2mm}

% ===================== SUBTABLE (b) =====================
\begin{subtable}{\linewidth}
\centering

\resizebox{\linewidth}{!}{
\setlength{\tabcolsep}{6pt}
\begin{tabular}{l p{1.8cm} p{1.8cm} p{1.8cm} p{1.8cm} p{1.8cm} p{1.5cm}}
\toprule
\multirow{3}{*}{\textbf{Method}} &
\multicolumn{2}{c}{\textbf{Image Quality}} &
\multicolumn{3}{c}{\textbf{Human Preference}} &
\multirow{2}{*}{\centering\arraybackslash\textbf{Average}} \\
\cmidrule(lr){2-3}
\cmidrule(lr){4-6}
&
\centering\textbf{Aesthetics} &
\centering\textbf{Aesthetics} &
\centering\textbf{ImageReward} &
\centering\textbf{HPS v2} &
\centering\textbf{PickScore} & \\
&
\centering\textbf{(ViT-14)} &
\centering\textbf{(ViT-16)} &
\centering\textbf{} &
\centering\textbf{} &
\centering\textbf{} & \\
\midrule

% -------- PASTE SECOND TABLE ROWS HERE --------
SFT
& \centering 44.8 & \centering 49.4 & \centering 50.4
& \centering 51.4 & \centering 50.4
& \centering\arraybackslash 49.28 \\

Diff.-DPO
& \centering 49.6 & \centering 46.0 & \centering 62.8
& \centering 72.0 & \centering 66.4
& \centering\arraybackslash 59.36 \\

SPO
& \centering 67.0 & \centering 65.8 & \centering 69.0
& \centering 84.6 & \centering \textbf{75.4}
& \centering\arraybackslash 72.36 \\

\\[-1em] \cdashline{1-7}[0.4pt/1.5pt] \\[-1em]

\textbf{Diffusion-DPO} & & & & & & \\

\hspace{5mm}w/ HPDv2
& \centering 50.8 & \centering 52.0 & \centering 52.4
& \centering 52.4 & \centering 52.6
& \centering\arraybackslash 52.04 \\

\hspace{5mm}w/ Pick-a-Pic-v2
& \centering 50.0 & \centering 50.8 & \centering 55.4
& \centering 52.2 & \centering 53.4
& \centering\arraybackslash 52.36 \\

\hspace{5mm}w/ OpenPref
& \centering 46.6 & \centering 47.2 & \centering 49.2
& \centering 46.4 & \centering 47.2
& \centering\arraybackslash 47.32 \\

\hspace{5mm}w/ LNR
& \centering 66.4 & \centering 67.2 & \centering 68.4
& \centering 84.6 & \centering 65.2
& \centering\arraybackslash 70.36 \\

\hspace{5mm}w/ LNR + RS
& \centering 59.4 & \centering 61.8 & \centering 66.8
& \centering 80.0 & \centering 61.2
& \centering\arraybackslash 65.84 \\

\hspace{5mm}w/ AugMix
& \centering 27.0 & \centering 44.4 & \centering 16.4
& \centering 6.6 & \centering 5.6
& \centering\arraybackslash 20.00 \\

\hspace{5mm}w/ AugMix-MN
& \centering 35.4 & \centering 53.0 & \centering 30.0
& \centering 24.2 & \centering 15.0
& \centering\arraybackslash 31.52 \\

\hspace{5mm}w/ \textsc{PrefLens}
& \centering \textbf{69.0}
& \centering \textbf{69.0}
& \centering \textbf{73.2}
& \centering \underline{87.6}
& \centering 69.6
& \centering\arraybackslash \textbf{73.68} \\

\\[-1em] \cdashline{1-7}[0.4pt/1.5pt] \\[-1em]

\textbf{DSPO} & & & & & & \\

\hspace{5mm}w/ HPDv2
& \centering 57.2 & \centering 53.2 & \centering 61.4
& \centering 61.2 & \centering 58.4
& \centering\arraybackslash 58.28 \\

\hspace{5mm}w/ Pick-a-Pic-v2
& \centering 49.8 & \centering 48.6 & \centering 48.0
& \centering 51.6 & \centering 53.0
& \centering\arraybackslash 50.20 \\

\hspace{5mm}w/ OpenPref
& \centering 49.2 & \centering 52.2 & \centering 50.4
& \centering 46.6 & \centering 51.8
& \centering\arraybackslash 50.04 \\

\hspace{5mm}w/ LNR
& \centering 64.2 & \centering 65.2 & \centering 65.0
& \centering 83.4 & \centering 68.2
& \centering\arraybackslash 69.20 \\

\hspace{5mm}w/ LNR + RS
& \centering \underline{68.4}
& \centering 67.0
& \centering 67.2
& \centering 86.8
& \centering 66.0
& \centering\arraybackslash 71.08 \\

\hspace{5mm}w/ AugMix
& \centering 31.2 & \centering 50.0 & \centering 23.6
& \centering 16.0 & \centering 12.4
& \centering\arraybackslash 26.64 \\

\hspace{5mm}w/ AugMix-MN
& \centering 39.6 & \centering 55.8 & \centering 35.4
& \centering 36.8 & \centering 15.6
& \centering\arraybackslash 36.64 \\

\hspace{5mm}w/ \textsc{PrefLens}
& \centering 67.2
& \centering \underline{68.4}
& \centering \underline{70.6}
& \centering \textbf{89.0}
& \centering \underline{72.4}
& \centering\arraybackslash \underline{73.52} \\

\bottomrule
\end{tabular}
}
\caption{SDXL-Base}
\label{tab:results_win_rate_b}
\end{subtable}
\end{table}

High anchor--neighbor similarities under CLIP and DINO-v2 (\textbf{0.80} and \textbf{0.77}) suggest that the recovered negatives remain semantically close to their anchors.

% Preamble:
% \usepackage{tikz}
% \usetikzlibrary{positioning}
% \usepackage{xcolor}
% \usepackage{graphicx}

\begin{figure*}[t]
\centering

\resizebox{0.8\linewidth}{!}{%
\begin{tikzpicture}[
    % ============================================================
    % LEFT PRINCIPLE STYLE
    % ============================================================
    % Change these values ONLY to modify the OpenPref side.
    leftprinciple/.style={
        rounded corners=0pt,
        minimum height=0.48cm, % <-- LEFT colored-box height
        text width=6.35cm,     % <-- LEFT colored-box/text width
        align=left,
        inner xsep=4pt,        % <-- LEFT horizontal padding
        inner ysep=2.0pt,      % <-- LEFT vertical padding
        font=\footnotesize     % <-- LEFT text size
    },
    % ============================================================
    % RIGHT PRINCIPLE STYLE
    % ============================================================
    % Change these values ONLY to modify the PrefLens side.
    rightprinciple/.style={
        rounded corners=0pt,
        minimum height=0.01cm, % <-- RIGHT colored-box height
        text width=6.15cm,     % <-- RIGHT colored-box/text width
        align=left,
        inner xsep=4pt,        % <-- RIGHT horizontal padding
        inner ysep=1.2pt,      % <-- RIGHT vertical padding
        font=\scriptsize       % <-- RIGHT text size
    },
    % ============================================================
    % OUTER PANEL STYLE
    % ============================================================
    box/.style={
        draw,
        rounded corners=1.5pt,
        very thick,
        inner sep=4pt
    },
    % ============================================================
    % LEGEND STYLE
    % ============================================================
    legendbox/.style={
        rounded corners=0pt,
        minimum height=0.58cm, % <-- LEGEND box height
        align=center,
        inner xsep=4pt,        % <-- LEGEND horizontal padding
        inner ysep=1.2pt,      % <-- LEGEND vertical padding
        font=\scriptsize       % <-- LEGEND text size
    }
]

% ============================================================
% EASY GLOBAL SIZE CONTROLS
% ============================================================

% Height of BOTH large outer panels
\def\panelheight{8.55cm}

% Vertical location of legend
% Less negative = move legend UP
% More negative = move legend DOWN
\def\legendY{-8.99}

% ============================================================
% Colors
% ============================================================

\definecolor{qualitycol}{RGB}{211,226,248}
\definecolor{compcol}{RGB}{218,239,211}
\definecolor{lightcol}{RGB}{238,216,246}
\definecolor{structcol}{RGB}{248,218,209}
\definecolor{promptcol}{RGB}{252,237,190}
\definecolor{artifactcol}{RGB}{228,228,228}

% ============================================================
% LEFT PANEL: OpenPref
% ============================================================

\node[
    box,
    anchor=north west,
    minimum width=7.0cm,
    minimum height=\panelheight
] (leftbox) at (0,0) {};

\node[
    font=\normalsize\bfseries,
    anchor=north
] at ([yshift=-0.18cm]leftbox.north)
{Preference principles from \textit{OpenPref}};

% 1 -- Composition
\node[leftprinciple,fill=compcol,anchor=north west]
at ([xshift=0.2cm,yshift=-0.72cm]leftbox.north west)
{\textbf{1.} More visually appealing composition.};

% 2 -- Prompt & Style
\node[leftprinciple,fill=promptcol,anchor=north west]
at ([xshift=0.2cm,yshift=-1.47cm]leftbox.north west)
{\textbf{2.} Friendlier, more inviting subject.};

% 3 -- Lighting & Color
\node[leftprinciple,fill=lightcol,anchor=north west]
at ([xshift=0.2cm,yshift=-2.22cm]leftbox.north west)
{\textbf{3.} Brighter and richer colors.};

% 4 -- Visual Quality
\node[leftprinciple,fill=qualitycol,anchor=north west]
at ([xshift=0.2cm,yshift=-2.97cm]leftbox.north west)
{\textbf{4.} Greater detail and sharpness of the subject.};

% 5 -- Composition
\node[leftprinciple,fill=compcol,anchor=north west]
at ([xshift=0.2cm,yshift=-3.72cm]leftbox.north west)
{\textbf{5.} Better integration of subject and background.};

% 6 -- Prompt & Style
\node[leftprinciple,fill=promptcol,anchor=north west]
at ([xshift=0.2cm,yshift=-4.47cm]leftbox.north west)
{\textbf{6.} Clean, polished, realistic art style.};

% 7 -- Lighting & Color
\node[leftprinciple,fill=lightcol,anchor=north west]
at ([xshift=0.2cm,yshift=-5.22cm]leftbox.north west)
{\textbf{7.} Captivating lighting effects and gradients.};

% 8 -- Structural Fidelity
\node[leftprinciple,fill=structcol,anchor=north west]
at ([xshift=0.2cm,yshift=-5.97cm]leftbox.north west)
{\textbf{8.} Greater emphasis on the subject's face and features.};

% 9 -- Composition
\node[leftprinciple,fill=compcol,anchor=north west]
at ([xshift=0.2cm,yshift=-6.72cm]leftbox.north west)
{\textbf{9.} Greater depth and atmospheric perspective.};

% 10 -- Prompt & Style
\node[leftprinciple,fill=promptcol,anchor=north west]
at ([xshift=0.2cm,yshift=-7.47cm]leftbox.north west)
{\textbf{10.} Stronger environmental storytelling and charm.};

% % 11 -- Lighting & Color
% \node[leftprinciple,fill=lightcol,anchor=north west]
% at ([xshift=0.2cm,yshift=-8.22cm]leftbox.north west)
% {\textbf{11.} More realistic lighting and subtle tones.};

% ============================================================
% RIGHT PANEL: PREFLENS
% ============================================================

\node[
    box,
    anchor=north west,
    minimum width=7.0cm,
    minimum height=\panelheight
] (rightbox) at (7.35,0) {};

\node[
    font=\normalsize\bfseries,
    anchor=north
] at ([yshift=-0.18cm]rightbox.north)
{Preference principles from \textsc{PrefLens}};

% 1 -- Prompt & Style
\node[rightprinciple,fill=promptcol,anchor=north west]
at ([xshift=0.30cm,yshift=-0.72cm]rightbox.north west)
{\textbf{1.} More accurate adherence to the prompt.};

% 2 -- Visual Quality
\node[rightprinciple,fill=qualitycol,anchor=north west]
at ([xshift=0.30cm,yshift=-1.15cm]rightbox.north west)
{\textbf{2.} Higher visual clarity.};

% 3 -- Lighting & Color
\node[rightprinciple,fill=lightcol,anchor=north west]
at ([xshift=0.30cm,yshift=-1.58cm]rightbox.north west)
{\textbf{3.} Better lighting and contrast.};

% 4 -- Composition
\node[rightprinciple,fill=compcol,anchor=north west]
at ([xshift=0.30cm,yshift=-2.01cm]rightbox.north west)
{\textbf{4.} Better composition and subject focus.};

% 5 -- Structural Fidelity
\node[rightprinciple,fill=structcol,anchor=north west]
at ([xshift=0.30cm,yshift=-2.44cm]rightbox.north west)
{\textbf{5.} More realistic proportions and structure.};

% 6 -- Artifact Avoidance
\node[rightprinciple,fill=artifactcol,anchor=north west]
at ([xshift=0.30cm,yshift=-2.87cm]rightbox.north west)
{\textbf{6.} Fewer visual distractions such as artifacts or noise.};

% 7 -- Visual Quality
\node[rightprinciple,fill=qualitycol,anchor=north west]
at ([xshift=0.30cm,yshift=-3.30cm]rightbox.north west)
{\textbf{7.} Better overall detail and resolution.};

% 8 -- Composition
\node[rightprinciple,fill=compcol,anchor=north west]
at ([xshift=0.30cm,yshift=-3.73cm]rightbox.north west)
{\textbf{8.} Better composition and balanced elements.};

% 9 -- Lighting & Color
\node[rightprinciple,fill=lightcol,anchor=north west]
at ([xshift=0.30cm,yshift=-4.16cm]rightbox.north west)
{\textbf{9.} Smoother color transitions and gradients.};

% 10 -- Prompt & Style
\node[rightprinciple,fill=promptcol,anchor=north west]
at ([xshift=0.30cm,yshift=-4.59cm]rightbox.north west)
{\textbf{10.} Clearly legible text and details.};

% 11 -- Structural Fidelity
\node[rightprinciple,fill=structcol,anchor=north west]
at ([xshift=0.30cm,yshift=-5.02cm]rightbox.north west)
{\textbf{11.} Better adherence to anatomical accuracy.};

% 12 -- Visual Quality
\node[rightprinciple,fill=qualitycol,anchor=north west]
at ([xshift=0.30cm,yshift=-5.45cm]rightbox.north west)
{\textbf{12.} Better-defined textures on surfaces.};

% 13 -- Artifact Avoidance
\node[rightprinciple,fill=artifactcol,anchor=north west]
at ([xshift=0.30cm,yshift=-5.88cm]rightbox.north west)
{\textbf{13.} Less pixelation and visual distortion.};

% 14 -- Composition
\node[rightprinciple,fill=compcol,anchor=north west]
at ([xshift=0.30cm,yshift=-6.31cm]rightbox.north west)
{\textbf{14.} Well-positioned and centered subject.};

% 15 -- Lighting & Color
\node[rightprinciple,fill=lightcol,anchor=north west]
at ([xshift=0.30cm,yshift=-6.74cm]rightbox.north west)
{\textbf{15.} More realistic lighting and reflections.};

% 16 -- Prompt & Style
\node[rightprinciple,fill=promptcol,anchor=north west]
at ([xshift=0.30cm,yshift=-7.17cm]rightbox.north west)
{\textbf{16.} Better adherence to a neon aesthetic.};

% 17 -- Structural Fidelity
\node[rightprinciple,fill=structcol,anchor=north west]
at ([xshift=0.30cm,yshift=-7.60cm]rightbox.north west)
{\textbf{17.} Intact and recognizable architectural details.};

% 18 -- Visual Quality
\node[rightprinciple,fill=qualitycol,anchor=north west]
at ([xshift=0.30cm,yshift=-8.03cm]rightbox.north west)
{\textbf{18.} More polished and higher visual quality.};

% % 19 -- Composition
% \node[rightprinciple,fill=compcol,anchor=north west]
% at ([xshift=0.30cm,yshift=-8.46cm]rightbox.north west)
% {\textbf{19.} Better emphasis on detail and spatial depth.};

% % 20 -- Lighting & Color
% \node[rightprinciple,fill=lightcol,anchor=north west]
% at ([xshift=0.30cm,yshift=-8.89cm]rightbox.north west)
% {\textbf{20.} Natural, aesthetically pleasing colors and lighting.};

% ============================================================
% LEGEND
% ============================================================

\node[
    legendbox,
    fill=qualitycol,
    text width=1.75cm
] at (1.25,\legendY)
{Visual Quality};

\node[
    legendbox,
    fill=compcol,
    text width=1.75cm
] at (3.50,\legendY)
{Composition};

\node[
    legendbox,
    fill=lightcol,
    text width=1.75cm
] at (5.75,\legendY)
{Lighting \& Color};

\node[
    legendbox,
    fill=structcol,
    text width=1.75cm
] at (8.00,\legendY)
{Structural Fidelity};

\node[
    legendbox,
    fill=promptcol,
    text width=1.85cm
] at (10.35,\legendY)
{Prompt \& Style};

\node[
    legendbox,
    fill=artifactcol,
    text width=1.85cm
] at (12.75,\legendY)
{Artifact Avoidance};

\end{tikzpicture}%
}

\caption{Representative preference principles extracted using ICAI from OpenPref and \textsc{PrefLens}-constructed preferences.}
\label{fig:icai_constitution_comparison}

\end{figure*}

\paragraph{\textbf{Q3: Does the multi-dimensional preference space provide complementary preference signals?}}
ImageReward and Aesthetic Score exhibit moderate correlation at both the image level (Pearson/Spearman: $0.68/0.68$) and preference-margin level ($0.65/0.66$), indicating complementary preference signals. The raw dimensions also have similar variances ($1.32$ and $1.52$), and the raw and $z$-score-normalized pipelines share $94.4\%$ of the selected samples, suggesting limited sensitivity to reward-scale differences. Geometric analyses are provided in Appendix~\ref{sec:preference_space_analysis}, and results for the normalized preference space are provided in Appendix~\ref{sec:additional_ablations} (Table~\ref{tab:normalization_ablation}).

\paragraph{\textbf{Q4: Can local preference neighborhoods provide richer supervision for preference alignment?}}

Table~\ref{tab:results_win_rate} shows that \textsc{PrefLens} achieves the strongest or near-strongest performance in most settings. Improvements are observed across both image-quality and human-preference metrics, indicating that the gains are not restricted to a single evaluation dimension. This trend is further supported by extended evaluation in Table~\ref{tab:human_vlm_eval} and the qualitative results in Figure~\ref{fig:qual_result}. Additional results are provided in
Appendices~\ref{sec:extended_benchmark_results} and~\ref{sec:additional_qualitative}.

\textbf{\textit{Image quality and human preference.}} Across both Diffusion-DPO and DSPO, \textsc{PrefLens} generally improves aesthetic quality as well as preference-oriented metrics. Qualitative examples in Figure~\ref{fig:qual_result} further show improved visual quality and coherence.

\begin{figure}[!tbh]
    \centering

    \begin{subfigure}[t]{0.48\linewidth}
        \centering
        \includegraphics[width=\linewidth]
        {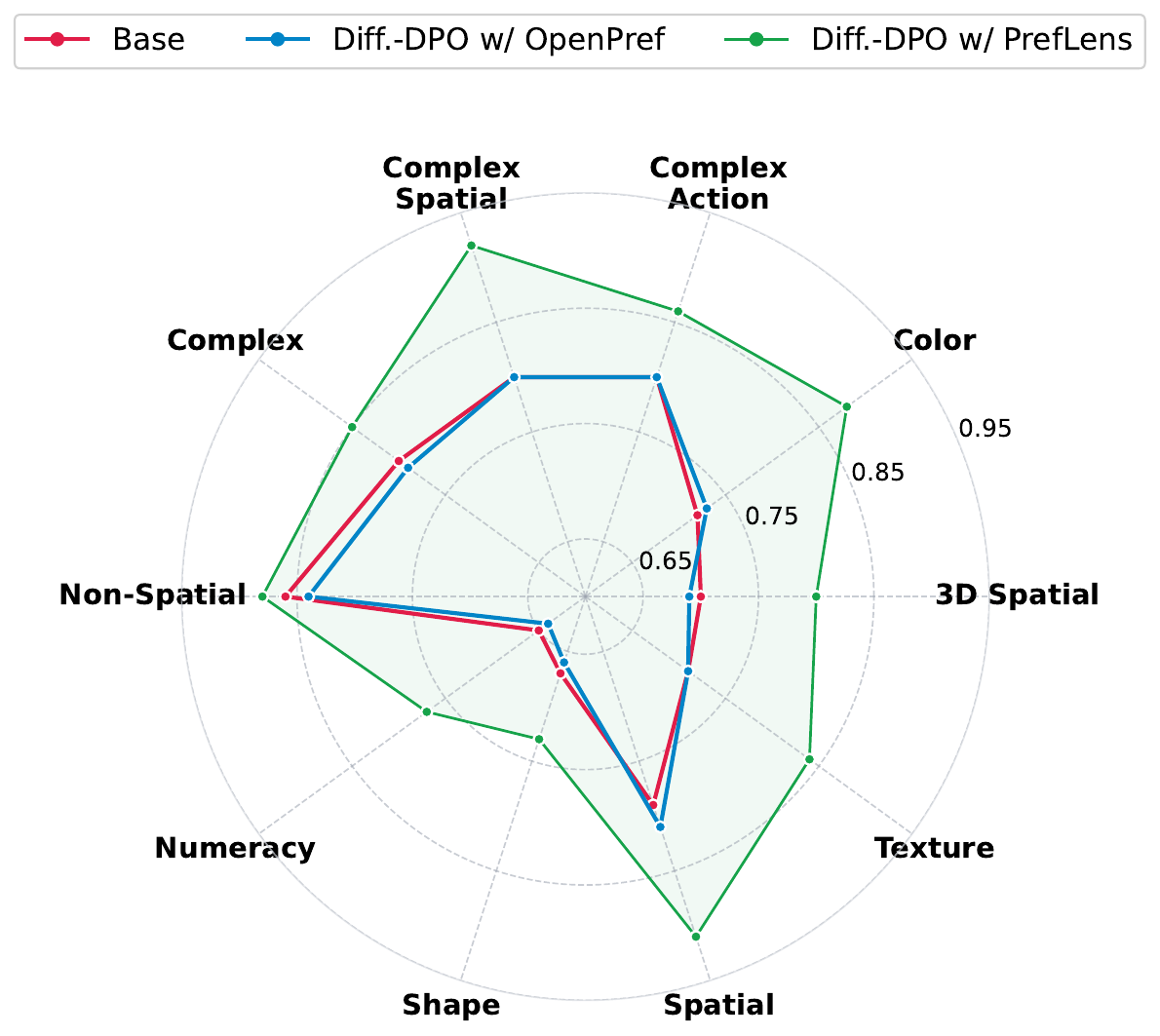}
        \caption{}
        \label{fig:tifa_blip}
    \end{subfigure}
    \hfill
    \begin{subfigure}[t]{0.48\linewidth}
        \centering
        \includegraphics[width=\linewidth]
        {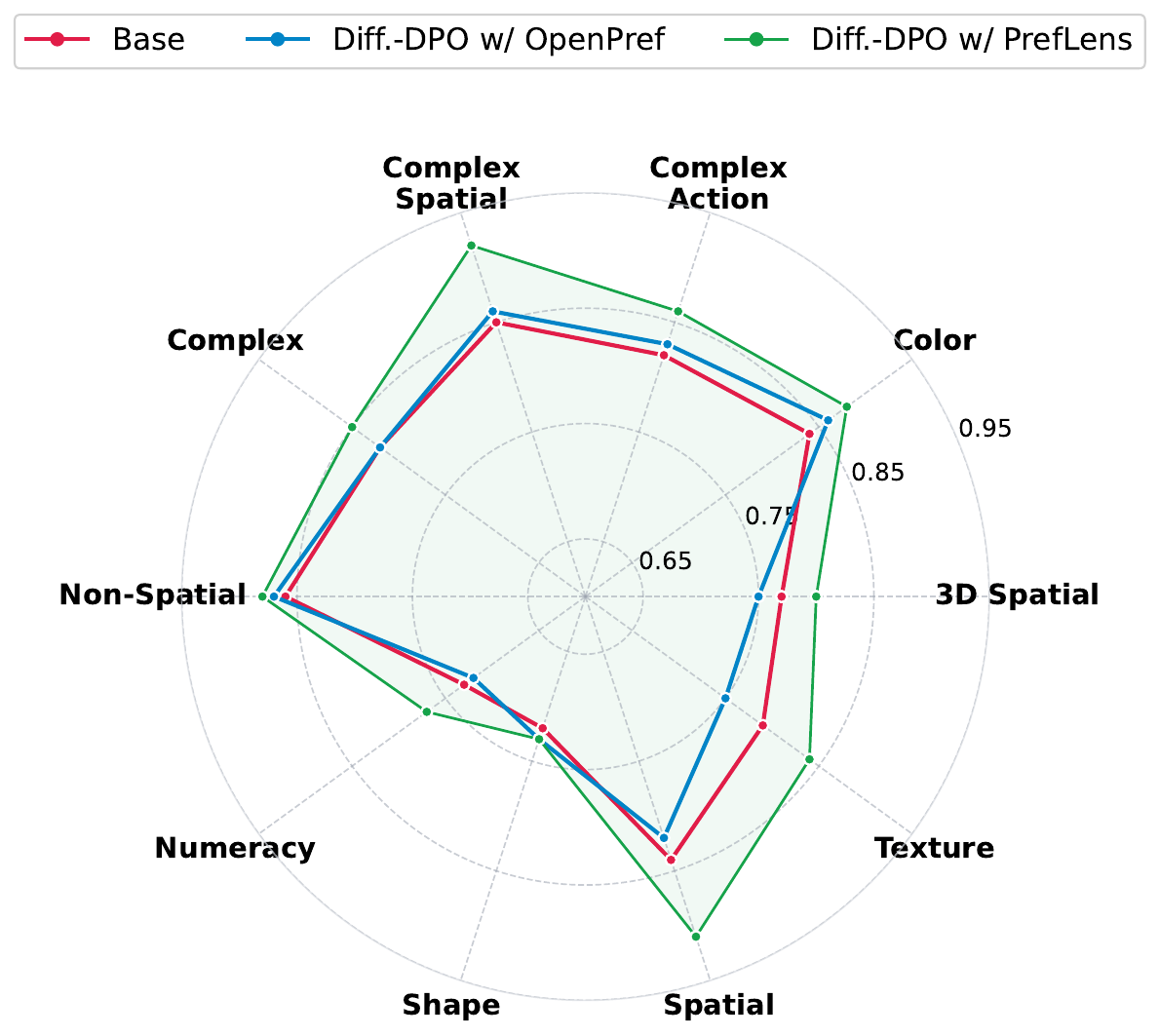}
        \caption{}
        \label{fig:tifa_git}
    \end{subfigure}

    \caption{
    \textbf{Compositional fidelity evaluation on T2I-CompBench++ for (a) SD v1.5 and (b) SDXL.}
    We report TIFA scores computed using BLIP-large as VQA model.
    % across multiple compositional categories.
    }
    \label{fig:compo_plot}
\end{figure}

\textbf{\textit{Compositional fidelity.}} Figure~\ref{fig:compo_plot} shows that the benefits of \textsc{PrefLens} extend to compositional fidelity, with improved TIFA scores across multiple compositional categories. Qualitative results in Figure~\ref{fig:qual_result} further exhibit improved adherence to compositional prompt attributes, including object counting and prompt specifications.

\begin{table}[!htb]
\centering
\caption{
Human and VLM-as-a-Judge evaluations.
\textbf{Top-1} denotes the percentage of samples ranked as the most preferred,
while \textbf{Borda} denotes the normalized Borda score across comparisons.
}
\label{tab:human_vlm_eval}
\small
\setlength{\tabcolsep}{6pt}
\renewcommand{\arraystretch}{1.15}

% =========================================================
% SUBTABLE (a): HUMAN EVALUATION
% =========================================================
\begin{subtable}{\linewidth}
\centering
\resizebox{\linewidth}{!}{%
\begin{tabular}{c l p{1.8cm} p{1.8cm} p{1.8cm} p{1.8cm}}
\toprule
\multirow{2}{*}{\textbf{Dataset}} &
\multirow{2}{*}{\textbf{Method}} &
\multicolumn{2}{c}{\textbf{SD v1.5}} &
\multicolumn{2}{c}{\textbf{SDXL}} \\
\cmidrule(lr){3-4}
\cmidrule(lr){5-6}

& &
\centering\textbf{Top-1 (\%) $\uparrow$} &
\centering\textbf{Borda $\uparrow$} &
\centering\textbf{Top-1 (\%) $\uparrow$} &
\centering\arraybackslash\textbf{Borda $\uparrow$} \\
\midrule

% ================= PASTE HUMAN-EVALUATION ROWS HERE =================
\multirow{5}{*}{\textbf{Pick-a-Pic-V2}}
& Base
& \centering 26.92
& \centering 62.82
& \centering 19.70
& \centering\arraybackslash 59.09 \\

\\[-1.2em]
\cdashline{2-6}[0.4pt/1.5pt]
\\[-1.2em]

& \textbf{Diffusion-DPO}
& & & & \\

& \hspace{5mm}w/ OpenPref
& \centering 23.72
& \centering 58.87
& \centering 20.00
& \centering\arraybackslash 58.69 \\

& \cellcolor{black!5}\hspace{5mm}w/ \textsc{PrefLens}
& \cellcolor{black!5}\centering \textbf{49.36}
& \cellcolor{black!5}\centering \textbf{78.31}
& \cellcolor{black!5}\centering \textbf{60.30}
& \cellcolor{black!5}\centering\arraybackslash \textbf{82.22} \\

\midrule

% ========================================================
% PartiPrompt
% ========================================================
\multirow{5}{*}{\textbf{PartiPrompt}}
& Base
& \centering 8.33
& \centering 51.07
& \centering 11.21
& \centering\arraybackslash 58.38 \\

\\[-1.2em]
\cdashline{2-6}[0.4pt/1.5pt]
\\[-1.2em]

& \textbf{Diffusion-DPO}
& & & & \\

& \hspace{5mm}w/ OpenPref
& \centering 13.78
& \centering 58.33
& \centering 19.39
& \centering\arraybackslash 56.97 \\

& \cellcolor{black!5}\hspace{5mm}w/ \textsc{PrefLens}
& \cellcolor{black!5}\centering \textbf{77.88}
& \cellcolor{black!5}\centering \textbf{90.60}
& \cellcolor{black!5}\centering \textbf{69.39}
& \cellcolor{black!5}\centering\arraybackslash \textbf{84.65} \\

\midrule

% ========================================================
% GenAIBench
% ========================================================
\multirow{5}{*}{\textbf{GenAIBench}}
& Base
& \centering 21.70
& \centering 61.11
& \centering 16.06
& \centering\arraybackslash 58.48 \\

\\[-1.2em]
\cdashline{2-6}[0.4pt/1.5pt]
\\[-1.2em]

& \textbf{Diffusion-DPO}
& & & & \\

& \hspace{5mm}w/ OpenPref
& \centering 15.72
& \centering 58.49
& \centering 21.21
& \centering\arraybackslash 57.27 \\

& \cellcolor{black!5}\hspace{5mm}w/ \textsc{PrefLens}
& \cellcolor{black!5}\centering \textbf{62.58}
& \cellcolor{black!5}\centering \textbf{80.40}
& \cellcolor{black!5}\centering \textbf{62.73}
& \cellcolor{black!5}\centering\arraybackslash \textbf{84.24} \\

\bottomrule
\end{tabular}%
}

\caption{Human evaluation}
\label{tab:human_eval}
\end{subtable}

\vspace{2mm}

% =========================================================
% SUBTABLE (b): VLM-AS-A-JUDGE
% =========================================================
\begin{subtable}{\linewidth}
\centering

\resizebox{\linewidth}{!}{%
\begin{tabular}{c l p{1.8cm} p{1.8cm} p{1.8cm} p{1.8cm}}
\toprule
\multirow{2}{*}{\textbf{Dataset}} &
\multirow{2}{*}{\textbf{Method}} &
\multicolumn{2}{c}{\textbf{SD v1.5}} &
\multicolumn{2}{c}{\textbf{SDXL}} \\
\cmidrule(lr){3-4}
\cmidrule(lr){5-6}

& &
\centering\textbf{Top-1 (\%) $\uparrow$} &
\centering\textbf{Borda (\%) $\uparrow$} &
\centering\textbf{Top-1 (\%) $\uparrow$} &
\centering\arraybackslash\textbf{Borda (\%) $\uparrow$} \\
\midrule

% ================= PASTE VLM-EVALUATION ROWS HERE =================
\multirow{8}{*}{\textbf{Pick-a-Pic-V2}}

% ---------------- Diffusion-DPO ----------------
& \textbf{Diffusion-DPO}
& & & & \\

& \hspace{5mm}Base
& \centering 26.6
& \centering 62.8
& \centering 25.4
& \centering\arraybackslash 62.6 \\

& \hspace{5mm}w/ OpenPref
& \centering 26.2
& \centering 63.7
& \centering 20.0
& \centering\arraybackslash 60.5 \\

&
\cellcolor{black!5}\hspace{5mm}w/ \textsc{PrefLens}
& \cellcolor{black!5}\centering \textbf{47.2}
& \cellcolor{black!5}\centering \textbf{73.5}
& \cellcolor{black!5}\centering \textbf{54.6}
& \cellcolor{black!5}\centering\arraybackslash \textbf{76.9} \\

\\[-1.2em]
\cdashline{2-6}[0.4pt/1.5pt]
\\[-1.2em]

% ---------------- DSPO ----------------
& \textbf{DSPO}
& & & & \\

& \hspace{5mm}Base
& \centering 22.6
& \centering 59.1
& \centering 22.0
& \centering\arraybackslash 61.1 \\

& \hspace{5mm}w/ OpenPref
& \centering 36.4
& \centering 69.7
& \centering 21.0
& \centering\arraybackslash 60.6 \\

&
\cellcolor{black!5}\hspace{5mm}w/ \textsc{PrefLens}
& \cellcolor{black!5}\centering \textbf{41.0}
& \cellcolor{black!5}\centering \textbf{71.3}
& \cellcolor{black!5}\centering \textbf{57.0}
& \cellcolor{black!5}\centering\arraybackslash \textbf{78.3} \\

\midrule

% ========================================================
% PartiPrompt
% ========================================================

\multirow{8}{*}{\textbf{PartiPrompt}}

% ---------------- Diffusion-DPO ----------------
& \textbf{Diffusion-DPO}
& & & & \\

& \hspace{5mm}Base
& \centering 24.4
& \centering 61.3
& \centering 24.6
& \centering\arraybackslash 63.3 \\

& \hspace{5mm}w/ OpenPref
& \centering 28.4
& \centering 65.3
& \centering 26.2
& \centering\arraybackslash 62.5 \\

&
\cellcolor{black!5}\hspace{5mm}w/ \textsc{PrefLens}
& \cellcolor{black!5}\centering \textbf{47.2}
& \cellcolor{black!5}\centering \textbf{73.3}
& \cellcolor{black!5}\centering \textbf{49.2}
& \cellcolor{black!5}\centering\arraybackslash \textbf{74.2} \\

\\[-1.2em]
\cdashline{2-6}[0.4pt/1.5pt]
\\[-1.2em]

% ---------------- DSPO ----------------
& \textbf{DSPO}
& & & & \\

& \hspace{5mm}Base
& \centering 24.0
& \centering 58.6
& \centering 26.0
& \centering\arraybackslash 63.1 \\

& \hspace{5mm}w/ OpenPref
& \centering 32.4
& \centering 68.8
& \centering 26.2
& \centering\arraybackslash 62.9 \\

&
\cellcolor{black!5}\hspace{5mm}w/ \textsc{PrefLens}
& \cellcolor{black!5}\centering \textbf{43.6}
& \cellcolor{black!5}\centering \textbf{72.6}
& \cellcolor{black!5}\centering \textbf{47.8}
& \cellcolor{black!5}\centering\arraybackslash \textbf{74.0} \\

\midrule

% ========================================================
% GenAIBench
% ========================================================

\multirow{8}{*}{\textbf{GenAIBench}}

% ---------------- Diffusion-DPO ----------------
& \textbf{Diffusion-DPO}
& & & & \\

& \hspace{5mm}Base
& \centering 25.6
& \centering 63.7
& \centering 29.2
& \centering\arraybackslash 63.9 \\

& \hspace{5mm}w/ OpenPref
& \centering 28.2
& \centering 63.7
& \centering 26.2
& \centering\arraybackslash 63.9 \\

&
\cellcolor{black!5}\hspace{5mm}w/ \textsc{PrefLens}
& \cellcolor{black!5}\centering \textbf{46.2}
& \cellcolor{black!5}\centering \textbf{72.7}
& \cellcolor{black!5}\centering \textbf{44.6}
& \cellcolor{black!5}\centering\arraybackslash \textbf{72.2} \\

\\[-1.2em]
\cdashline{2-6}[0.4pt/1.5pt]
\\[-1.2em]

% ---------------- DSPO ----------------
& \textbf{DSPO}
& & & & \\

& \hspace{5mm}Base
& \centering 23.6
& \centering 59.9
& \centering 25.0
& \centering\arraybackslash 62.5 \\

& \hspace{5mm}w/ OpenPref
& \centering 32.8
& \centering 68.2
& \centering 30.6
& \centering\arraybackslash 64.9 \\

&
\cellcolor{black!5}\hspace{5mm}w/ \textsc{PrefLens}
& \cellcolor{black!5}\centering \textbf{43.6}
& \cellcolor{black!5}\centering \textbf{71.9}
& \cellcolor{black!5}\centering \textbf{44.4}
& \cellcolor{black!5}\centering\arraybackslash \textbf{72.5} \\

\bottomrule
\end{tabular}%
}

\caption{VLM-as-a-Judge evaluation}
\label{tab:vlm_eval}
\end{subtable}
\end{table}

\paragraph{\textbf{Q5: What factors explain the behavior of different preference-supervision strategies?}}
The results reveal a quantity--quality trade-off in preference supervision. SFT remains weak, consistent with positive-only training providing less informative supervision than pairwise comparisons. OpenPref provides high-quality but limited comparisons, whereas larger datasets such as HPDv2 and Pick-a-Pic-v2 offer greater coverage but may contain redundant or weakly informative pairs. Importantly, \textsc{PrefLens} does not discard the human preference signal: the human-preferred image defines each anchor, around which additional localized comparisons are constructed.

The neighborhood baselines further show that pair construction matters. \emph{LNR} generally improves over OpenPref, while \emph{LNR+RS} and retaining all recovered neighbors are less consistent than \textsc{PrefLens}, indicating that expansion alone is insufficient. Matched-budget random and uniform sampling similarly support the benefit of the proposed selection strategy; detailed results, including sampling-budget sensitivity, are provided in Appendix~\ref{sec:additional_ablations} (Figure~\ref{fig:ablation}).

AugMix and AugMix-MN generally underperform the other alternatives. Manual inspection shows that many AugMix negatives remain visually close to their anchors, potentially yielding ambiguous preference supervision. Their varying behavior across SD v1.5 and SDXL, together with differences between Diffusion-DPO and DSPO, suggests that effective preference-pair construction can depend on both the diffusion backbone and optimization objective.

\begin{figure}[t]
    \centering
    
    \begin{subfigure}{\linewidth}
        \centering
\renewcommand{\arraystretch}{1.2}
\setlength{\tabcolsep}{2.5pt}
\resizebox{\linewidth}{!}{%
\begin{tabular}{>{\raggedright\arraybackslash}m{2.25cm} *{4}{>{\centering\arraybackslash}m{2.7cm}}}

& \makecell{
\textbf{\fontsize{10}{10}\selectfont Three demons}\\[-2pt]
\textbf{\fontsize{10}{10}\selectfont together}\\[-2pt]
\textbf{\fontsize{10}{10}\selectfont holding swords}\\[-2pt]
\textbf{\fontsize{10}{10}\selectfont around hell}
}
& \makecell{
\textbf{\fontsize{10}{10}\selectfont Light blue}\\[-2pt]
\textbf{\fontsize{10}{10}\selectfont haired anime}\\[-2pt]
\textbf{\fontsize{10}{10}\selectfont girl ocean}\\[-2pt]
\textbf{\fontsize{10}{10}\selectfont themed isopod}\\[-2pt]
\textbf{\fontsize{10}{10}\selectfont antenna twintails}
}
& \makecell{
\textbf{\fontsize{10}{10}\selectfont A red sphere}\\[-2pt]
\textbf{\fontsize{10}{10}\selectfont on top of}\\[-2pt]
\textbf{\fontsize{10}{10}\selectfont a yellow box}
}
& \makecell{
\textbf{\fontsize{10}{10}\selectfont A pineapple}\\[-2pt]
\textbf{\fontsize{10}{10}\selectfont surfing on a}\\[-2pt]
\textbf{\fontsize{10}{10}\selectfont wave}
} \\[-1pt]

% Row 1 – Base (SD v1.5)
\makecell[c]{\textbf{Base} \\ \textbf{(SD v1.5)}}
& \includegraphics[width=\linewidth]{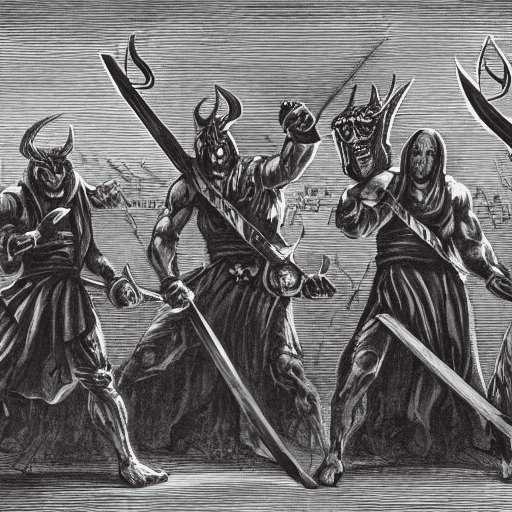}
& \includegraphics[width=\linewidth]{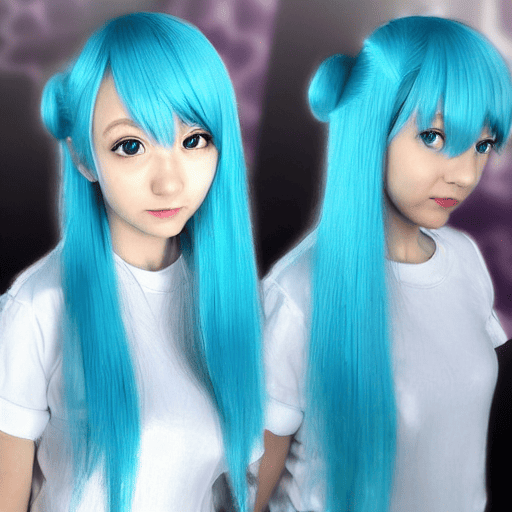}
& \includegraphics[width=\linewidth]{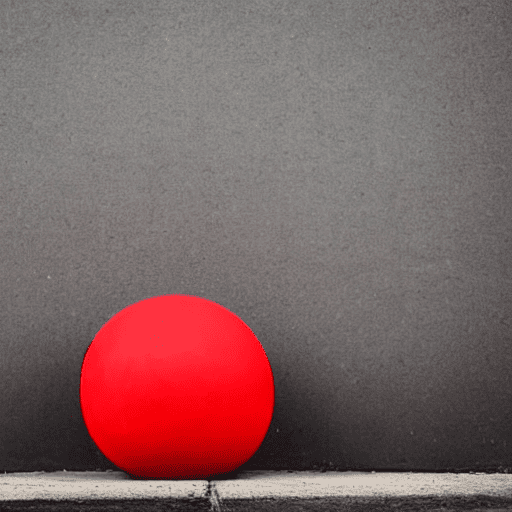}
& \includegraphics[width=\linewidth]{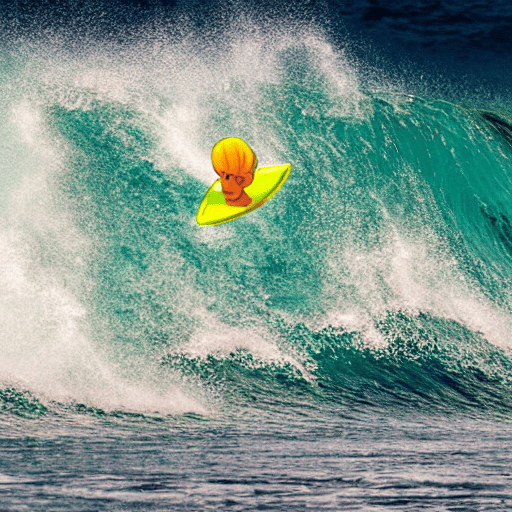} \\

% Row 2 – Diff-DPO
\makecell[c]{\textbf{Diff-DPO}\\\textbf{with}\\\textbf{\textit{OpenPref}}} 
& \includegraphics[width=\linewidth]{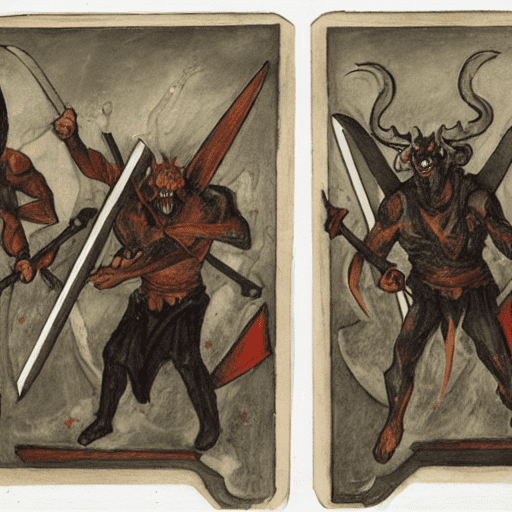}
& \includegraphics[width=\linewidth]{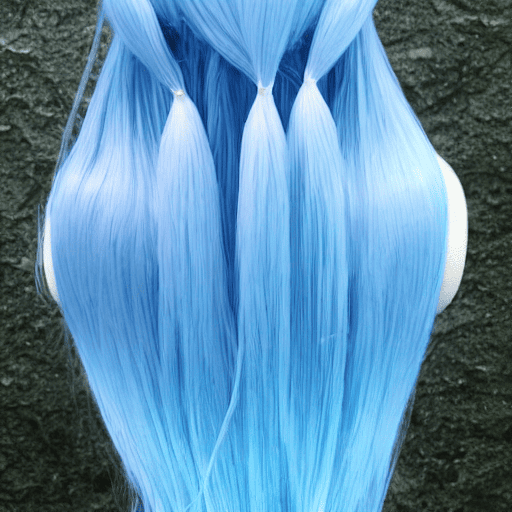}
& \includegraphics[width=\linewidth]{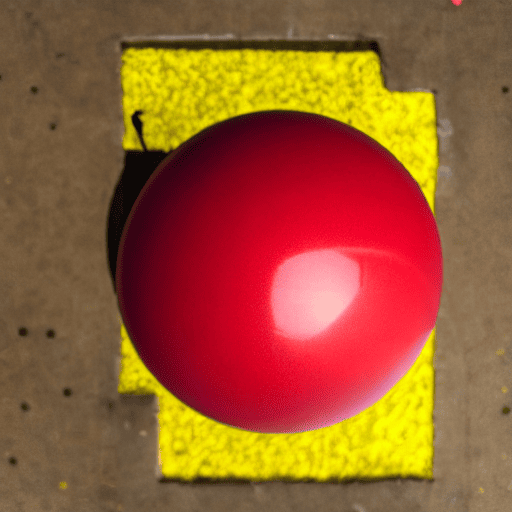}
& \includegraphics[width=\linewidth]{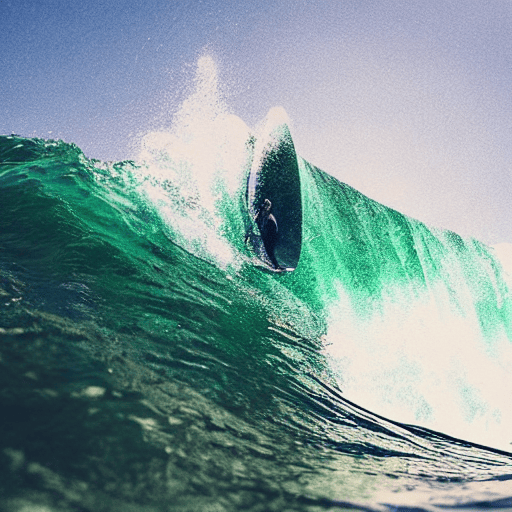} \\

% Row 3 – Ours with Diff-DPO
\makecell[c]{\textbf{Diff-DPO}\\\textbf{with}\\\textbf{\textsc{PrefLens}}} 
& \includegraphics[width=\linewidth]{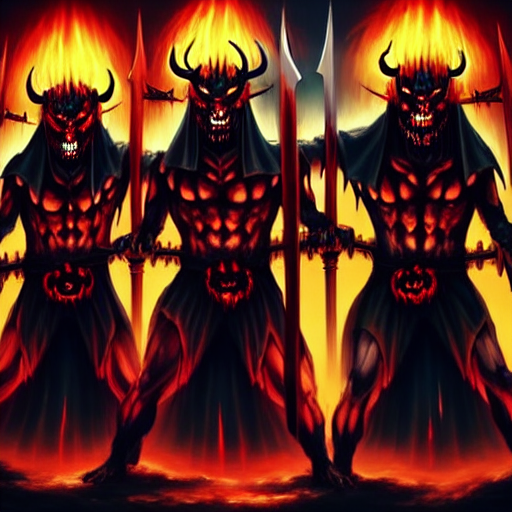}
& \includegraphics[width=\linewidth]{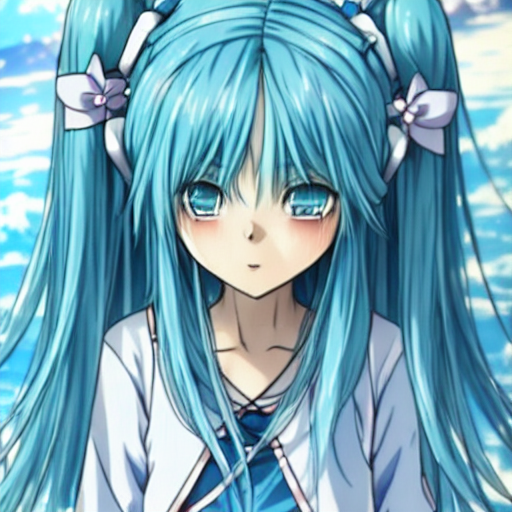}
& \includegraphics[width=\linewidth]{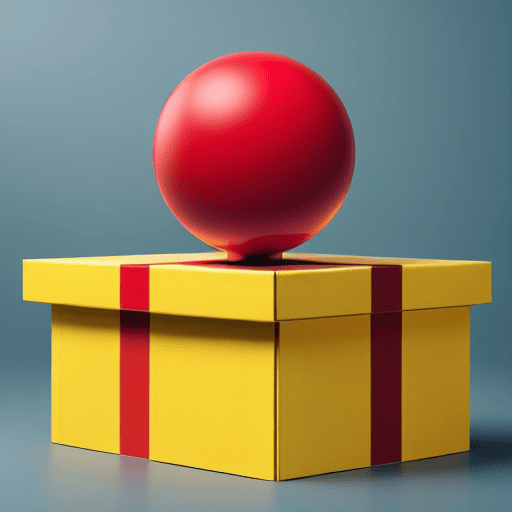}
& \includegraphics[width=\linewidth]{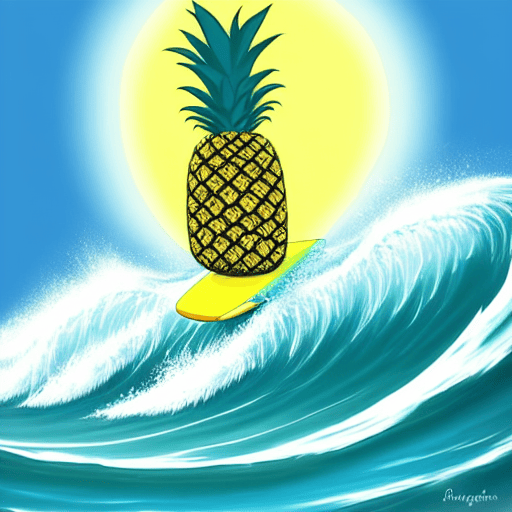} \\

\end{tabular}}

        % \caption{SD v1.5}
        % \label{fig:ablation-sd15}
\end{subfigure}

    \caption{Qualitative results for SD v1.5 backbone.}
    \label{fig:qual_result}
\end{figure}

\paragraph{\textbf{Q6: How do perturbation groups and strength affect preference alignment?}}
Combining all three perturbation groups gives the best performance in most cases, suggesting that the different families provide complementary preference signals. For SDXL, Group 3 alone performs noticeably worse than the combined setting, unlike SD v1.5, indicating that its effectiveness in isolation is backbone-dependent. For perturbation strength, sampling across the full range outperforms fixing the strength at either the minimum or maximum, suggesting that exposure to varying perturbation intensities provides more effective supervision. Detailed results are provided in Appendix~\ref{sec:additional_ablations} (Figures~\ref{fig:perturbation_ablation} and~\ref{fig:strength_ablation}).

\begin{figure}[!htb]
    \centering
    \includegraphics[width=\linewidth]{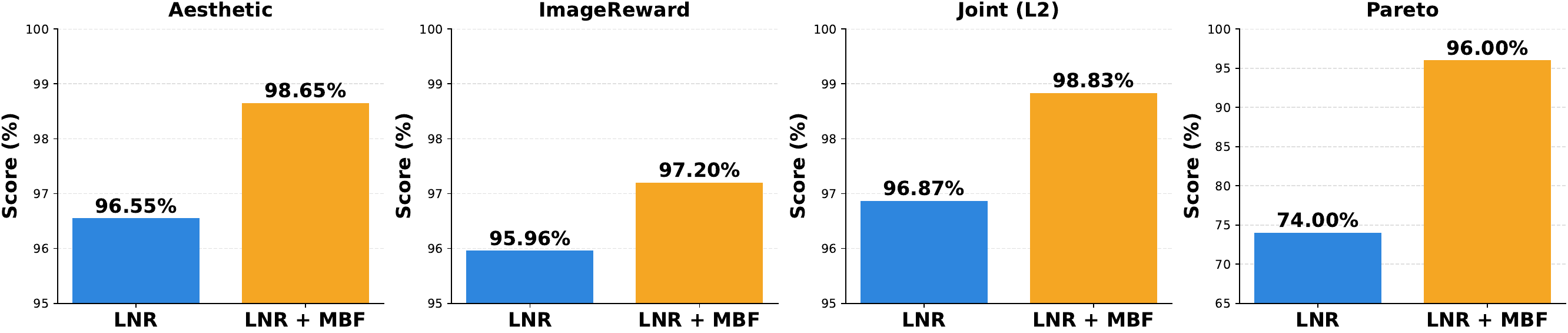}
   \caption{
    Effect of MBF on preference quality.
    }
    \label{fig:epsilon_vis}
\end{figure}

\paragraph{\textbf{Q7: How does Margin-Based Filtering (MBF) affect neighborhood quality?}}
We compare neighborhoods before and after MBF under both 1D and joint 2D preference representations. As shown in Fig.~\ref{fig:epsilon_vis}, MBF consistently improves preference correctness and increases the fraction of Pareto-optimal samples~\cite{miettinen1999nonlinear}, indicating more reliable and informative retained neighbors. We also vary the margin threshold and find that $\epsilon=0.5$ performs best among $\epsilon\in\{0.1,0.5,1.0\}$. Detailed ablation results are provided in Appendix~\ref{sec:additional_ablations} (Figure~\ref{fig:epsilon_abalation}).

% \paragraph{Q8. What is the computational overhead of \textsc{PrefLens}?}
% \textsc{PrefLens} performs neighborhood construction as an offline preprocessing step prior to training. As summarized in Table~\ref{tab:runtime}, the complete preprocessing pipeline requires approximately \textbf{23.4 hours} using a CPU-only implementation. Moreover, because perturbation generation and reward-model inference are highly parallelizable, the preprocessing time can be substantially reduced through parallel execution or GPU acceleration.

% \begin{table}[!htb]
% \centering
% \small
% \begin{tabular}{lc}
% \toprule
% \textbf{Stage} & \textbf{Runtime (hours)} \\
% \midrule
% Perturbation Generation & 9.0 \\
% Aesthetic Score Inference & 6.4 \\
% ImageReward Inference & 5.8 \\
% MBF + Neighbor Sampling & 2.2 \\
% \midrule
% \textbf{Total} & \textbf{23.4} \\
% \bottomrule
% \end{tabular}
% \caption{Computational cost of \textsc{PrefLens}.}
% \label{tab:runtime}
% \end{table}

\section{Conclusion}
We introduced \textsc{PrefLens}, which enriches sparse preference supervision by constructing local neighborhoods around human-preferred images and selecting informative comparisons in a reward-defined preference space. Diagnostic analyses show that the recovered pairs preserve the intended preference ordering and remain semantically close to their anchors, while downstream evaluations demonstrate consistent alignment gains. Our results indicate that both neighborhood expansion and reward-guided informative selection contribute to these improvements. Future work will explore richer preference representations and perturbation strategies guided by model failure modes. 
% See Section~\ref{sec:limitations} for limitations and future-work directions.

\section{Limitations and Future Work}
\label{sec:limitations}

\textbf{\textit{Subjectivity of human preferences.}}
Human preferences are inherently subjective and may reflect multiple, sometimes competing, criteria that vary across individuals and contexts. Consequently, a single preference ordering or evaluator may not capture the full reasoning underlying a judgment. Related directions such as ICAI~\cite{findeis2025inverse} and Democratic ICAI~\cite{kingslin2026democratic} similarly motivate recovering multiple underlying principles and rationales from pairwise preferences. Future work could incorporate evaluator disagreement, uncertainty, and diverse preference principles to construct more robust and potentially personalized preference neighborhoods.

\textbf{\textit{Preference-space representation.}}
\textsc{PrefLens} currently represents preference structure using a fixed set of pretrained reward models. Although complementary reward signals provide a practical low-dimensional space for neighborhood construction, they cannot capture every perceptual or semantic factor relevant to human preference. Future work could explore richer, learned preference representations that adapt to the data and capture interactions among multiple preference dimensions.

\textbf{\textit{Coverage of local variations.}}
The current neighborhood construction relies on a predefined family of controlled perceptual perturbations. While this enables systematic and interpretable exploration around each preferred anchor, a fixed transformation set cannot represent every meaningful variation or failure mode of a generative model. Future work could develop adaptive, model-informed perturbations derived from observed generation failure modes, enabling neighborhoods that better reflect the weaknesses of the model being aligned.

% \clearpage
% \input{Figures/noise_abalation}
% \input{Figures/minmax_abalation}
% \input{Figures/epsilon_plot}
% \input{Figures/ablation_plot2}
% \input{Table/icai_data}
% \input{Table/hyper_sd15}
% \input{Figures/compo_plot_sdxl}

% \clearpage

\section*{LLM Usage Statement}

We used ChatGPT to assist with language editing and improving
the clarity of the manuscript. In our experiments, GPT-4o
was used to generate evaluation questions for TIFA, while
Qwen3-VL and LLaVA were used as automated evaluators for
image preferences and consistency checks. We also employed
LLM-assisted inverse constitutional analysis to examine
preference characteristics. The authors reviewed the
generated outputs and remain responsible for the methodology,
experimental results, and conclusions.

%\clearpage

{
   \small
   \bibliographystyle{ieeenat_fullname}
   \bibliography{main}

@String(CVPR  = {IEEE Conf. Comput. Vis. Pattern Recog.})

@String(NeurIPS = {Adv. Neural Inform. Process. Syst.})

@String(CVPR  = {CVPR})

@String(NeurIPS = {NeurIPS})

@inproceedings{wallace2024diffusion,
  title={Diffusion model alignment using direct preference optimization},
  author={Wallace, Bram and Dang, Meihua and Rafailov, Rafael and Zhou, Linqi and Lou, Aaron and Purushwalkam, Senthil and Ermon, Stefano and Xiong, Caiming and Joty, Shafiq and Naik, Nikhil},
  booktitle={Proceedings of the IEEE/CVF Conference on Computer Vision and Pattern Recognition},
  pages={8228--8238},
  year={2024}
}

@article{li2024aligning,
  title={Aligning diffusion models by optimizing human utility},
  author={Li, Shufan and Kallidromitis, Konstantinos and Gokul, Akash and Kato, Yusuke and Kozuka, Kazuki},
  journal={arXiv preprint arXiv:2404.04465},
  year={2024}
}

@InProceedings{Rombach_2022_CVPR,
    author    = {Rombach, Robin and Blattmann, Andreas and Lorenz, Dominik and Esser, Patrick and Ommer, Bj\"orn},
    title     = {High-Resolution Image Synthesis With Latent Diffusion Models},
    booktitle = {Proceedings of the IEEE/CVF Conference on Computer Vision and Pattern Recognition (CVPR)},
    month     = {June},
    year      = {2022},
    pages     = {10684-10695}
}

@article{podell2023sdxl,
  title={Sdxl: Improving latent diffusion models for high-resolution image synthesis},
  author={Podell, Dustin and English, Zion and Lacey, Kyle and Blattmann, Andreas and Dockhorn, Tim and M{\"u}ller, Jonas and Penna, Joe and Rombach, Robin},
  journal={arXiv preprint arXiv:2307.01952},
  year={2023}
}

@article{kirstain2023pick,
  title={Pick-a-pic: An open dataset of user preferences for text-to-image generation},
  author={Kirstain, Yuval and Polyak, Adam and Singer, Uriel and Matiana, Shahbuland and Penna, Joe and Levy, Omer},
  journal={Advances in Neural Information Processing Systems},
  volume={36},
  pages={36652--36663},
  year={2023}
}

@article{li2024genai,
  title={Genai-bench: Evaluating and improving compositional text-to-visual generation},
  author={Li, Baiqi and Lin, Zhiqiu and Pathak, Deepak and Li, Jiayao and Fei, Yixin and Wu, Kewen and Ling, Tiffany and Xia, Xide and Zhang, Pengchuan and Neubig, Graham and others},
  journal={arXiv preprint arXiv:2406.13743},
  year={2024}
}

@article{yu2022scaling,
  title={Scaling autoregressive models for content-rich text-to-image generation},
  author={Yu, Jiahui and Xu, Yuanzhong and Koh, Jing Yu and Luong, Thang and Baid, Gunjan and Wang, Zirui and Vasudevan, Vijay and Ku, Alexander and Yang, Yinfei and Ayan, Burcu Karagol and others},
  journal={arXiv preprint arXiv:2206.10789},
  volume={2},
  number={3},
  pages={5},
  year={2022}
}

@misc{schuhmann2022laion,
  author       = {Christoph Schuhmann},
  title        = {{LAION-Aesthetics}},
  year         = {2022},
  howpublished = {\url{https://laion.ai/blog/laion-aesthetics/}},
  note         = {Accessed: 2023-11-10}
}

@inproceedings{xu2023imagereward,
  title={ImageReward: learning and evaluating human preferences for text-to-image generation},
  author={Xu, Jiazheng and Liu, Xiao and Wu, Yuchen and Tong, Yuxuan and Li, Qinkai and Ding, Ming and Tang, Jie and Dong, Yuxiao},
  booktitle={Proceedings of the 37th International Conference on Neural Information Processing Systems},
  pages={15903--15935},
  year={2023}
}

@article{saharia2022photorealistic,
  title={Photorealistic text-to-image diffusion models with deep language understanding},
  author={Saharia, Chitwan and Chan, William and Saxena, Saurabh and Li, Lala and Whang, Jay and Denton, Emily L and Ghasemipour, Kamyar and Gontijo Lopes, Raphael and Karagol Ayan, Burcu and Salimans, Tim and others},
  journal={Advances in neural information processing systems},
  volume={35},
  pages={36479--36494},
  year={2022}
}

@article{betker2023improving,
  title={Improving image generation with better captions (2023)},
  author={Betker, James and Goh, Gabriel and Jing, Li and Brooks, Tim and Wang, Jianfeng and Li, Linjie and Ouyang, Long and Zhuang, Juntang and Lee, Joyce and Guo, Yufei and others},
  journal={URL https://cdn. openai. com/papers/dall-e-3. pdf},
  year={2023}
}

@article{nichol2021glide,
  title={Glide: Towards photorealistic image generation and editing with text-guided diffusion models},
  author={Nichol, Alex and Dhariwal, Prafulla and Ramesh, Aditya and Shyam, Pranav and Mishkin, Pamela and McGrew, Bob and Sutskever, Ilya and Chen, Mark},
  journal={arXiv preprint arXiv:2112.10741},
  year={2021}
}

@inproceedings{rombach2022high,
  title={High-resolution image synthesis with latent diffusion models},
  author={Rombach, Robin and Blattmann, Andreas and Lorenz, Dominik and Esser, Patrick and Ommer, Bj{\"o}rn},
  booktitle={Proceedings of the IEEE/CVF conference on computer vision and pattern recognition},
  pages={10684--10695},
  year={2022}
}

@article{fan2023dpok,
  title={Dpok: Reinforcement learning for fine-tuning text-to-image diffusion models},
  author={Fan, Ying and Watkins, Olivia and Du, Yuqing and Liu, Hao and Ryu, Moonkyung and Boutilier, Craig and Abbeel, Pieter and Ghavamzadeh, Mohammad and Lee, Kangwook and Lee, Kimin},
  journal={Advances in Neural Information Processing Systems},
  volume={36},
  pages={79858--79885},
  year={2023}
}

@article{christiano2017deep,
  title={Deep reinforcement learning from human preferences},
  author={Christiano, Paul F and Leike, Jan and Brown, Tom and Martic, Miljan and Legg, Shane and Amodei, Dario},
  journal={Advances in neural information processing systems},
  volume={30},
  year={2017}
}

@article{black2023training,
  title={Training diffusion models with reinforcement learning},
  author={Black, Kevin and Janner, Michael and Du, Yilun and Kostrikov, Ilya and Levine, Sergey},
  journal={arXiv preprint arXiv:2305.13301},
  year={2023}
}

@inproceedings{liang2024rich,
  title={Rich human feedback for text-to-image generation},
  author={Liang, Youwei and He, Junfeng and Li, Gang and Li, Peizhao and Klimovskiy, Arseniy and Carolan, Nicholas and Sun, Jiao and Pont-Tuset, Jordi and Young, Sarah and Yang, Feng and others},
  booktitle={Proceedings of the IEEE/CVF Conference on Computer Vision and Pattern Recognition},
  pages={19401--19411},
  year={2024}
}

@inproceedings{zhu2025dspo,
  title={DSPO: Direct score preference optimization for diffusion model alignment},
  author={Zhu, Huaisheng and Xiao, Teng and Honavar, Vasant G},
  booktitle={The Thirteenth International Conference on Learning Representations}
}

@inproceedings{sun2024dreamsync,
  title={DreamSync: Aligning Text-to-Image Generation with Image Understanding Feedback},
  author={Sun, Jiao and Fu, Deqing and Hu, Yushi and Wang, Su and Rassin, Royi and Juan, Da-Cheng and Alon, Dana and Herrmann, Charles and van Steenkiste, Sjoerd and Krishna, Ranjay and others},
  booktitle={Synthetic Data for Computer Vision Workshop@ CVPR 2024}
}

@article{ho2020denoising,
  title={Denoising diffusion probabilistic models},
  author={Ho, Jonathan and Jain, Ajay and Abbeel, Pieter},
  journal={Advances in neural information processing systems},
  volume={33},
  pages={6840--6851},
  year={2020}
}

@article{wu2023human,
  title={Human Preference Score v2: A Solid Benchmark for Evaluating Human Preferences of Text-to-Image Synthesis},
  author={Wu, Xiaoshi and Hao, Yiming and Sun, Keqiang and Chen, Yixiong and Zhu, Feng and Zhao, Rui and Li, Hongsheng},
  journal={CoRR},
  year={2023}
}

@article{wu2024multimodal,
  title={Multimodal large language models make text-to-image generative models align better},
  author={Wu, Xun and Huang, Shaohan and Wang, Guolong and Xiong, Jing and Wei, Furu},
  journal={Advances in Neural Information Processing Systems},
  volume={37},
  pages={81287--81323},
  year={2024}
}

@inproceedings{lusmoothed,
  title={Smoothed Preference Optimization via ReNoise Inversion for Aligning Diffusion Models with Varied Human Preferences},
  author={Lu, Yunhong and Wang, Qichao and Cao, Hengyuan and Xu, Xiaoyin and Zhang, Min},
  booktitle={Forty-second International Conference on Machine Learning}
}

@inproceedings{singh2024effective,
  title={Effective Text-to-Image Alignment with Quality Aware Pair Ranking},
  author={Singh, Kunal and Khanna, Mukund and Moturi, Pradeep},
  booktitle={NeurIPS 2024 Workshop on Fine-Tuning in Modern Machine Learning: Principles and Scalability}
}

@inproceedings{croitoru2025curriculum,
  title={Curriculum direct preference optimization for diffusion and consistency models},
  author={Croitoru, Florinel-Alin and Hondru, Vlad and Ionescu, Radu Tudor and Sebe, Nicu and Shah, Mubarak},
  booktitle={Proceedings of the Computer Vision and Pattern Recognition Conference},
  pages={2824--2834},
  year={2025}
}

@article{hendrycks2019augmix,
  title={Augmix: A simple data processing method to improve robustness and uncertainty},
  author={Hendrycks, Dan and Mu, Norman and Cubuk, Ekin D and Zoph, Barret and Gilmer, Justin and Lakshminarayanan, Balaji},
  journal={arXiv preprint arXiv:1912.02781},
  year={2019}
}

@article{gu2024diffusion,
  title={Diffusion-rpo: Aligning diffusion models through relative preference optimization},
  author={Gu, Yi and Wang, Zhendong and Yin, Yueqin and Xie, Yujia and Zhou, Mingyuan},
  journal={arXiv preprint arXiv:2406.06382},
  year={2024}
}

@misc{huggingfaceDataisbettertogetheropenimagepreferencesv1binarizedDatasets,
	author = {},
	title = {data-is-better-together/open-image-preferences-v1-binarized · {D}atasets at {H}ugging {F}ace --- huggingface.co},
	howpublished = {\url{https://huggingface.co/datasets/data-is-better-together/open-image-preferences-v1-binarized}},
	year = {},
	note = {[Accessed 10-05-2025]},
}

@misc{huggingfaceBlackforestlabsFLUX1devHugging,
	author = {},
	title = {black-forest-labs/{F}{L}{U}{X}.1-dev · {H}ugging {F}ace --- huggingface.co},
	howpublished = {\url{https://huggingface.co/black-forest-labs/FLUX.1-dev}},
	year = {},
	note = {[Accessed 10-05-2025]},
}

@inproceedings{esser2024scaling,
  title={Scaling rectified flow transformers for high-resolution image synthesis},
  author={Esser, Patrick and Kulal, Sumith and Blattmann, Andreas and Entezari, Rahim and M{\"u}ller, Jonas and Saini, Harry and Levi, Yam and Lorenz, Dominik and Sauer, Axel and Boesel, Frederic and others},
  booktitle={Forty-first international conference on machine learning},
  year={2024}
}

@inproceedings{lee2025calibrated,
  title={Calibrated multi-preference optimization for aligning diffusion models},
  author={Lee, Kyungmin and Li, Xiahong and Wang, Qifei and He, Junfeng and Ke, Junjie and Yang, Ming-Hsuan and Essa, Irfan and Shin, Jinwoo and Yang, Feng and Li, Yinxiao},
  booktitle={Proceedings of the Computer Vision and Pattern Recognition Conference},
  pages={18465--18475},
  year={2025}
}

@inproceedings{lu2025inpo,
  title={InPO: Inversion Preference Optimization with Reparametrized DDIM for Efficient Diffusion Model Alignment},
  author={Lu, Yunhong and Wang, Qichao and Cao, Hengyuan and Wang, Xierui and Xu, Xiaoyin and Zhang, Min},
  booktitle={Proceedings of the Computer Vision and Pattern Recognition Conference},
  pages={28629--28639},
  year={2025}
}

@inproceedings{yangdense,
  title={A Dense Reward View on Aligning Text-to-Image Diffusion with Preference},
  author={Yang, Shentao and Chen, Tianqi and Zhou, Mingyuan},
  booktitle={Forty-first International Conference on Machine Learning}
}

@article{wu2019survey,
  title={Survey of visual just noticeable difference estimation},
  author={Wu, Jinjian and Shi, Guangming and Lin, Weisi},
  journal={Frontiers of Computer Science},
  volume={13},
  number={1},
  pages={4--15},
  year={2019},
  publisher={Springer}
}

@inproceedings{karthik2025scalable,
  title={Scalable ranked preference optimization for text-to-image generation},
  author={Karthik, Shyamgopal and Coskun, Huseyin and Akata, Zeynep and Tulyakov, Sergey and Ren, Jian and Kag, Anil},
  booktitle={Proceedings of the IEEE/CVF International Conference on Computer Vision},
  pages={18399--18410},
  year={2025}
}

@article{jang2025multi,
  title={Multi-dimensional Preference Alignment by Conditioning Reward Itself},
  author={Jang, Jiho and Kim, Jinyoung and Baek, Kyungjune and Kwak, Nojun},
  journal={arXiv preprint arXiv:2512.10237},
  year={2025}
}

@article{li2026preference,
  title={Preference-Guided Prompt Optimization for Text-to-Image Generation},
  author={Li, Zhipeng and Liao, Yi-Chi and Holz, Christian},
  journal={arXiv preprint arXiv:2602.13131},
  year={2026}
}

@article{xian2025free,
  title={Free Lunch Alignment of Text-to-Image Diffusion Models without Preference Image Pairs},
  author={Xian, Jia Jun Cheng and Li, Muchen and Yang, Haotian and Tao, Xin and Wan, Pengfei and Sigal, Leonid and Liao, Renjie},
  journal={arXiv preprint arXiv:2509.25771},
  year={2025}
}

@inproceedings{schroff2015facenet,
  title={Facenet: A unified embedding for face recognition and clustering},
  author={Schroff, Florian and Kalenichenko, Dmitry and Philbin, James},
  booktitle={Proceedings of the IEEE conference on computer vision and pattern recognition},
  pages={815--823},
  year={2015}
}

@article{erkut1990discrete,
  title={The discrete p-dispersion problem},
  author={Erkut, Erhan},
  journal={European Journal of Operational Research},
  volume={46},
  number={1},
  pages={48--60},
  year={1990},
  publisher={Elsevier}
}

@inproceedings{ravi1991facility,
  title={Facility dispersion problems: Heuristics and special cases},
  author={Ravi, Sekharipuram S and Rosenkrantz, Daniel J and Tayi, Giri Kumar},
  booktitle={Workshop on Algorithms and Data Structures},
  pages={355--366},
  year={1991},
  organization={Springer}
}

@article{ouyang2022training,
  title={Training language models to follow instructions with human feedback},
  author={Ouyang, Long and Wu, Jeffrey and Jiang, Xu and Almeida, Diogo and Wainwright, Carroll and Mishkin, Pamela and Zhang, Chong and Agarwal, Sandhini and Slama, Katarina and Ray, Alex and others},
  journal={Advances in neural information processing systems},
  volume={35},
  pages={27730--27744},
  year={2022}
}

@book{miettinen1999nonlinear,
  title={Nonlinear multiobjective optimization},
  author={Miettinen, Kaisa},
  volume={12},
  year={1999},
  publisher={Springer Science \& Business Media}
}

@article{spearman1961proof,
  title={The proof and measurement of association between two things.},
  author={Spearman, Charles},
  year={1961},
  publisher={Appleton-Century-Crofts}
}

@article{pearson1894iii,
  title={III. Contributions to the mathematical theory of evolution},
  author={Pearson, Karl},
  journal={Proceedings of the Royal Society of London},
  volume={54},
  number={326-330},
  pages={329--333},
  year={1894},
  publisher={The Royal Society London}
}

@article{cover1967nearest,
  title={Nearest neighbor pattern classification},
  author={Cover, Thomas and Hart, Peter},
  journal={IEEE transactions on information theory},
  volume={13},
  number={1},
  pages={21--27},
  year={1967},
  publisher={IEEE}
}

@article{emerson2013original,
  title={The original Borda count and partial voting},
  author={Emerson, Peter},
  journal={Social Choice and Welfare},
  volume={40},
  number={2},
  pages={353--358},
  year={2013},
  publisher={Springer}
}

@inproceedings{hu2023tifa,
  title={Tifa: Accurate and interpretable text-to-image faithfulness evaluation with question answering},
  author={Hu, Yushi and Liu, Benlin and Kasai, Jungo and Wang, Yizhong and Ostendorf, Mari and Krishna, Ranjay and Smith, Noah A},
  booktitle={Proceedings of the IEEE/CVF International Conference on Computer Vision},
  pages={20406--20417},
  year={2023}
}

@article{mcleod2005kendall,
  title={Kendall rank correlation and Mann-Kendall trend test},
  author={McLeod, A Ian},
  journal={R package Kendall},
  volume={602},
  pages={1--10},
  year={2005}
}

@article{hurst2024gpt,
  title={Gpt-4o system card},
  author={Hurst, Aaron and Lerer, Adam and Goucher, Adam P and Perelman, Adam and Ramesh, Aditya and Clark, Aidan and Ostrow, AJ and Welihinda, Akila and Hayes, Alan and Radford, Alec and others},
  journal={arXiv preprint arXiv:2410.21276},
  year={2024}
}

@inproceedings{li2022blip,
  title={Blip: Bootstrapping language-image pre-training for unified vision-language understanding and generation},
  author={Li, Junnan and Li, Dongxu and Xiong, Caiming and Hoi, Steven},
  booktitle={International conference on machine learning},
  pages={12888--12900},
  year={2022},
  organization={PMLR}
}

@article{wang2022git,
  title={GIT: A Generative Image-to-text Transformer for Vision and Language},
  author={Wang, Jianfeng and Yang, Zhengyuan and Hu, Xiaowei and Li, Linjie and Lin, Kevin and Gan, Zhe and Liu, Zicheng and Liu, Ce and Wang, Lijuan},
  year={2022}
}

@article{bai2025qwen3,
  title={Qwen3-vl technical report},
  author={Bai, Shuai and Cai, Yuxuan and Chen, Ruizhe and Chen, Keqin and Chen, Xionghui and Cheng, Zesen and Deng, Lianghao and Ding, Wei and Gao, Chang and Ge, Chunjiang and others},
  journal={arXiv preprint arXiv:2511.21631},
  year={2025}
}

@article{ravi1994heuristic,
  title={Heuristic and special case algorithms for dispersion problems},
  author={Ravi, Sekharipuram S and Rosenkrantz, Daniel J and Tayi, Giri Kumar},
  journal={Operations research},
  volume={42},
  number={2},
  pages={299--310},
  year={1994},
  publisher={INFORMS}
}

@article{bai2025towards,
  title={Towards better optimization for listwise preference in diffusion models},
  author={Bai, Jiamu and Yu, Xin and Xu, Meilong and Lu, Weitao and Pan, Xin and Maeng, Kiwan and Kifer, Daniel and Wang, Jian and Wang, Yu},
  journal={arXiv preprint arXiv:2510.01540},
  year={2025}
}

@inproceedings{findeis2025inverse,
  title={Inverse constitutional ai: Compressing preferences into principles},
  author={Findeis, Arduin and Kaufmann, Timo and H{\"u}llermeier, Eyke and Albanie, Samuel and Mullins, Robert},
  booktitle={International Conference on Learning Representations},
  volume={2025},
  pages={52687--52730},
  year={2025}
}

@article{surana2026mass,
  title={MASS-DPO: Multi-negative Active Sample Selection for Direct Policy Optimization},
  author={Surana, Rohan and Li, Xintong and Yu, Sheldon and Shen, Yiran Jenny and Wang, Chuhan and Yu, Tong and Ammanabrolu, Prithviraj and Shang, Jingbo and McAuley, Julian and Wu, Junda},
  journal={arXiv preprint arXiv:2605.10784},
  year={2026}
}

@inproceedings{wu2025ranking,
  title={Ranking-based Preference Optimization for Diffusion Models from Implicit User Feedback},
  author={Wu, Yi-Lun and Ruan, Bo-Kai and Tseng, Chiang and Shuai, Hong-Han},
  booktitle={The Thirty-ninth Annual Conference on Neural Information Processing Systems},
  year={2025}
}

@article{lee2026mm,
  title={MM-JudgeBias: A Benchmark for Evaluating Compositional Biases in MLLM-as-a-Judge},
  author={Lee, Sua and Park, Sanghee and Im, Jinbae},
  journal={arXiv preprint arXiv:2604.18164},
  year={2026}
}

@article{gentner1983structure,
  title={Structure-mapping: A theoretical framework for analogy},
  author={Gentner, Dedre},
  journal={Cognitive science},
  volume={7},
  number={2},
  pages={155--170},
  year={1983},
  publisher={Elsevier}
}

@article{gentner1997structure,
  title={Structure mapping in analogy and similarity.},
  author={Gentner, Dedre and Markman, Arthur B},
  journal={American psychologist},
  volume={52},
  number={1},
  pages={45},
  year={1997},
  publisher={American Psychological Association}
}

@article{sagi2012difference,
  title={What difference reveals about similarity},
  author={Sagi, Eyal and Gentner, Dedre and Lovett, Andrew},
  journal={Cognitive science},
  volume={36},
  number={6},
  pages={1019--1050},
  year={2012},
  publisher={Wiley Online Library}
}

@article{kingslin2026democratic,
  title={Democratic ICAI: Debating Our Way to Steering Principles from Preferences},
  author={Kingslin, Kevin and Natekar, Anish and Ranjan, Ashutosh and Srivastava, Vivek and Bhat, Savita and Karande, Shirish},
  journal={arXiv preprint arXiv:2606.28294},
  year={2026}
}

@inproceedings{zou2026vision,
  title={When Vision-Language Models Judge Without Seeing: Exposing Informativeness Bias},
  author={Zou, Xiaohan and Sridhar, Roshan and Safarzadeh, Mohammadtaher and Roth, Dan},
  booktitle={Proceedings of the 64th Annual Meeting of the Association for Computational Linguistics (Volume 1: Long Papers)},
  pages={15417--15448},
  year={2026}
}

@inproceedings{laskar2025judging,
  title={Judging the Judges: Can Large Vision-Language Models Fairly Evaluate Chart Comprehension and Reasoning?},
  author={Laskar, Md Tahmid Rahman and Islam, Mohammed Saidul and Mahbub, Ridwan and Masry, Ahmed and Rahman, Mizanur and Bhuiyan, Amran and Nayeem, Mir Tafseer and Joty, Shafiq and Hoque, Enamul and Huang, Jimmy},
  booktitle={Proceedings of the 63rd Annual Meeting of the Association for Computational Linguistics (Volume 6: Industry Track)},
  pages={1203--1216},
  year={2025}
}

@article{zhang2024seppo,
  title={Seppo: Semi-policy preference optimization for diffusion alignment},
  author={Zhang, Daoan and Lan, Guangchen and Han, Dong-Jun and Yao, Wenlin and Pan, Xiaoman and Zhang, Hongming and Li, Mingxiao and Chen, Pengcheng and Dong, Yu and Brinton, Christopher and others},
  journal={arXiv preprint arXiv:2410.05255},
  year={2024}
}

@article{liang2024step,
  title={Step-aware preference optimization: Aligning preference with denoising performance at each step},
  author={Liang, Zhanhao and Yuan, Yuhui and Gu, Shuyang and Chen, Bohan and Hang, Tiankai and Li, Ji and Zheng, Liang},
  journal={arXiv preprint arXiv:2406.04314},
  volume={2},
  number={5},
  pages={7},
  year={2024}
}

@article{yuan2024self,
  title={Self-play fine-tuning of diffusion models for text-to-image generation},
  author={Yuan, Huizhuo and Chen, Zixiang and Ji, Kaixuan and Gu, Quanquan},
  journal={Advances in Neural Information Processing Systems},
  volume={37},
  pages={73366--73398},
  year={2024}
}

@inproceedings{wang2025diffusion,
  title={Diffusion-npo: Negative preference optimization for better preference aligned generation of diffusion models},
  author={Wang, Fu-Yun and Shui, Yunhao and Piao, Jingtan and Sun, Keqiang and Li, Hongsheng},
  booktitle={International Conference on Learning Representations},
  volume={2025},
  pages={71440--71462},
  year={2025}
}

@article{huang2025t2i,
  title={T2i-compbench++: An enhanced and comprehensive benchmark for compositional text-to-image generation},
  author={Huang, Kaiyi and Duan, Chengqi and Sun, Kaiyue and Xie, Enze and Li, Zhenguo and Liu, Xihui},
  journal={IEEE Transactions on Pattern Analysis and Machine Intelligence},
  volume={47},
  number={5},
  pages={3563--3579},
  year={2025},
  publisher={IEEE}
}

@article{oquab2023dinov2,
  title={Dinov2: Learning robust visual features without supervision},
  author={Oquab, Maxime and Darcet, Timoth{\'e}e and Moutakanni, Th{\'e}o and Vo, Huy and Szafraniec, Marc and Khalidov, Vasil and Fernandez, Pierre and Haziza, Daniel and Massa, Francisco and El-Nouby, Alaaeldin and others},
  journal={arXiv preprint arXiv:2304.07193},
  year={2023}
}

@inproceedings{radford2021learning,
  title={Learning transferable visual models from natural language supervision},
  author={Radford, Alec and Kim, Jong Wook and Hallacy, Chris and Ramesh, Aditya and Goh, Gabriel and Agarwal, Sandhini and Sastry, Girish and Askell, Amanda and Mishkin, Pamela and Clark, Jack and others},
  booktitle={International conference on machine learning},
  pages={8748--8763},
  year={2021},
  organization={PmLR}
}
}

\clearpage
\appendix

\onecolumn

\startcontents[appendices]
\begin{center}

    {\Large\bfseries Appendix\par}
\end{center}

\addcontentsline{toc}{section}{Contents}

\printcontents[appendices]{}{1}{}

\clearpage
\twocolumn

\raggedbottom
\setlength{\floatsep}{8pt plus 2pt minus 2pt}
\setlength{\textfloatsep}{10pt plus 2pt minus 2pt}
\setlength{\dblfloatsep}{8pt plus 2pt minus 2pt}
\setlength{\dbltextfloatsep}{10pt plus 2pt minus 2pt}
\renewcommand{\topfraction}{0.95}
\renewcommand{\dbltopfraction}{0.95}
\renewcommand{\textfraction}{0.05}
\makeatletter
\setlength{\@fptop}{0pt}
\setlength{\@fpsep}{10pt}
\setlength{\@fpbot}{0pt plus 1fil}
\setlength{\@dblfptop}{0pt}
\setlength{\@dblfpsep}{10pt}
\setlength{\@dblfpbot}{0pt plus 1fil}
\makeatother

\section{Background}
\label{sec:background}

\subsection{Denoising Diffusion Probabilistic Models} Denoising Diffusion Probabilistic Models~\cite{ho2020denoising} define a forward noising process that incrementally corrupts a clean image $\mathbf{x}_0\!\sim\! p_{\mathrm{data}}$ in $T$ discrete steps via a fixed variance schedule $\{\beta_t\}_{t=1}^T$. In each step:

\begin{equation}
q(\mathbf{x}_t\mid\mathbf{x}_{t-1}) = \mathcal{N}\!\bigl(\sqrt{1-\beta_t}\,\mathbf{x}_{t-1},\,\beta_t\mathbf{I}\bigr),
\end{equation}

so that overall

\begin{equation}
q(\mathbf{x}_{1:T}\mid\mathbf{x}_0) = \prod_{t=1}^T q(\mathbf{x}_t\mid\mathbf{x}_{t-1})
\end{equation}

The model then learns the reverse Markov chain $p_\theta(\mathbf{x}_{t-1}\mid\mathbf{x}_t)$, parameterized by a neural network $\epsilon_\theta(\mathbf{x}_t,t)$ that predicts the added noise:

\begin{equation}
\begin{aligned}
p_\theta(\mathbf{x}_{t-1}\mid\mathbf{x}_t) &= \mathcal{N}\!\Bigl(
  \mathbf{x}_{t-1};
  \frac{\mathbf{x}_t - \dfrac{\beta_t}{\sqrt{1-\bar\alpha_t}}\,\epsilon_\theta(\mathbf{x}_t,t)}{\sqrt{1-\beta_t}},\\&
  \sigma_t^2\mathbf{I}
\Bigr),\\[4pt]
\text{where } \sigma_t^2 &= \frac{1-\bar\alpha_{t-1}}{1-\bar\alpha_t}\beta_t,\qquad
\bar\alpha_t = \prod_{s=1}^t (1-\beta_s)
\end{aligned}
\end{equation}

Training is performed by minimizing the re-weighted mean-squared error

\begin{equation}
L(\theta) = \mathbb{E}_{t,\mathbf{x}_0,\epsilon}\!\Bigl[\lambda(t)\,\lVert\epsilon - \epsilon_\theta(\mathbf{x}_t,t)\rVert^2\Bigr],
\end{equation}

where $\mathbf{x}_t = \sqrt{\bar\alpha_t}\,\mathbf{x}_0 + \sqrt{1-\bar\alpha_t}\,\epsilon$ with $\epsilon\sim\mathcal{N}(\mathbf{0},\mathbf{I})$.

\subsection{RLHF for T2I Diffusion Models} Reinforcement Learning from Human Feedback (RLHF) adapts a pretrained diffusion model using human preferences. First, a reward model \(r(c, x)\) is trained on comparison data \((x_w, x_l)\) using a Bradley--Terry (BT) likelihood:

\begin{equation}
P(x_w \succ x_l \mid c) = \sigma\!\bigl(r(c, x_w) - r(c, x_l)\bigr),
\end{equation}

where \(c\) is the text prompt and \(\sigma\) the sigmoid function. Training minimizes the negative log-likelihood of observed preferences.

Next, the diffusion model, viewed as a multi-step Markov Decision Process (MDP) producing trajectories \((x_{0:T})\), is fine-tuned to maximize cumulative reward while staying close to its pretrained behavior:

\begin{equation}
\begin{split}
\mathcal{L}_{\text{RLHF}}
&=
\mathbb{E}_{c,\,x_{0:T}\sim p_\theta(\cdot\mid c)}
\left[\sum_{t=0}^{T-1} r(x_t, c)\right] \\
&\quad
-\lambda\,D_{\mathrm{KL}}
\bigl(
p_\theta(x_{0:T}\mid c)
\,\|\,
p_{\mathrm{pref}}(x_{0:T}\mid c)
\bigr)
\end{split}
\end{equation}

where \(p_{\mathrm{pref}}\) is the pretrained diffusion distribution, and \(\lambda\) balances reward versus KL regularization.

\subsection{Preference Alignment with Diffusion DPO} Diffusion-DPO~\cite{wallace2024diffusion} adapts Direct Preference Optimization to diffusion models by defining a reward on the full reverse trajectory and optimizing for alignment with human preferences, without requiring a reward model.

\noindent\textbf{Trajectory-level reward.}
Given a prompt \( c \) and a preferred sample \( x_0^w \succ x_0^l \), we define the preference-based reward on the clean image \( x_0 \) as the expected reward over the reverse diffusion trajectory \( x_{1:T} \sim p_\theta(\cdot \mid x_0, c) \):

\begin{equation}
r(c, x_0) = \mathbb{E}_{x_{1:T} \sim p_\theta(\cdot \mid x_0, c)} \left[ R(c, x_{0:T}) \right],
\label{eq:reward}
\end{equation}

where \( R(c, x_{0:T}) \) is a reward function over the entire denoising path.

\noindent\textbf{Final Diffusion-DPO training objective.}
Using a preference dataset \( D = \{(c, x_0^w, x_0^l)\} \), the loss encourages higher reward for preferred samples over less preferred ones. After approximation and simplification, the final training loss becomes:

\begin{equation}
\begin{aligned}
\mathcal{L}(\theta)
&=
- \mathbb{E}_{(x_0^w, x_0^l)\sim D,\ t\sim\mathcal U(0,T)}
\log \sigma \Big(
-\beta T\,\omega(\lambda_t)
\Big[
\\
&\qquad
\|\epsilon^w-\epsilon_\theta(x_t^w,t)\|^2
-
\|\epsilon^w-\epsilon_{\mathrm{ref}}(x_t^w,t)\|^2
\\
&\qquad
-
\|\epsilon^l-\epsilon_\theta(x_t^l,t)\|^2
+
\|\epsilon^l-\epsilon_{\mathrm{ref}}(x_t^l,t)\|^2
\Big]
\Big),
\end{aligned}
\end{equation}

where \( x_t^\ast = \alpha_t x_0^\ast + \sigma_t \epsilon^\ast \), with \( \epsilon^\ast \sim \mathcal{N}(0, I) \), and \( \lambda_t = \alpha_t^2 / \sigma_t^2 \) is the signal-to-noise ratio at step \( t \). The loss aligns the model's denoising trajectory with human preferences, relative to a frozen reference model.

\subsection{Direct Score Preference Optimization (DSPO)}
DSPO~\cite{zhu2025dspo} aligns diffusion models with pairwise human
preferences through score matching, thereby keeping the fine-tuning
objective consistent with the score-matching objective used during
diffusion pretraining. Given a preference pair
$(c,x_0^w,x_0^l)$, let $x_t^w$ and $x_t^l$ denote the corresponding
noisy samples at timestep $t$. DSPO treats human preference as an
additional condition $y$ and defines the preference-conditioned score
using Bayes' rule:
\begin{equation}
\begin{aligned}
\nabla_{x_t}\log p_\theta(x_t\mid c,y)
&=
\nabla_{x_t}\log p_\theta(x_t\mid c) \\
&\quad+
\nabla_{x_t}\log p(y\mid x_t,c).
\end{aligned}
\end{equation}
The preference condition is modeled using the Bradley--Terry
formulation
\begin{equation}
p(y\mid x_t^w,c)
=
\sigma\!\left(
r(c,x_t^w)-r(c,x_t^l)
\right),
\end{equation}
where $y$ denotes that $x_t^w$ is preferred to $x_t^l$. Based on this
formulation, DSPO fine-tunes the diffusion model by matching its score
to a human-preferred target score:
\begin{equation}
\begin{aligned}
\mathcal{L}_{\mathrm{DSPO}}
=
\mathbb{E}_{(c,x_0^w,x_0^l)\sim D,\,t}
\Big[
\omega(t)
\big\|
&\nabla_{x_t^w}\log p_\theta(x_t^w\mid c)
-\\ &
\nabla_{x_t^w}\log p_{\mathrm{data}}(x_t^w\mid c)
\\
-
\gamma\nabla_{x_t^w}
\log p(y\mid x_t^w,c)
\big\|_2^2
\Big],
\end{aligned}
\end{equation}
where $\omega(t)$ is a timestep-dependent score-matching weight and
$\gamma$ controls the strength of the preference condition. Rather
than training an additional preference model, DSPO uses the implicit
reward induced by the current and reference diffusion models,
following the DPO formulation. The resulting objective can be
expressed directly in terms of the noise predictions
$\epsilon_\theta$ and $\epsilon_{\mathrm{ref}}$, allowing preference
fine-tuning within the same denoising parameterization used during
diffusion pretraining. DSPO further shows that, under specific choices
of the weighting parameters, its optimization direction is equivalent
to that of the corresponding RLHF objective.
\section{Summary of Notations}
\label{sec:notation_data}
In this section, we summarize the key symbols and terms used throughout the paper. These notations describe the main components of our formulation, including prompts, images, perturbations, and the structure of training data (see Table~\ref{tab:notation}).

\begin{table}[!htb]
\centering
\scriptsize
\setlength{\tabcolsep}{4pt}
\begin{tabular}{p{0.20\linewidth} p{0.70\linewidth}}
\toprule
\textbf{Symbol} & \textbf{Description} \\
\midrule

$\mathcal D$ &
Human preference dataset of triplets $(p_i,x_i^+,x_i^-)$ \\

$p_i$ &
Text prompt associated with the $i$-th preference pair \\

$x_i^+$ &
Human-preferred image for prompt $p_i$; used as the anchor \\

$x_i^-$ &
Original less-preferred image paired with $x_i^+$ \\

$\mathcal N(x^+)$ &
Final local preference neighborhood associated with anchor $x^+$ \\

$\mathcal X$ &
Image space \\

$\mathcal T$ &
Family of controlled perceptual perturbation operators \\

$T_j$ &
A perturbation operator from $\mathcal T$ \\

$\tilde{x}^{-}$ &
Candidate negative obtained by composing perturbations around $x^+$ \\

$\mathcal C$ &
Candidate pool $\{\tilde{x}_1^-,\ldots,\tilde{x}_N^-\}$ generated around an anchor \\

$N$ &
Number of generated candidate neighbors in $\mathcal C$ \\

$\mathcal Q\subset\mathbb{R}^{d}$ &
Reward-defined preference space induced by pretrained reward models \\

$q:\mathcal X\rightarrow\mathcal Q$ &
Mapping from image space to the reward-defined preference space \\

$q(x)$ &
Preference representation of image $x$, i.e.,
$[r_1(x),\ldots,r_d(x)]$ \\

$r_i(x)$ &
Score assigned to image $x$ by the $i$-th reward model \\

$d$ &
Dimensionality of the reward-defined preference space \\

$d(q_i,q_j)$ &
Euclidean distance between two preference representations \\

$\epsilon$ &
Margin threshold used for Margin-Based Filtering (MBF) \\

$\mathcal C^{\epsilon}$ &
Candidate set retained after MBF:
$\{x\in\mathcal C:d(q(x),q(x^+))\geq\epsilon\}$ \\

$C_\ell$ &
$\ell$-th variation mode (cluster) in $\mathcal C^{\epsilon}$ \\

$L$ &
Number of variation modes (clusters) \\

$K$ &
Sampling ratio controlling the fraction of candidates selected from each cluster \\

$k_\ell$ &
Sampling budget for cluster $C_\ell$,
$k_\ell=\lfloor K|C_\ell|\rfloor$ \\

$S_\ell^*$ &
Subset of informative neighbors selected from cluster $C_\ell$ by the max--min dispersion objective \\

$\lambda_{\min},\lambda_{\max}$ &
Minimum and maximum perturbation strengths used for range-based perturbation sampling \\

\bottomrule
\end{tabular}
\caption{Summary of notation used in \textsc{PrefLens}.}
\label{tab:notation}
\end{table}

\section{Theoretical Analysis}
\label{sec:theory_appendix}

In this section, we provide theoretical justification for two key components of
\textsc{PrefLens}. First, under a bounded semantic-displacement assumption, we
show that compositions of controlled perceptual perturbations remain within a
bounded semantic neighborhood of the preferred anchor. Second, we show that
neighborhoods with broader coverage admit more accurate approximations of local
human preferences in the reward-defined preference space.

\subsection{Locality Preservation under Sequential Perturbations.}
The proposed neighborhood recovery procedure generates candidate negatives by
sequentially composing controlled perceptual perturbation operators. To analyze
their locality, we introduce a bounded semantic-displacement assumption in an
embedding space.

\begin{assumption}[Bounded Semantic Displacement]
Let $\phi:\mathcal X\rightarrow\mathcal Z$ denote a semantic embedding function
mapping images into a semantic space $\mathcal Z$. For the purpose of analysis,
we assume that every perturbation operator $T$ used by \textsc{PrefLens}
induces bounded semantic displacement, i.e.,
\[
|\phi(T(x))-\phi(x)|_2\leq \delta,
\]
for all images $x\in\mathcal X$, where $\delta>0$ is a small constant.
\end{assumption}

The above assumption formalizes the intuition that each controlled perceptual
perturbation induces only a bounded change in semantic representation. In the
following theorem, ``semantic-preserving'' refers to perturbations satisfying
this bounded-displacement condition rather than requiring exact semantic
invariance.

\begin{theorem}[Locality Preservation]
Let
\[
\tilde{x}^{-}=T_k\circ T_{k-1}\circ\cdots\circ T_1(x^+)
\]
denote a candidate negative generated by composing $k$ semantic-preserving perturbations. Then,
\[
|\phi(\tilde{x}^{-})-\phi(x^+)|_2
\leq k\delta
\]
Consequently, sequential perturbations remain within a bounded semantic neighborhood of the anchor.
\end{theorem}

\begin{proof}
Define the intermediate images
\[
x_0=x^+,
\qquad
x_i=T_i\circ T_{i-1}\circ\cdots\circ T_1(x^+),
\quad i=1,..,k
\]
Observe that $x_k=\tilde{x}^{-}$.

Using a telescoping sum,
\[
\phi(x_k)-\phi(x_0)
=
\sum_{i=1}^{k}
\left(
\phi(x_i)-\phi(x_{i-1})
\right)
\]

Since each perturbation is semantic-preserving and
$x_i=T_i(x_{i-1})$, we have
\begin{equation}
\begin{aligned}
\|\phi(\tilde{x}^{-})-\phi(x^+)\|_2
&\leq
\sum_{i=1}^{k}
\|\phi(x_i)-\phi(x_{i-1})\|_2 \\
&\leq
\sum_{i=1}^{k}\delta=
k\delta
\end{aligned}
\end{equation}
Hence, the composed perturbation remains within a bounded semantic neighborhood of the anchor image.
\end{proof}

\subsection{Neighborhood Coverage Improves Preference Approximation.}
We next provide theoretical motivation for the diversity-aware neighborhood
selection strategy employed by \textsc{PrefLens} in the reward-defined
preference space.

\begin{assumption}[Local Preference Smoothness]
Let $\mathcal Q$ denote the reward-defined preference space induced by the
reward models. We assume that human preferences within a local neighborhood are
governed by an unknown preference function
\[
f:\mathcal Q\rightarrow\mathbb R,
\]
where larger values indicate stronger preference. Furthermore, we assume that $f$ is locally Lipschitz, i.e.,
\[
|f(\mathbf q_i)-f(\mathbf q_j)|
\leq
L\, d(\mathbf q_i,\mathbf q_j),
\]
for all $\mathbf q_i,\mathbf q_j$ within a local neighborhood, where $L>0$ is a Lipschitz constant.
\end{assumption}

This assumption formalizes the intuition that nearby samples in the
reward-defined preference space should exhibit locally similar preference
values.

\begin{definition}[Covering Radius]
Let $\mathcal N_{\mathbf q}\subset\mathcal Q$ denote the representation of a
recovered local preference neighborhood in the reward-defined preference space,
and let $S\subseteq\mathcal N_{\mathbf q}$ denote a selected subset of
informative neighbors. The covering radius of $S$ is defined as
\[
\rho(S)
=
\sup_{\mathbf q\in\mathcal N_{\mathbf q}}
\min_{\mathbf s\in S}
d(\mathbf q,\mathbf s)
\]
The covering radius measures the largest distance between any point in the neighborhood and its nearest selected representative.
\end{definition}

\begin{theorem}[Neighborhood Coverage Improves Preference Approximation]
Let $S\subseteq\mathcal N_{\mathbf q}$ be a selected neighborhood and define the nearest-neighbor approximation
\[
\widehat f(\mathbf q)
=
f\!\left(
\arg\min_{\mathbf s\in S}
d(\mathbf q,\mathbf s)
\right)
\]
Then,
\[
\sup_{\mathbf q\in\mathcal N_{\mathbf q}}
|f(\mathbf q)-\widehat f(\mathbf q)|
\leq
L\,\rho(S)
\]
Consequently, neighborhoods with smaller covering radius provide more accurate approximations of local human preferences.
\end{theorem}

\begin{proof}
Fix an arbitrary point $\mathbf q\in\mathcal N_{\mathbf q}$ and let
\[
\mathbf s^*(\mathbf q)
=
\arg\min_{\mathbf s\in S}
d(\mathbf q,\mathbf s)
\]
denote its nearest selected neighbor. By definition,
\[
\widehat f(\mathbf q)
=
f(\mathbf s^*(\mathbf q))
\]

Since $f$ is locally Lipschitz,
\[
|f(\mathbf q)-\widehat f(\mathbf q)|
=
|f(\mathbf q)-f(\mathbf s^*(\mathbf q))|
\leq
L\,d(\mathbf q,\mathbf s^*(\mathbf q))
\]

By definition of the covering radius,
\[
d(\mathbf q,\mathbf s^*(\mathbf q))
\leq
\rho(S)
\]
Therefore,
\[
|f(\mathbf q)-\widehat f(\mathbf q)|
\leq
L\,\rho(S)
\]

Since the above argument holds for every $\mathbf q\in\mathcal N_{\mathbf q}$, taking the supremum over all neighborhood points yields
\[
\sup_{\mathbf q\in\mathcal N_{\mathbf q}}
|f(\mathbf q)-\widehat f(\mathbf q)|
\leq
L\,\rho(S)
\]
\end{proof}
Although the max--min selection objective used by \textsc{PrefLens} does not
explicitly minimize the covering radius $\rho(S)$, it serves a closely related
practical purpose under a fixed sampling budget. By discouraging the selection
of redundant neighbors that occupy similar regions of the reward-defined
preference space, max--min selection allocates the available samples across
distinct variation modes and preference-space regions. This allows the selected
subset to capture more of the information present in the recovered neighborhood
with the same number of training pairs. In this sense, diversity-aware
selection improves the representativeness and information efficiency of the
finite neighborhood subset, while the result above provides theoretical
motivation for why broader coverage of the underlying neighborhood can improve
local preference approximation.

\section{Hyperparameters}
\label{sec:hyper_data}

Table~\ref{tab:params-for-training} summarizes the training configurations
used for SD v1.5 and SDXL-Base. We use a generic training implementation
that supports multiple preference-optimization algorithms through the
\texttt{training\_method} argument. Consequently, some hyperparameters
listed in the table are method-specific and are only used when required by
the corresponding optimization objective; otherwise, they have no effect
on training. Unless otherwise specified, the remaining optimization and
LoRA settings are shared across methods. For \textbf{DPO} and \textbf{DSPO}, the $\beta$ hyperparameter is configured separately for the two diffusion backbones. For \textbf{SDXL}, we use $\beta=5000$ for both DPO and DSPO. For \textbf{SD v1.5}, we use $\beta=5000$ for DPO and $\beta=0.001$ for DSPO.
\begin{table}[htbp]
\centering
\small

\begin{subtable}[t]{0.49\textwidth}
\centering
\footnotesize
\begin{tabular}{p{3.35cm}|p{2.65cm}}
\hline
\textbf{Parameter} & \textbf{Value} \\
\hline
Model & Stable Diffusion v1.5 \\
VAE & -- \\
Mixed Precision & -- \\
Resolution & 512 \\
Batch Size & 16 \\
Grad. Accumulation & 2 \\
Gradient Checkpointing & True \\
8-bit Adam & True \\
LoRA Rank & 128 \\
Learning Rate & $5\times10^{-5}$ \\
LR Scheduler & constant \\
Warmup Steps & 0 \\
Max Training Steps & 300 \\
Checkpointing Steps & 50 \\
$\eta_{\mathrm{DPO}}$ & 0.001 \\
$\alpha_{\mathrm{DMPO}}$ & 0.01 \\
Training Method & method-dependent \\
Seed & 0 \\
\hline
\end{tabular}
\caption{SD v1.5 configuration}
\end{subtable}%
\\
\vspace{12pt}
\begin{subtable}[t]{0.49\textwidth}
\centering
\footnotesize
\begin{tabular}{p{3.35cm}|p{2.65cm}}
\hline
\textbf{Parameter} & \textbf{Value} \\
\hline
Model & SDXL-Base \\
VAE & SDXL fp16 VAE \\
Mixed Precision & fp16 \\
Resolution & 512 \\
Batch Size & 16 \\
Grad. Accumulation & 2 \\
Gradient Checkpointing & True \\
8-bit Adam & True \\
LoRA Rank & 128 \\
Learning Rate & $5\times10^{-5}$ \\
LR Scheduler & constant \\
Warmup Steps & 0 \\
Max Training Steps & 303 \\
Checkpointing Steps & 50 \\
$\eta_{\mathrm{DPO}}$ & 0.001 \\
% $\alpha_{\mathrm{DMPO}}$ & 0.01 \\
Training Method & method-dependent \\
Seed & 0 \\
\hline
\end{tabular}
\caption{SDXL-Base configuration}
\end{subtable}

\caption{Training-script configurations for SD v1.5 and SDXL-Base. 
The same training implementation supports multiple preference-optimization
algorithms; therefore, some method-specific hyperparameters are active only
for the corresponding training objective.}
\label{tab:params-for-training}
\end{table}

\section{Perturbation Categorization}
\label{sec:perturbations_desc}

We use a total of 15 controlled perceptual perturbations, organized into
three categories according to the dominant property of the image that they
modify: \emph{appearance}, \emph{spatial structure}, or \emph{visual
fidelity}. This categorization provides an interpretable decomposition of
the local variations explored around each preferred anchor and enables us
to separately analyze the contribution of qualitatively different sources
of perceptual variation.

\textbf{\textit{Photometric}} perturbations primarily modify the
appearance statistics of an image while leaving its spatial coordinate
structure largely unchanged. In other words, the locations and shapes of
objects are preserved, while pixel intensities, colors, contrast, or
channel relationships are altered. This family therefore captures
variations in how the same underlying scene is visually rendered. It
includes Gaussian noise, salt-and-pepper noise, posterization, channel
manipulation, and color jitter.

\textbf{\textit{Geometric}} perturbations instead act primarily on the
spatial mapping of image content. They modify pixel coordinates and can
therefore alter local shape, orientation, scale, or spatial arrangement
while largely preserving the visual appearance of the underlying content.
This family captures sensitivity to structural and spatial distortions and
includes shear, elastic deformation, swirl, sine-wave warping, twist
deformation, and radial zoom.

\textbf{\textit{Fidelity Reduction}} perturbations primarily reduce the
amount or precision of visual information available in the image. Rather
than changing only color statistics or spatial coordinates, these
operations suppress fine-scale detail, introduce lossy reconstruction, or
remove localized visual evidence. For example, Gaussian blur removes high-frequency
detail and pixelation reduces spatial resolution. This family therefore captures variations in the quality and completeness with which the underlying scene is represented.

The three families can consequently be viewed as perturbing complementary
axes of the preferred image: \emph{how it looks} (Photometric),
\emph{where and in what spatial form its content appears} (Geometric), and
\emph{how much visual information is retained} (Fidelity Reduction).
Although individual operators may affect more than one property, we assign
each perturbation according to its dominant perceptual effect. The
categorization is therefore intended as an operational and interpretable
taxonomy rather than a claim that the three families are mathematically
disjoint.

\begin{figure*}[!tp]
    \centering
    \begin{minipage}{\linewidth}
\captionsetup{type=table}
\centering
\footnotesize
\setlength{\tabcolsep}{4pt}
\renewcommand{\arraystretch}{1.1}
\begin{tabularx}{\linewidth}{p{0.19\linewidth} X p{0.27\linewidth}}
\toprule
\textbf{Perturbation} & \textbf{Description} & \textbf{Hyperparameters} \\
\midrule
% \multicolumn{3}{c}{\cellcolor{grey!20}\textbf{Global Perturbations}}\\
% \midrule
Gaussian Blur & Smooths the image using a random Gaussian kernel, simulating loss of fine detail. & Kernel size: odd $\in$ [3, 13] \\
\hline
Gaussian Noise & Adds mild random Gaussian noise to mimic texture artifacts or sensor-like noise. & Std. dev. $\sim \mathcal{U}(5, 40)$ \\
\hline
Salt-and-Pepper Noise & Introduces sparse white and black pixels, imitating binary pixel errors. & Amount $\sim \mathcal{U}(0.002, 0.05)$ \\
\hline
Channel Swapping & Randomly permutes or drops RGB channels, introducing unnatural color compositions. & Action $\in$ \{swap, drop, swap\_bw\} \\
\hline
Shearing & Applies affine shear to simulate mild geometric warping in horizontal and vertical directions. & Shear factor in X and Y $\in \mathcal{U}(-0.25, 0.25)$\\
\hline
Posterization & Reduces color depth per channel, producing flat, banded color regions. & Bits $\in$ [1, 6] per channel \\
\hline
Elastic Transformation & Applies smoothed random displacement fields, causing flexible geometric deformations. & $\alpha \in [30, 80]$, $\sigma \propto$ image size \\
\hline
JPEG Compression & Emulates medium-quality JPEG artifacts due to lossy encoding and blocking. & Quality $\in$ [1, 40] \\
\hline
Color Jitter & Randomly alters brightness and contrast, simulating lighting or post-processing inconsistencies. & Contrast $\alpha \sim \mathcal{U}(0.8, 1.6)$, Brightness $\beta \sim \mathcal{U}(-20, 20)$ \\
\hline
Pixelation & Downsamples and upsamples random regions to simulate blocky, low-resolution textures. & Block size: $10\%\sim30\%$, Pixel size $\in$ [4, 20] \\
\hline
Erase with Inpainting & Random regions are masked and filled using nearby content, mimicking over-smooth reconstruction. & Num regions $\in$ [1, 3], Shape $\in$ \{circle, rectangle\} \\
\hline
Swirl Perturbation & Applies local rotational warping around a random center, resembling twisted spatial patterns. & Strength $\in [10, 20]$, Radius $\in [100, 300]$ \\
\hline
Twist (Polar Rotation) & Applies center-weighted angular perturbation, creating spiraling or warped structures. & Strength = 5.0 (fixed) \\
\hline
Radial Zoom & Simulates zoom-like radial expansion centered in the image, exaggerating scale and geometry locally. & Factor = 0.001 (fixed) \\
\hline
Sine Wave Perturbation & Displaces pixels using sinusoidal horizontal shifts, imitating spatial wobble or ripple artifacts. & Amplitude = 20, Wavelength = 50 \\
\bottomrule
\end{tabularx}

\caption{List of perturbations used to simulate alignable differences.}
\label{tab:global_perturbations}
\end{minipage}
    \par\vspace{8pt}
    % Perturbation-family colors
\definecolor{photometric}{RGB}{0,114,178}
\definecolor{geometric}{RGB}{213,94,0}
\definecolor{fidelity}{RGB}{0,135,105}
\begin{minipage}{\linewidth}
\captionsetup{type=figure}
\small
\centering

%---- First row ----%
\begin{subfigure}[t]{\linewidth}
    \centering
    \setlength{\tabcolsep}{1.5pt}
    \renewcommand{\arraystretch}{1.0}

    \begin{tabular}{>{\centering\arraybackslash}m{0.10cm}
                    *{8}{>{\centering\arraybackslash}m{1.6cm}}}

    % Fidelity Reduction
    & \scalebox{0.8}{{\fontsize{7}{2}\selectfont
      \makecell{\textcolor{fidelity}{\textbf{Gaussian}}\\[0.5pt]
                \textcolor{fidelity}{\textbf{Blur}}}}}

    % Photometric
    & \scalebox{0.8}{{\fontsize{7}{2}\selectfont
      \makecell{\textcolor{photometric}{\textbf{Gaussian}}\\[0.5pt]
                \textcolor{photometric}{\textbf{Noise}}}}}

    % Photometric
    & \scalebox{0.8}{{\fontsize{7}{2}\selectfont
      \makecell{\textcolor{photometric}{\textbf{Salt-and-}}\\[0.5pt]
                \textcolor{photometric}{\textbf{Pepper Noise}}}}}

    % Photometric
    & \scalebox{0.8}{{\fontsize{7}{2}\selectfont
      \makecell{\textcolor{photometric}{\textbf{Channel}}\\[0.5pt]
                \textcolor{photometric}{\textbf{Swapping}}}}}

    % Geometric
    & \scalebox{0.8}{{\fontsize{7}{2}\selectfont
      \makecell{\textcolor{geometric}{\textbf{Shearing}}}}}

    % Fidelity Reduction
    & \scalebox{0.8}{{\fontsize{7}{2}\selectfont
      \makecell{\textcolor{fidelity}{\textbf{JPEG}}\\[0.5pt]
                \textcolor{fidelity}{\textbf{Compression}}}}}

    % Photometric
    & \scalebox{0.8}{{\fontsize{7}{2}\selectfont
      \makecell{\textcolor{photometric}{\textbf{Posterization}}}}}

    % Geometric
    & \scalebox{0.8}{{\fontsize{7}{2}\selectfont
      \makecell{\textcolor{geometric}{\textbf{Elastic}}\\[0.5pt]
                \textcolor{geometric}{\textbf{Transformation}}}}} \\

    \makebox[0pt][r]{\rotatebox{90}{\fontsize{7}{7}\textit{Original}}}
    & \includegraphics[width=\linewidth]{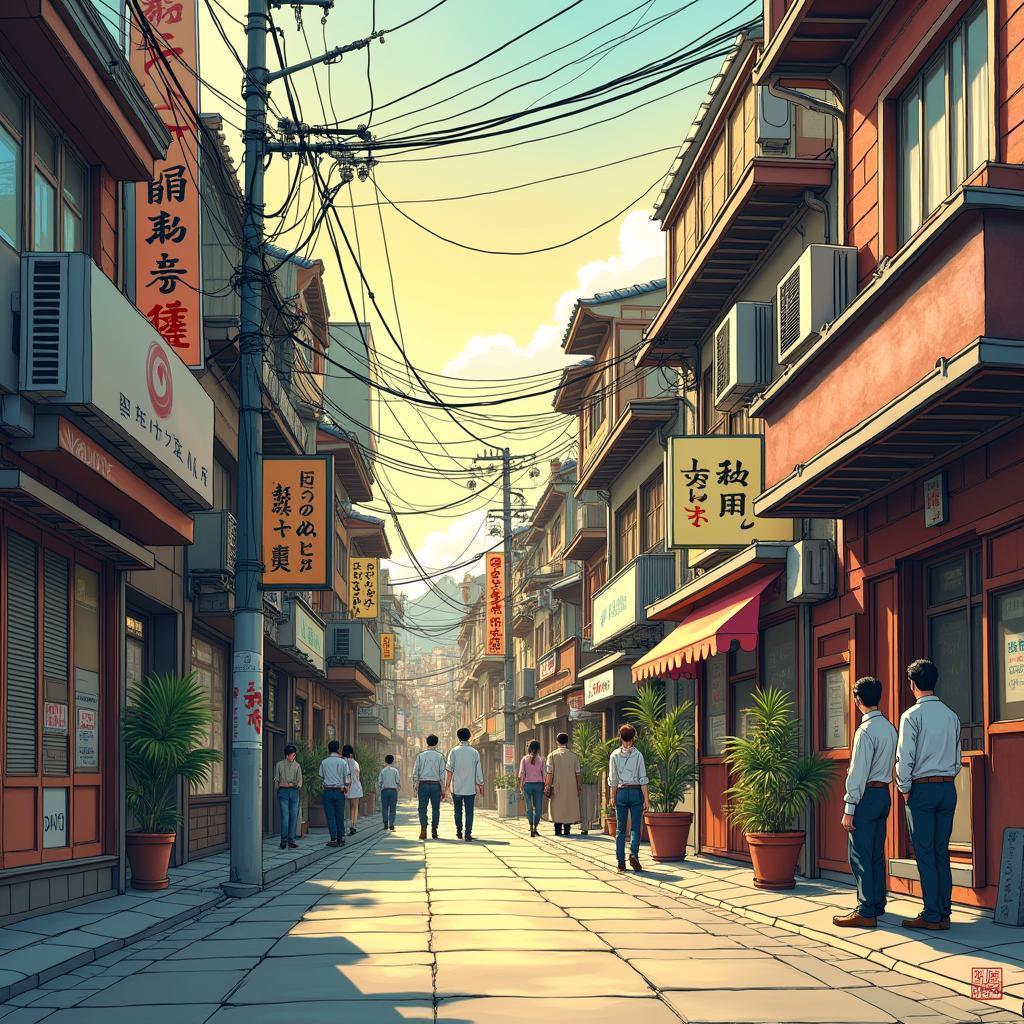}
    & \includegraphics[width=\linewidth]{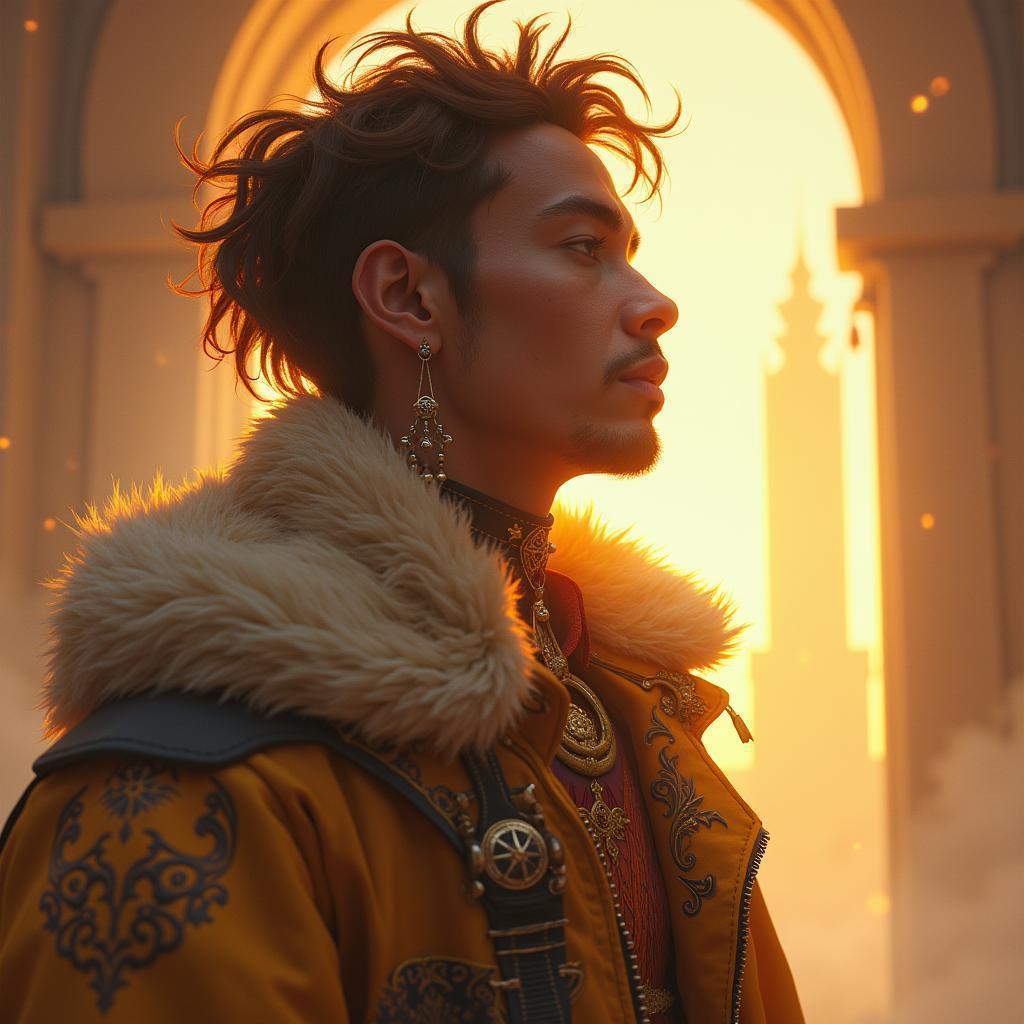}
    & \includegraphics[width=\linewidth]{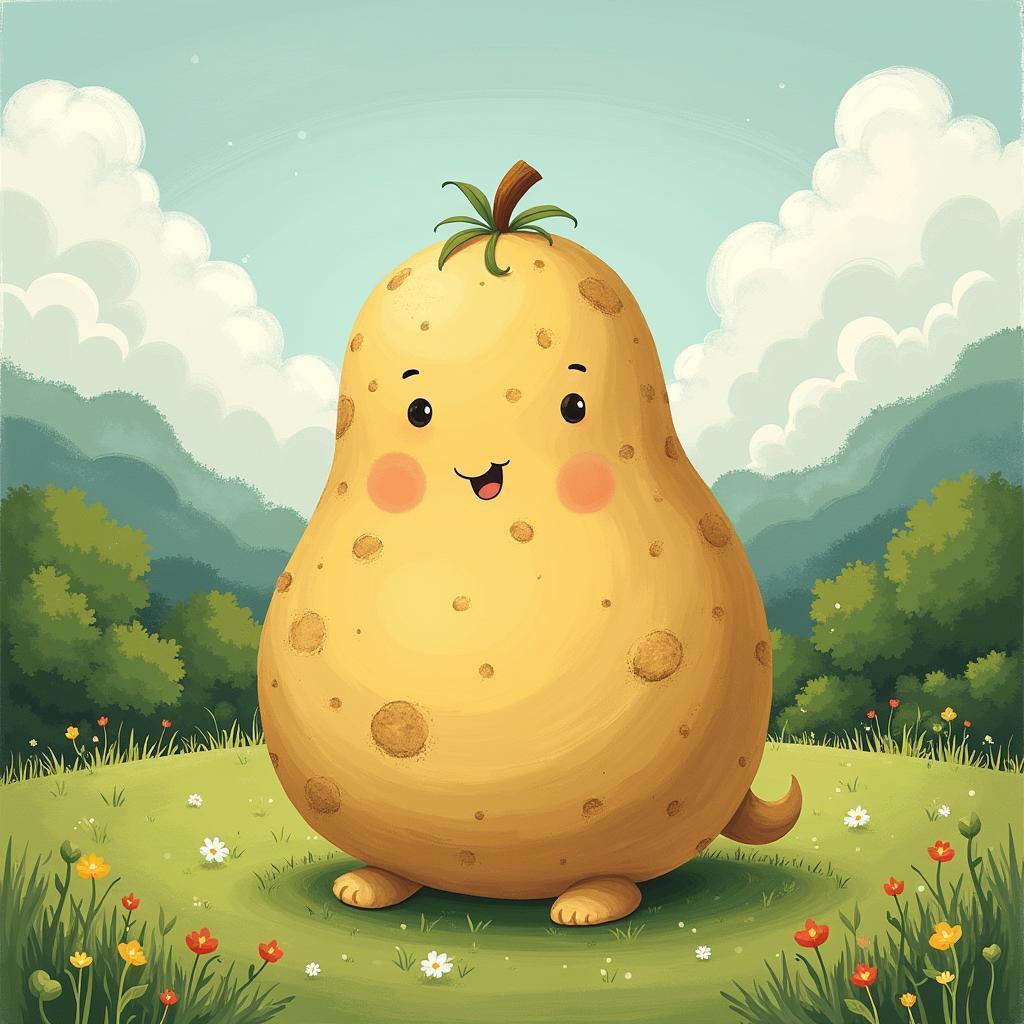}
    & \includegraphics[width=\linewidth]{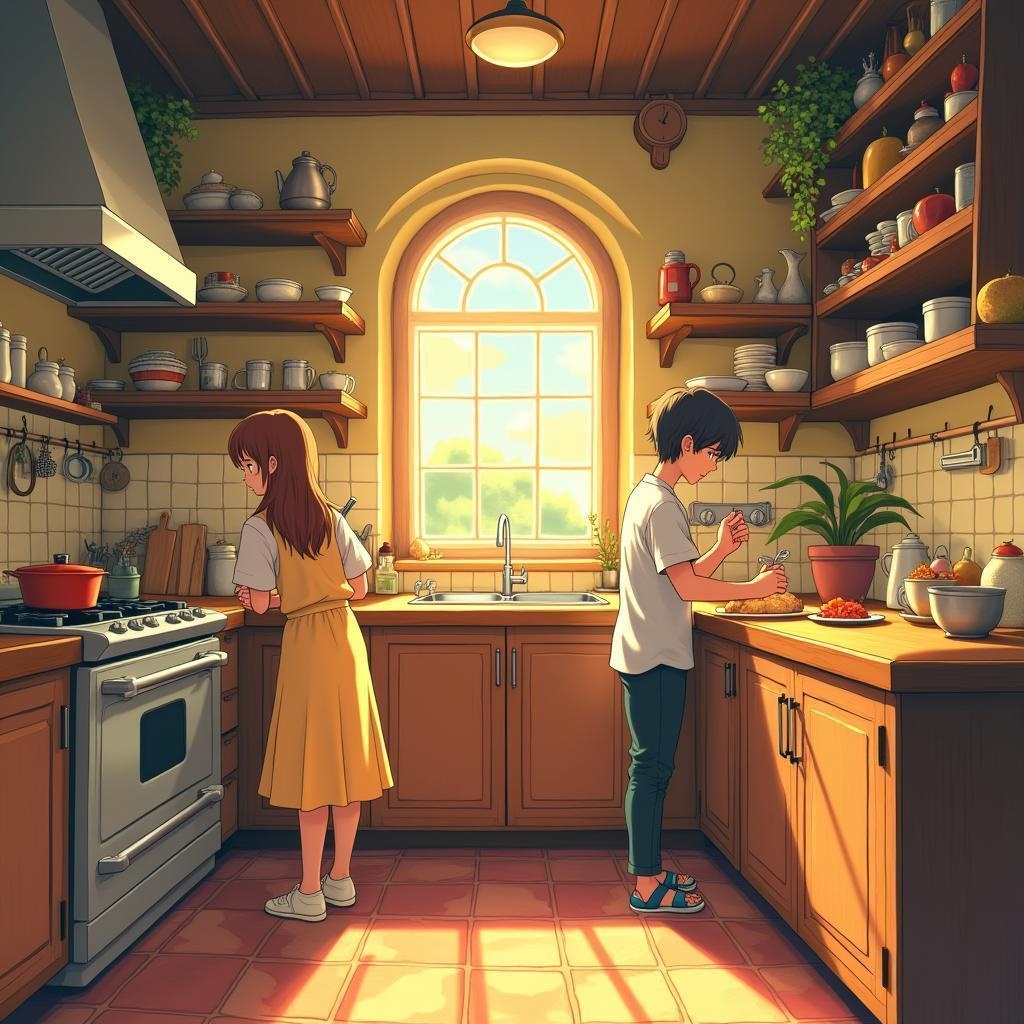}
    & \includegraphics[width=\linewidth]{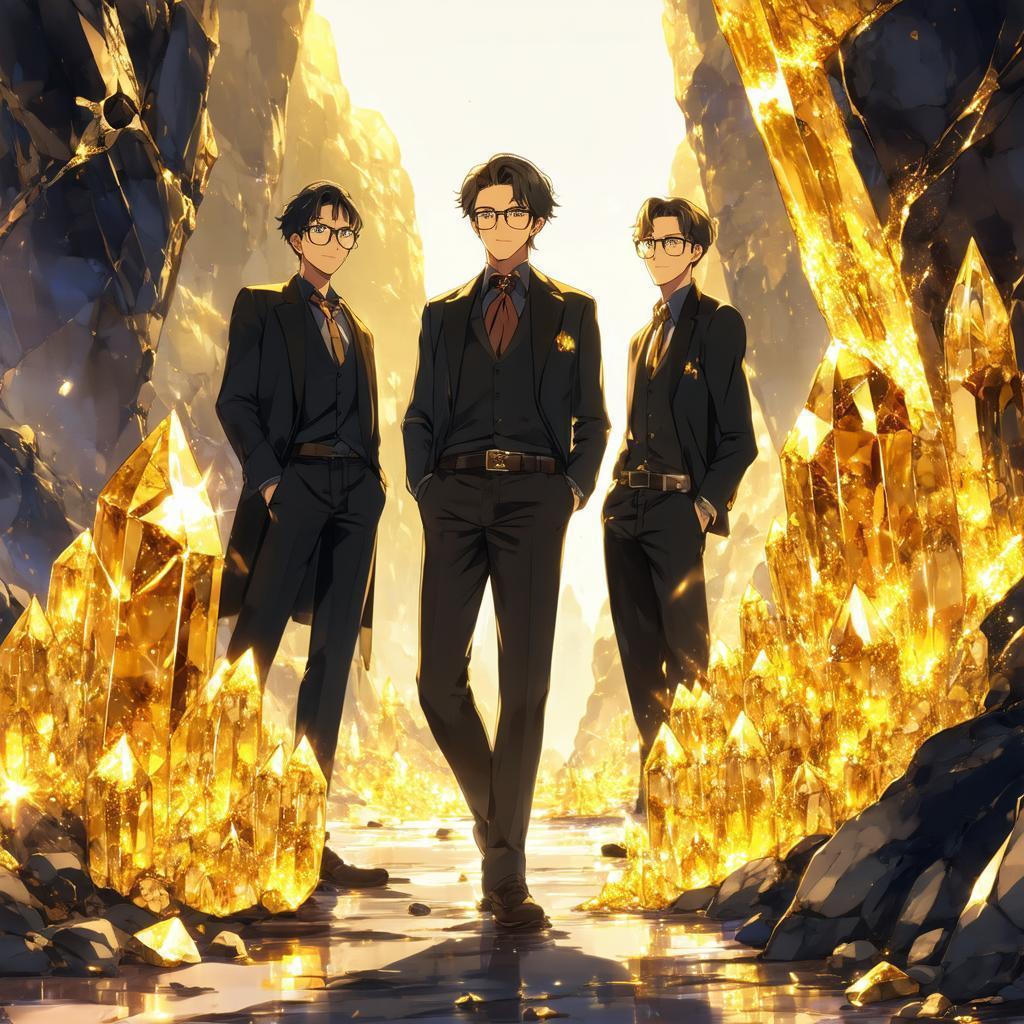}
    & \includegraphics[width=\linewidth]{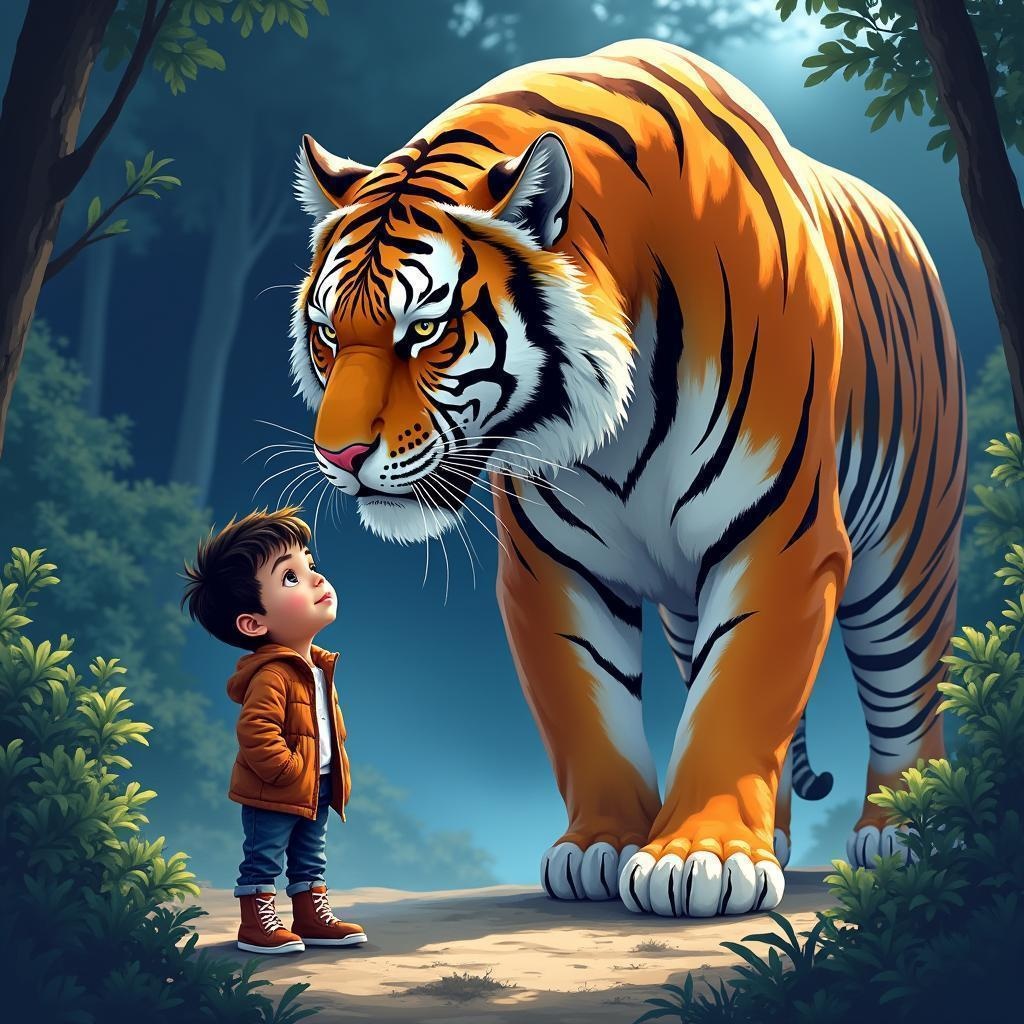}
    & \includegraphics[width=\linewidth]{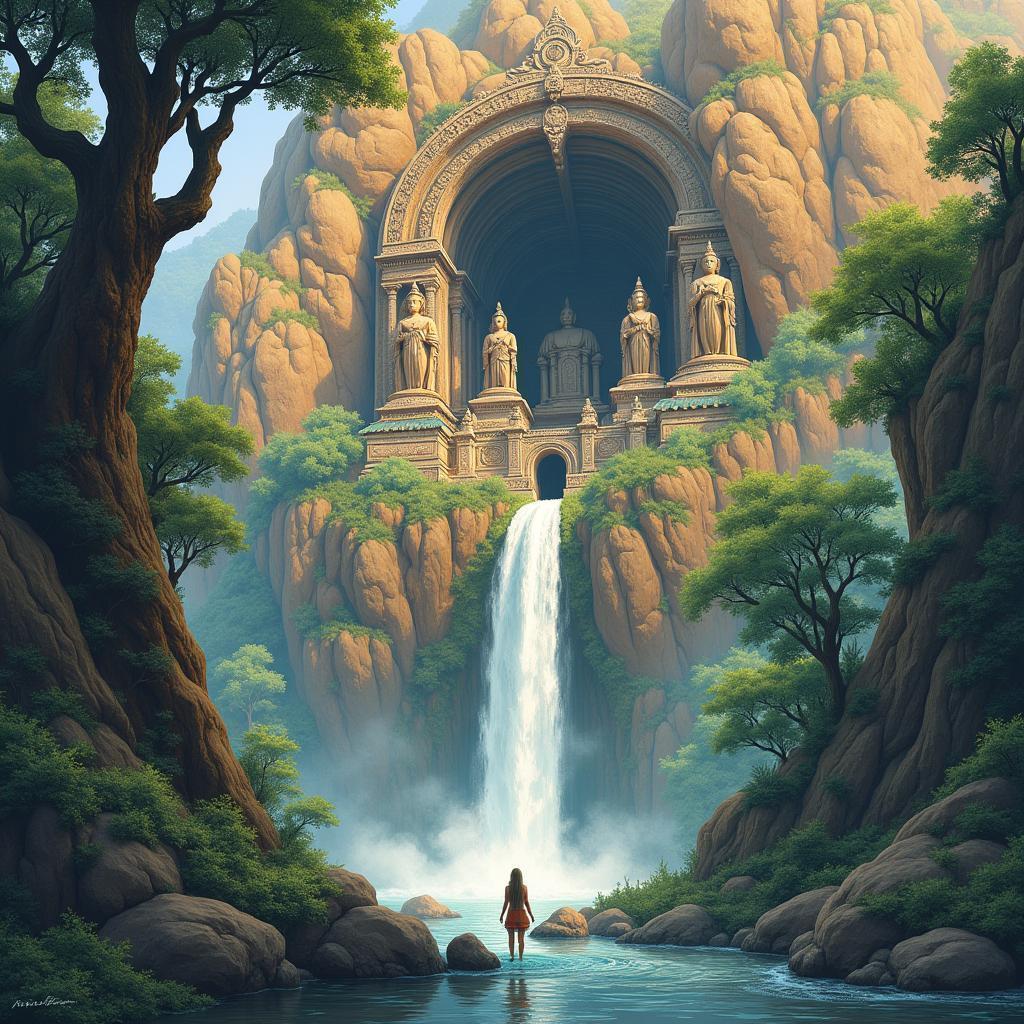}
    & \includegraphics[width=\linewidth]{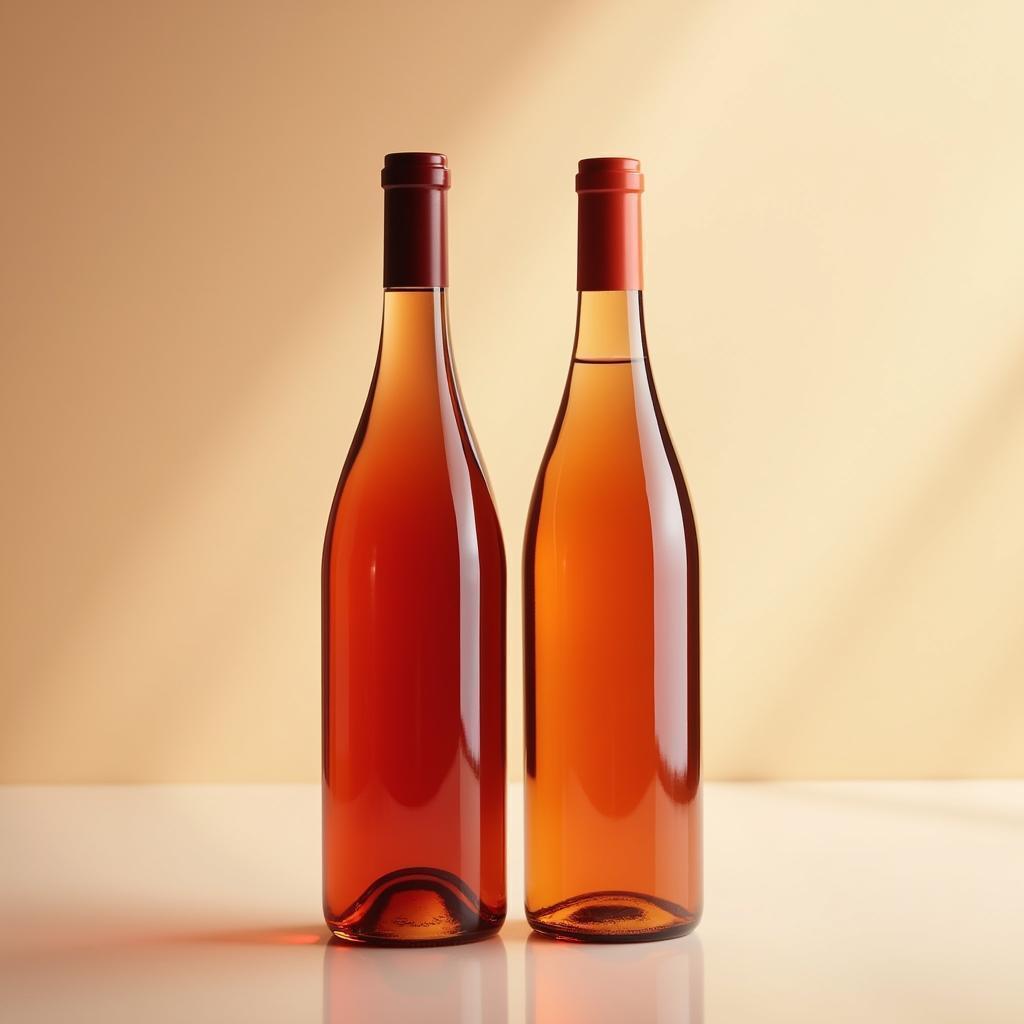} \\[-0.3em]

    \makebox[0pt][r]{\rotatebox{90}{\fontsize{7}{7}\textit{Perturbed}}}
    & \includegraphics[width=\linewidth]{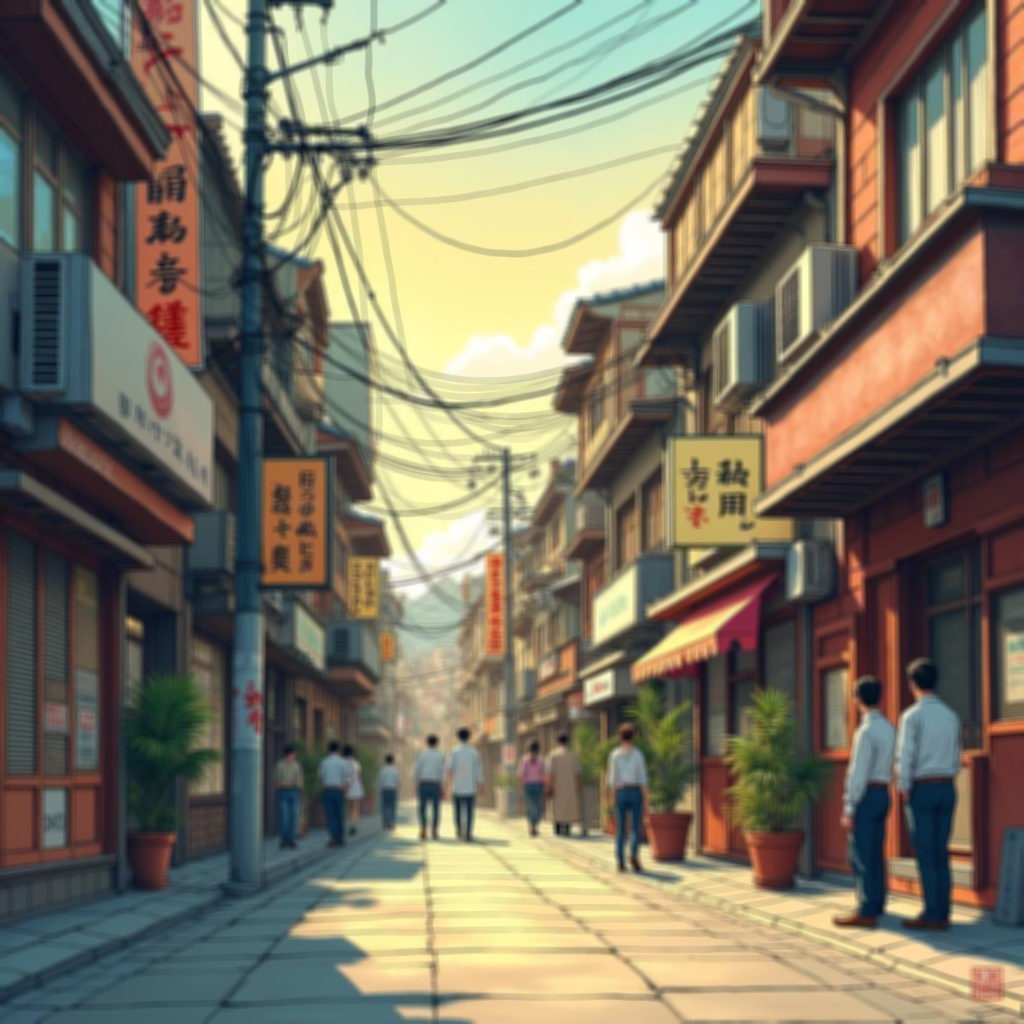}
    & \includegraphics[width=\linewidth]{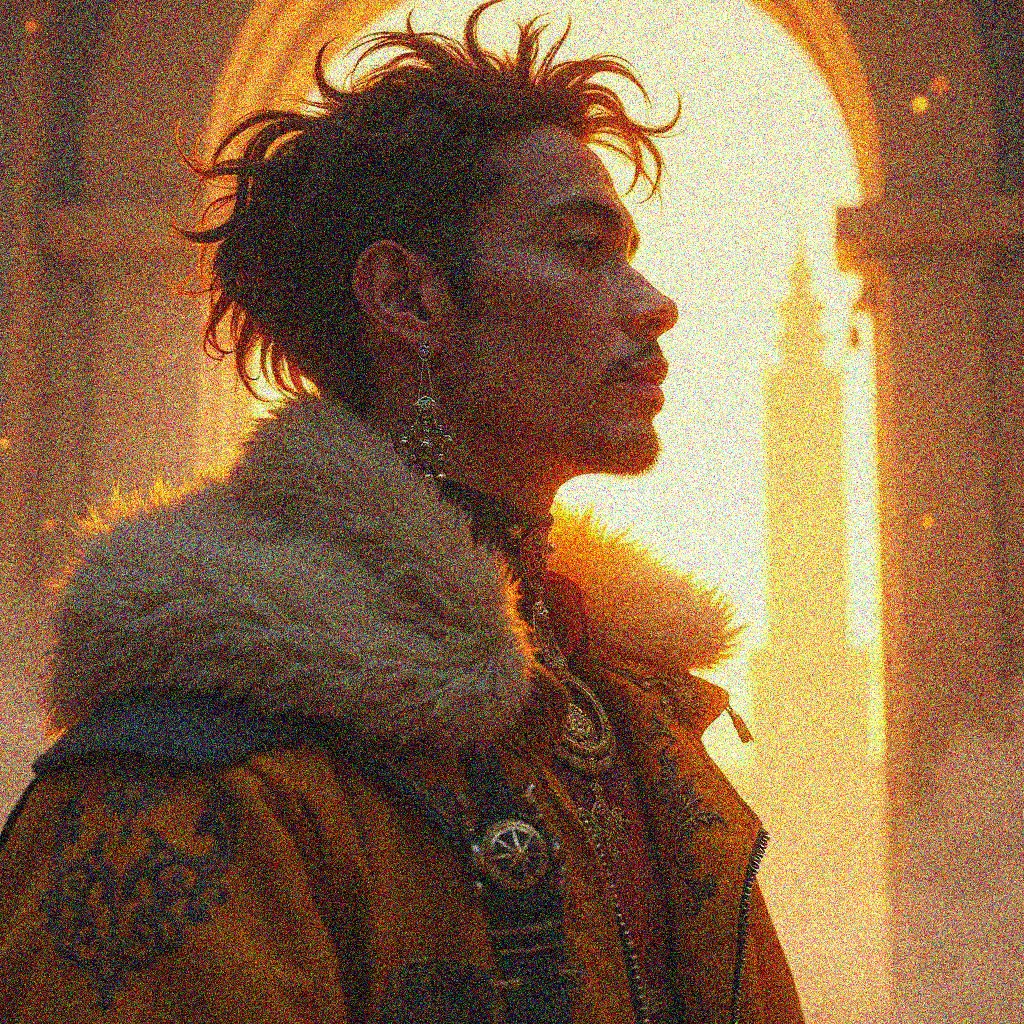}
    & \includegraphics[width=\linewidth]{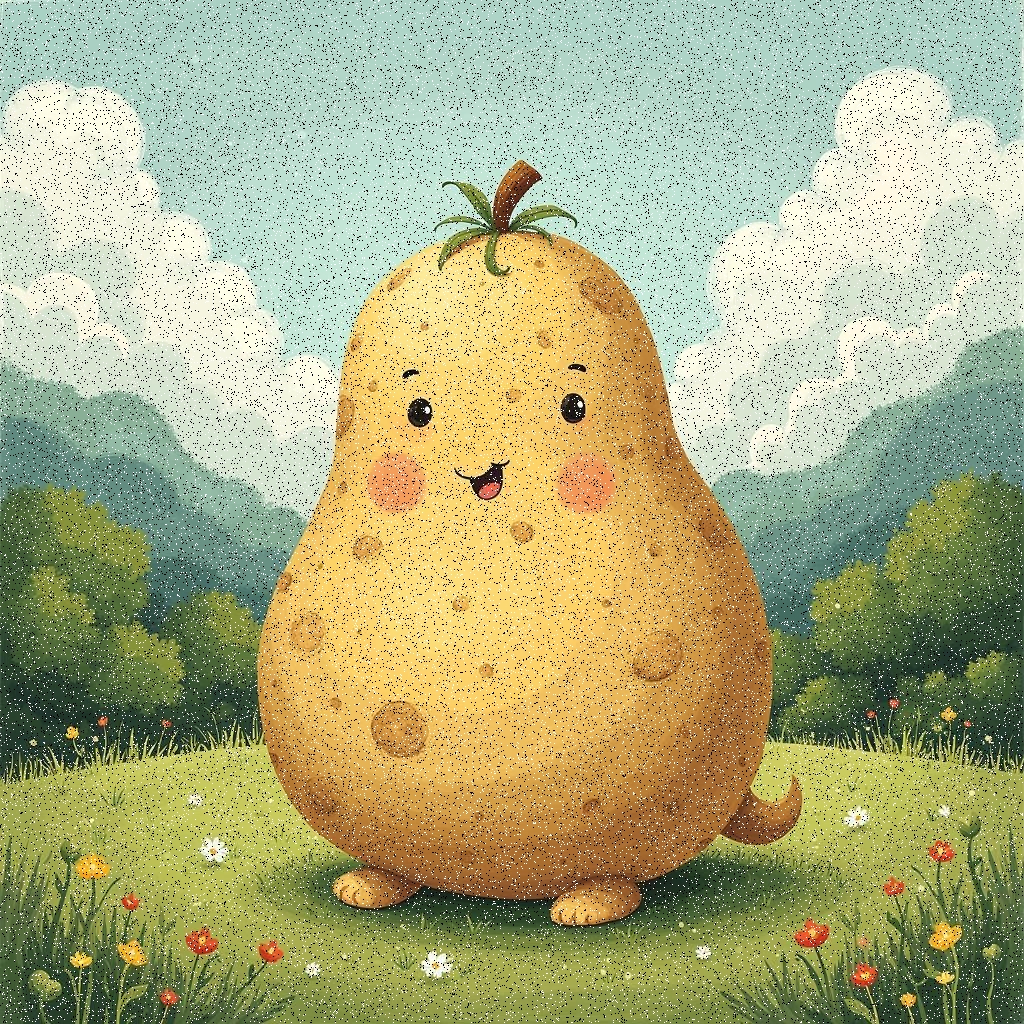}
    & \includegraphics[width=\linewidth]{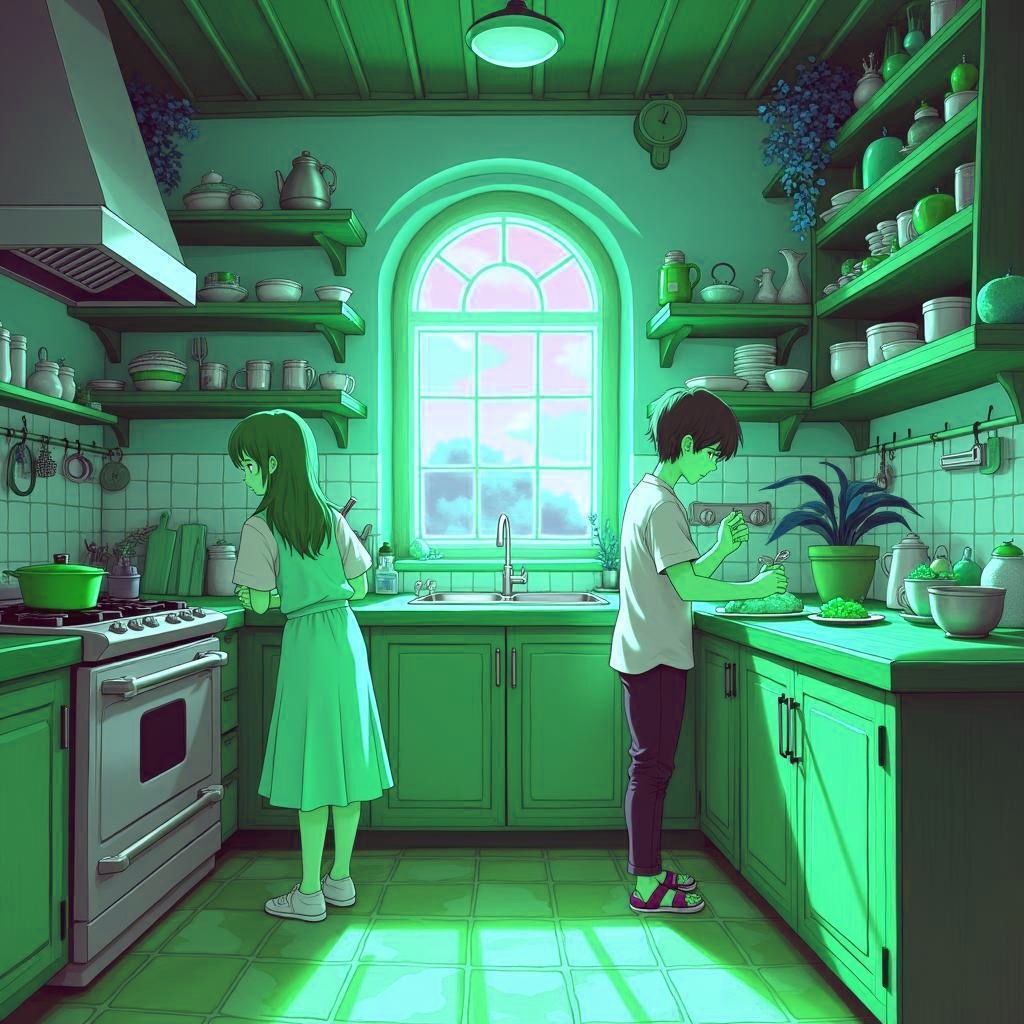}
    & \includegraphics[width=\linewidth]{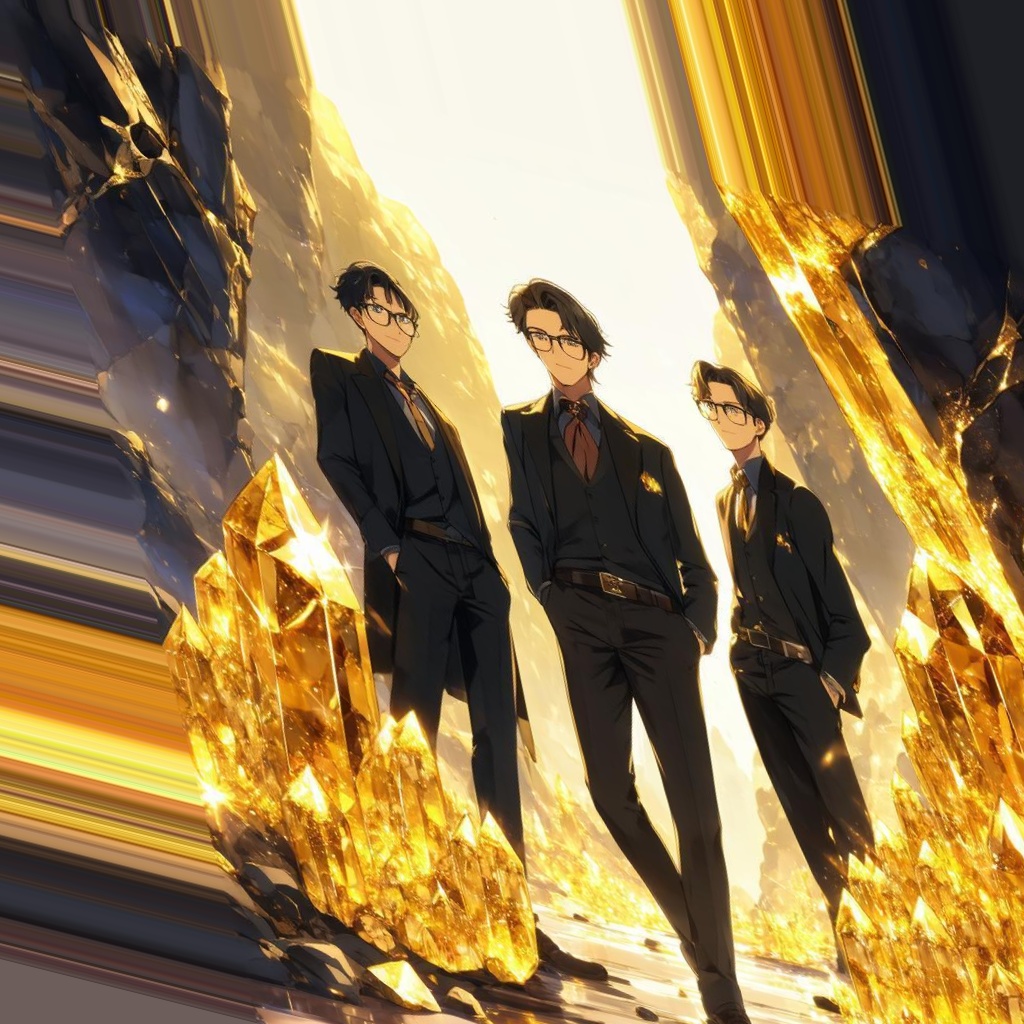}
    & \includegraphics[width=\linewidth]{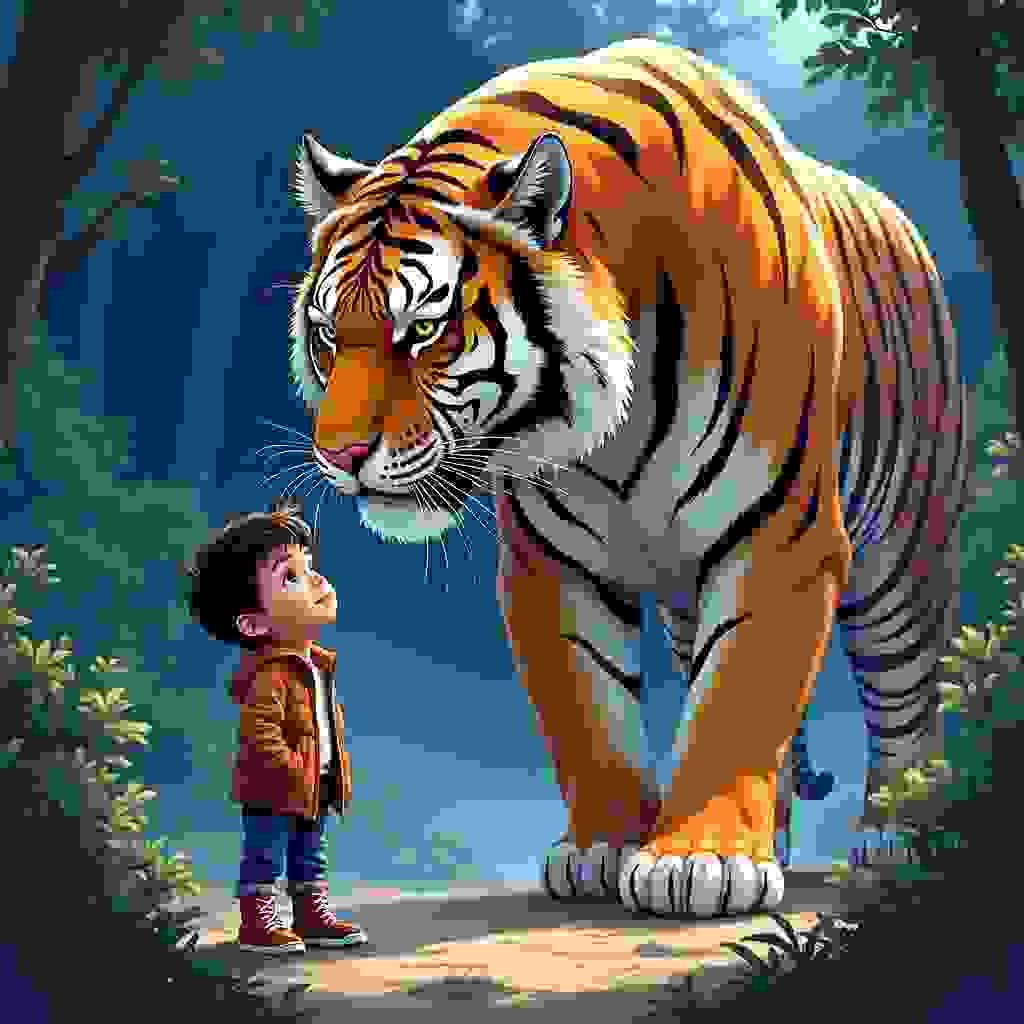}
    & \includegraphics[width=\linewidth]{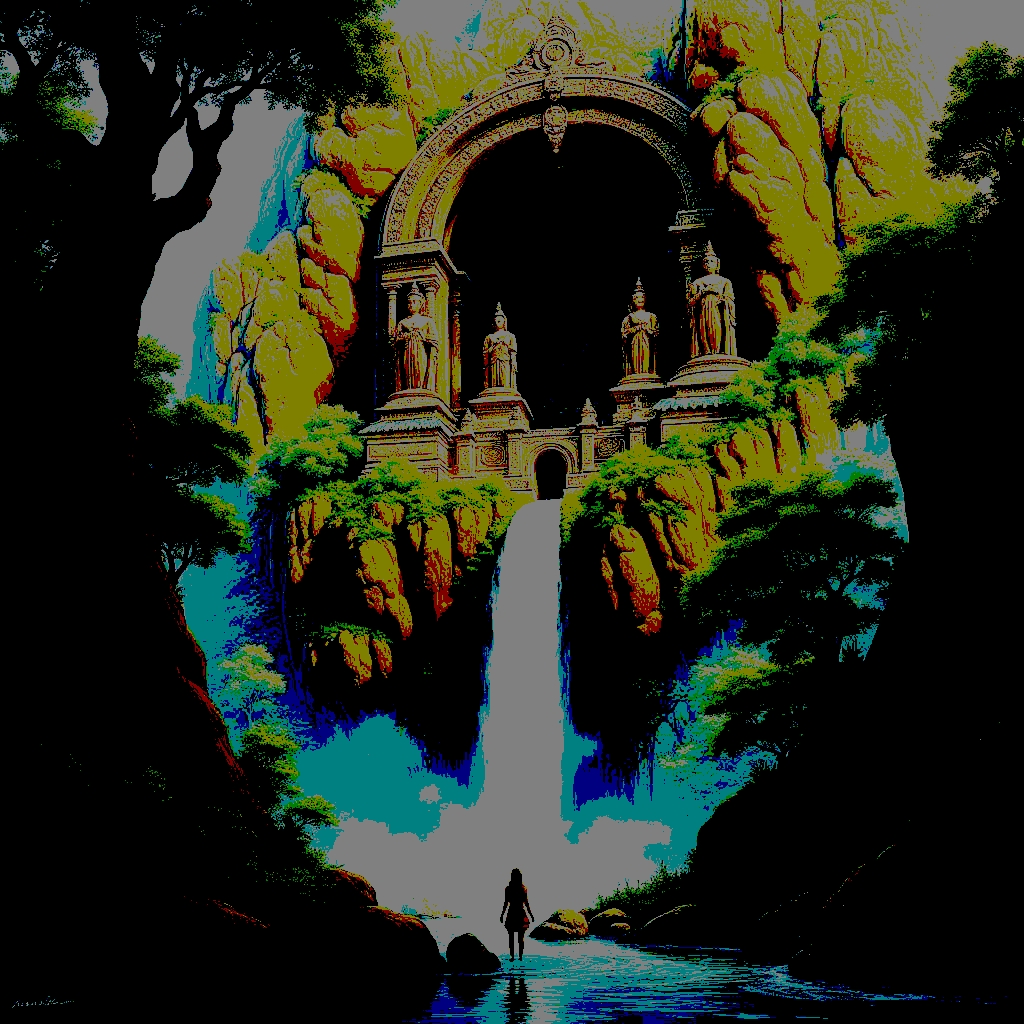}
    & \includegraphics[width=\linewidth]{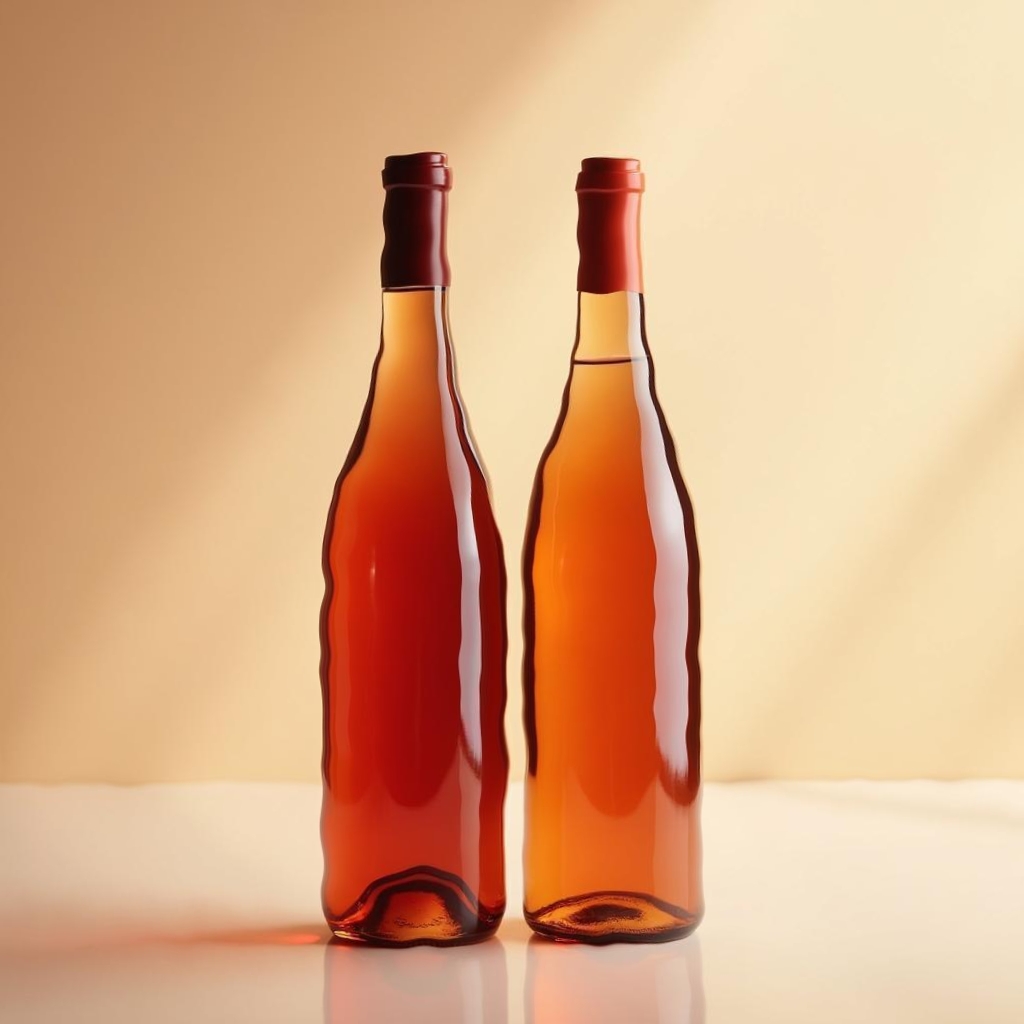}
    \end{tabular}
\end{subfigure}

\vspace{4pt}

%---- Second row ----%
\centering
\setlength{\tabcolsep}{10pt}

\begin{subfigure}[t]{\linewidth}
\centering
\setlength{\tabcolsep}{2pt}
\renewcommand{\arraystretch}{1.3}

\begin{tabular}[t]{>{\centering\arraybackslash}m{0.10cm}
                *{7}{>{\centering\arraybackslash}m{1.6cm}}}

% Geometric
& \scalebox{0.8}{{\fontsize{7}{2}\selectfont
  \makecell{\textcolor{geometric}{\textbf{Swirl}}\\[0.5pt]
            \textcolor{geometric}{\textbf{Distortion}}}}}

% Geometric
& \scalebox{0.8}{{\fontsize{7}{2}\selectfont
  \makecell{\textcolor{geometric}{\textbf{Twist (Polar}}\\[0.5pt]
            \textcolor{geometric}{\textbf{Rotation)}}}}}

% Geometric
& \scalebox{0.8}{{\fontsize{7}{2}\selectfont
  \makecell{\textcolor{geometric}{\textbf{Radial}}\\[0.5pt]
            \textcolor{geometric}{\textbf{Zoom}}}}}

% Geometric
& \scalebox{0.8}{{\fontsize{7}{2}\selectfont
  \makecell{\textcolor{geometric}{\textbf{Sine Wave}}\\[0.5pt]
            \textcolor{geometric}{\textbf{Distortion}}}}}

% Fidelity Reduction
& \scalebox{0.8}{{\fontsize{7}{2}\selectfont
  \makecell{\textcolor{fidelity}{\textbf{Pixelation}}}}}

% Photometric
& \scalebox{0.8}{{\fontsize{7}{2}\selectfont
  \makecell{\textcolor{photometric}{\textbf{Color}}\\[0.5pt]
            \textcolor{photometric}{\textbf{Jitter}}}}}

% Fidelity Reduction
& \scalebox{0.8}{{\fontsize{7}{2}\selectfont
  \makecell{\textcolor{fidelity}{\textbf{Erase with}}\\[0.5pt]
            \textcolor{fidelity}{\textbf{Inpainting}}}}} \\

\makebox[0pt][r]{\rotatebox{90}{\fontsize{7}{7}\textit{Original}}}
& \includegraphics[width=\linewidth]{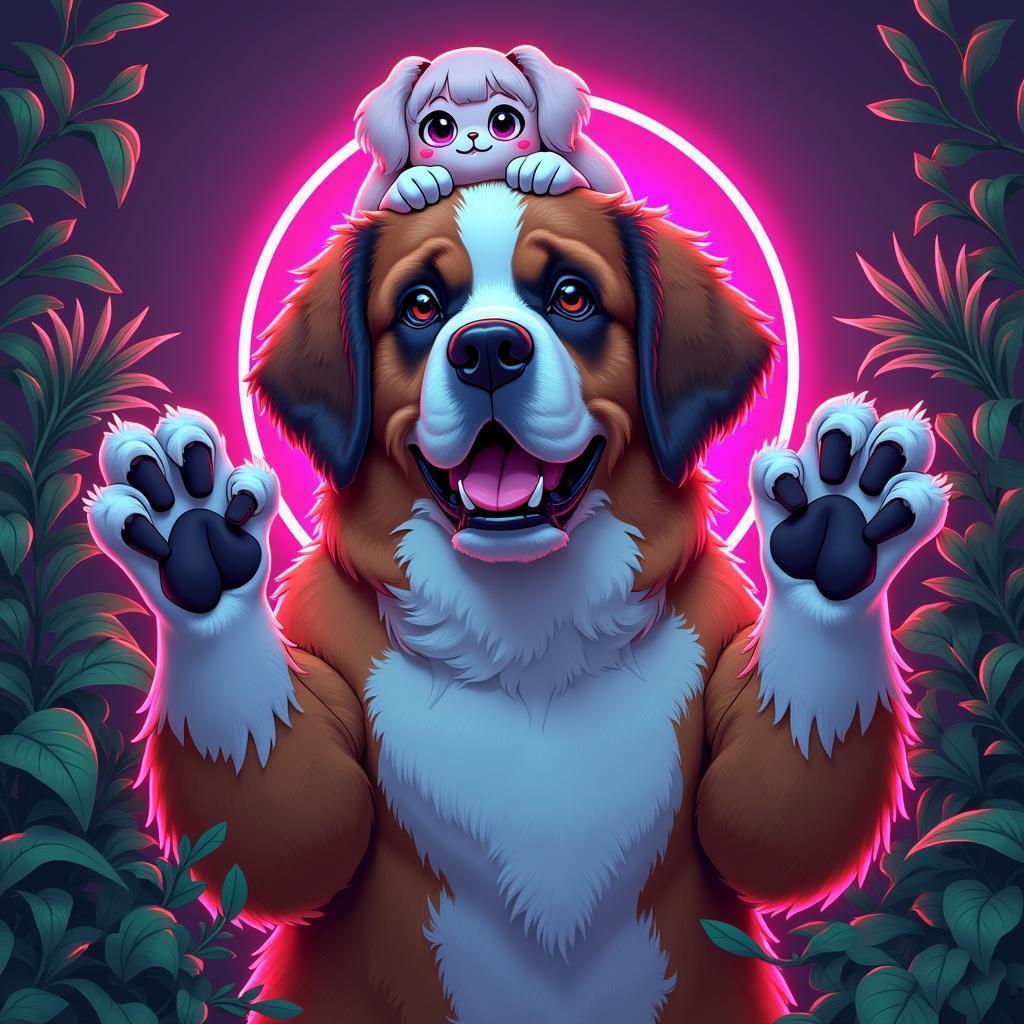}
& \includegraphics[width=\linewidth]{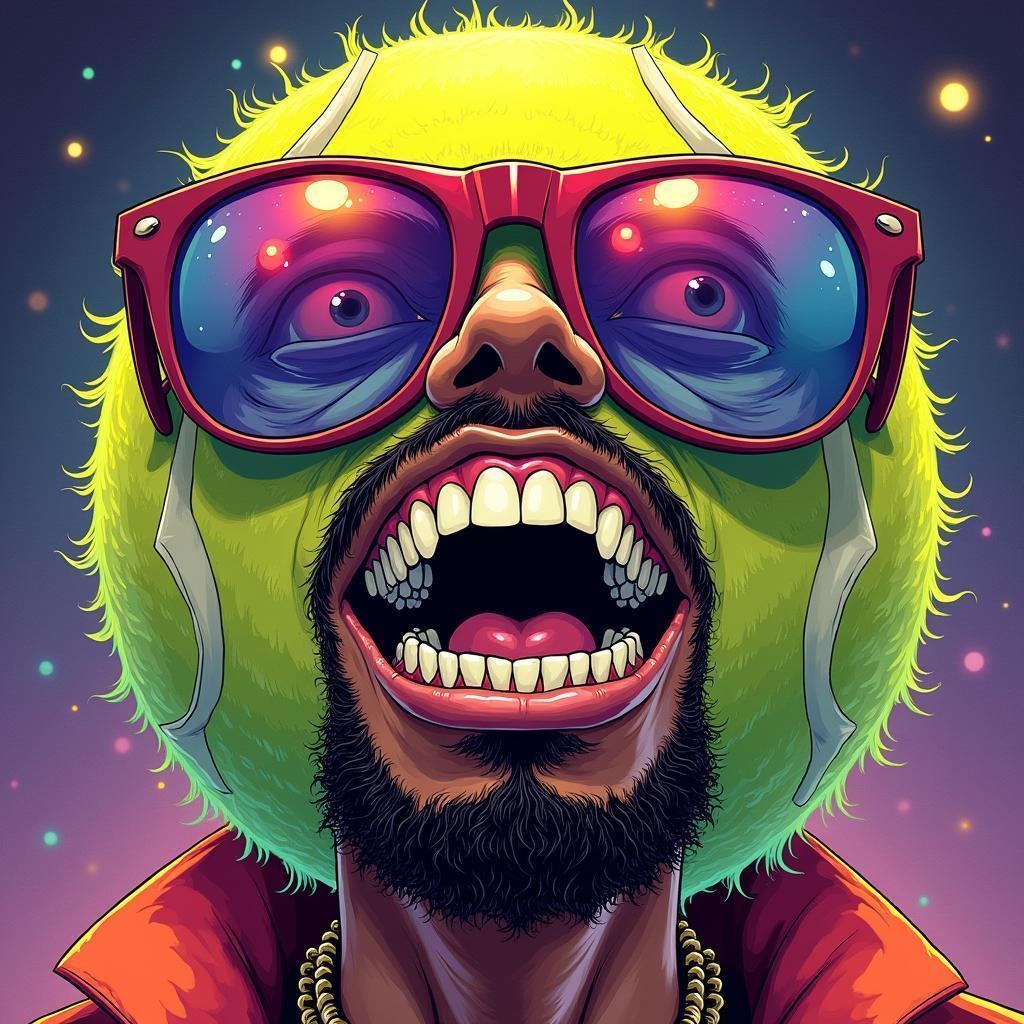}
& \includegraphics[width=\linewidth]{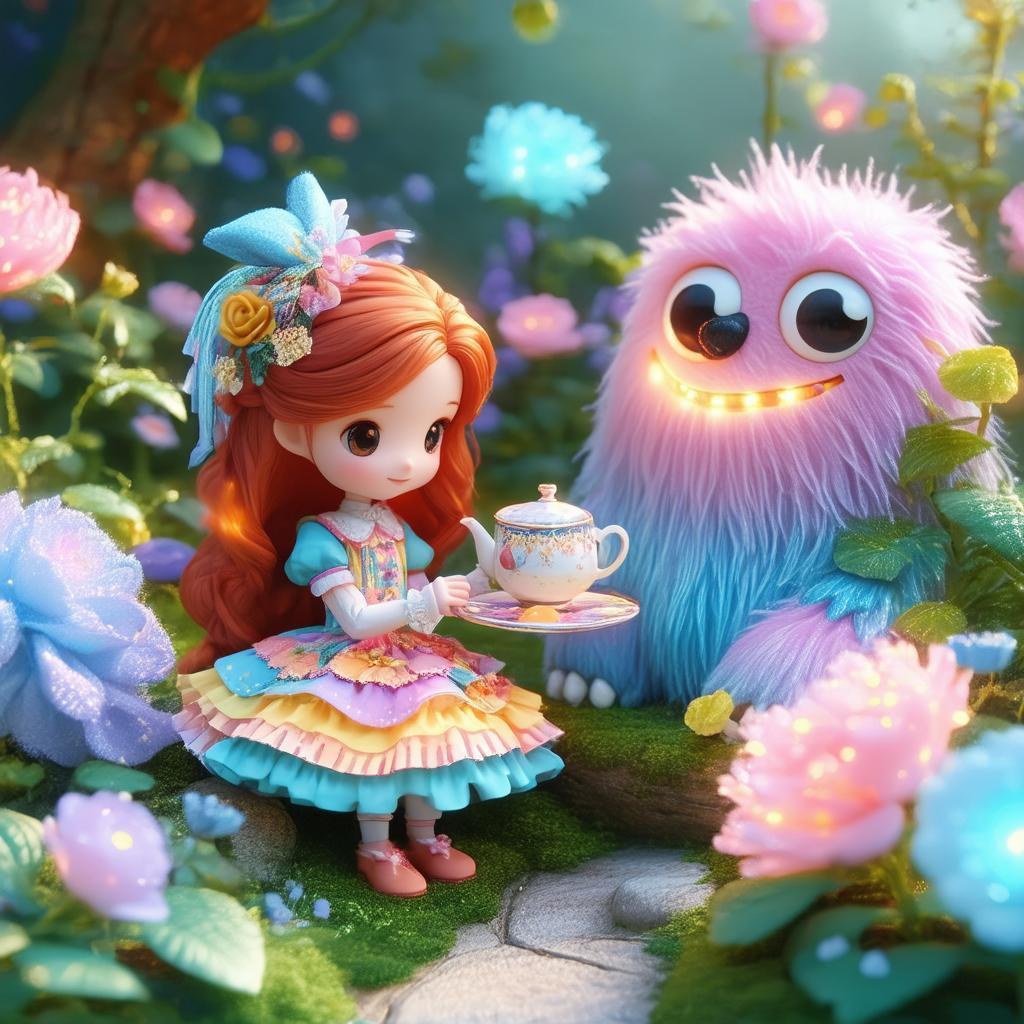}
& \includegraphics[width=\linewidth]{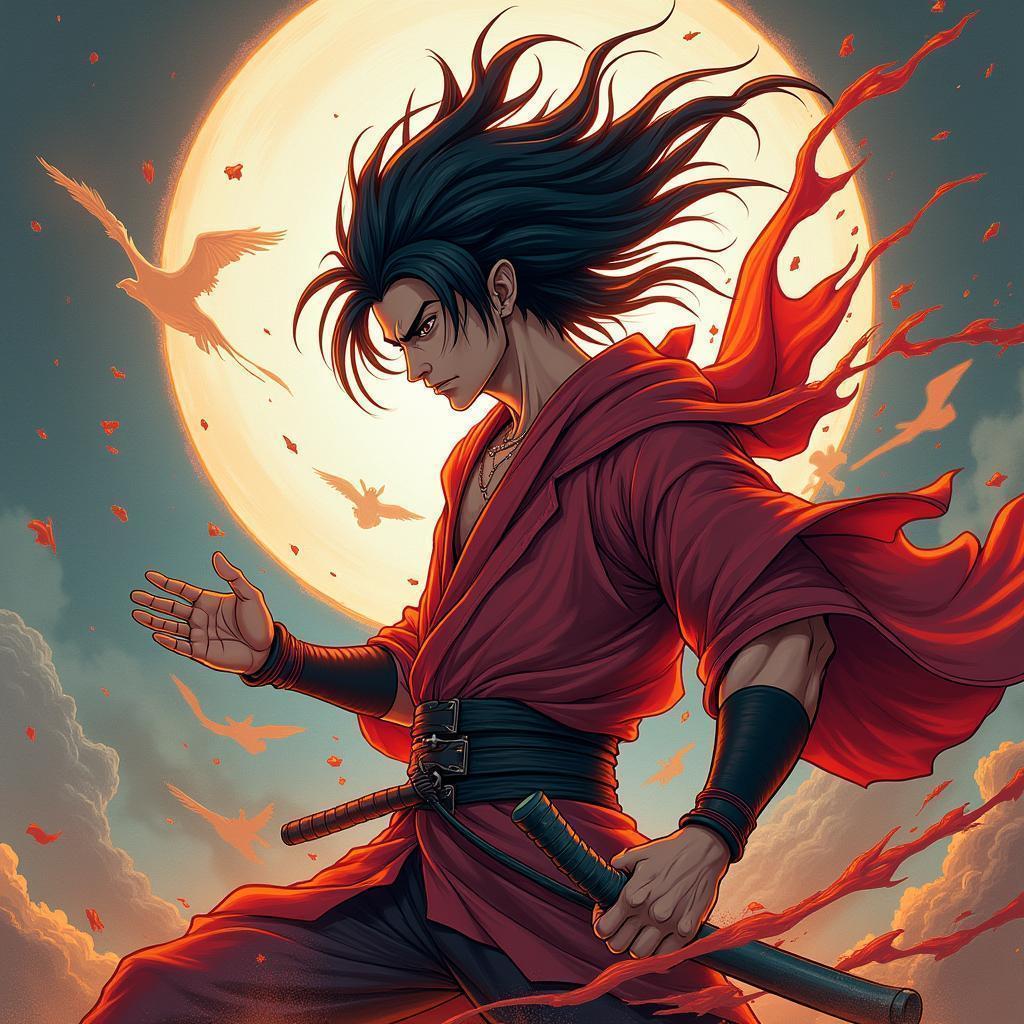} 
& \includegraphics[width=\linewidth]{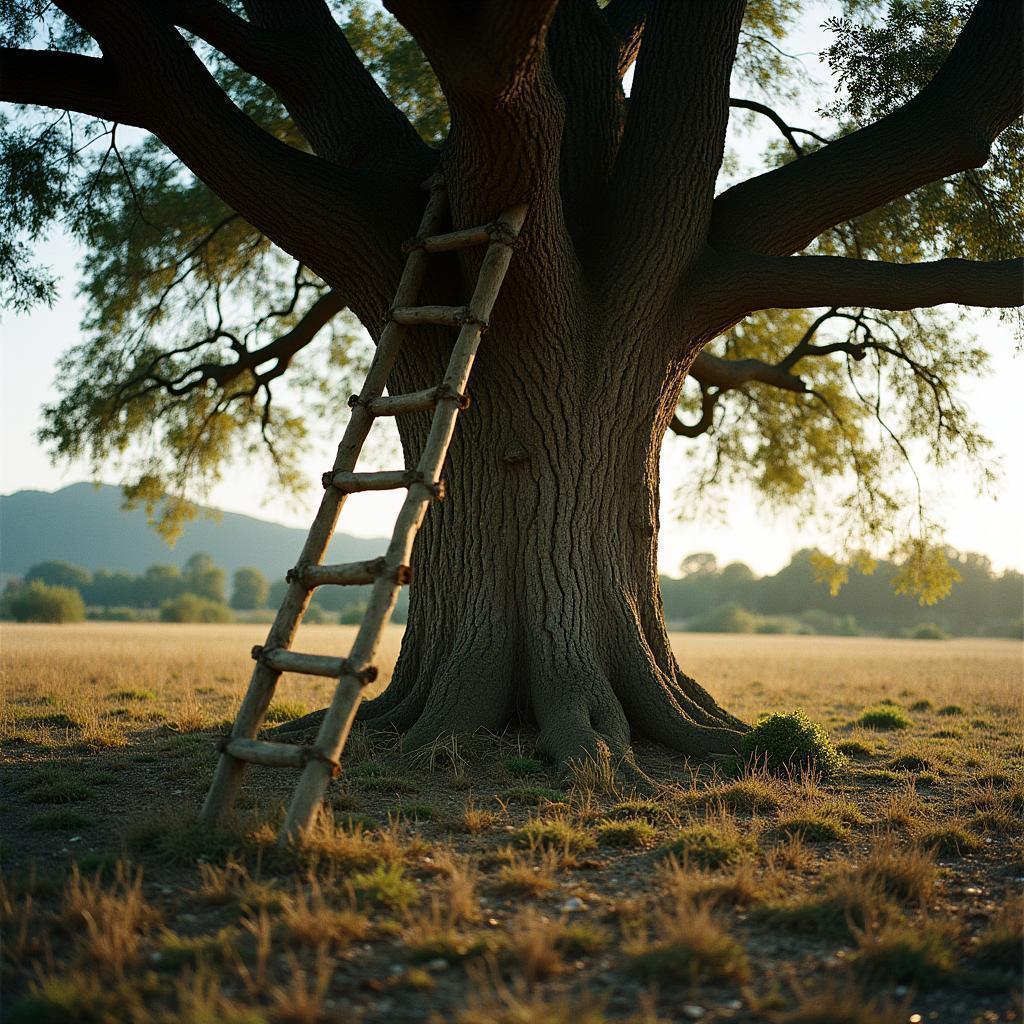}
& \includegraphics[width=\linewidth]{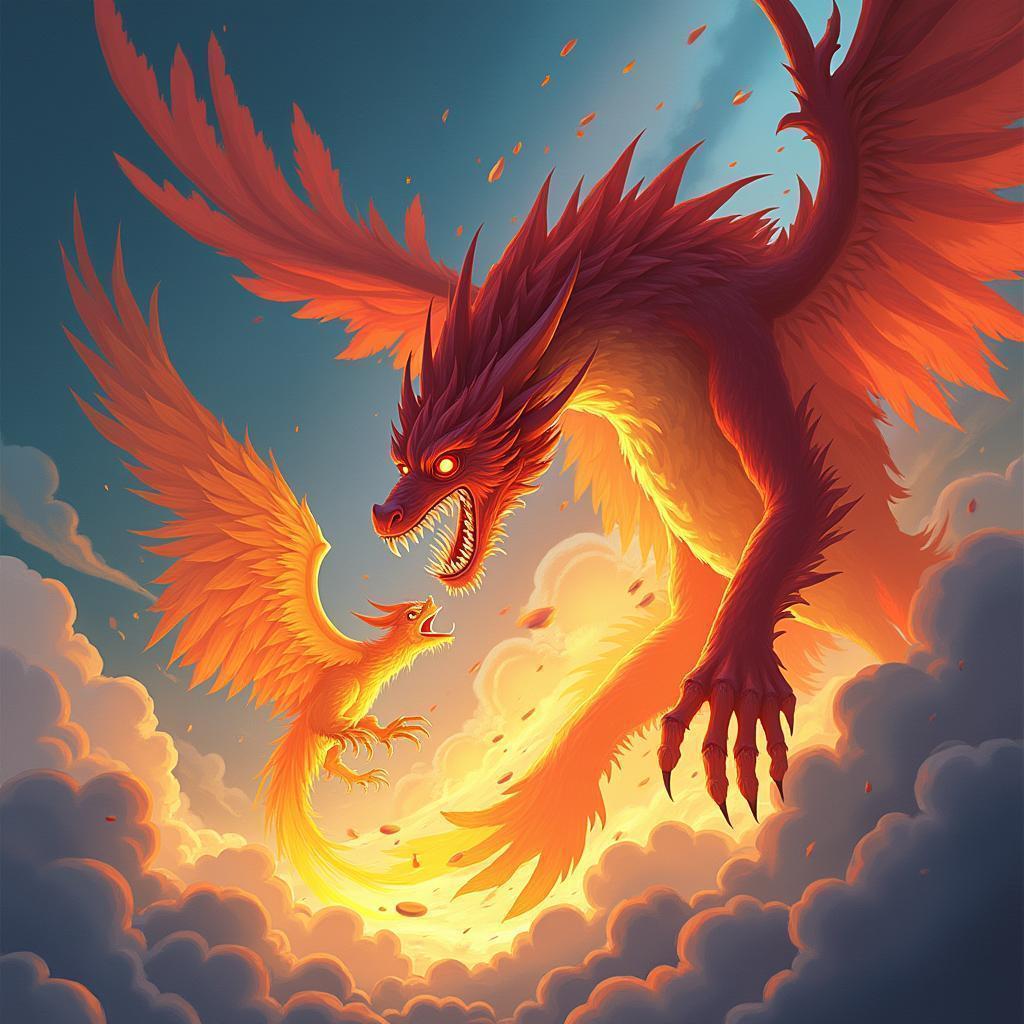}
& \includegraphics[width=\linewidth]{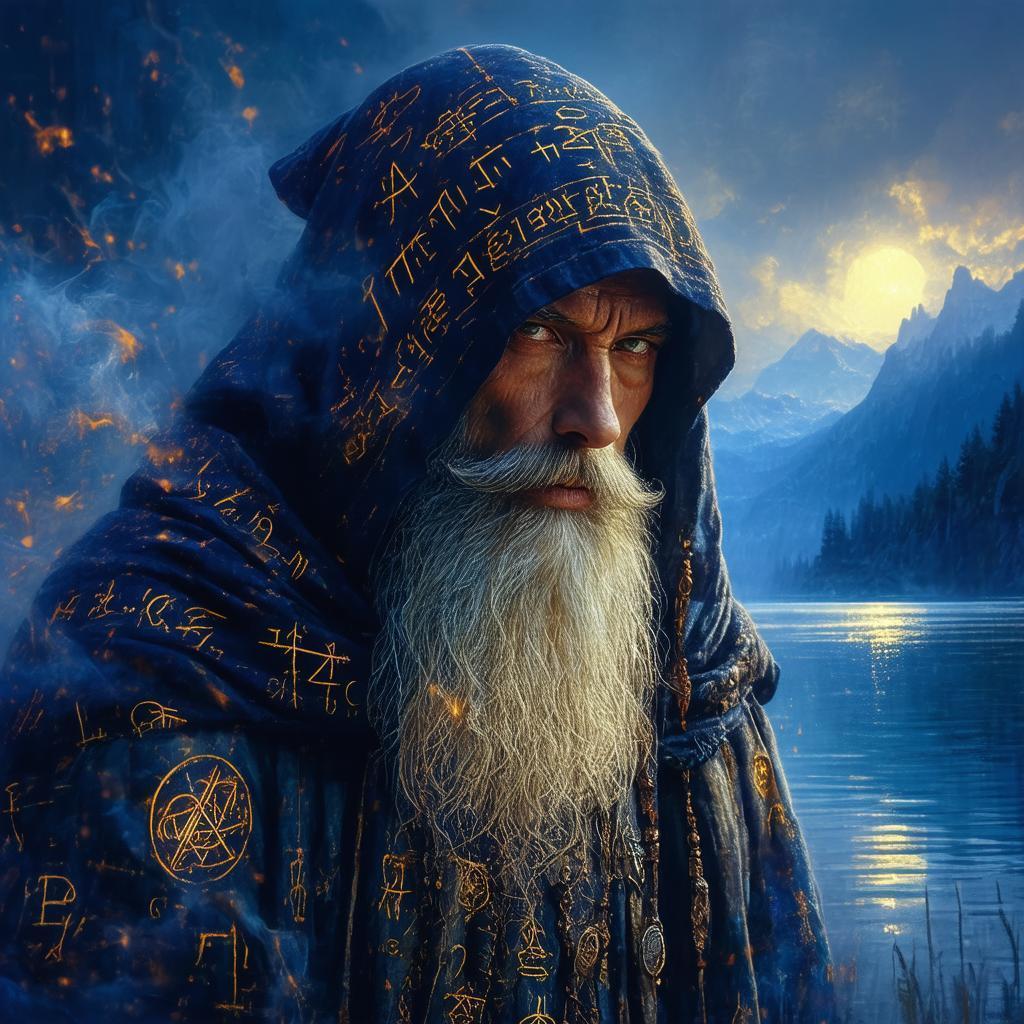} \\[-0.3em]

\makebox[0pt][r]{\rotatebox{90}{\fontsize{7}{7}\textit{Perturbed}}}
& \includegraphics[width=\linewidth]{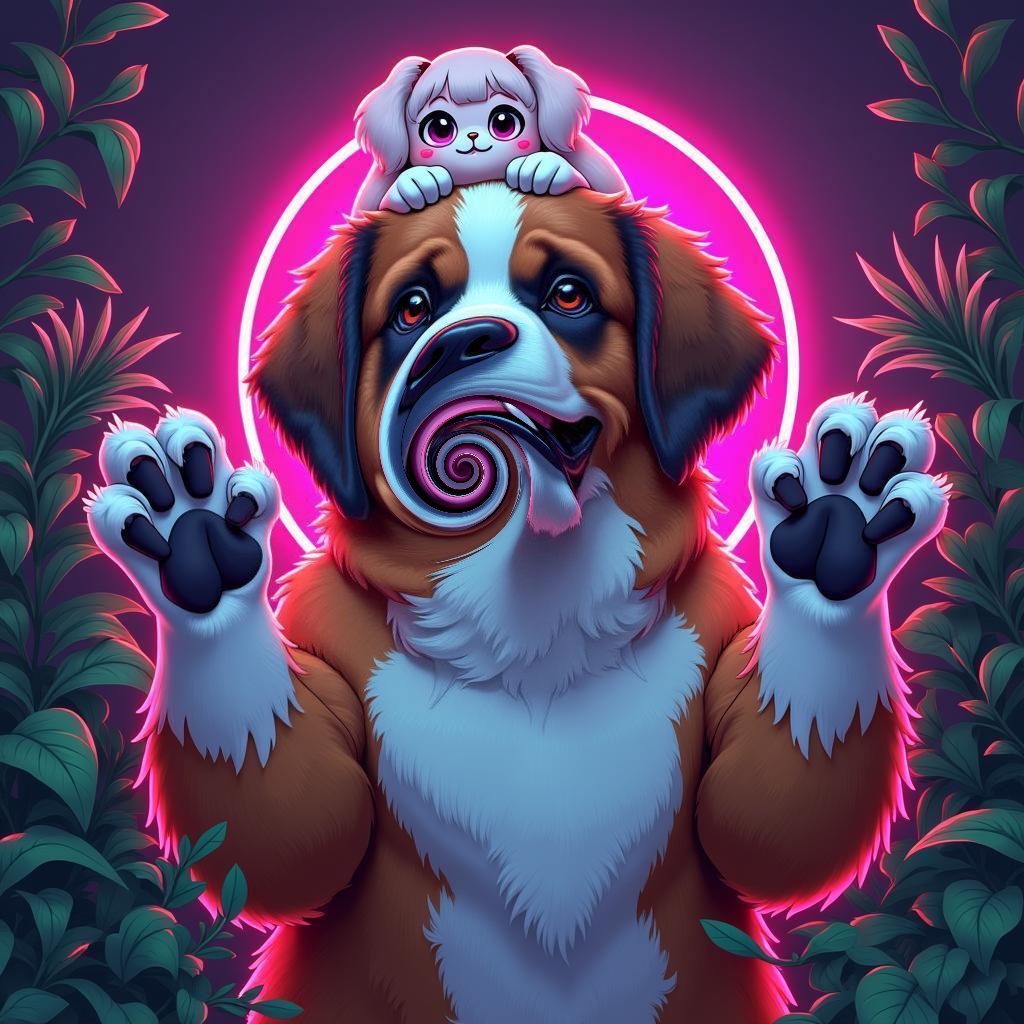}
& \includegraphics[width=\linewidth]{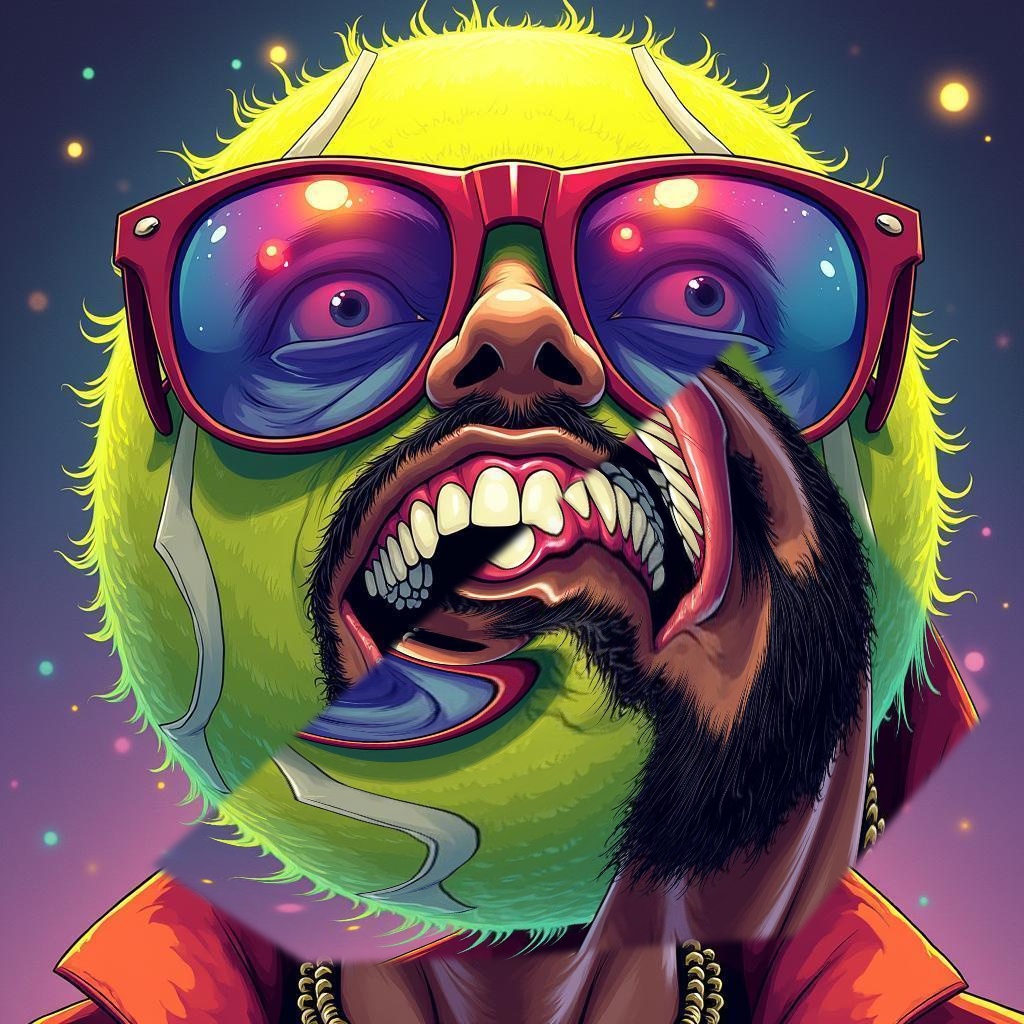}
& \includegraphics[width=\linewidth]{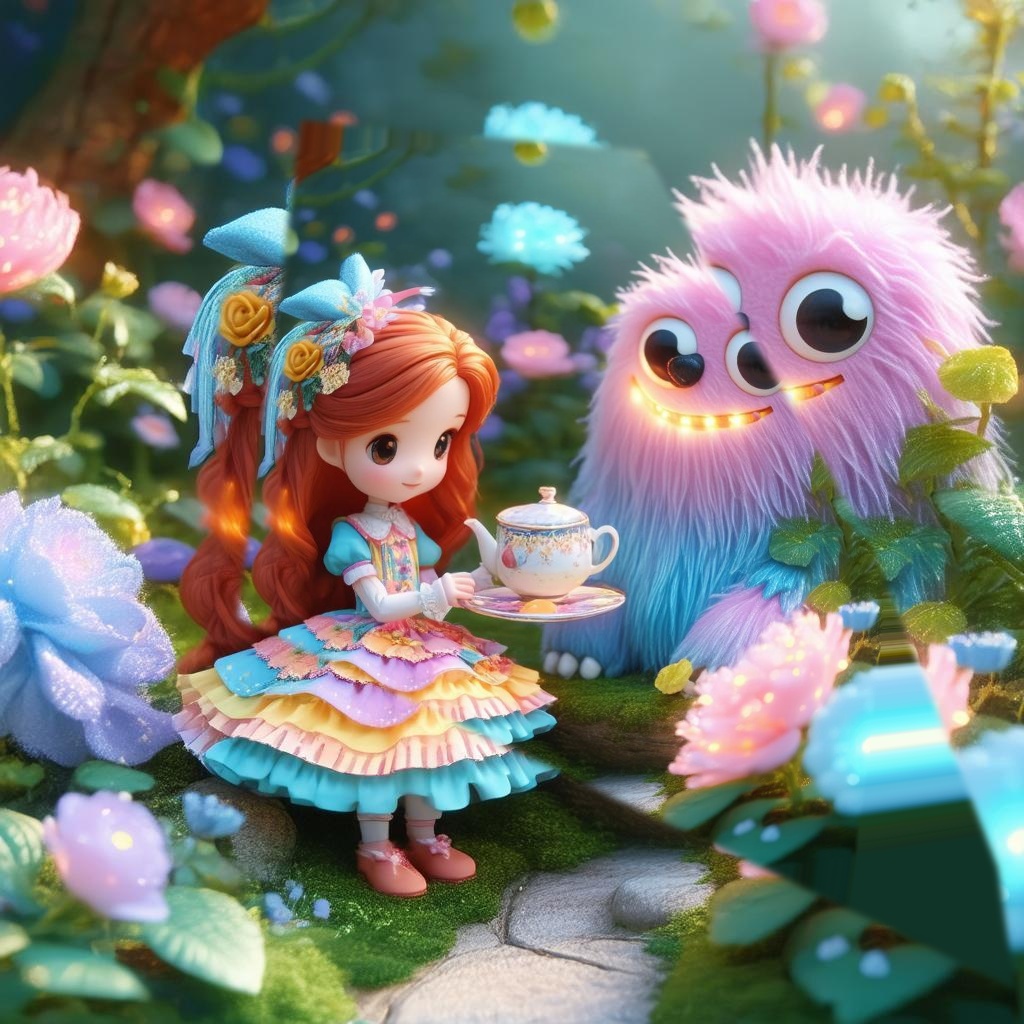}
& \includegraphics[width=\linewidth]{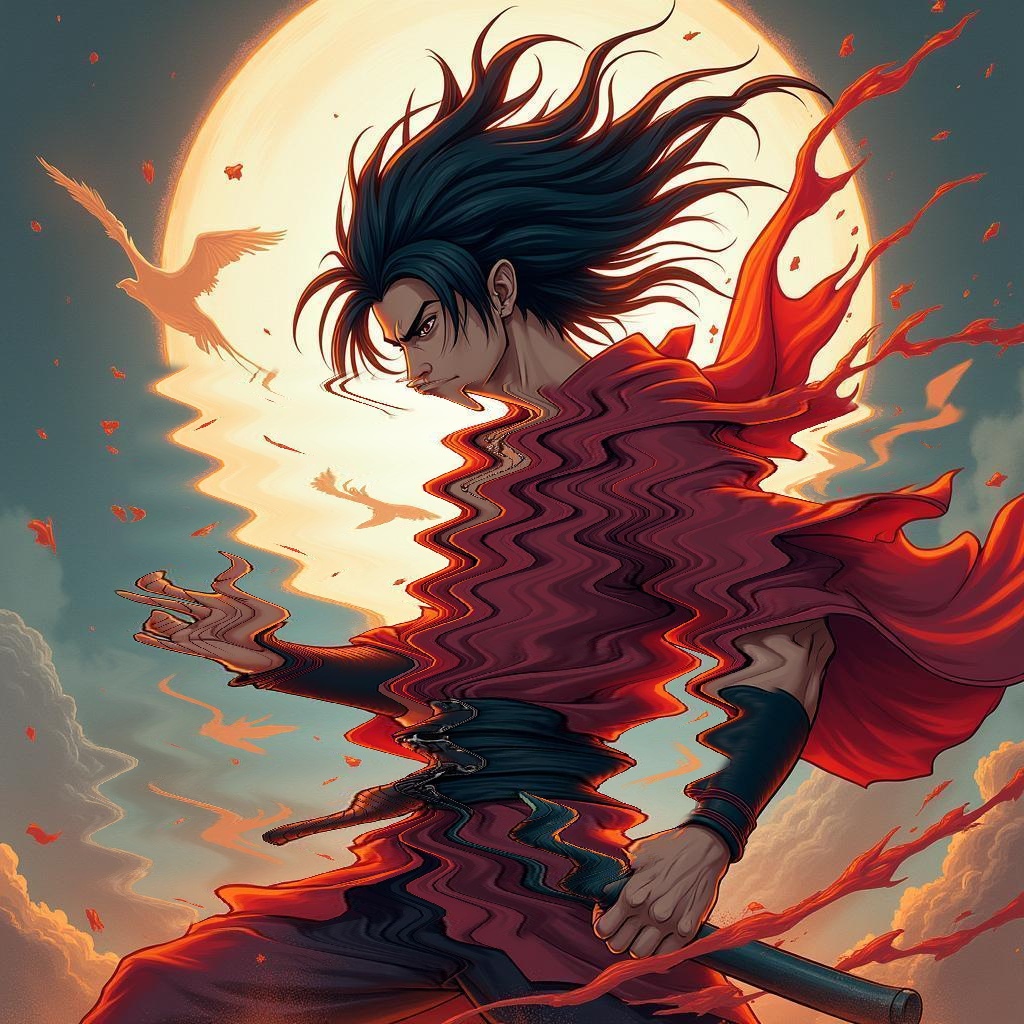}
& \includegraphics[width=\linewidth]{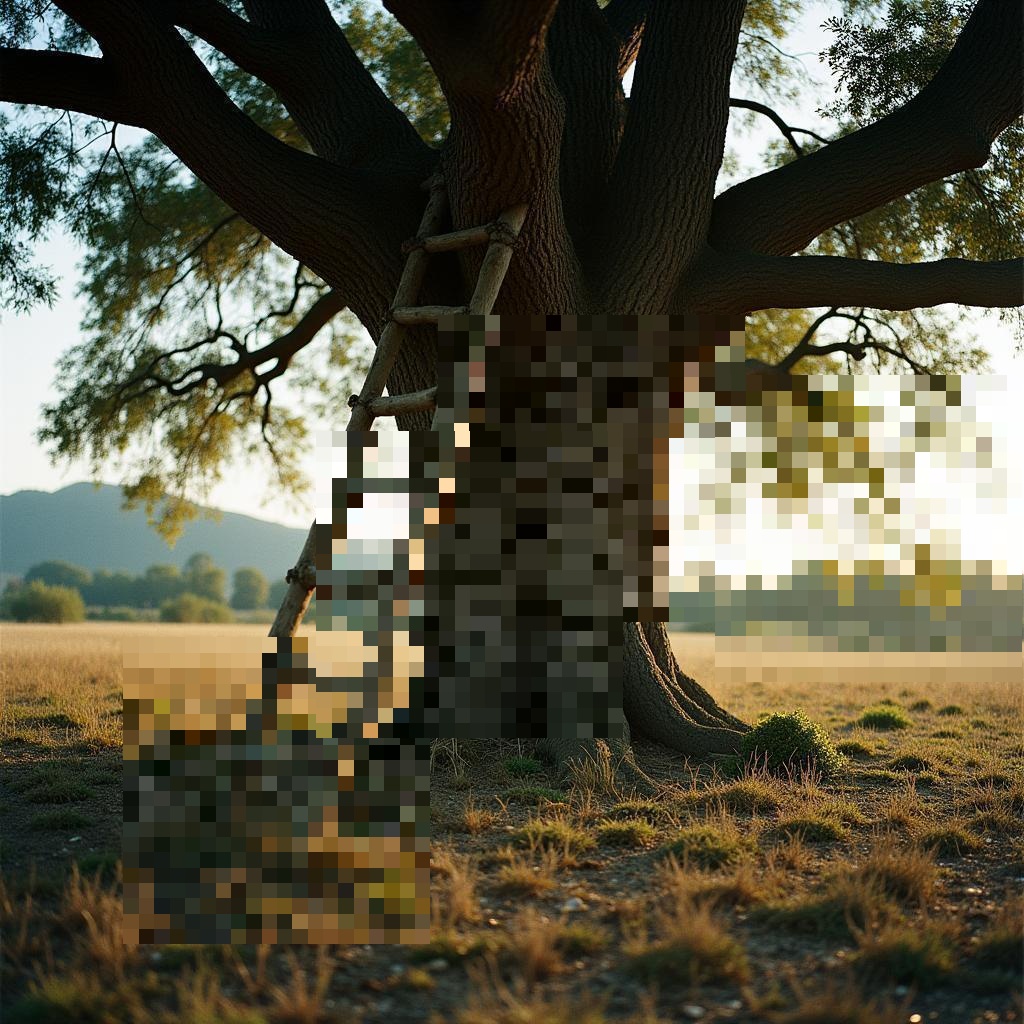}
& \includegraphics[width=\linewidth]{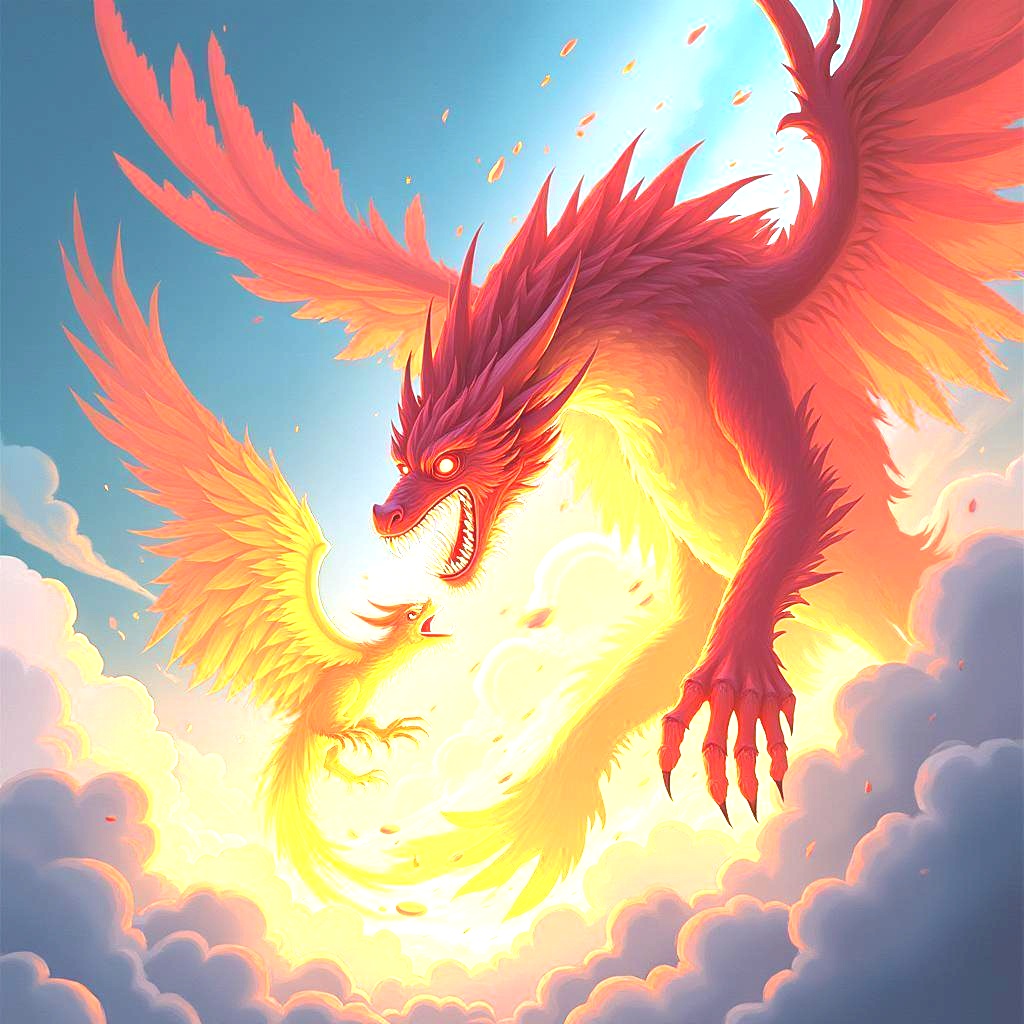}
& \includegraphics[width=\linewidth]{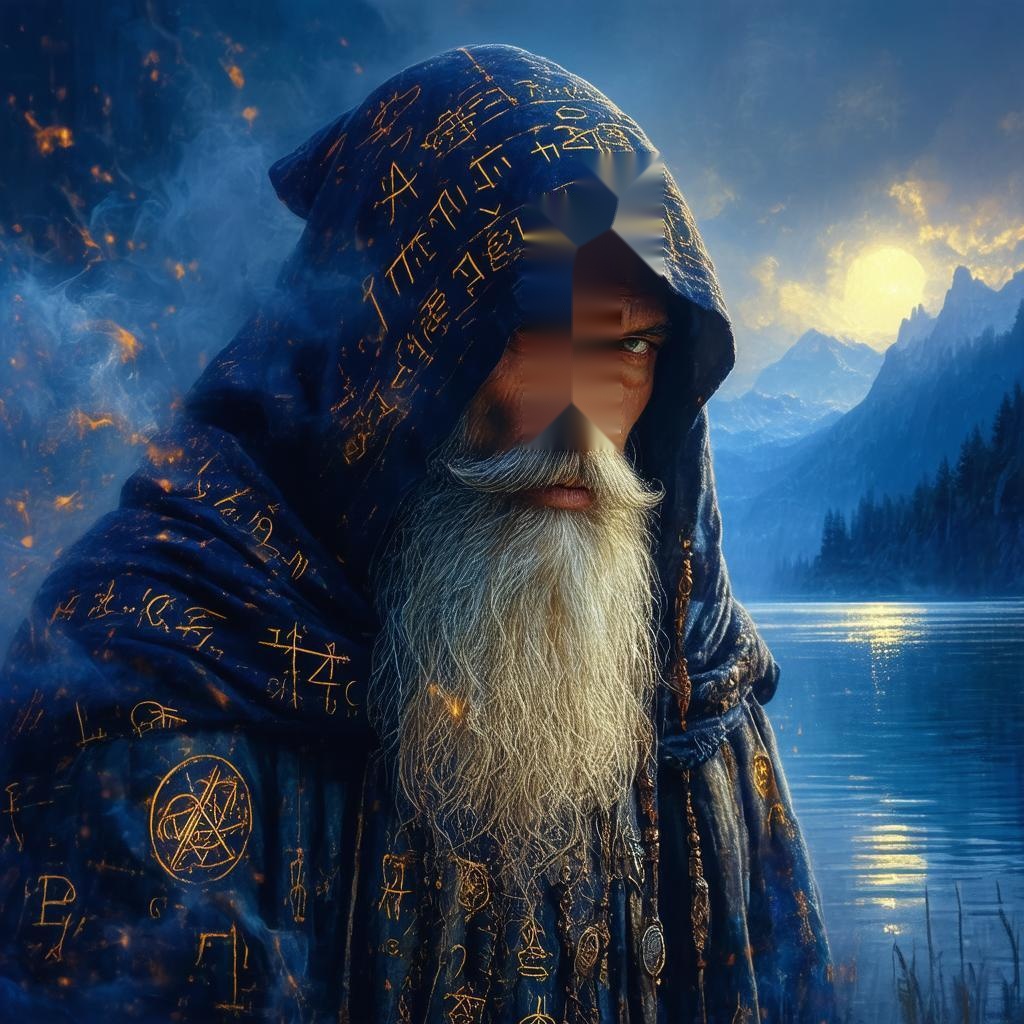}
\end{tabular}

\end{subfigure}

\caption{
Illustration of the controlled perceptual perturbations used for local
neighborhood recovery. Perturbation names are color-coded according to
the three families used in our analysis:
\textcolor{photometric}{\textbf{Photometric}},
\textcolor{geometric}{\textbf{Geometric}}, and
\textcolor{fidelity}{\textbf{Fidelity Reduction}}. \textbf{We recommend zooming in on the figure for better visualization
of the local variations.}
}

\label{fig:perturbation_examples}
\end{minipage}
\end{figure*}

Together, these perturbations provide diverse variations in appearance,
spatial structure, and visual fidelity around the preferred anchor image.
Table~\ref{tab:global_perturbations} provides a brief description of the
visual effect of each perturbation, while
Figure~\ref{fig:perturbation_examples} illustrates representative examples.

\section{Visualization of Recovered Local Preference Neighborhoods}
\label{sec:neighborhood_visualization}

Figure~\ref{fig:neighbourhood_examples} visualizes four representative
local preference neighborhoods constructed by \textsc{PrefLens}. In each
example, the human-preferred image $x^{+}$ serves as the anchor, while
the surrounding samples $x_i^{-}\in\mathcal{N}(x^{+})$ are recovered
through different compositions of controlled perceptual perturbations.
These neighbors introduce diverse variations in appearance, structure,
and visual fidelity while largely retaining the content of the anchor.
The irregular boundaries and spatial placement are schematic and do not
represent distances in the reward-defined preference space.

% ============================================================
% Required packages
% ============================================================
% \usepackage{graphicx}
% \usepackage{tikz}
% \usetikzlibrary{calc}

% ============================================================
% Global image-size controls
% ============================================================
\newcommand{\AnchorImgWidth}{4.0cm}
\newcommand{\NeighborImgWidth}{3.05cm}

% ============================================================
% Irregular neighborhood boundary
%
% #1 = random seed
% #2 = base radius
% #3 = background color
% ============================================================
\newcommand{\DrawIrregularBoundary}[3]{%
    \pgfmathsetseed{#1}%
    \def\boundarycoords{}%

    \foreach \ang in {0,24,...,336}{%
        \pgfmathsetmacro{\rad}{#2 + 0.42*rnd}%
        \xdef\boundarycoords{%
            \boundarycoords
            (\ang:\rad cm)
        }%
    }%

    \draw[
        fill=#3,
        draw=gray!60,
        dashed,
        line width=0.8pt
    ]
    plot[
        smooth cycle,
        tension=0.72
    ]
    coordinates {\boundarycoords};
}

% ============================================================
% Individual recovered neighbor
%
% #1 = neighborhood prefix
% #2 = neighbor index
% #3 = image path
% #4 = angle
% #5 = radius
% ============================================================
\newcommand{\Neighbor}[5]{%

    \coordinate (#1p#2) at (#4:#5);

    % --------------------------------------------------------
    % Neighbor image
    % --------------------------------------------------------
    \node[
        draw=gray!65,
        line width=0.5pt,
        inner sep=1pt,
        fill=white
    ] (#1n#2) at (#1p#2) {
        \includegraphics[width=\NeighborImgWidth]{#3}
    };

    % --------------------------------------------------------
    % Connection line
    % --------------------------------------------------------
    \draw[
        gray!58,
        line width=0.65pt
    ]
    (#1anchor) -- (#1n#2);

    % --------------------------------------------------------
    % Neighbor label
    % --------------------------------------------------------
    \node[
        font=\scriptsize,
        anchor=north,
        fill=white,
        inner sep=1pt
    ] at ([yshift=-3pt]#1n#2.south) {
        $x^{-}_{#2}$
    };
}

% ============================================================
% Complete local preference neighborhood
%
% #1 = x position
% #2 = y position
% #3 = unique prefix
% #4 = anchor image
% #5 = random boundary seed
% #6 = boundary base radius
% #7 = background color
% #8 = neighbor commands
% ============================================================
\newcommand{\NeighborhoodPlot}[8]{%
\begin{scope}[shift={(#1,#2)}]

    % --------------------------------------------------------
    % Irregular neighborhood
    % --------------------------------------------------------
    \DrawIrregularBoundary{#5}{#6}{#7}

    % --------------------------------------------------------
    % Anchor image
    % --------------------------------------------------------
    \coordinate (#3center) at (0,0);

    \node[
        draw=black,
        line width=1.2pt,
        inner sep=1pt,
        fill=white
    ] (#3anchor) at (#3center) {
        \includegraphics[width=\AnchorImgWidth]{#4}
    };

    % --------------------------------------------------------
    % Anchor label
    % --------------------------------------------------------
    \node[
        font=\scriptsize\bfseries,
        anchor=north,
        fill=white,
        inner sep=1pt
    ] at ([yshift=-3pt]#3anchor.south) {
        Anchor $x^{+}$
    };

    % --------------------------------------------------------
    % Recovered neighbors
    % --------------------------------------------------------
    #8

\end{scope}%
}

% ============================================================
% Manual panel prompt
%
% #1 = x position
% #2 = y position
% #3 = font size
% #4 = line spacing
% #5 = text width
% #6 = panel label + prompt
% ============================================================
\newcommand{\PanelPrompt}[6]{%
    \node[
        font=\fontsize{#3}{#4}\selectfont,
        text width=#5,
        align=center,
        anchor=center
    ] at (#1,#2) {
        #6
    };
}

% ============================================================
% Figure
% ============================================================
\begin{figure*}[!htb]
\centering

\resizebox{0.8\textwidth}{!}{%
\begin{tikzpicture}

% ============================================================
% (a) Neighborhood -- light blue
% ============================================================
\NeighborhoodPlot
{-6.15cm}{8.15cm}
{A}
{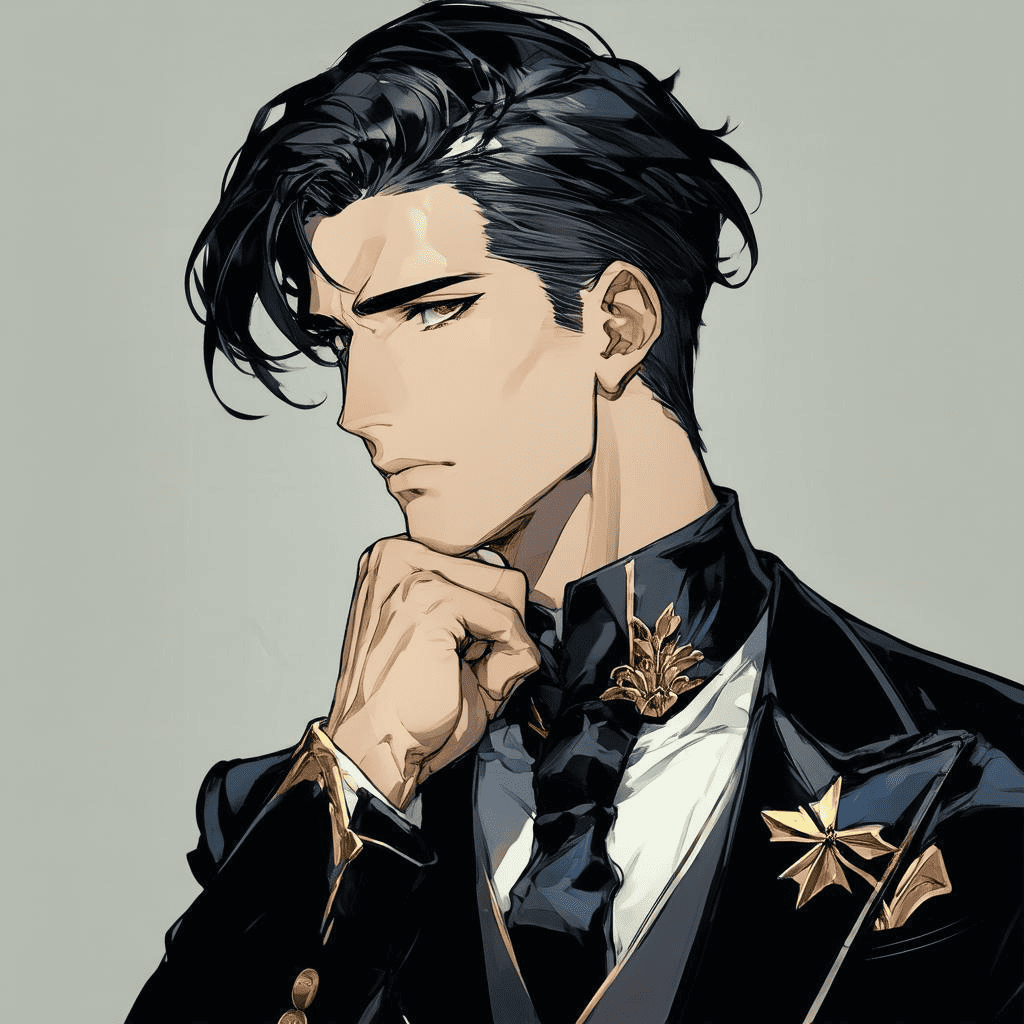}
{101}
{8.0}
{black!4}
{
    \Neighbor{A}{1}
    {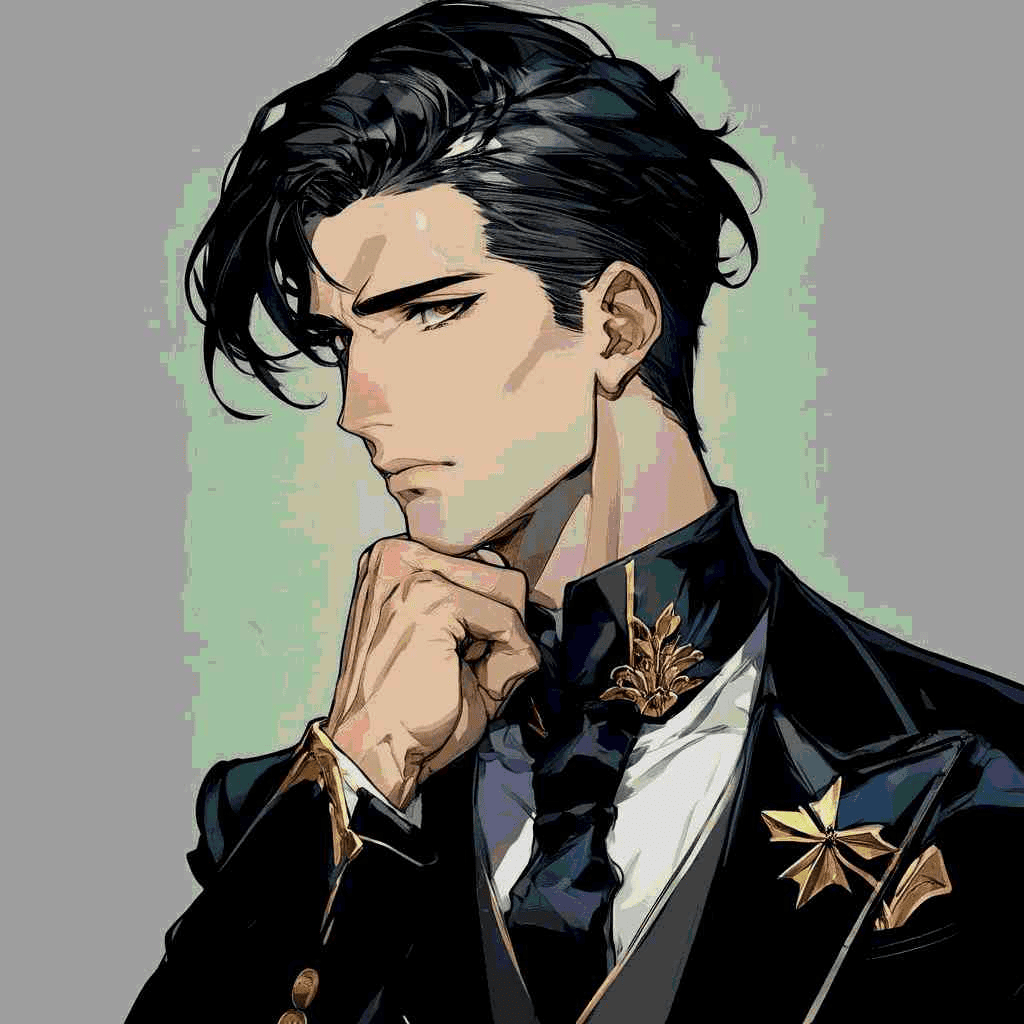}
    {92}{6.10cm}

    \Neighbor{A}{2}
    {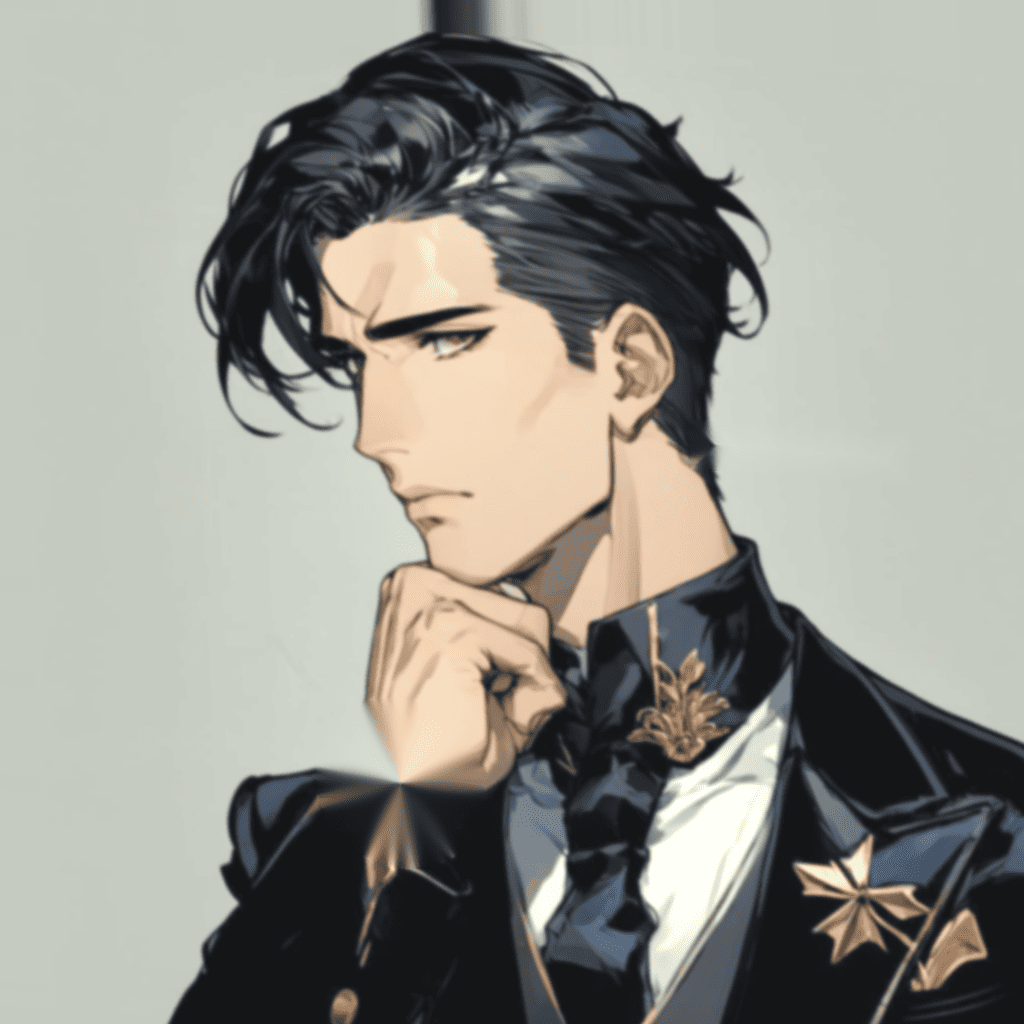}
    {36}{6.35cm}

    \Neighbor{A}{3}
    {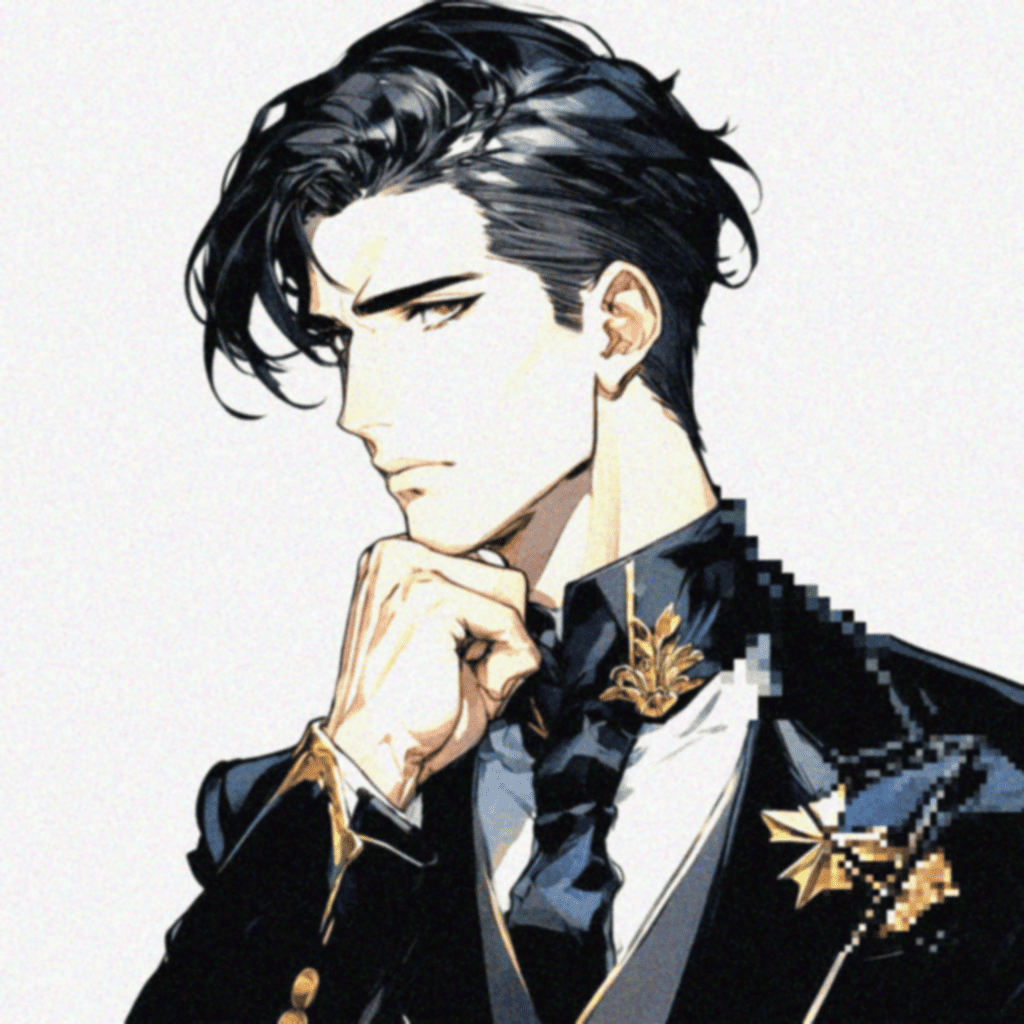}
    {-18}{5.95cm}

    \Neighbor{A}{4}
    {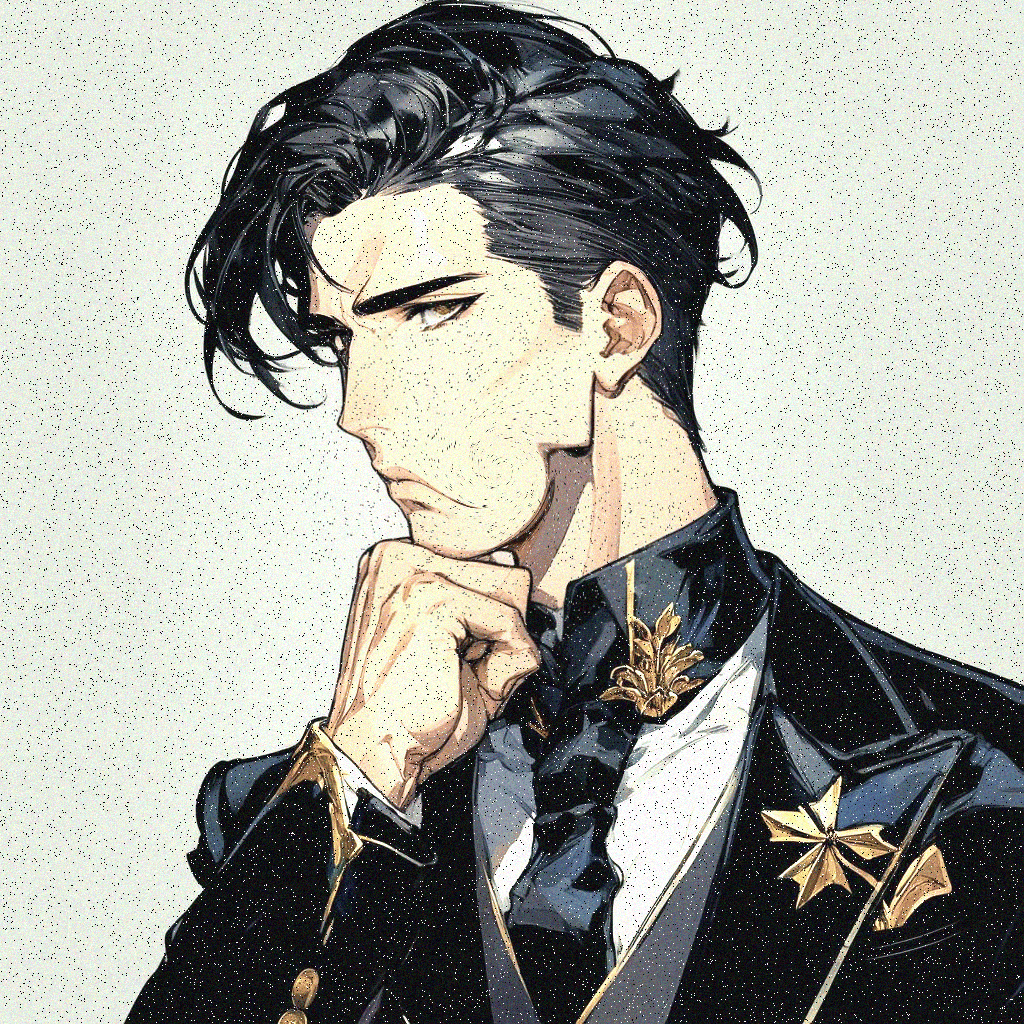}
    {-76}{6.25cm}

    \Neighbor{A}{5}
    {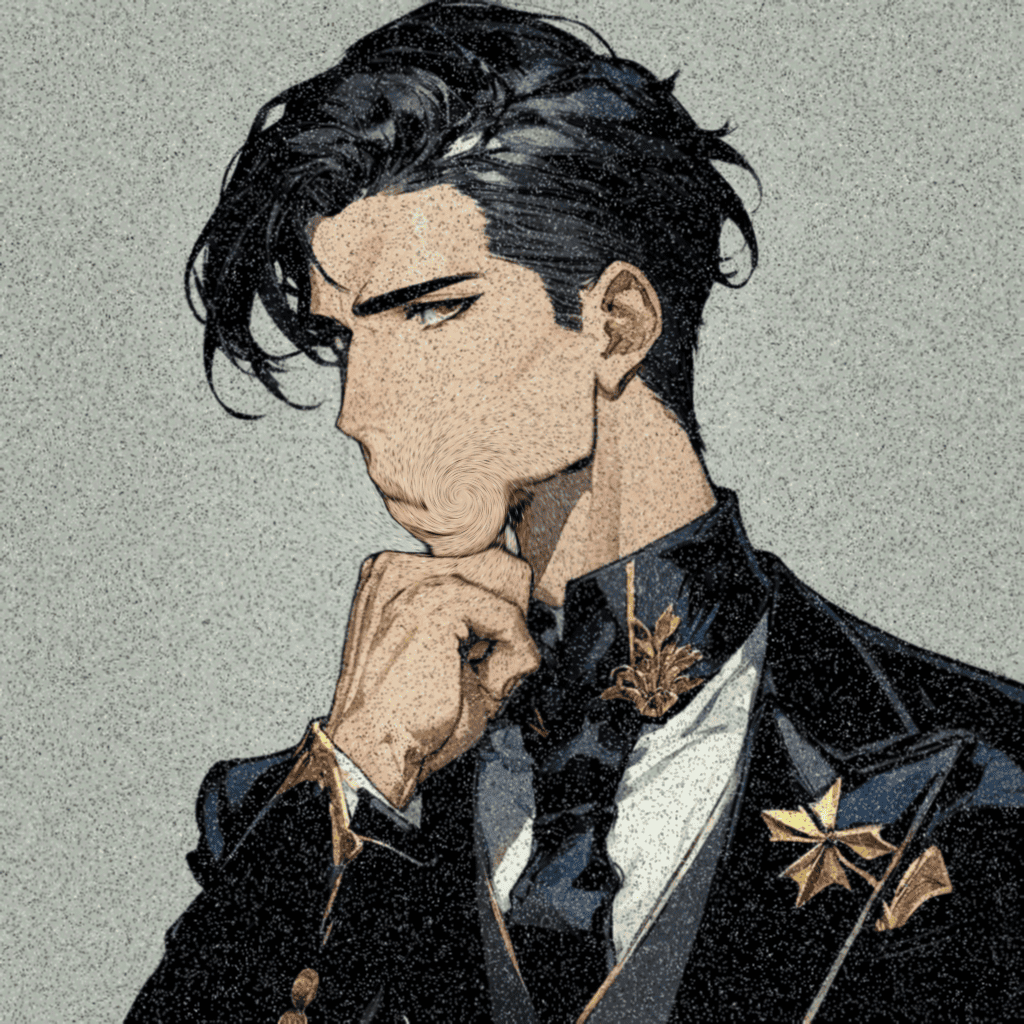}
    {-137}{5.90cm}

    \Neighbor{A}{6}
    {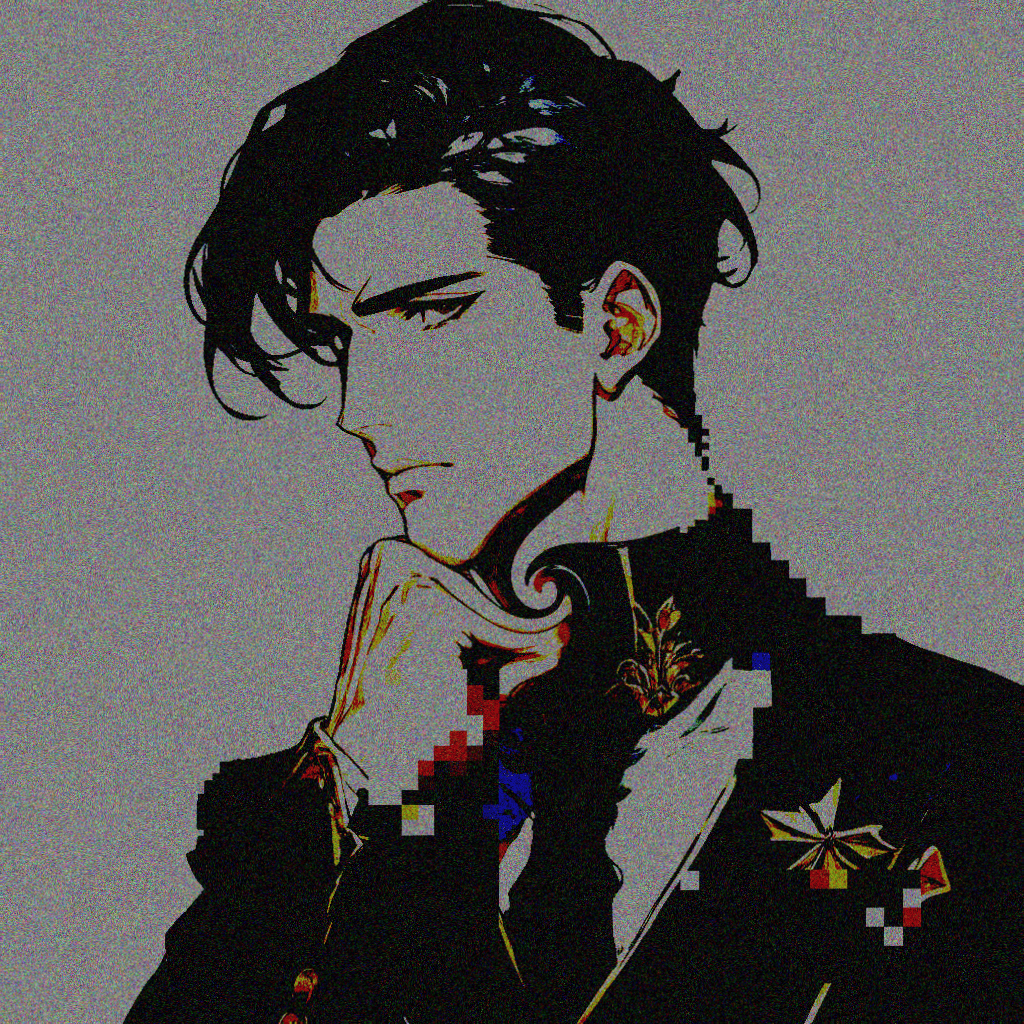}
    {161}{6.30cm}
}

% ============================================================
% (b) Neighborhood -- light green
% ============================================================
\NeighborhoodPlot
{9.85cm}{-0.70cm}
{B}
{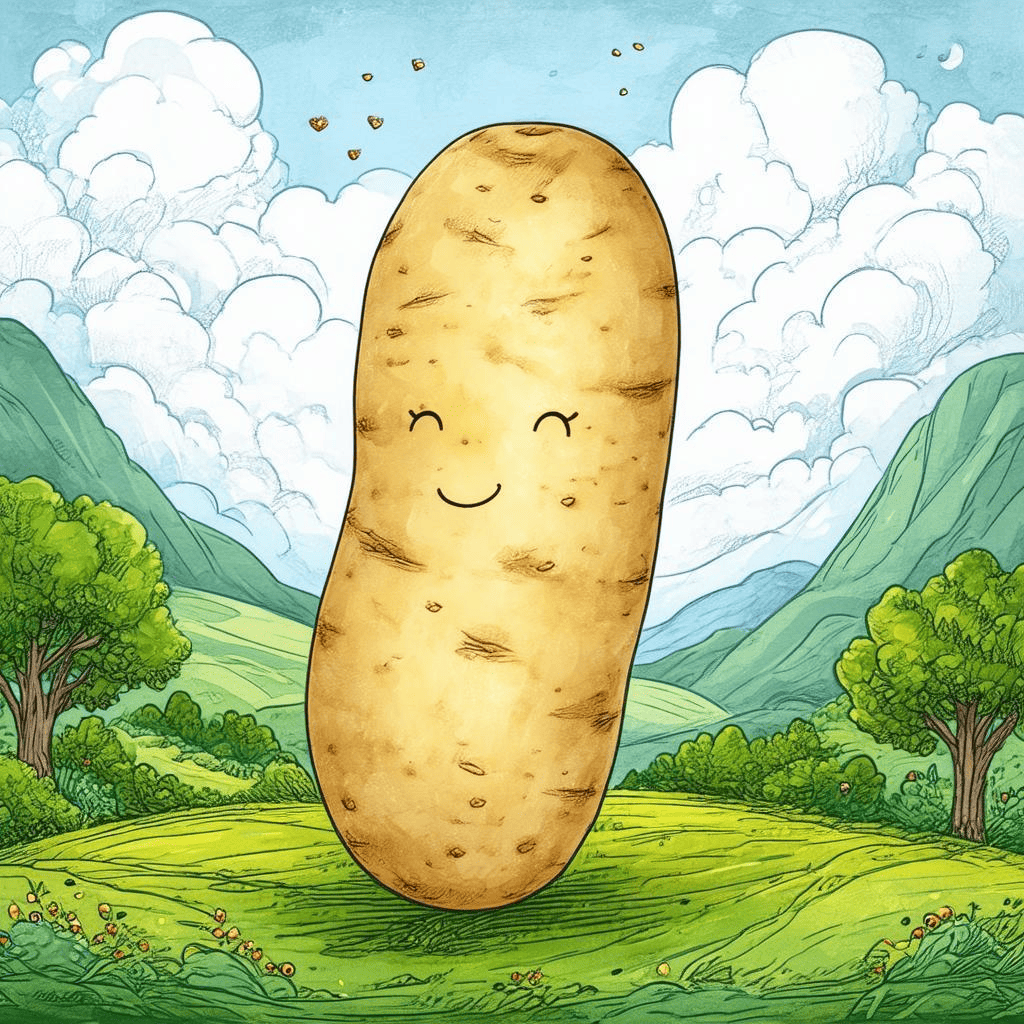}
{202}
{8.0}
{red!6}
{
    \Neighbor{B}{1}
    {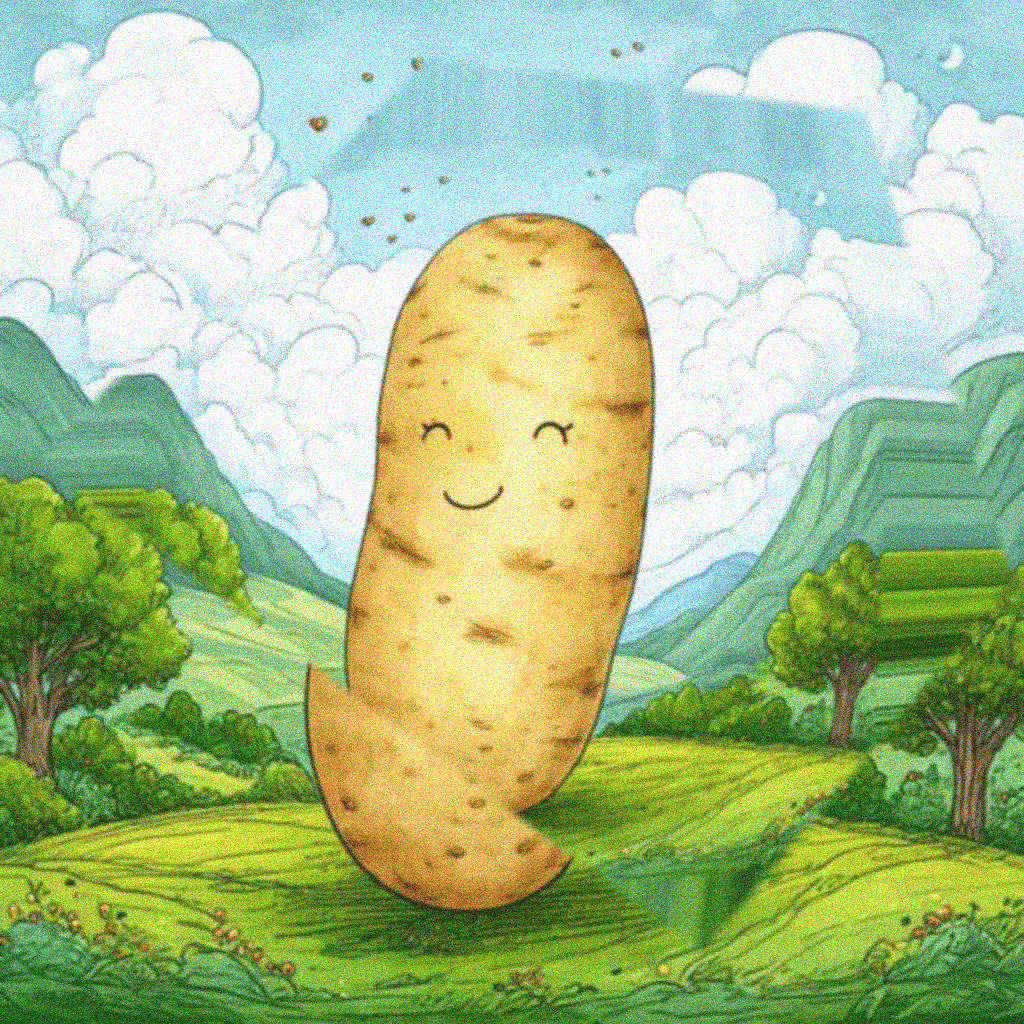}
    {104}{6.15cm}

    \Neighbor{B}{2}
    {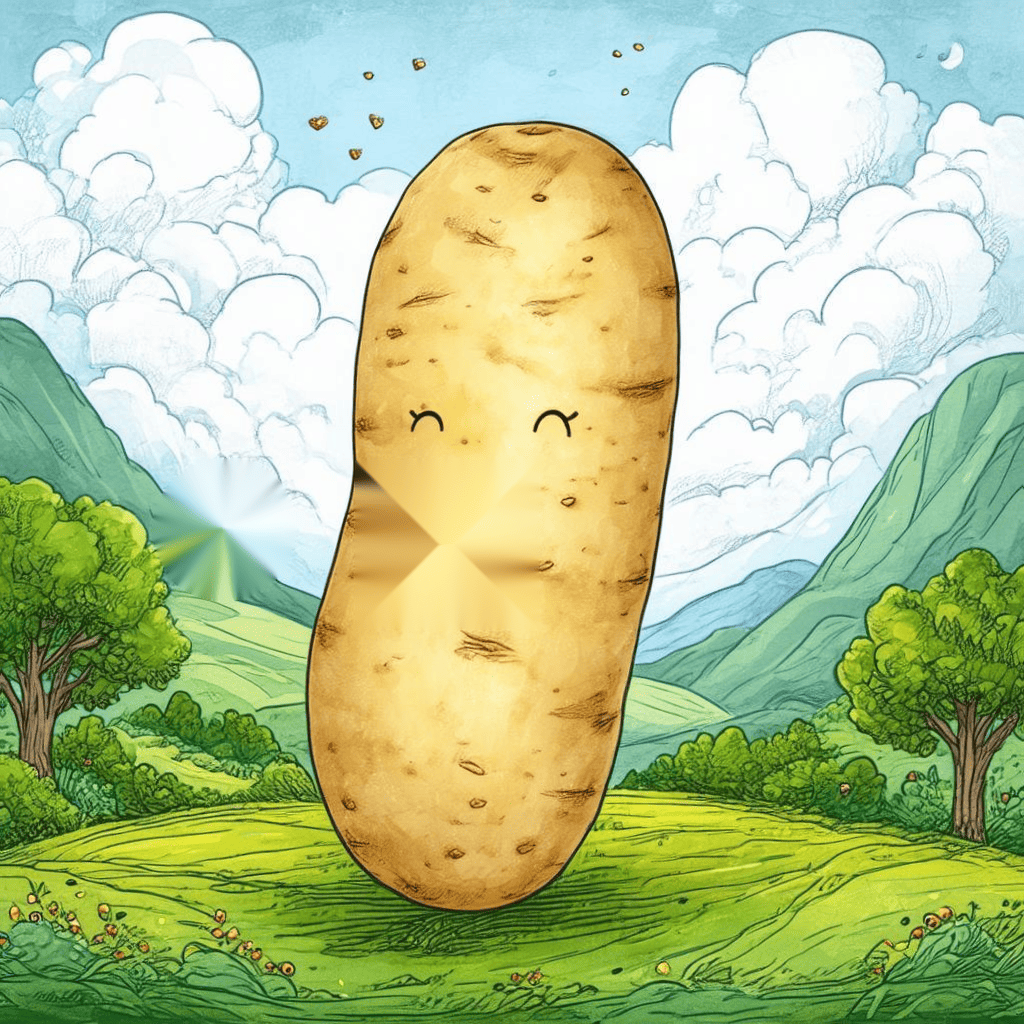}
    {38}{6.35cm}

    \Neighbor{B}{3}
    {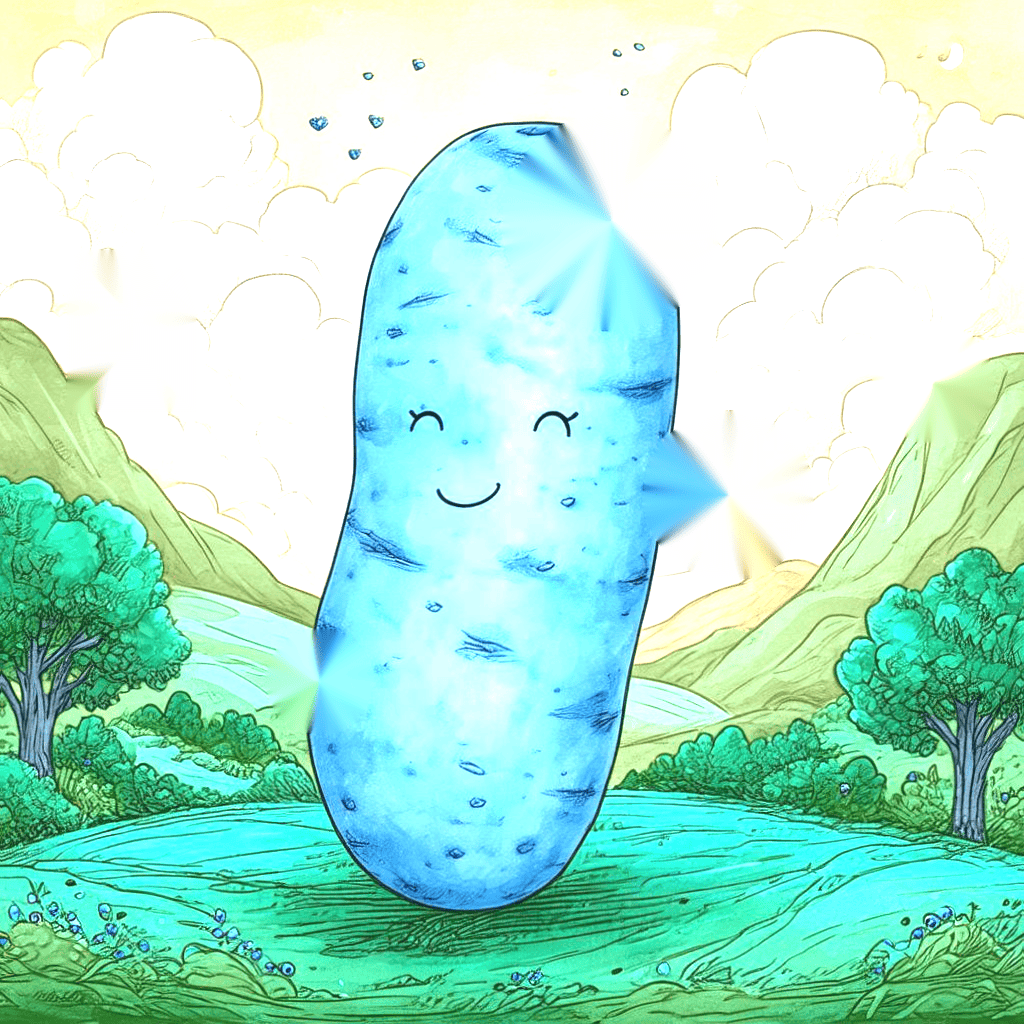}
    {-28}{6.05cm}

    \Neighbor{B}{4}
    {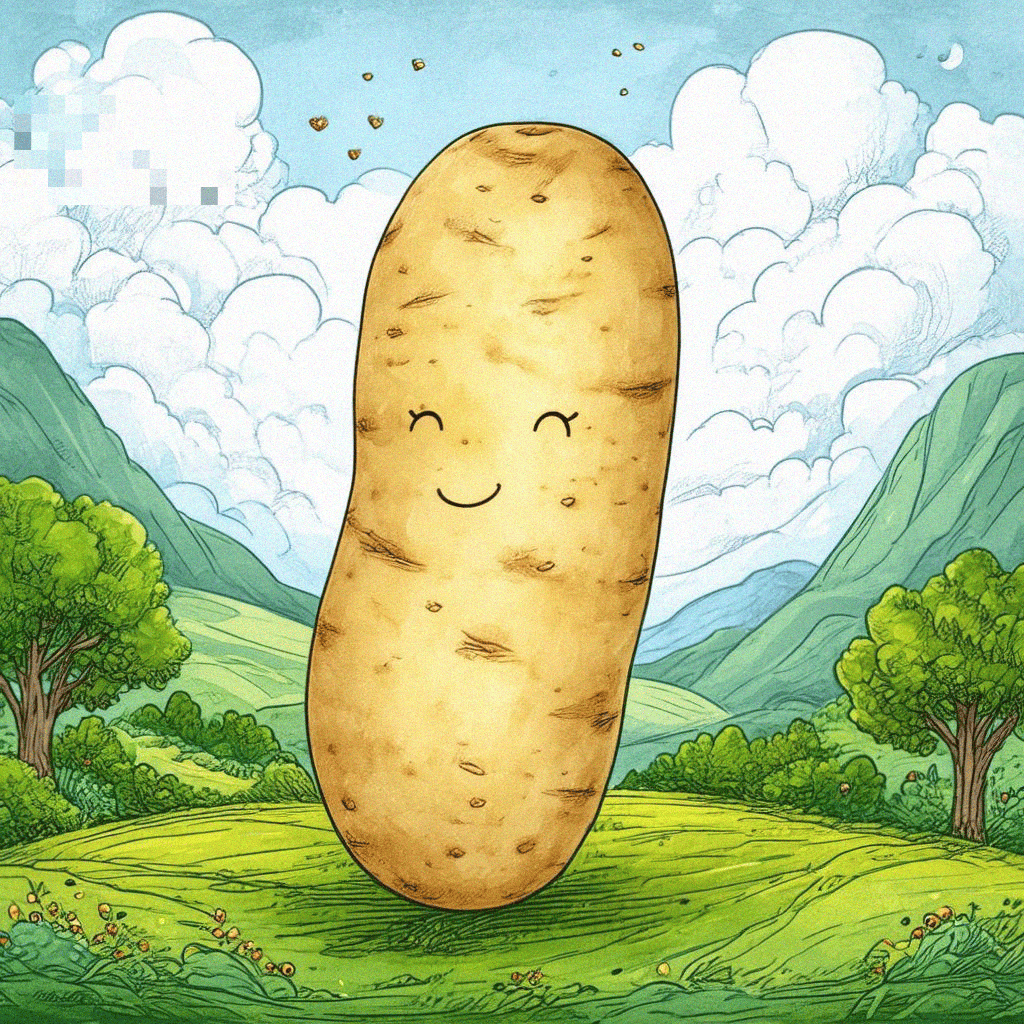}
    {-104}{6.20cm}

    \Neighbor{B}{5}
    {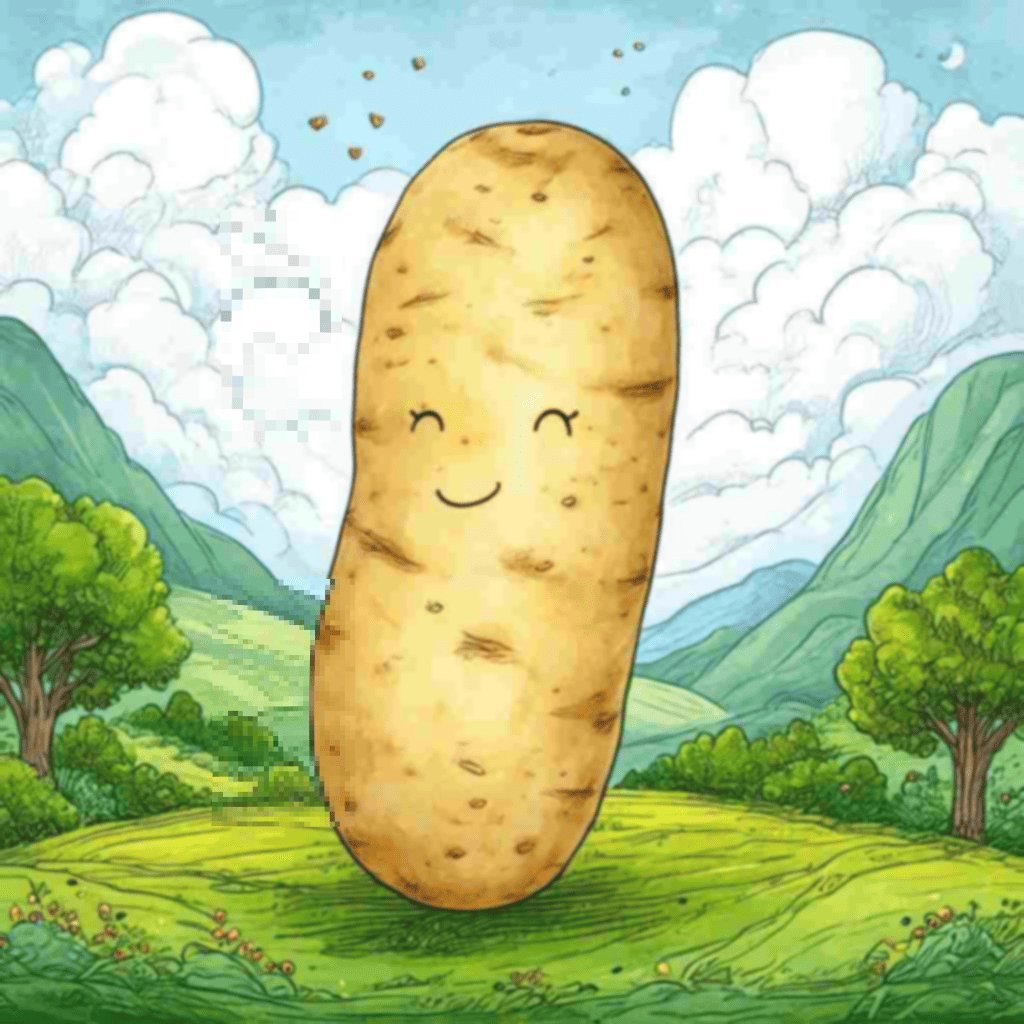}
    {166}{6.00cm}
}

% ============================================================
% (c) Neighborhood -- light red/pink
% ============================================================
\NeighborhoodPlot
{-6.15cm}{-15.15cm}
{C}
{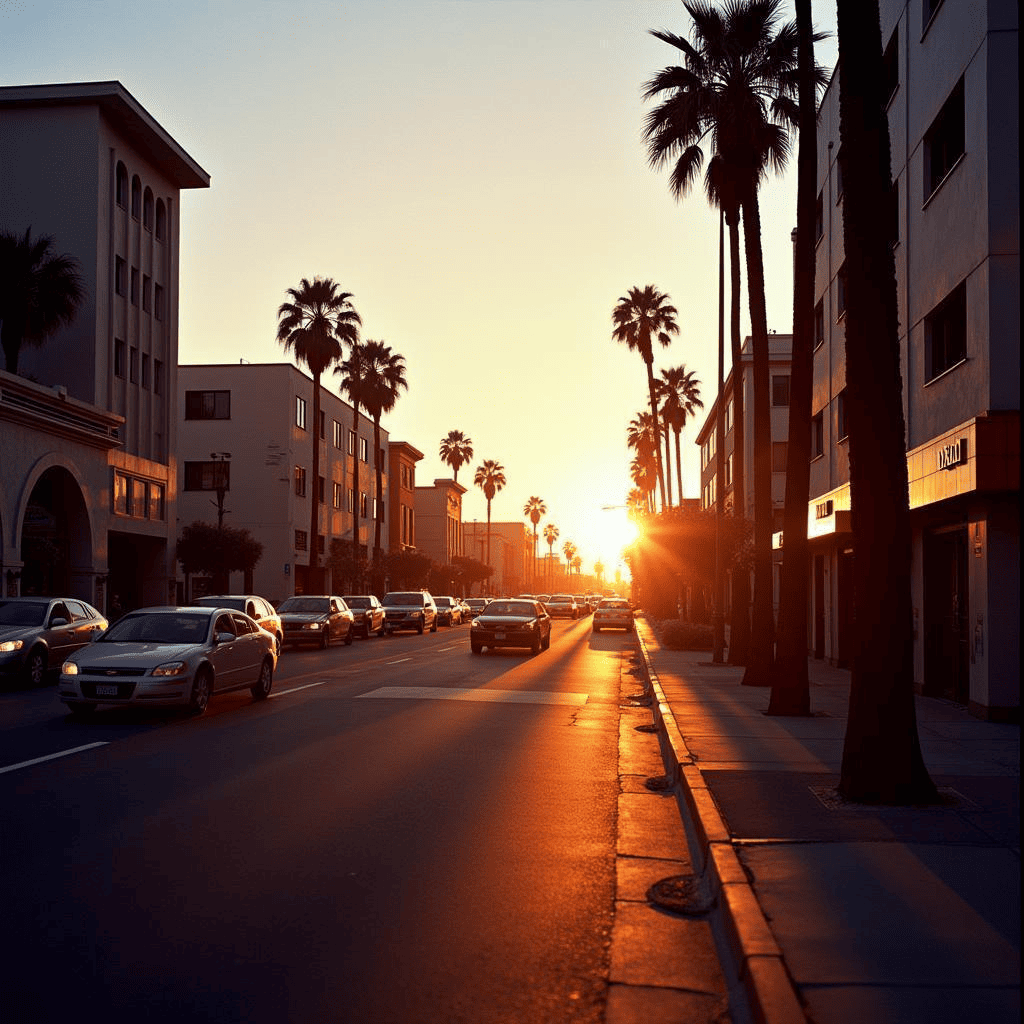}
{303}
{8.0}
{green!5}
{
    \Neighbor{C}{1}
    {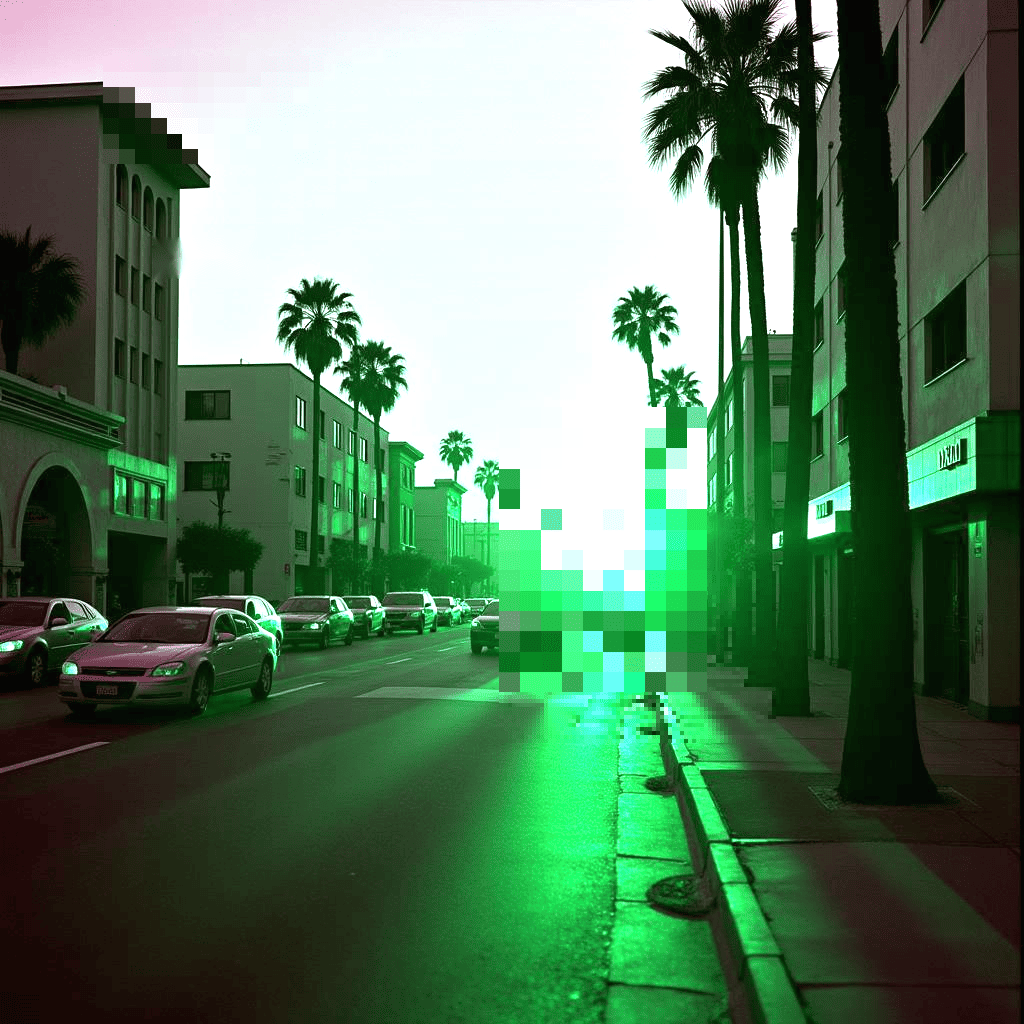}
    {78}{6.25cm}

    \Neighbor{C}{2}
    {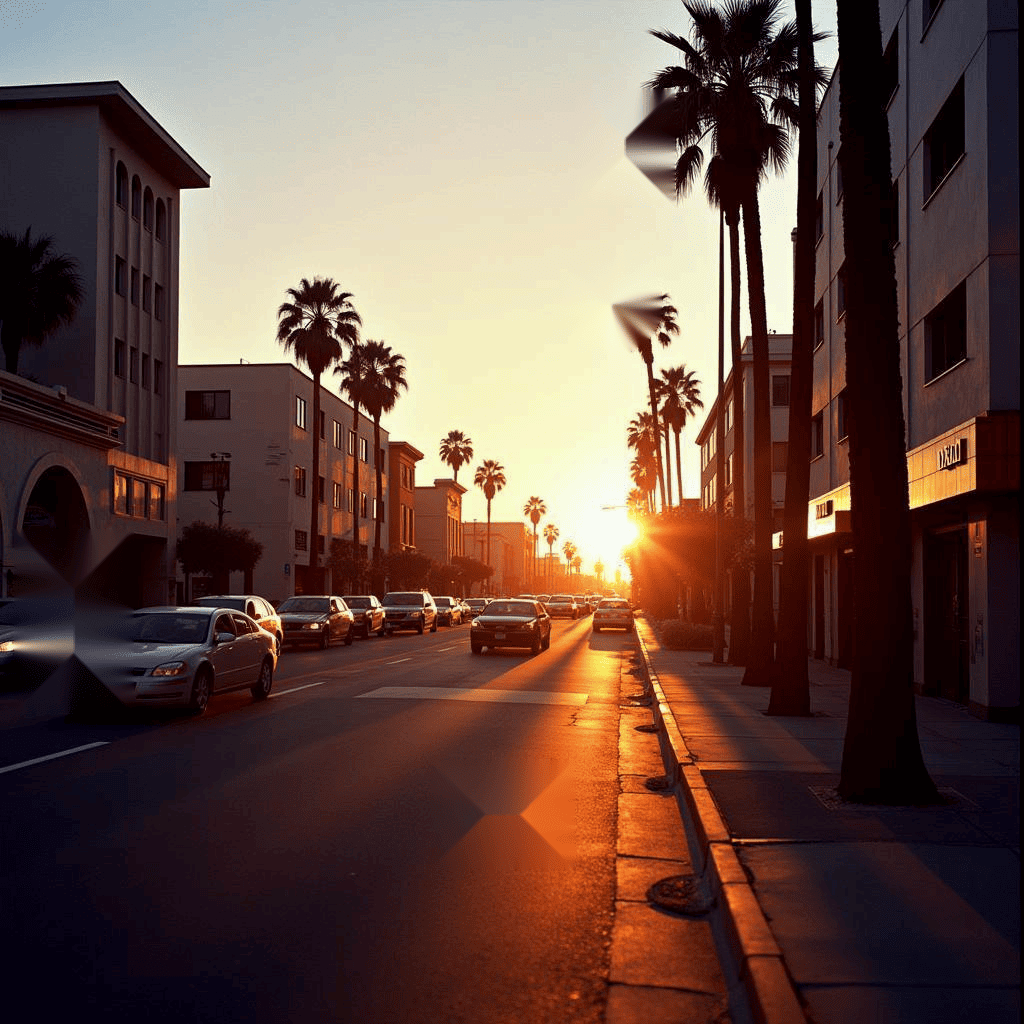}
    {-8}{6.40cm}

    \Neighbor{C}{3}
    {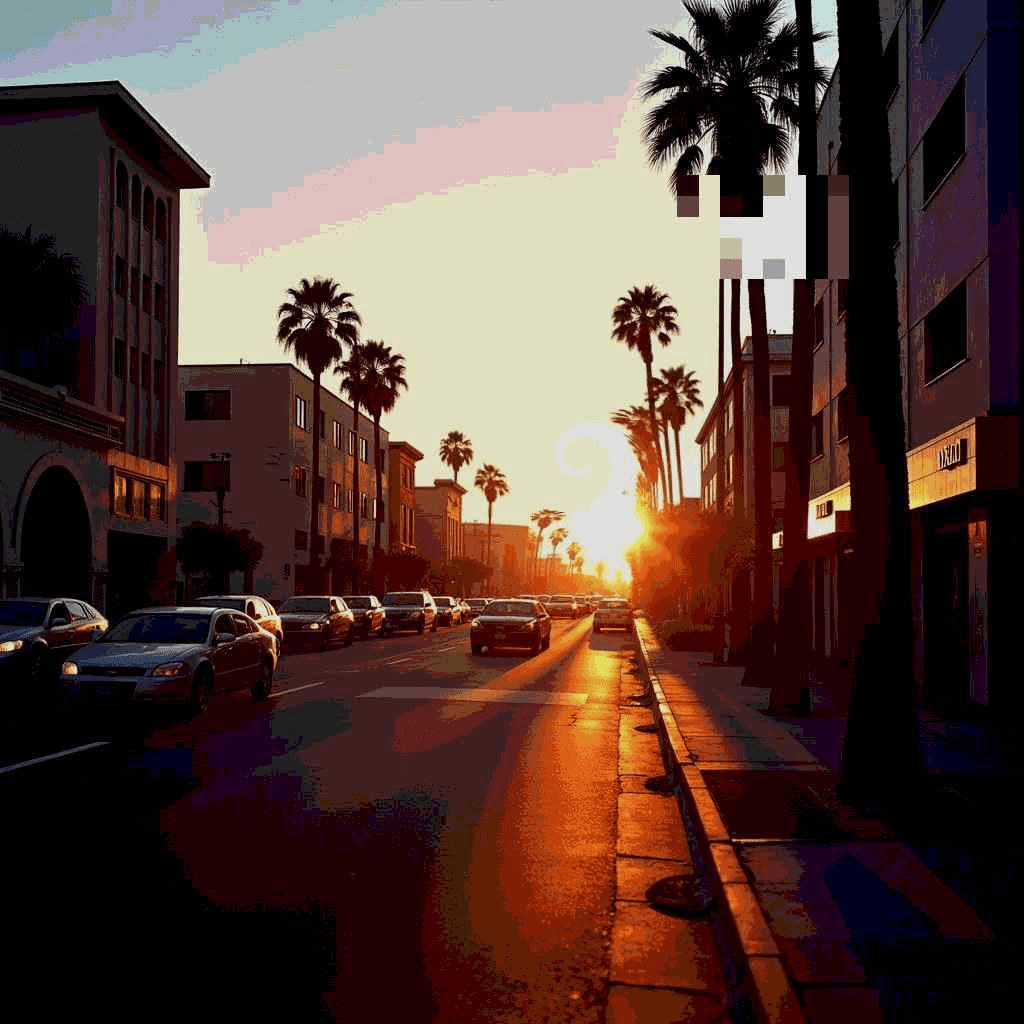}
    {-105}{6.10cm}

    \Neighbor{C}{4}
    {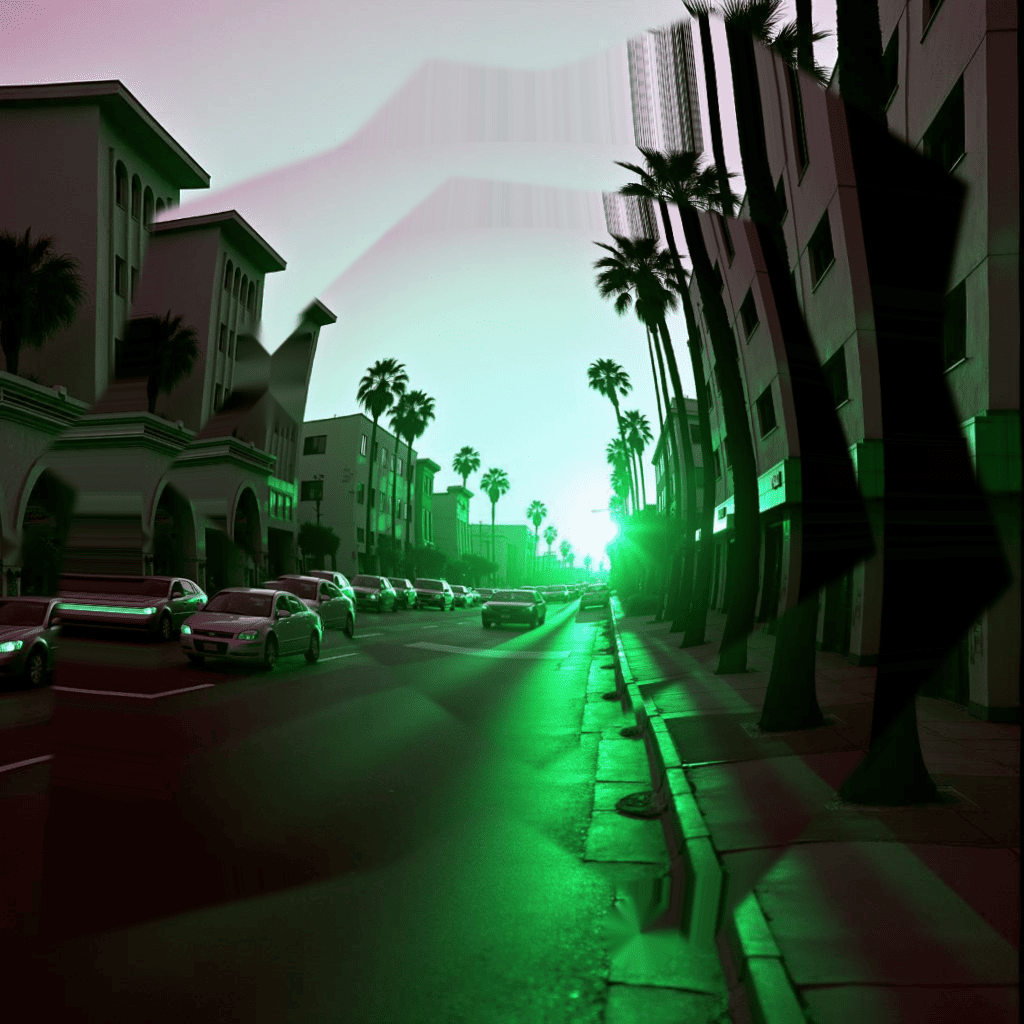}
    {170}{6.20cm}
}

% ============================================================
% (d) Neighborhood -- light yellow
% ============================================================
\NeighborhoodPlot
{9.85cm}{-22.60cm}
{D}
{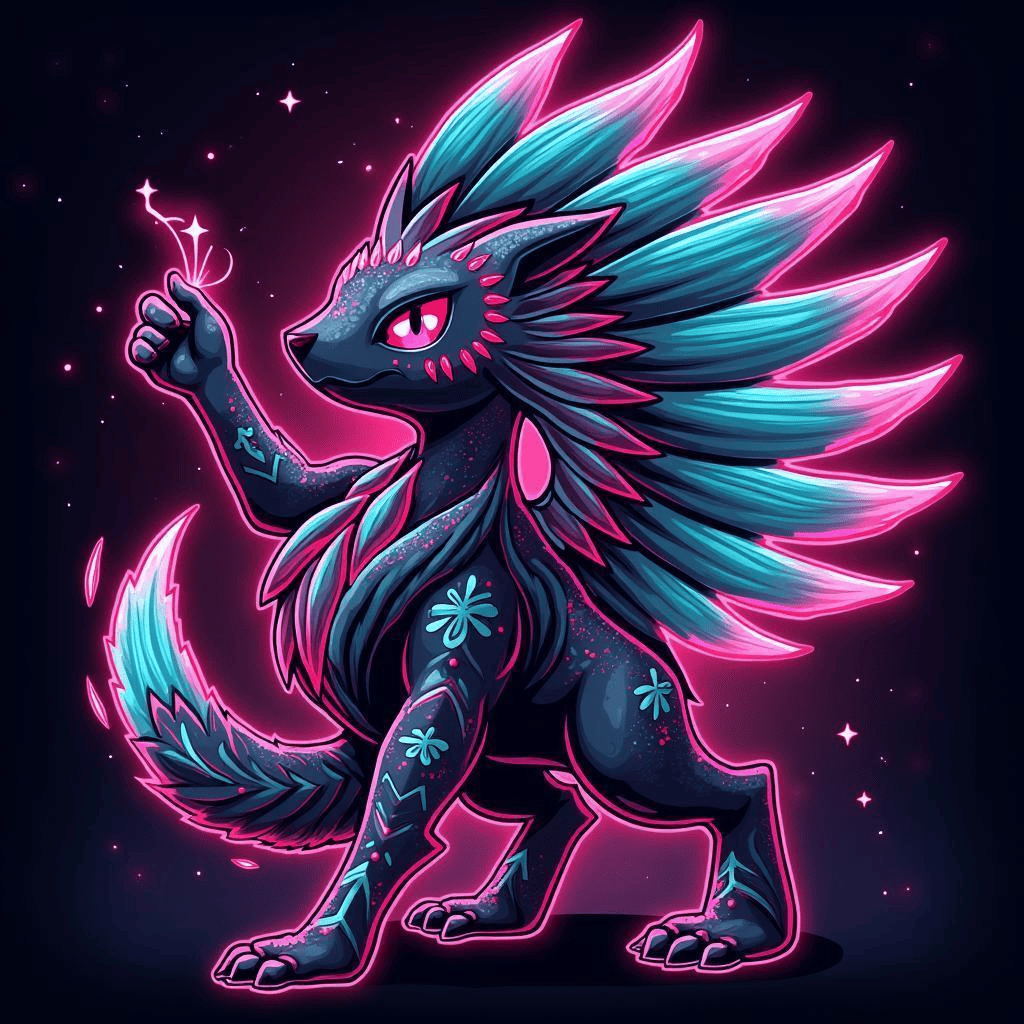}
{404}
{8.0}
{cyan!4}
{
    \Neighbor{D}{1}
    {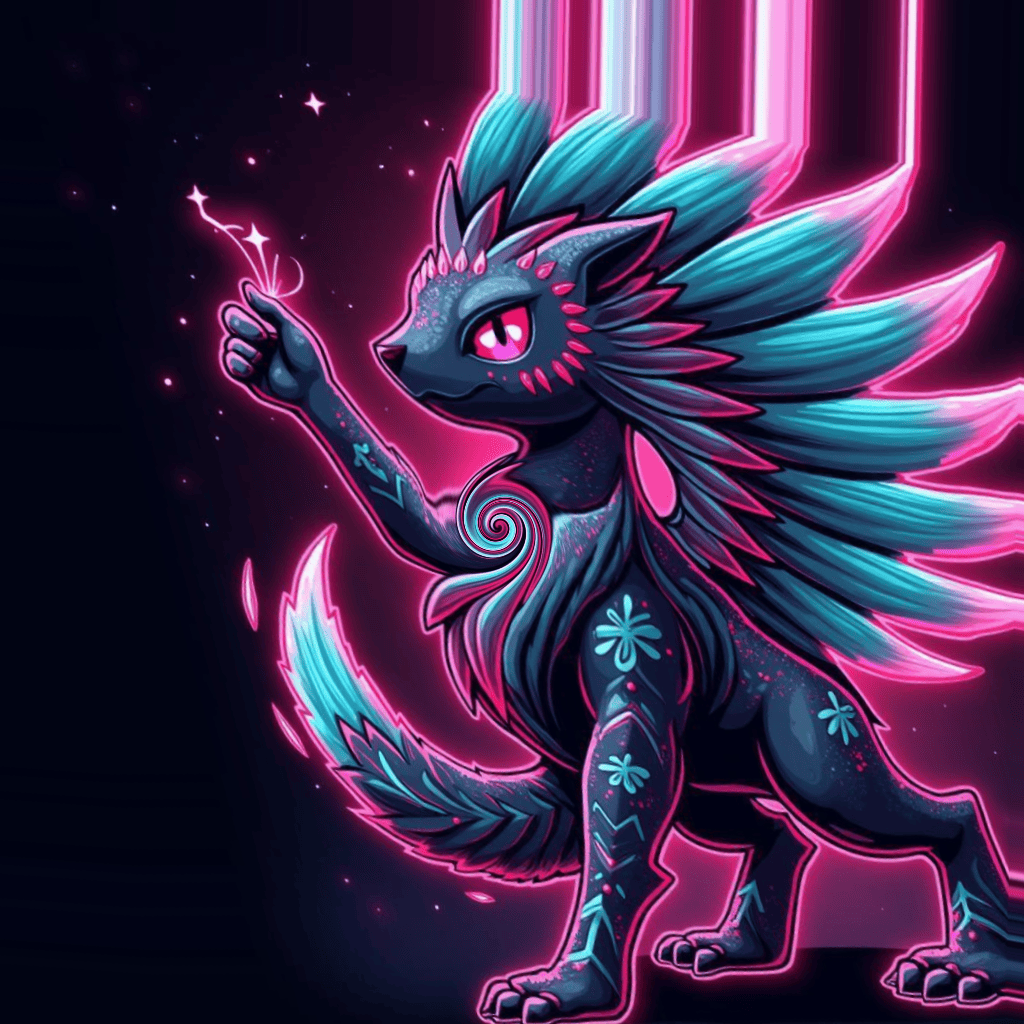}
    {96}{6.15cm}

    \Neighbor{D}{2}
    {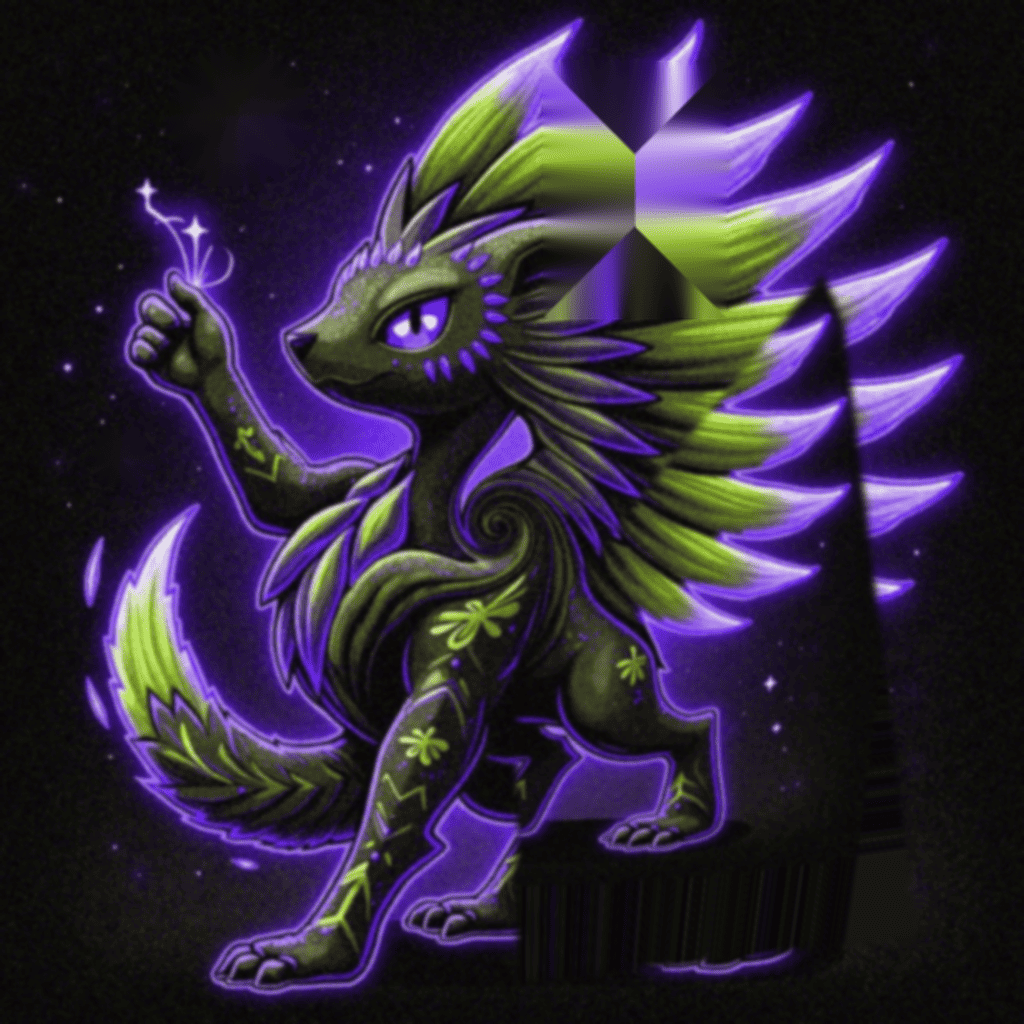}
    {32}{6.35cm}

    \Neighbor{D}{3}
    {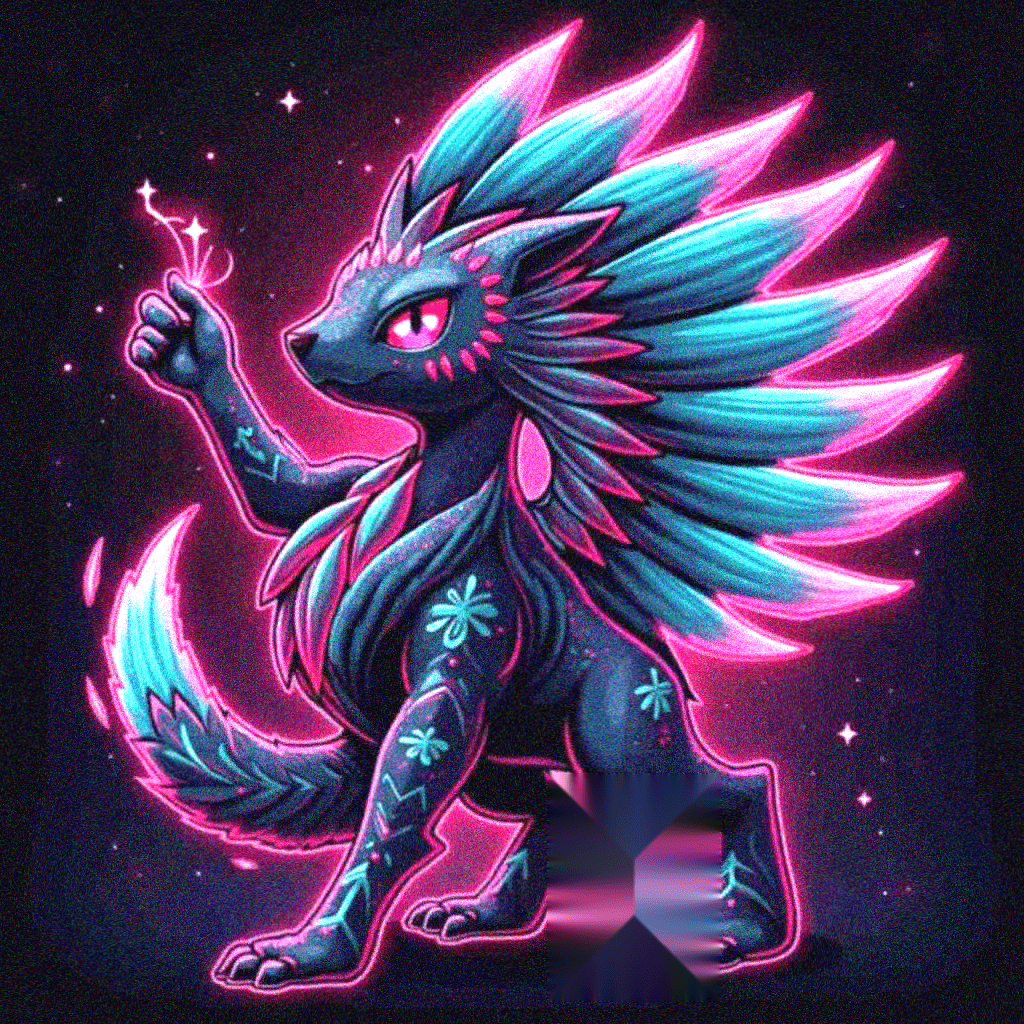}
    {-35}{6.05cm}

    \Neighbor{D}{4}
    {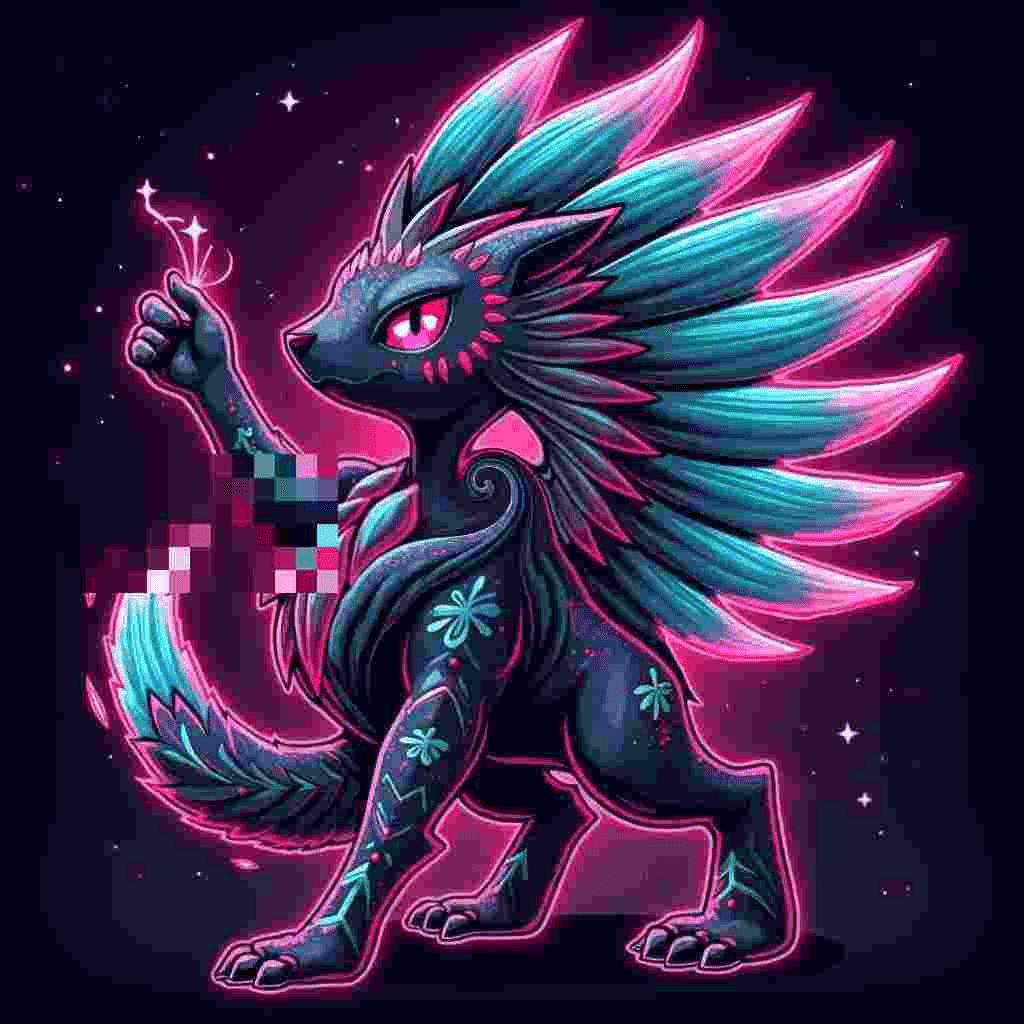}
    {-108}{6.25cm}

    \Neighbor{D}{5}
    {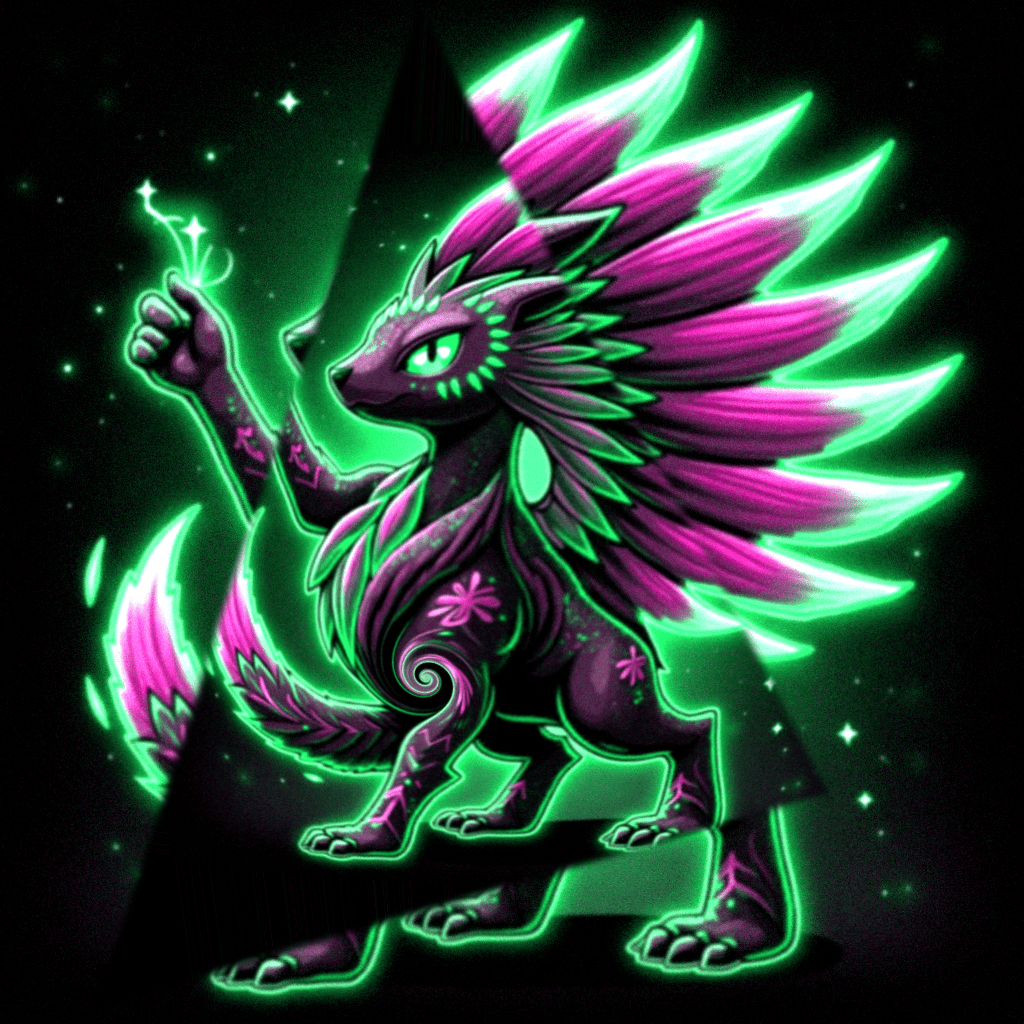}
    {164}{6.10cm}
}

% ============================================================
% MANUAL PROMPTS
%
% Every prompt is controlled independently:
%
% {x position}
% {y position}
% {font size}
% {line spacing}
% {text width}
% {text}
% ============================================================

% ------------------------------------------------------------
% (a) Prompt
% ------------------------------------------------------------
\PanelPrompt
{-6.15cm}     % x position
{-2.60cm}     % y position
{20pt}        % font size
{13pt}        % line spacing
{13cm}        % text width
{\textbf{(a)}~\textit{
Portrait of Ernest Khalimov, thinking, sigma male, black hair,
Nordic crown, black suit; elegant anime style, vibrant shading,
dramatic lighting, monochromatic high contrast.
}}

% ------------------------------------------------------------
% (b) Prompt
% ------------------------------------------------------------
\PanelPrompt
{9.85cm}      % x position
{9.70cm}      % y position
{20pt}        % font size
{13pt}        % line spacing
{13cm}        % text width
{\textbf{(b)}~\textit{
Portrait of a potato with intricate line work, against a cloudy sky
and lush landscape, in a whimsical children's-book style with soft,
vibrant colors.
}}

% ------------------------------------------------------------
% (c) Prompt
% ------------------------------------------------------------
\PanelPrompt
{-6.15cm}     % x position
{-25.20cm}    % y position
{20pt}        % font size
{13pt}        % line spacing
{13cm}        % text width
{\textbf{(c)}~\textit{
Downtown LA at sunrise, Velvia film, vivid colors, sharp contrast,
crisp details, soft golden-hour light, dynamic perspective.
}}

% ------------------------------------------------------------
% (d) Prompt
% ------------------------------------------------------------
\PanelPrompt
{9.85cm}      % x position
{-12.60cm}    % y position
{20pt}        % font size
{13pt}        % line spacing
{13cm}        % text width
{\textbf{(d)}~\textit{
Neonpunk Pokémon with tribal patterns and Indian art, vibrant neon
glow, moon markings, powerful posture, and vaporwave style.
}}

\end{tikzpicture}%
}

\caption{
Representative recovered local preference neighborhoods for four
human-preferred anchor images. In each example, $x^{+}$ denotes the
preferred anchor, while the surrounding
$x_i^{-}\in\mathcal{N}(x^{+})$ denote recovered local alternatives.
The irregular boundaries and spatial positions are schematic and do not
encode distances in the reward-defined preference space.
\textbf{We recommend zooming in on the figure for better visualization
of the local variations.}
}
\label{fig:neighbourhood_examples}

\end{figure*}

\section{Additional Dataset Examples}
\label{sec:dataset_examples}

\begin{figure*}[p]
    \centering

    % ============================================================
    % ROW 1
    % ============================================================
    \promptpair{
    Ethereal fantasy emoticons, foxes in whimsical poses,
    floating in luminescent mist with glowing flora, soft golden-hour
    light, vibrant colors, ....
    }
    {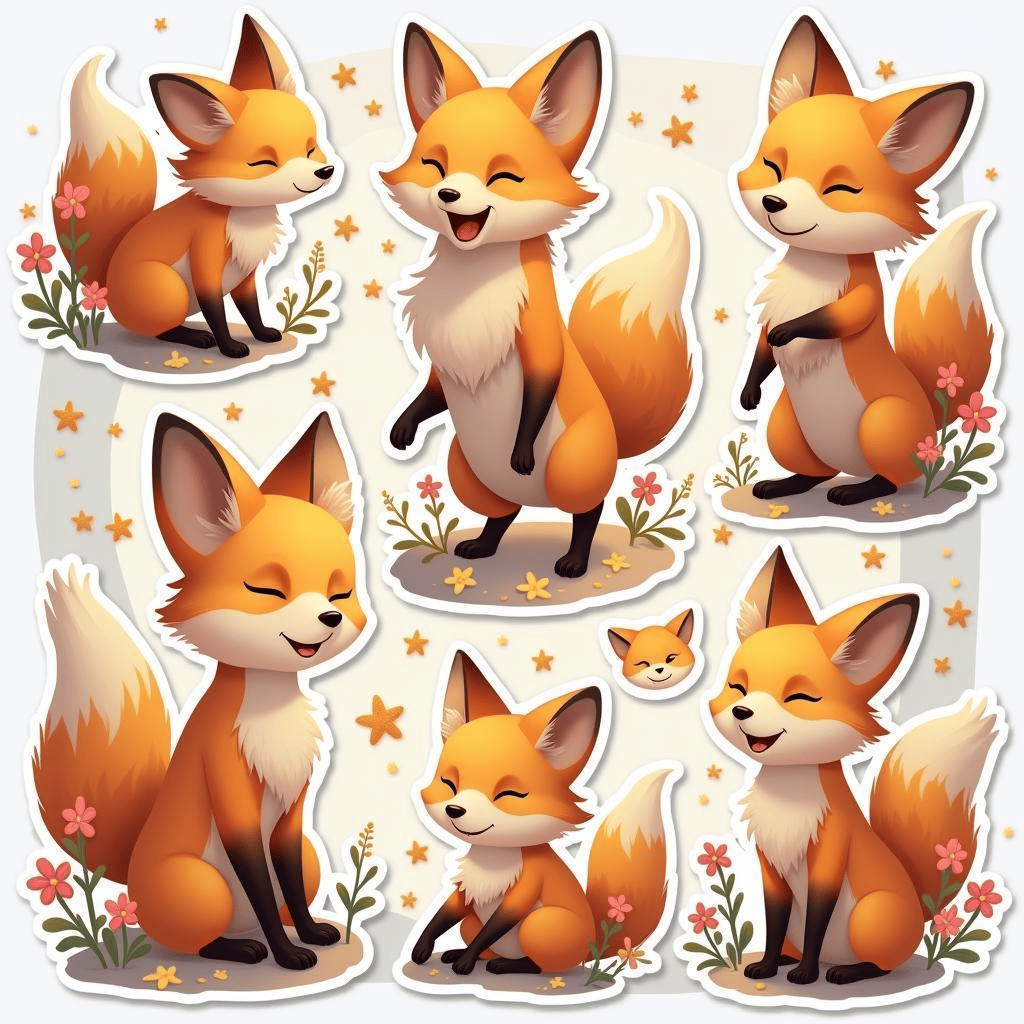}
    {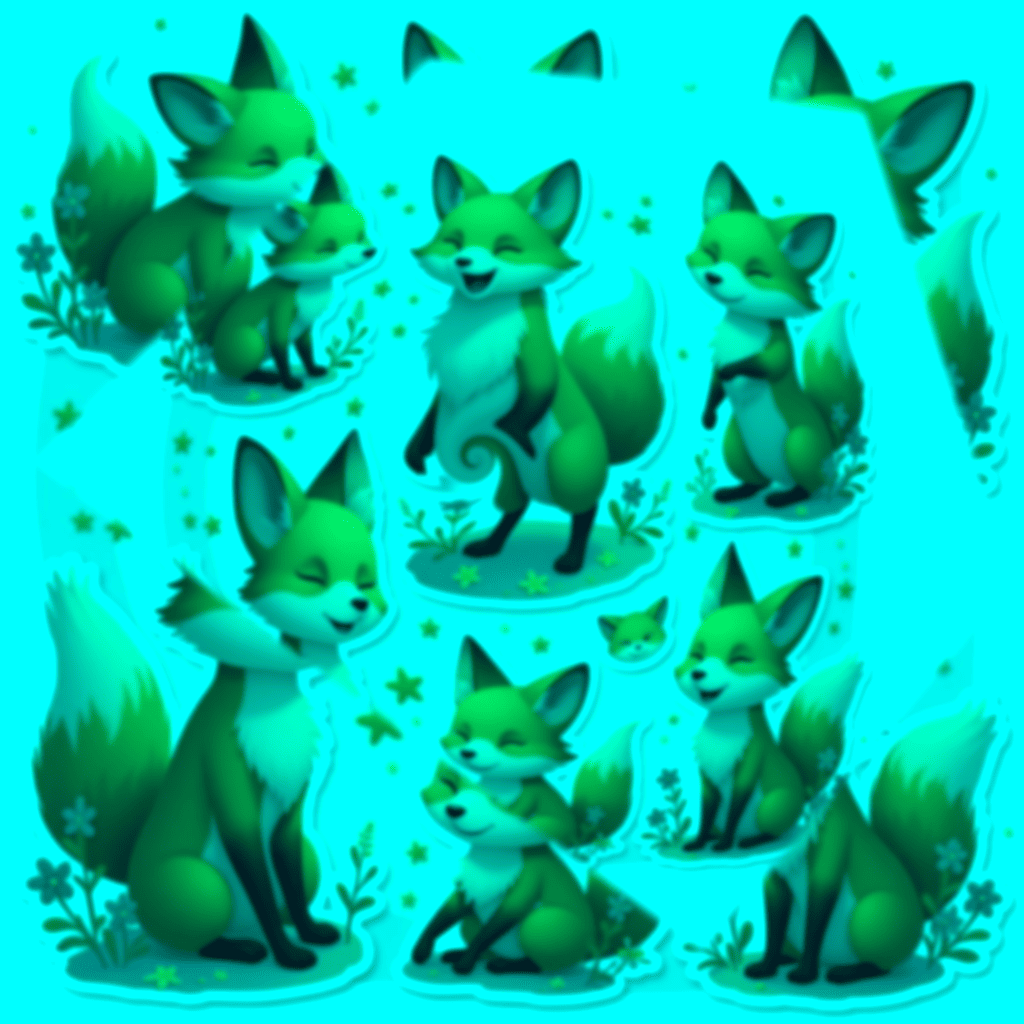}
    \hfill
    \promptpair{
    A pile of cash on a stone floor, anime key visual,
    vibrant colors, dynamic shadows, intricate details,
    dramatic chiaroscuro lighting, ....
    }
    {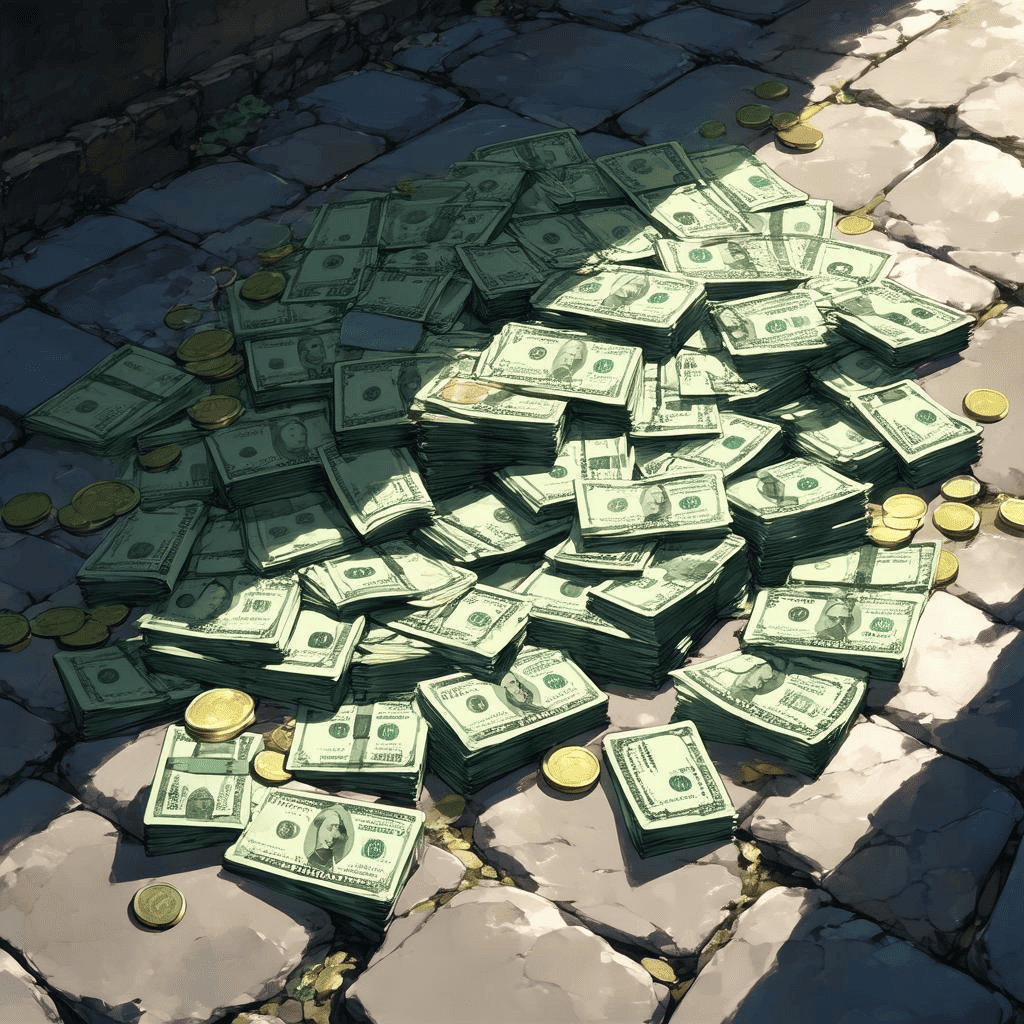}
    {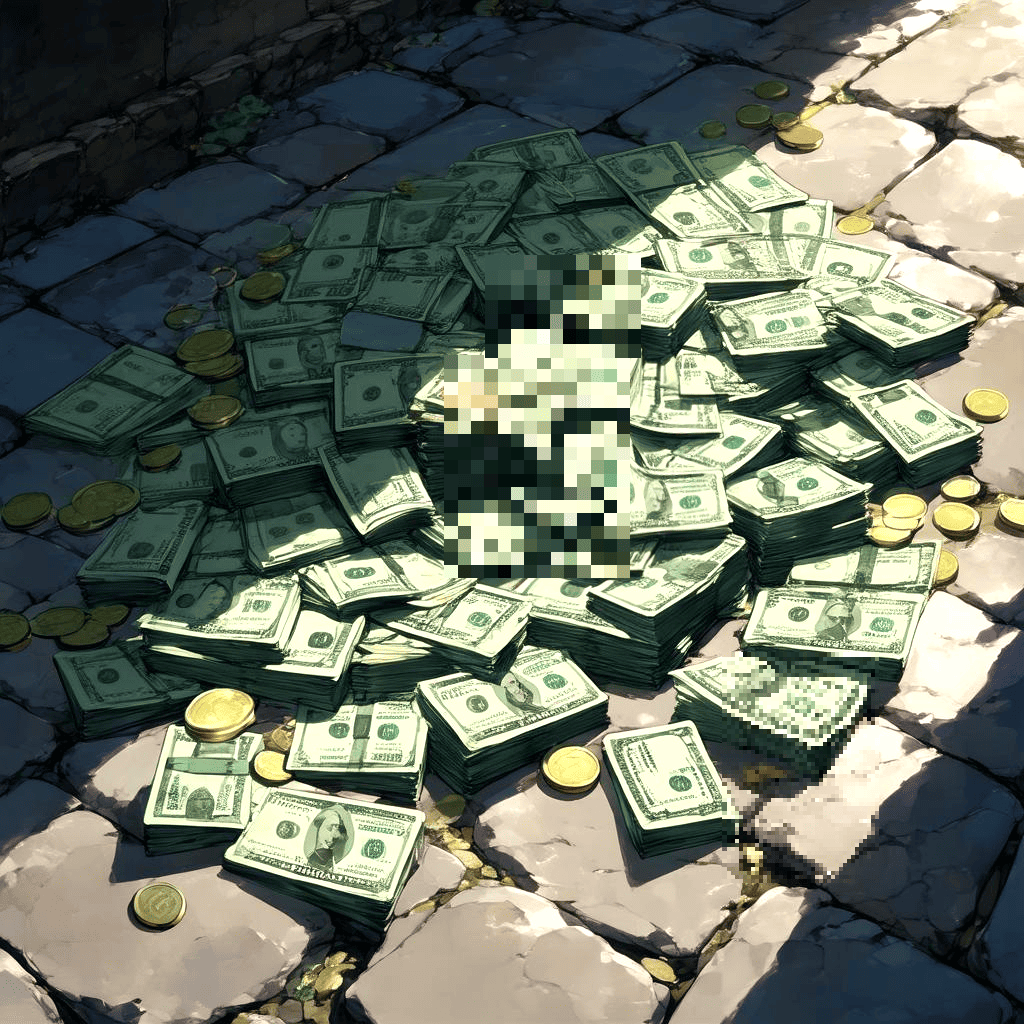}
    \hfill
    \promptpair{
    A golf bag with clubs, painted with vibrant
    digital brushstrokes.
    }
    {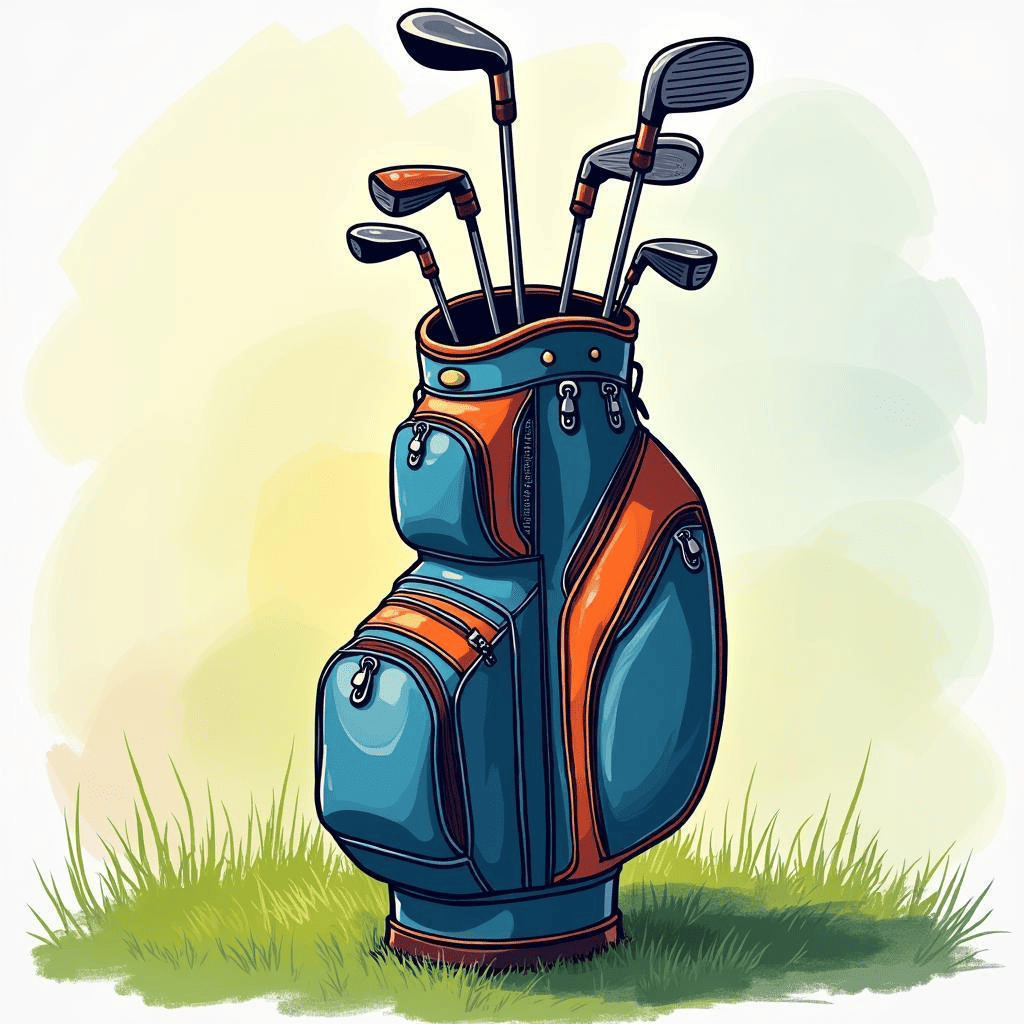}
    {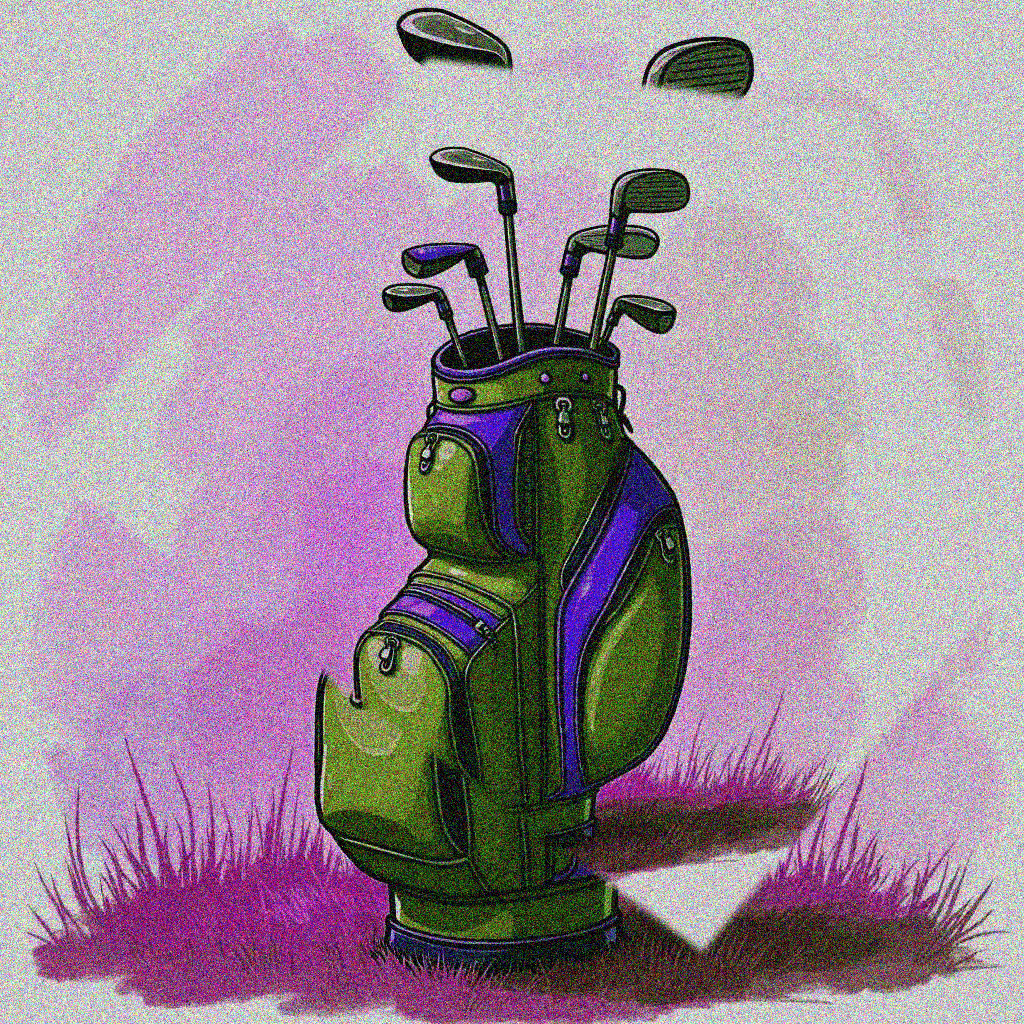}

    \vspace{29pt}

    % ============================================================
    % ROW 2
    % ============================================================
    \promptpair{
    Cinematic moody steampunk baroque Mad Max cars,
    Rocket League, GTA 5 cover, dynamic perspective,
    intricate details, vibrant colors, ....
    }
    {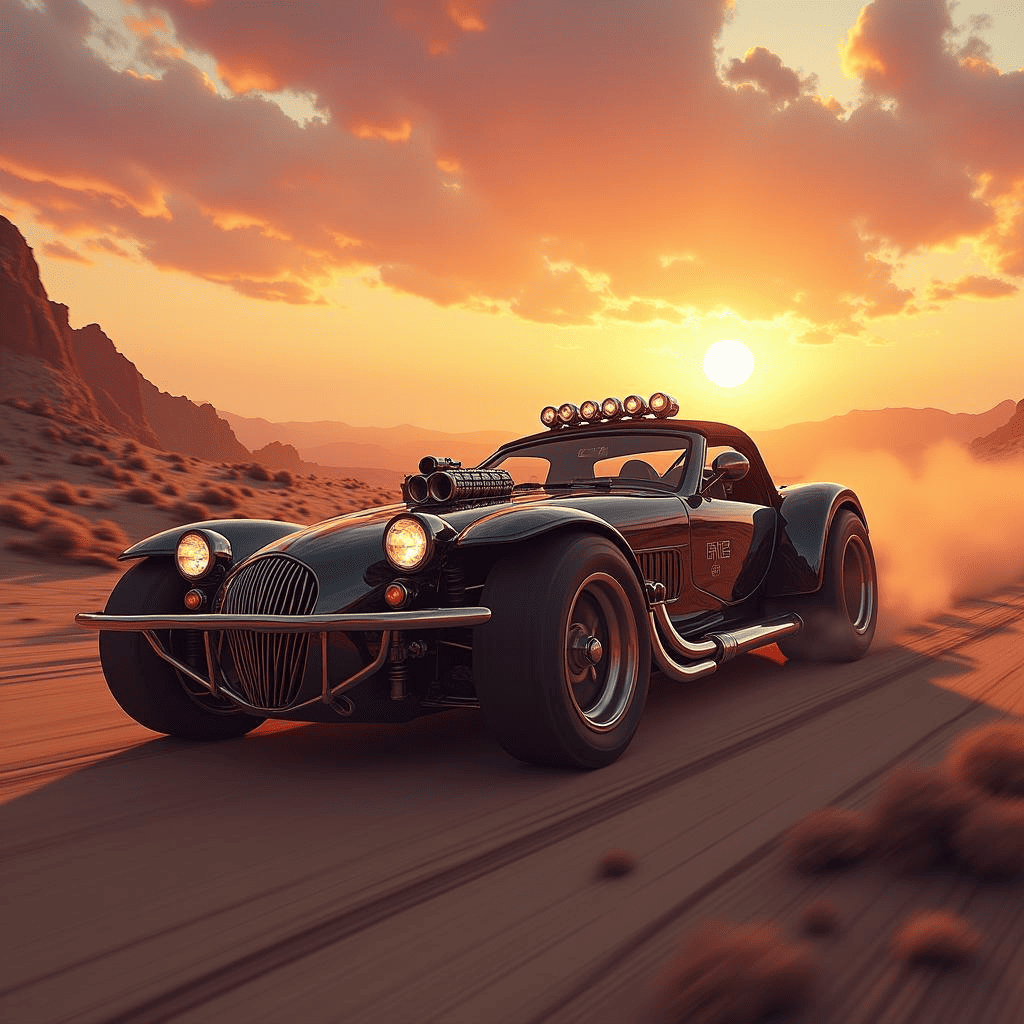}
    {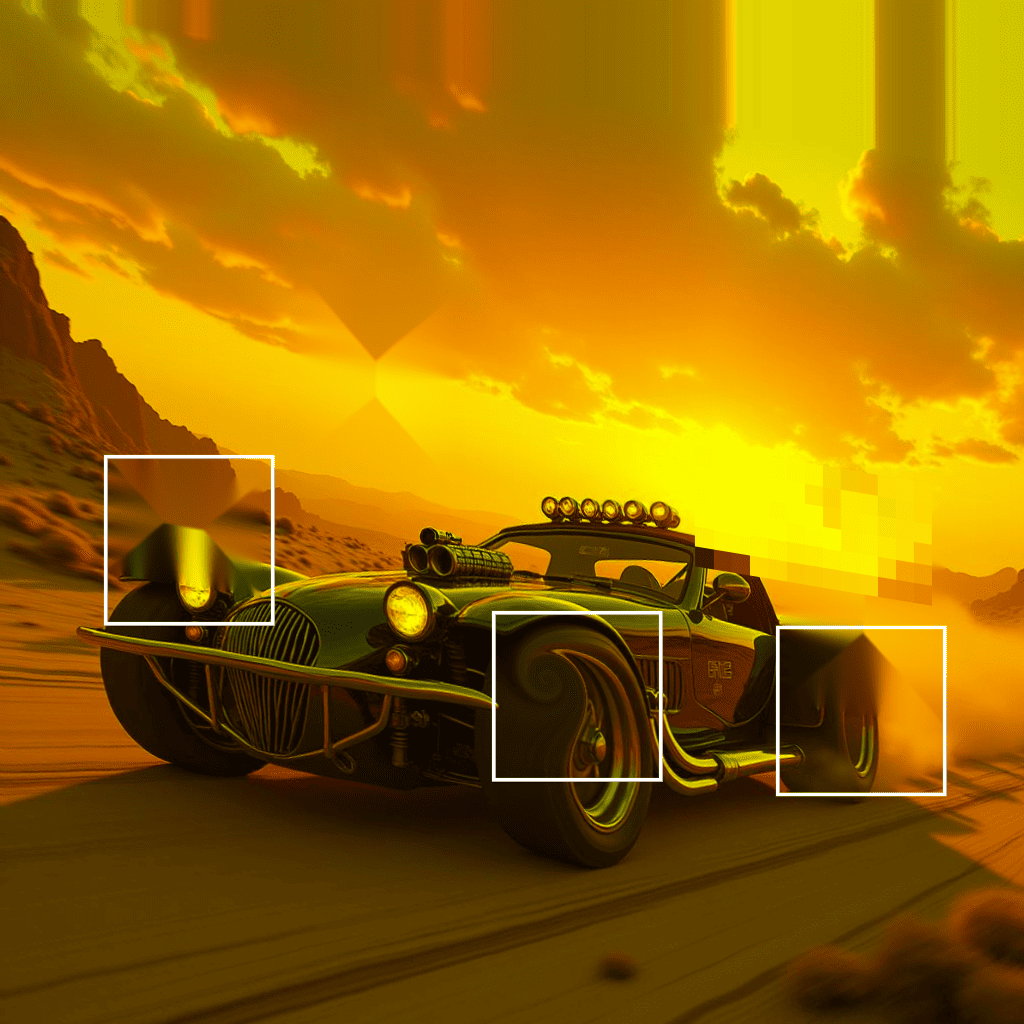}
    \hfill
    \promptpair{
    An intricate neonpunk phonograph crafted from layered
    cardboard, glowing with vibrant neon lights against
    a cybernetic backdrop, ....
    }
    {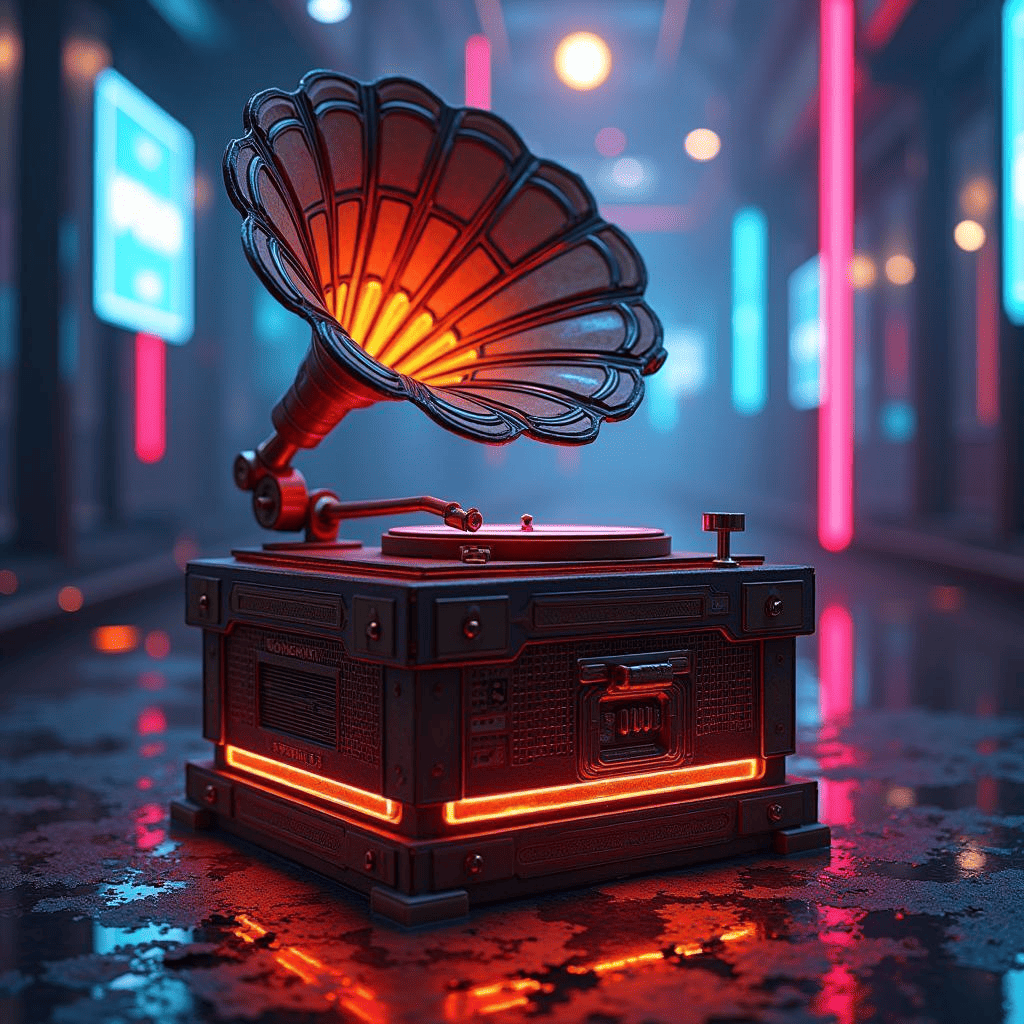}
    {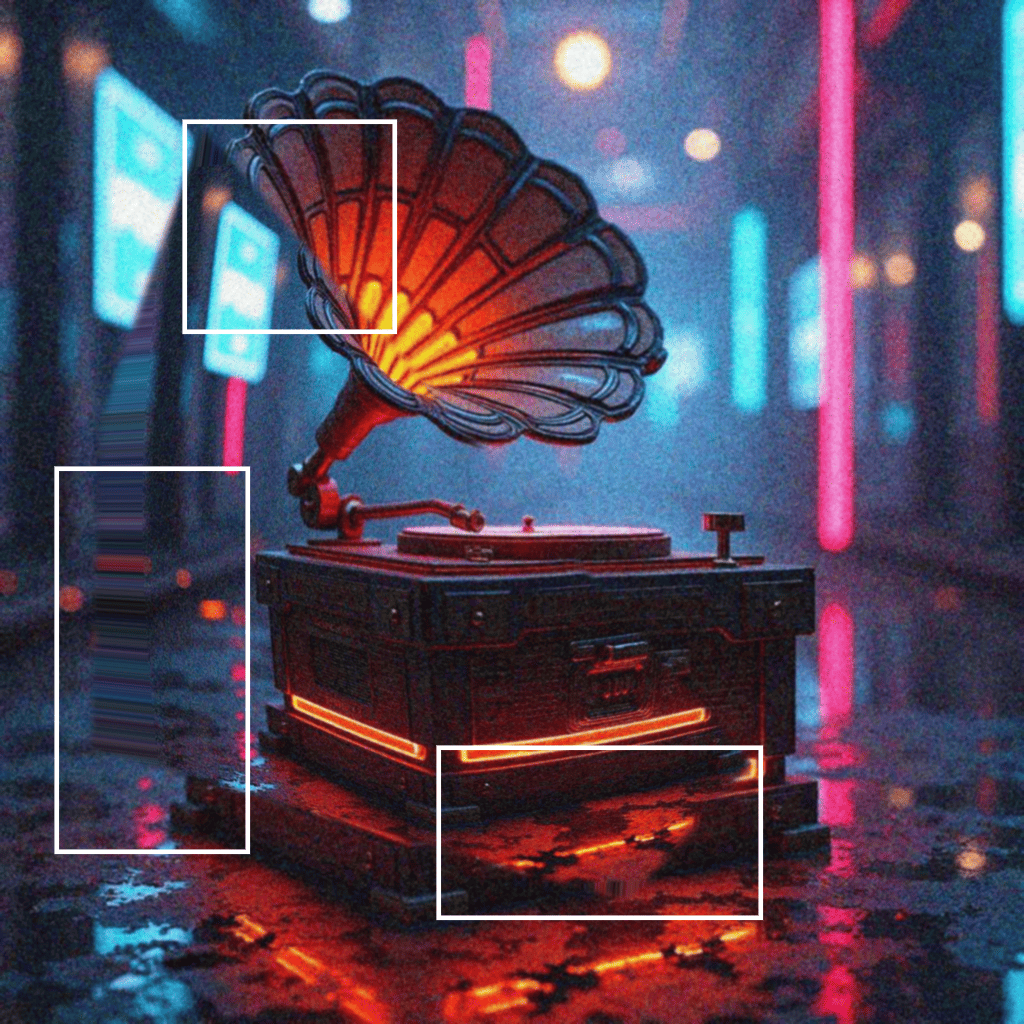}
    \hfill
    \promptpair{
    A fantasy poster of three ancient trees,
    glowing with a mystical aura in an enchanted forest.
    }
    {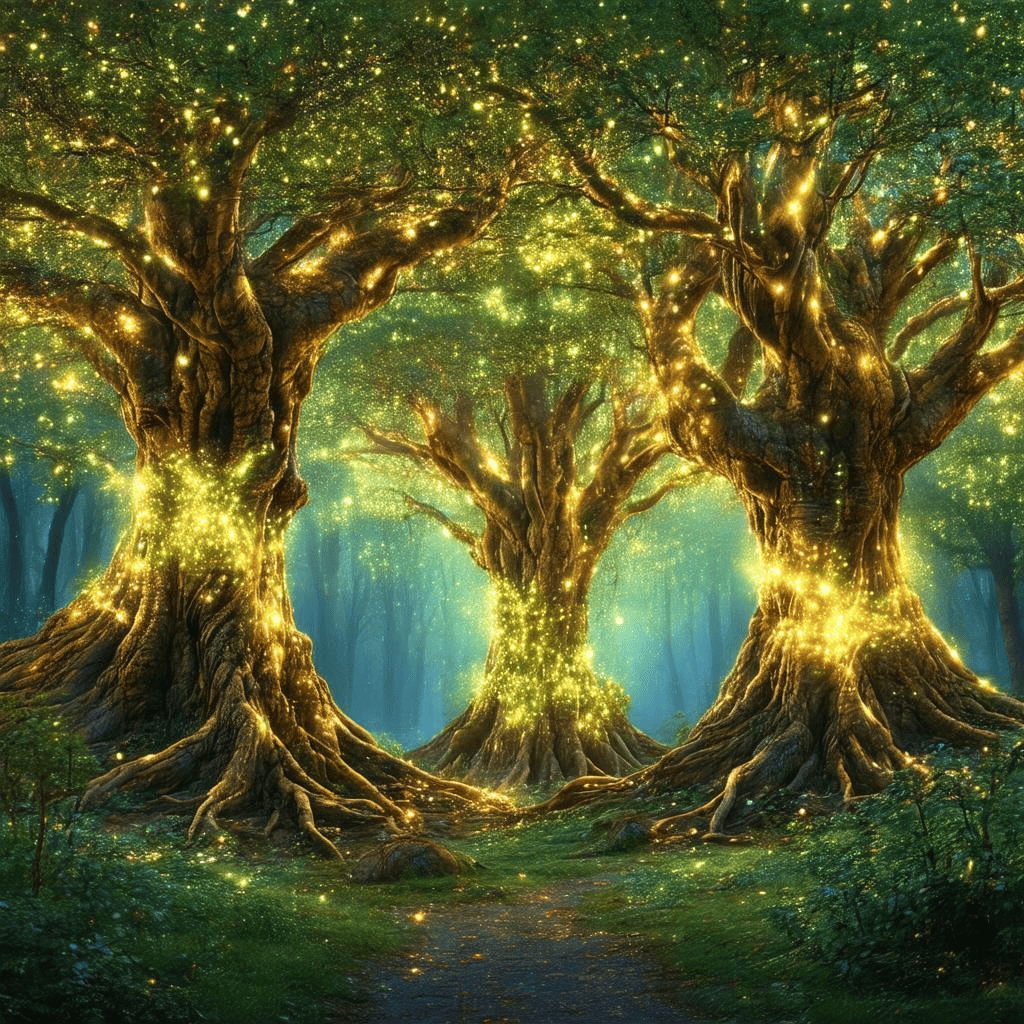}
    {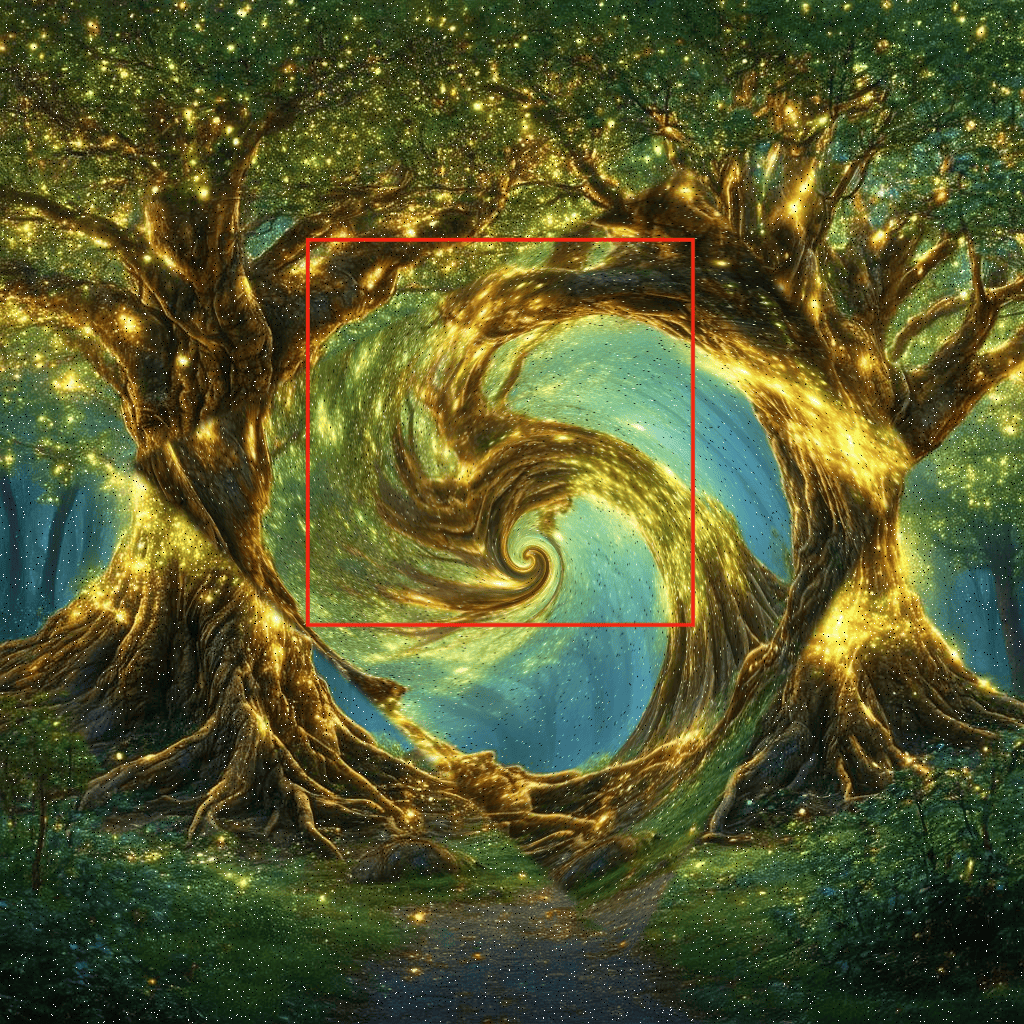}

    \vspace{29pt}

    % ============================================================
    % ROW 3
    % ============================================================
    \promptpair{
    Fantasy art of the Grim Reaper with a cloak of
    fireflies and flames, golden glowing scythe,
    intricate stained-glass background, ....
    }
    {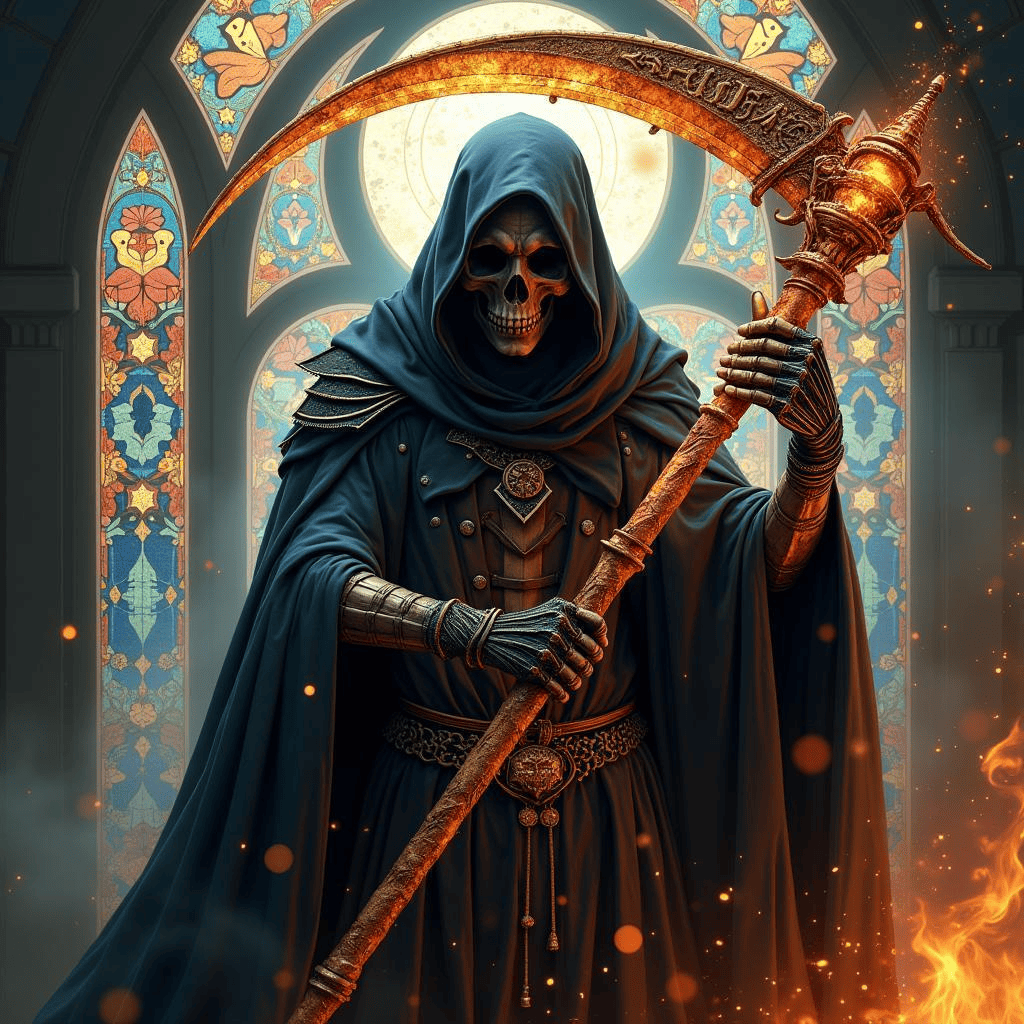}
    {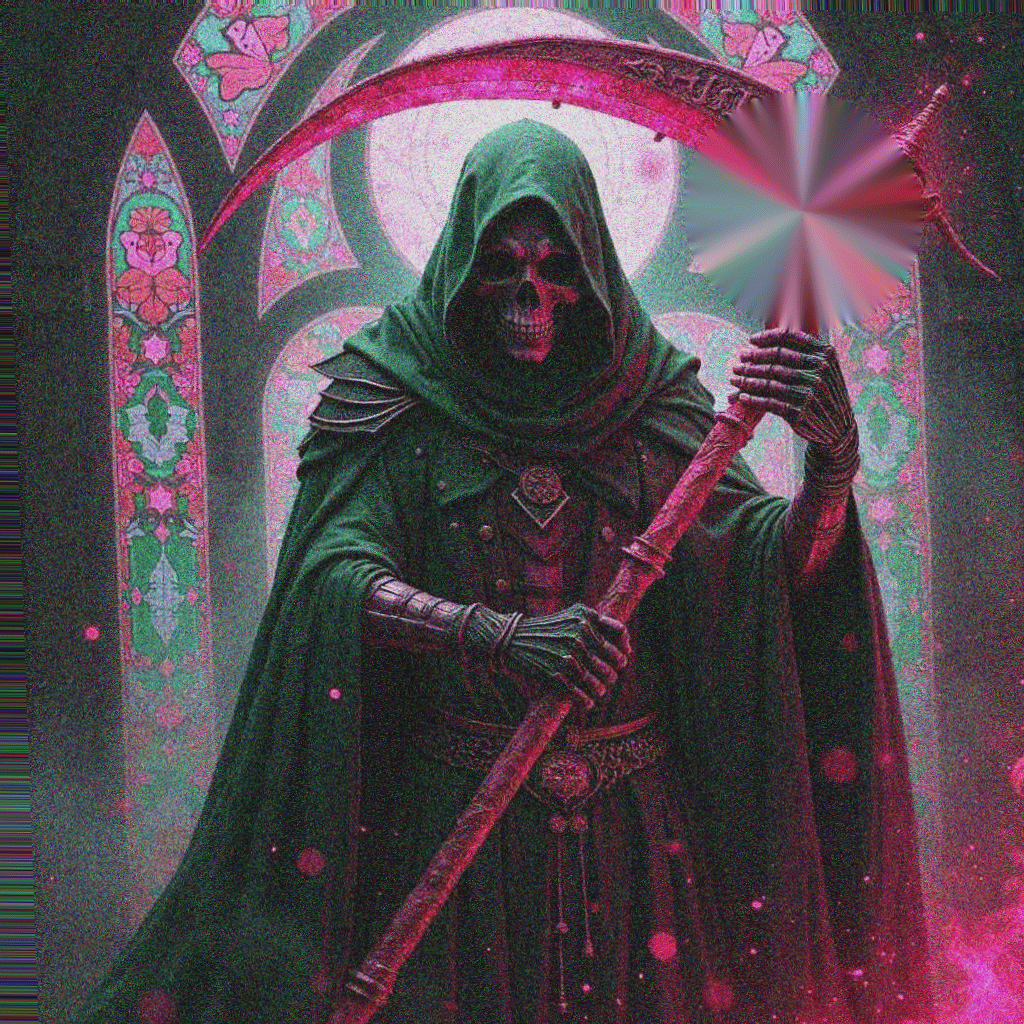}
    \hfill
    \promptpair{
    A vibrant matte painting of a coiled green snake,
    its scales gleaming with iridescent hues against
    a surreal otherworldly landscape.
    }
    {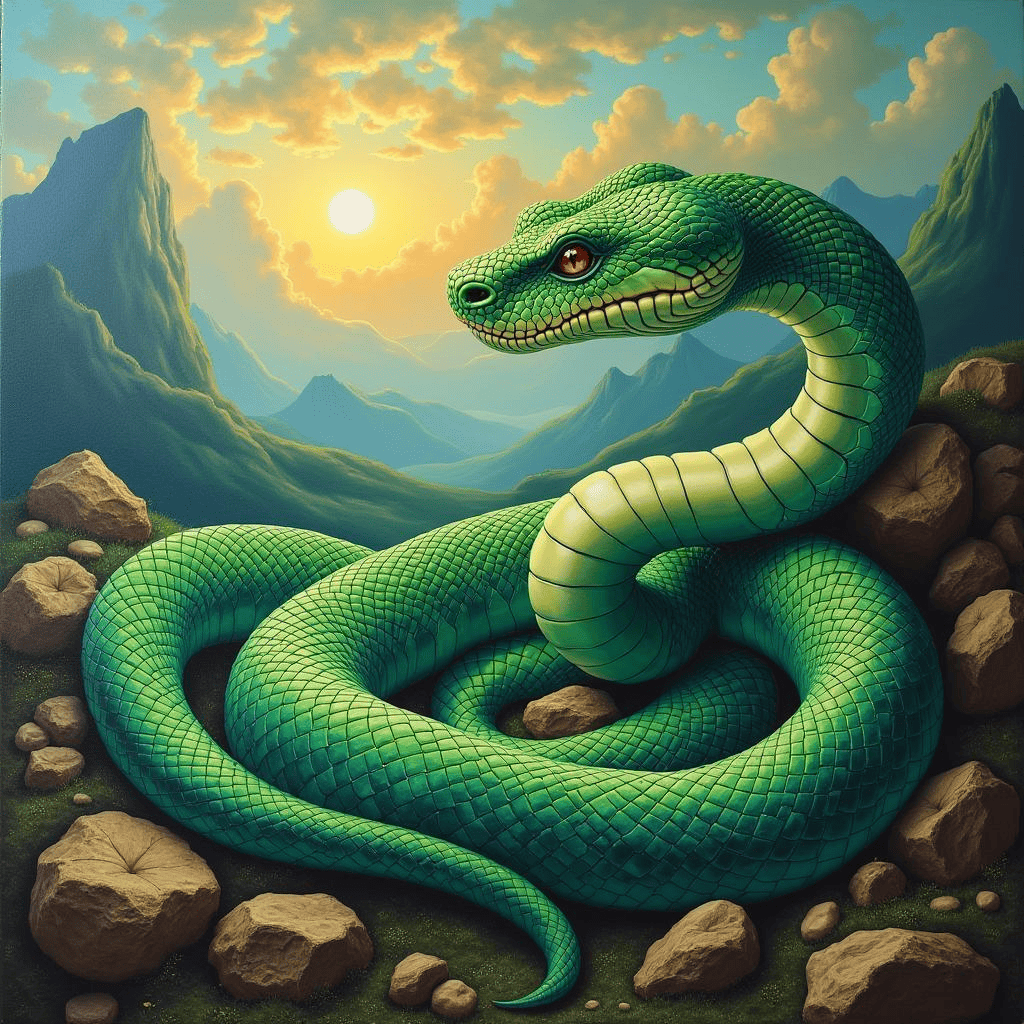}
    {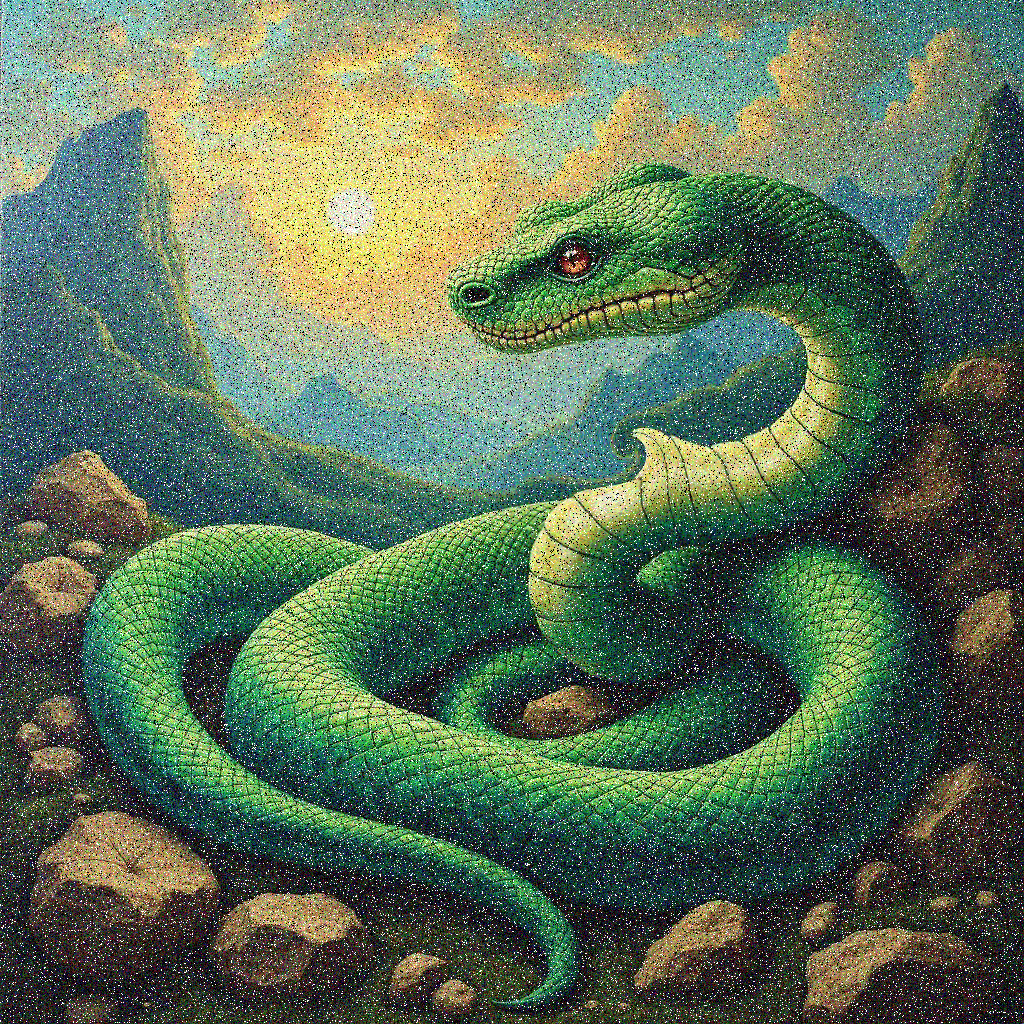}
    \hfill
    \promptpair{
    A detailed manga-style giant Lovecraftian mechanized
    turtle on a desert road, full-body illustration,
    soft golden-hour light, intricate linework, ....
    }
    {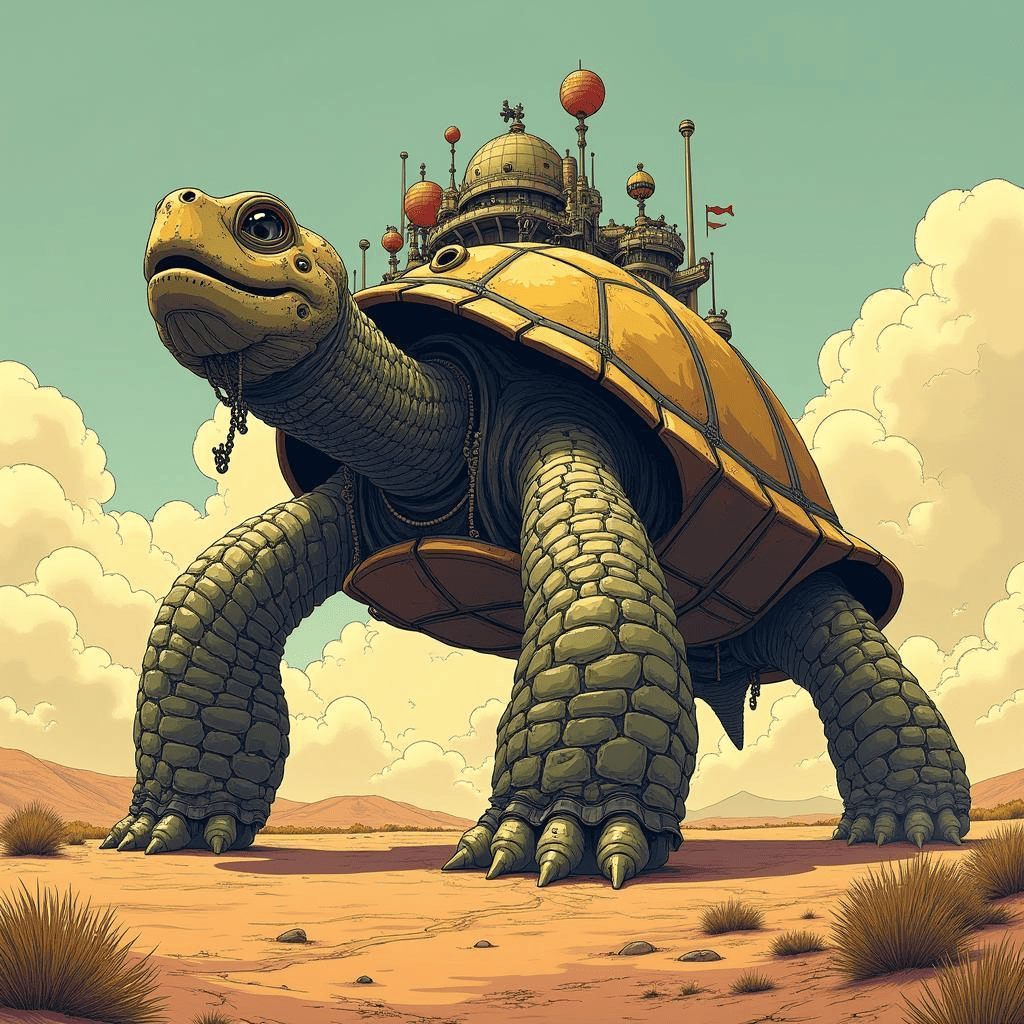}
    {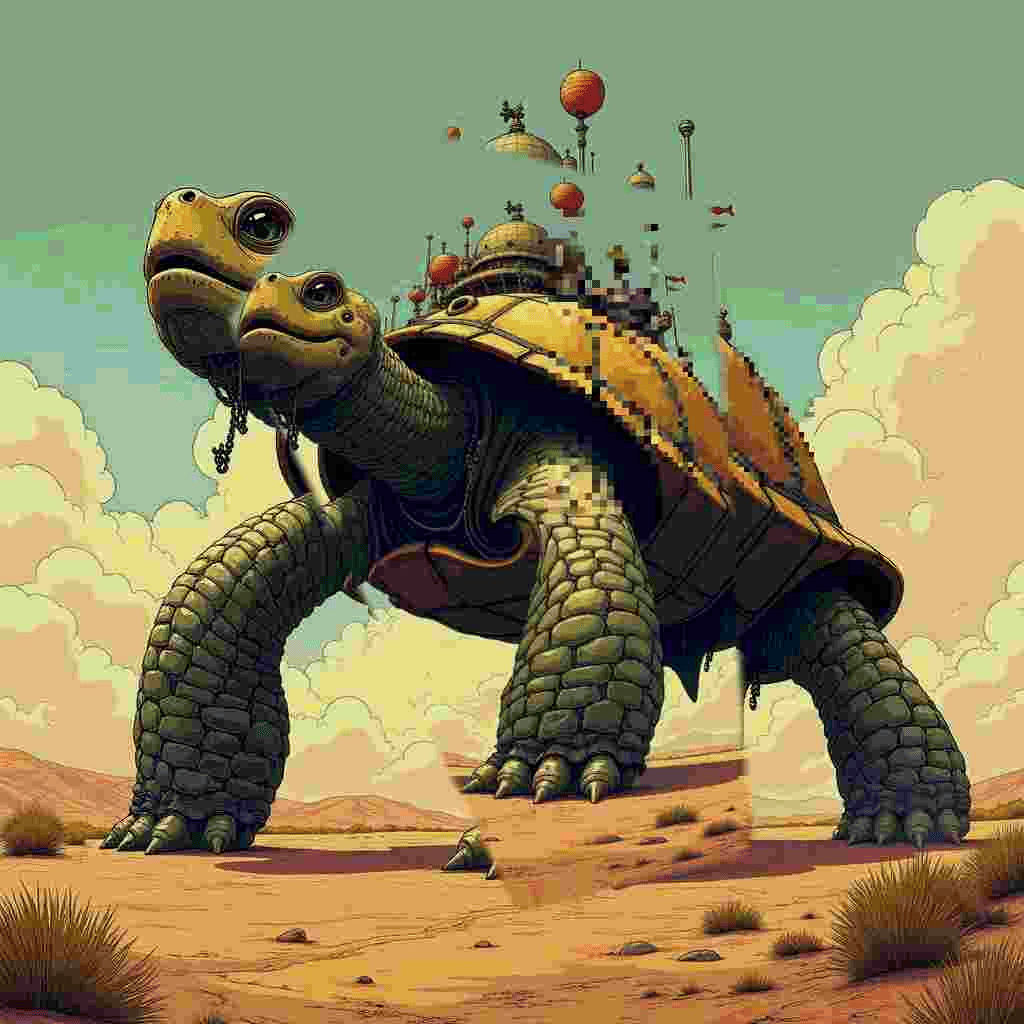}

    \vspace{29pt}

    % ============================================================
    % ROW 4
    % ============================================================
    \promptpair{
    A digital illustration of an orange cartoon cat
    with a looping tail on a light-blue starry backdrop,
    with ``CUTRA CHARDLE'' text below, ....
    }
    {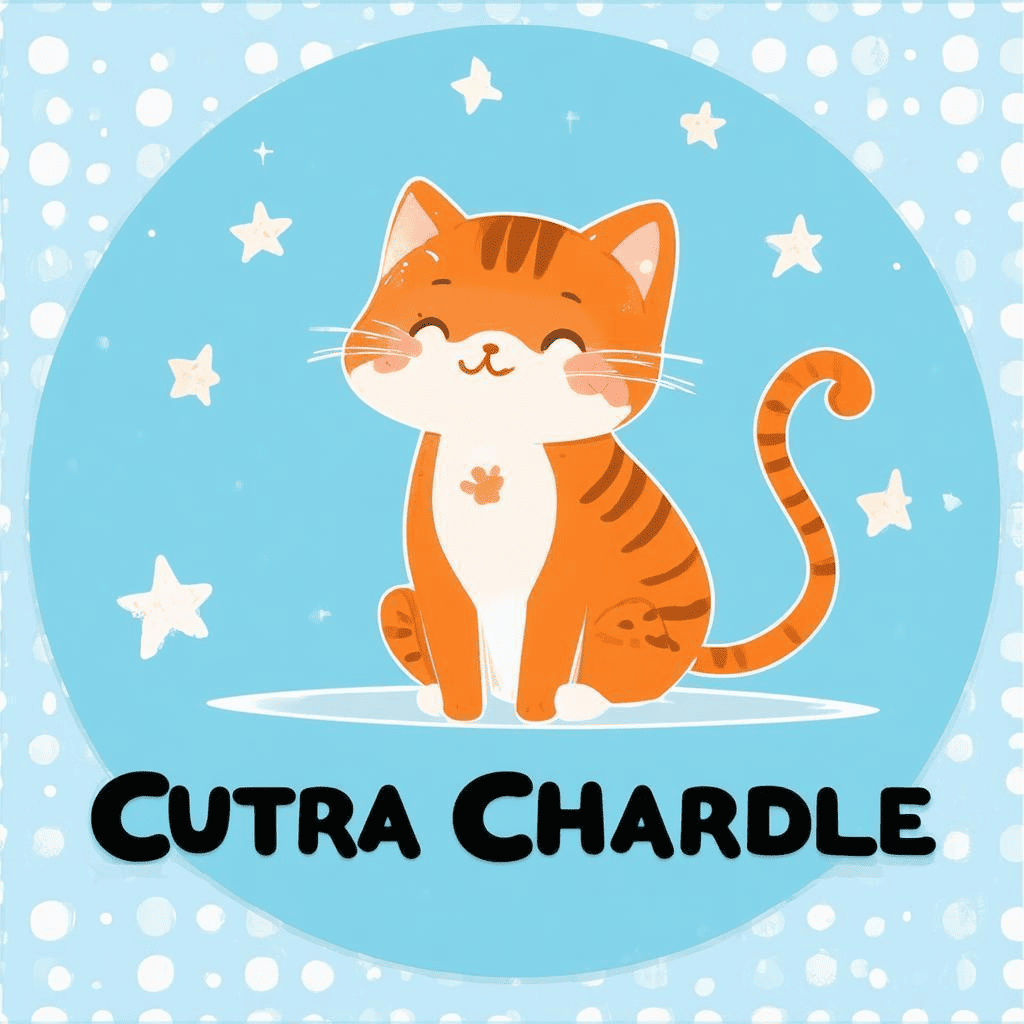}
    {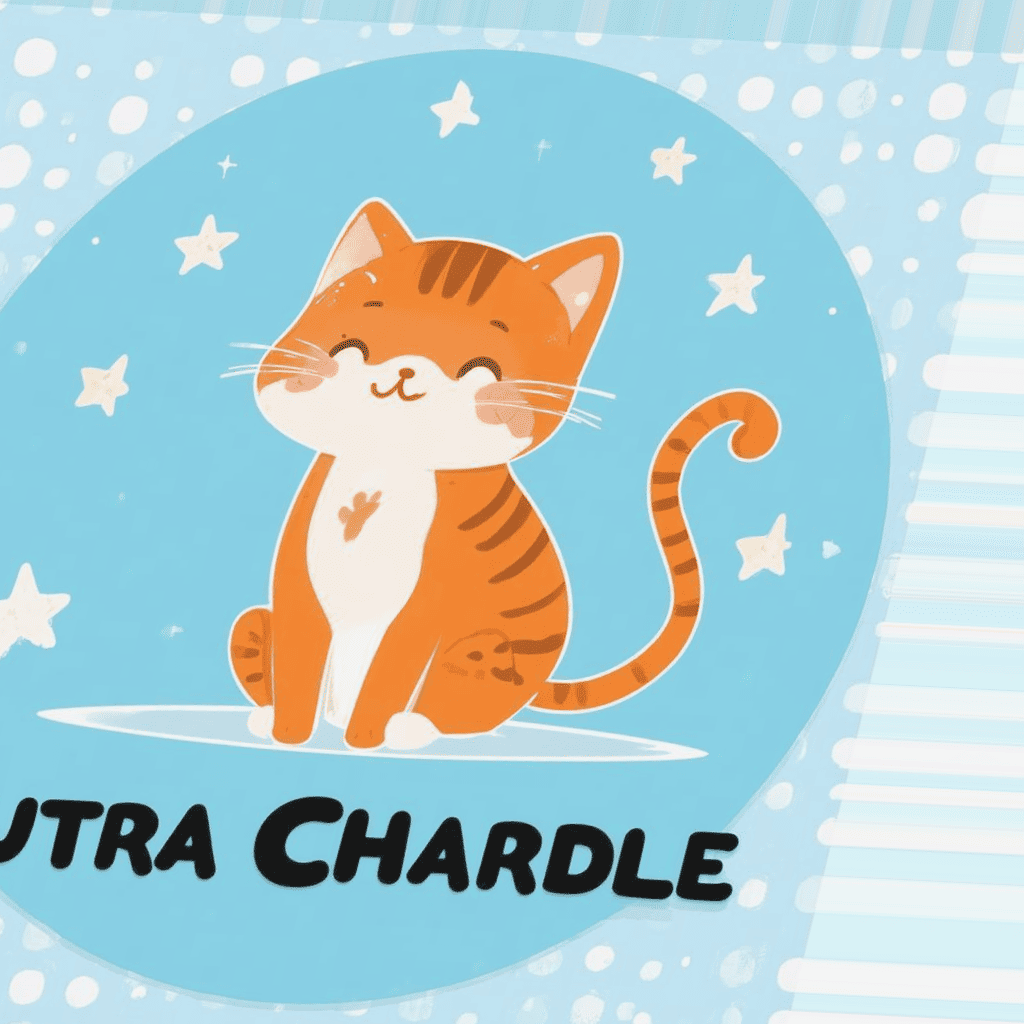}
    \hfill
    \promptpair{
    Kommissar Tarsis, anime style, pale skin, navy hat,
    imperial guard, grimdark, sharp eyes, symmetrical face,
    vibrant colors, ....
    }
    {Images/output_images/win_579912.png}
    {Images/output_images/lose_579912.png}
    \hfill
    \promptpair{
    A manga-style illustration of a whimsical straw figure
    with a broom, surrounded by swirling leaves against
    a vibrant yellow backdrop, ....
    }
    {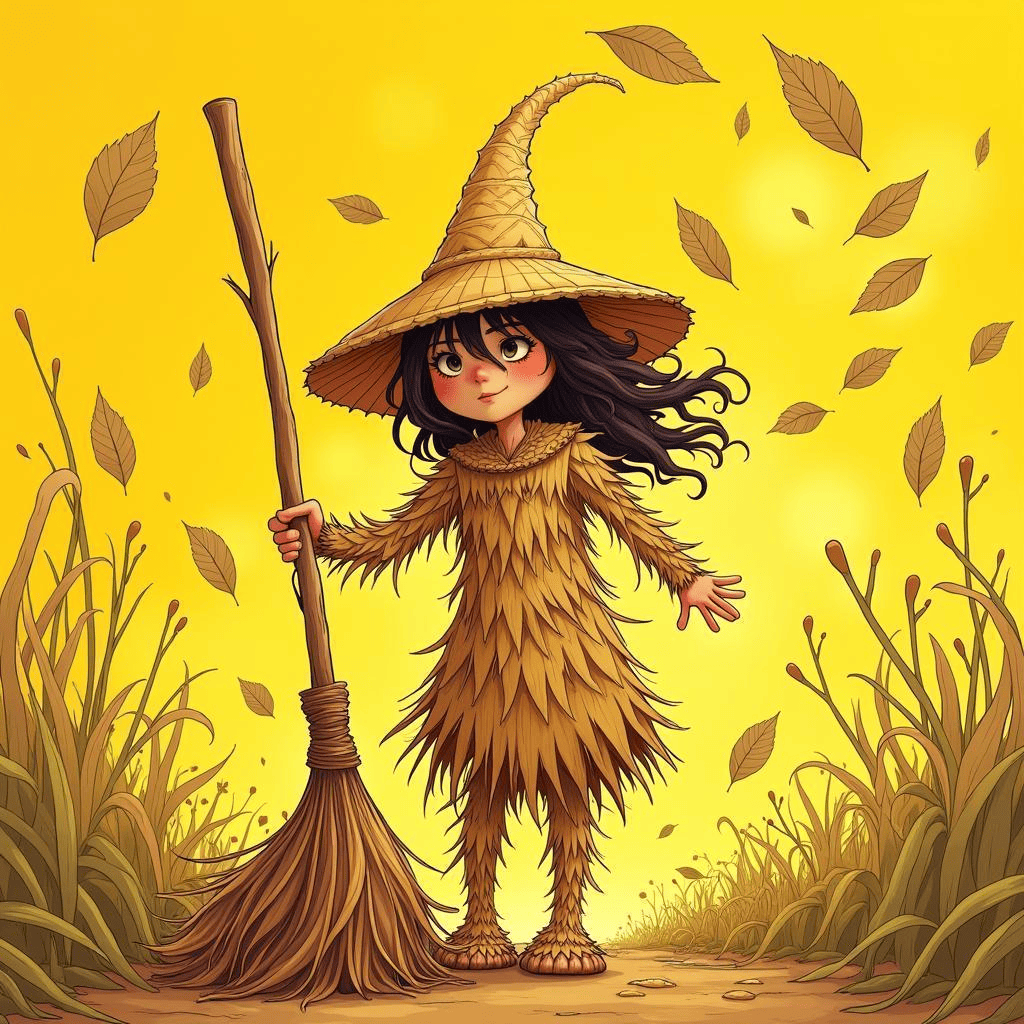}
    {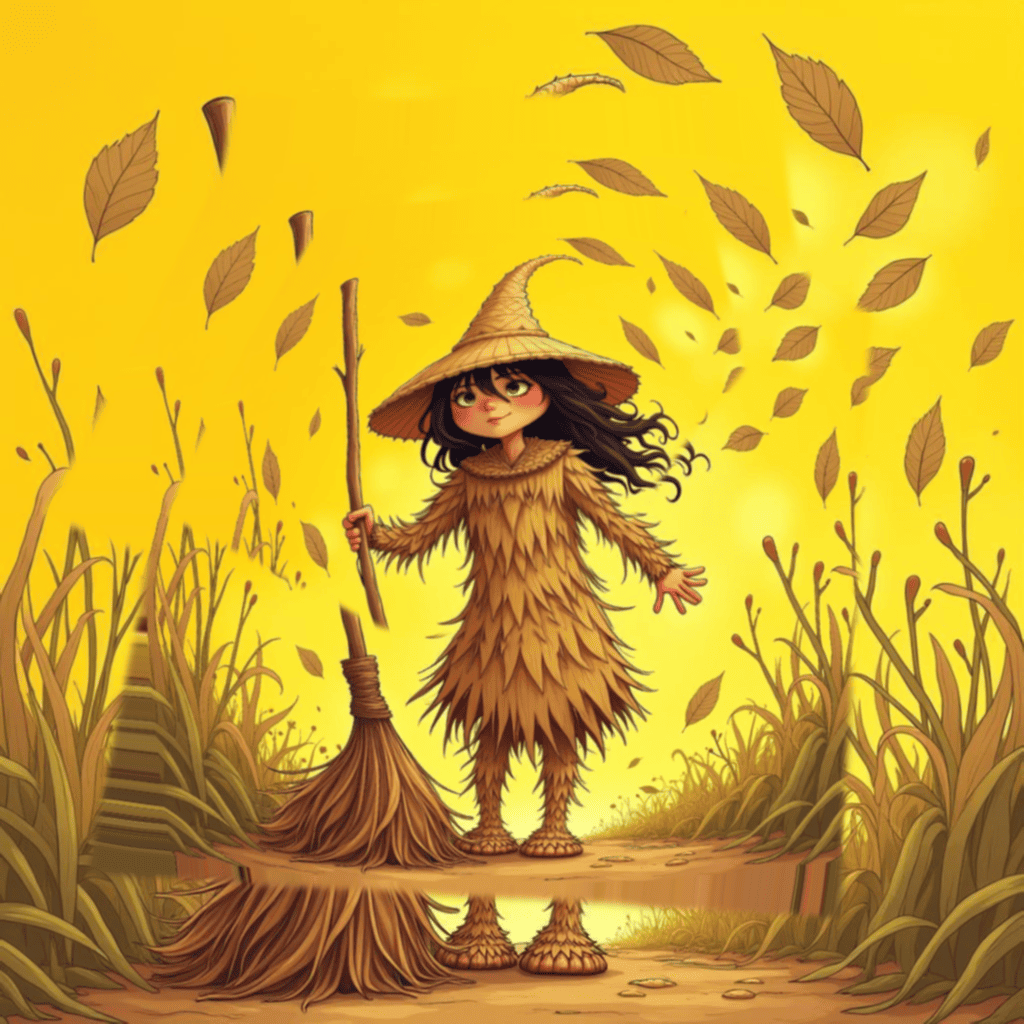}

    % ============================================================
    % ROW 5 -- ONLY TWO PAIRS
    % ============================================================
    \vspace{29pt}

    \makebox[\textwidth][c]{%
        \promptpair{A 3D model of a scientist, dramatic chiaroscuro lighting, intricate textures, .....}{Images/output_images/win_126500.png}{Images/output_images/lose_126500.png}
    \hspace{12pt}
    \promptpair{A yellow wall with Hello, world. in bold, cinematic letters .....}{Images/output_images/win_8996.png}{Images/output_images/lose_8996.png}
    }

    \caption{
    \textbf{Examples of preference pairs constructed with
    \textsc{PrefLens} across diverse prompts.}
    Each example shows the human-preferred anchor (Win) together with a
    recovered local alternative (Lose), illustrating perceptual differences
    in appearance, structure, and visual fidelity while largely retaining
    the underlying semantic content. \textbf{We recommend zooming in on the figure for better visualization
of the local variations.}
    }
    \label{fig:dataset_examples_appendix}
\end{figure*}

We provide additional examples from the \textsc{PrefLens} dataset in
Figure~\ref{fig:dataset_examples_appendix}, complementing the representative
examples shown in the main paper. Each example illustrates a human-preferred
anchor together with recovered local alternatives selected for preference
training. The examples span different prompts and visual content, and show
how the constructed pairs capture variations in appearance, structure, and
visual fidelity while largely retaining the underlying content of the
preferred image.

\clearpage

\clearpage

\section{Additional Quantitative Results}
\label{sec:additional_quantitative}

\subsection{Human-relevant preference principles}

A key question is whether the preference comparisons recovered by
\textsc{PrefLens} reflect meaningful human preference structure beyond the
reward signals used during neighborhood construction. To examine this, we
use ICAI to recover interpretable preference principles from different
human-preference datasets and compare their resulting category
distributions. This analysis provides a complementary view of the data by
characterizing the types of preference criteria represented in each
dataset, rather than only measuring downstream model performance.

To organize the recovered ICAI principles, we group them into six
interpretable preference factors according to the dominant criterion
expressed by each principle. \emph{Visual Quality} captures overall
clarity and perceptual quality; \emph{Composition} covers layout, balance,
and spatial arrangement; \emph{Lighting \& Color} captures illumination,
contrast, and color characteristics; \emph{Structural Fidelity} concerns
object geometry, anatomy, and structural correctness; \emph{Prompt \&
Style} covers semantic adherence, attributes, and requested visual style;
and \emph{Artifact Avoidance} captures undesirable generation artifacts
and visual defects. Each principle is assigned to a single primary factor
based on its predominant preference criterion.

\begin{table}[H]
\centering

\resizebox{\linewidth}{!}{
\begin{tabular}{lcccc}
\toprule
\textbf{Preference Factor} &
\textbf{HPDv2} &
\textbf{Pick-a-Pic-v2} &
\textbf{OpenPref} &
\textbf{\textsc{PrefLens}} \\
\midrule

Visual Quality
& 8.33\%
& 14.29\%
& 4.08\%
& 18.60\% \\

Composition
& 25.00\%
& 17.86\%
& 28.57\%
& 18.60\% \\

Lighting \& Color
& 25.00\%
& 28.57\%
& 28.57\%
& 22.09\% \\

Structural Fidelity
& 8.33\%
& 10.71\%
& 8.16\%
& 17.44\% \\

Prompt \& Style
& 25.00\%
& 25.00\%
& 30.61\%
& 13.95\% \\

Artifact Avoidance
& 8.33\%
& 3.57\%
& 0.00\%
& 9.30\% \\

\midrule
\textbf{Total principles ($N$)}
& \textbf{12}
& \textbf{56}
& \textbf{49}
& \textbf{86} \\

\bottomrule
\end{tabular}
}

\caption{
\textbf{Distribution of ICAI preference principles across datasets.}
Each extracted principle is assigned to one primary preference factor
according to its dominant preference criterion. Entries report the
percentage of extracted principles associated with each factor, normalized
independently within each dataset. The final row reports the total number
of extracted principles for each dataset.
}
\label{tab:icai_principle_distribution}
\end{table}

In Table~\ref{tab:icai_principle_distribution}, we provide the complete
category-wise distributions of preference principles recovered using ICAI
from HPDv2, Pick-a-Pic-v2, OpenPref, and \textsc{PrefLens}. While the main
paper presents representative principles and their broad correspondence,
the results reported here provide a more complete view of the preference
characteristics identified across the four datasets.

\subsection{Extended benchmark results.}
\label{sec:extended_benchmark_results}
We provide additional quantitative results complementing the evaluations
reported in the main paper. In particular, we report results on
\emph{PartiPrompts} and \emph{GenAIBench}, in addition to
Pick-a-Pic-v2. We evaluate SD v1.5 and SDXL-Base using Aesthetic Score,
ImageReward, HPSv2, and PickScore. Across
benchmarks and metrics, \textsc{PrefLens}-based training consistently
improves over training with the original OpenPref comparisons. The gains
are also observed on Aesthetic Score (ViT-B/16), HPSv2, and PickScore,
which are not used for preference-space construction, indicating that the
improvements extend beyond the reward signals used to construct the
neighborhoods.

\begin{figure}[!htb]
    \centering

    \begin{subfigure}[t]{0.49\linewidth}
        \centering
        \includegraphics[width=\linewidth]
        {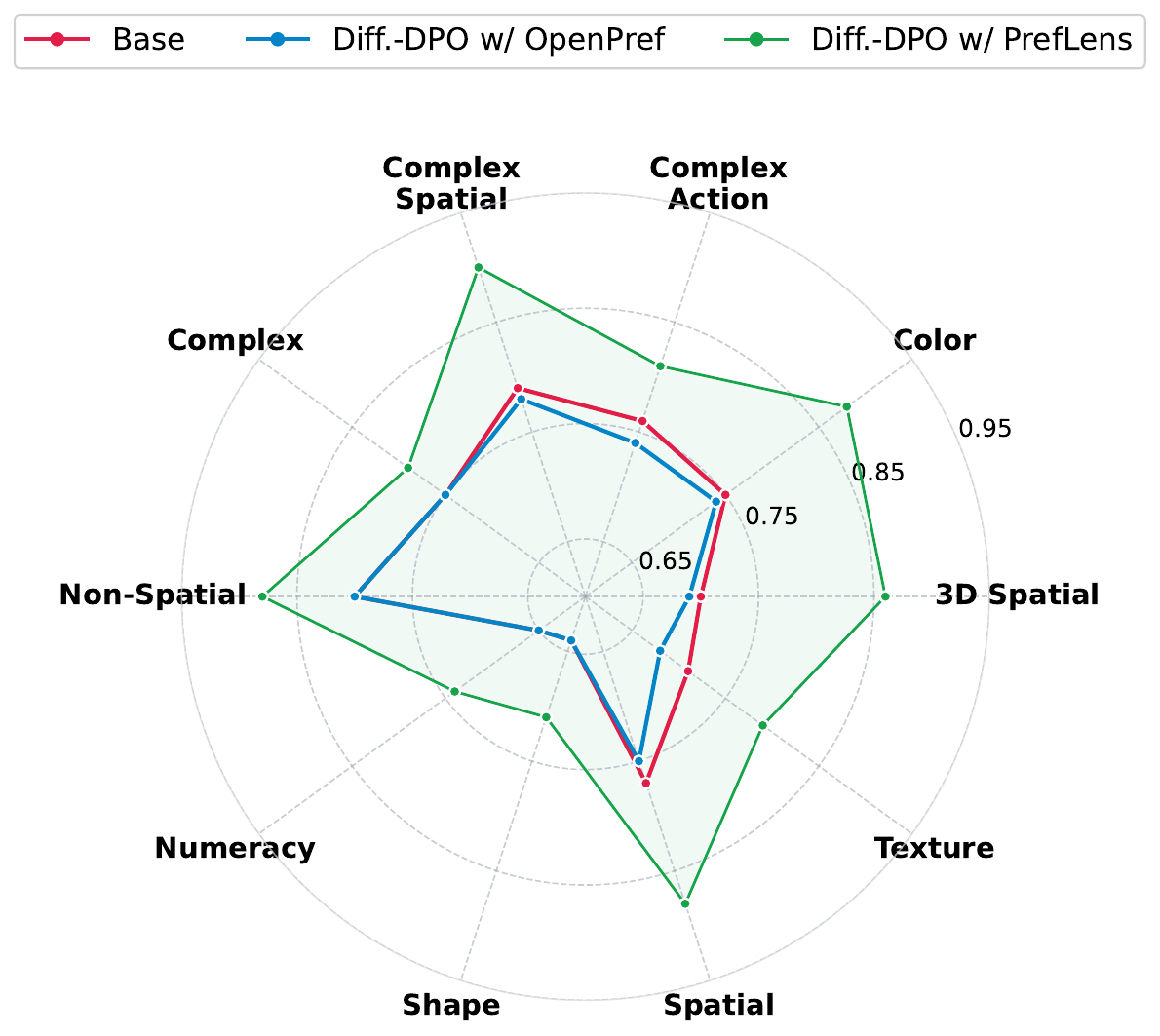}
        \caption{}
        \label{fig:tifa_blip}
    \end{subfigure}
    \hfill
    \begin{subfigure}[t]{0.48\linewidth}
        \centering
        \includegraphics[width=\linewidth]
        {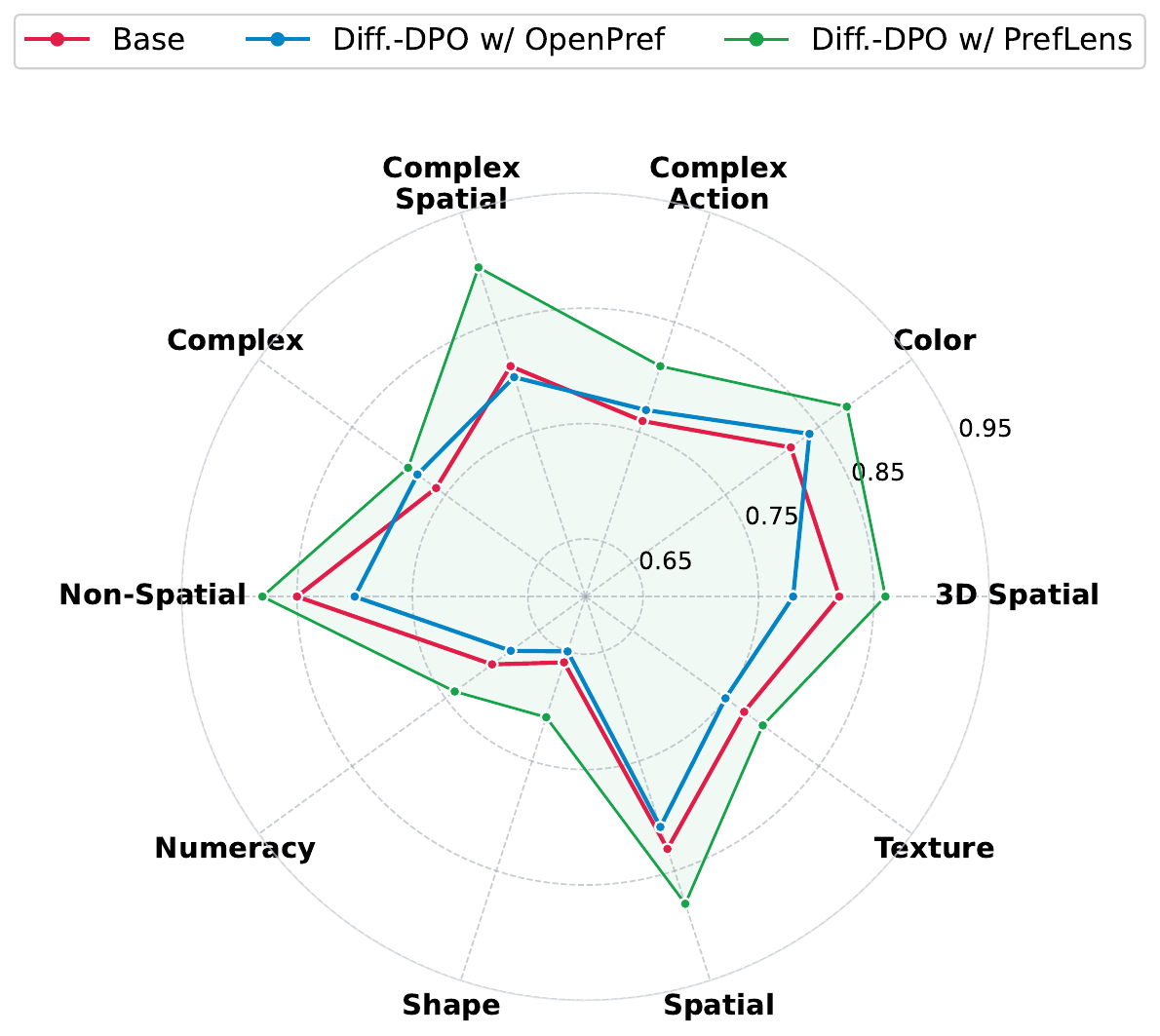}
        \caption{}
        \label{fig:tifa_git}
    \end{subfigure}

    \caption{
    \textbf{Compositional fidelity evaluation on T2I-CompBench++ for (a) SD v1.5 and (b) SDXL.}
    We report TIFA scores computed using GIT-large as VQA model.
    % across multiple compositional categories.
    }
    \label{fig:compo_plot_sdxl}
\end{figure}

We extend the compositional-fidelity evaluation reported in
the main paper. While the main paper reports results using BLIP-large as
the VQA model, Figure~\ref{fig:compo_plot_sdxl} reports the corresponding
evaluation using GIT-large.

We additionally evaluate compositional alignment using \texttt{Qwen3-VL-8B-Instruct}~\cite{bai2025qwen3} as a
VLM-as-a-Judge on 500 GenAIBench prompts. For each prompt, VLM jointly
ranks generations from the base SD v1.5 model, Diffusion-DPO trained with
OpenPref, and Diffusion-DPO trained with \textsc{PrefLens}, based on
prompt adherence and compositional correctness. \textsc{PrefLens} achieves
a Top-1 win rate of $43.8\%$, compared with $29.8\%$ for the base model
and $26.4\%$ for Diffusion-DPO with OpenPref. The same trend is reflected
in the normalized Borda score, where \textsc{PrefLens} obtains $70.6$,
compared with $65.3$ and $64.1$ for the base model and Diffusion-DPO with
OpenPref, respectively. These results provide complementary evidence that
the alignment gains of \textsc{PrefLens} extend to compositional prompt
adherence under a joint VLM-based ranking evaluation.

% \clearpage
\begin{table*}[p]
\centering
\resizebox{0.72\linewidth}{!}{
\setlength{\tabcolsep}{6pt}
\begin{tabular}{
c l
p{1.8cm} p{1.8cm}
p{1.8cm} p{1.8cm} p{1.8cm}
p{1.5cm}}
\toprule
\multirow{3}{*}{\textbf{Benchmark}} &
\multirow{3}{*}{\textbf{Method}} &
\multicolumn{2}{c}{\textbf{Image Quality}} &
\multicolumn{3}{c}{\textbf{Human Preference}} &
\multirow{2}{*}{\centering\arraybackslash\textbf{Average}} \\
\cmidrule(lr){3-4}
\cmidrule(lr){5-7}
& &
\centering\textbf{Aesthetics} &
\centering\textbf{Aesthetics} &
\centering\textbf{ImageReward} &
\centering\textbf{HPS v2} &
\centering\textbf{PickScore} & \\
& &
\centering\textbf{(ViT-14)} &
\centering\textbf{(ViT-16)} &
\centering\textbf{} &
\centering\textbf{} &
\centering\textbf{} & \\
\midrule

% ============================================================
% Pick-a-Pic-v2
% ============================================================
\multirow{23}{*}{\textbf{Pick-a-Pic-v2}}

& SFT
& \centering 46.8
& \centering 47.8
& \centering 49.2
& \centering 50.6
& \centering 49.8
& \centering\arraybackslash 48.8 \\

& SPIN
& \centering 64.2
& \centering 65.2
& \centering 62.6
& \centering 71.2
& \centering 66.8
& \centering\arraybackslash 66.0 \\

& SePPO
& \centering 65.6
& \centering 62.8
& \centering 63.2
& \centering 76.2
& \centering 67.4
& \centering\arraybackslash 67.0 \\

& SPO
& \centering 61.2
& \centering 62.0
& \centering 56.6
& \centering 72.8
& \centering 66.4
& \centering\arraybackslash 63.8 \\

& Diff.-DPO
& \centering 50.2
& \centering 52.2
& \centering 56.4
& \centering 61.6
& \centering 64.2
& \centering\arraybackslash 56.9 \\

\\[-1em] \cdashline{2-8}[0.4pt/1.5pt] \\[-1em]

& \textbf{Diffusion-DPO}
& & & & & & \\

& \hspace{5mm}w/ HPDv2
& \centering 50.6
& \centering 49.4
& \centering 54.4
& \centering 60.0
& \centering 53.0
& \centering\arraybackslash 53.5 \\

& \hspace{5mm}w/ Pick-a-Pic-v2
& \centering 50.6
& \centering 51.6
& \centering 50.6
& \centering 55.6
& \centering 51.8
& \centering\arraybackslash 52.0 \\

& \hspace{5mm}w/ OpenPref
& \centering 47.2
& \centering 49.4
& \centering 49.8
& \centering 53.6
& \centering 50.4
& \centering\arraybackslash 50.1 \\

& \hspace{5mm}w/ LNR
& \centering 58.6
& \centering 61.0
& \centering 58.8
& \centering 66.6
& \centering 61.8
& \centering\arraybackslash 61.4 \\

& \hspace{5mm}w/ LNR + RS
& \centering 62.2
& \centering 54.0
& \centering 60.4
& \centering 68.0
& \centering 63.2
& \centering\arraybackslash 61.6 \\

& \hspace{5mm}w/ AugMix
& \centering 42.0
& \centering 55.0
& \centering 49.0
& \centering 57.8
& \centering 37.0
& \centering\arraybackslash 48.2 \\

& \hspace{5mm}w/ AugMix-MN
& \centering 42.0
& \centering 52.6
& \centering 53.2
& \centering 63.4
& \centering 37.6
& \centering\arraybackslash 49.8 \\

& \hspace{5mm}w/ \textsc{PrefLens}
& \centering \textbf{68.8}
& \centering \underline{67.0}
& \centering \underline{66.6}
& \centering 72.4
& \centering \textbf{71.0}
& \centering\arraybackslash \underline{69.2} \\

\\[-1em] \cdashline{2-8}[0.4pt/1.5pt] \\[-1em]

& \textbf{DSPO}
& & & & & & \\

& \hspace{5mm}w/ HPDv2
& \centering 56.8
& \centering 53.8
& \centering 61.0
& \centering 71.6
& \centering 63.0
& \centering\arraybackslash 61.2 \\

& \hspace{5mm}w/ Pick-a-Pic-v2
& \centering 59.6
& \centering 58.2
& \centering 62.8
& \centering 74.4
& \centering 62.0
& \centering\arraybackslash 63.4 \\

& \hspace{5mm}w/ OpenPref
& \centering 64.2
& \centering 63.2
& \centering 59.8
& \centering 74.8
& \centering 65.8
& \centering\arraybackslash 65.6 \\

& \hspace{5mm}w/ LNR
& \centering 64.8
& \centering 65.0
& \centering 63.8
& \centering \underline{77.6}
& \centering 68.0
& \centering\arraybackslash 67.8 \\

& \hspace{5mm}w/ LNR + RS
& \centering 62.0
& \centering 59.2
& \centering 62.4
& \centering 74.8
& \centering 62.2
& \centering\arraybackslash 64.1 \\

& \hspace{5mm}w/ AugMix
& \centering 62.0
& \centering 66.0
& \centering 60.0
& \centering 75.8
& \centering 62.6
& \centering\arraybackslash 65.3 \\

& \hspace{5mm}w/ AugMix-MN
& \centering 60.0
& \centering 66.2
& \centering 61.8
& \centering 75.4
& \centering 63.0
& \centering\arraybackslash 65.3 \\

& \hspace{5mm}w/ \textsc{PrefLens}
& \centering \underline{68.0}
& \centering \textbf{71.0}
& \centering \textbf{66.8}
& \centering \textbf{80.6}
& \centering \underline{70.4}
& \centering\arraybackslash \textbf{71.4} \\

\midrule

% ============================================================
% PartiPrompt
% ============================================================
\multirow{23}{*}{\textbf{PartiPrompt}}

& SFT
& \centering 50.2
& \centering 52.8
& \centering 44.8
& \centering 48.2
& \centering 45.2
& \centering\arraybackslash 48.2 \\

& SPIN
& \centering \underline{67.6}
& \centering 66.8
& \centering 60.0
& \centering 70.8
& \centering 61.0
& \centering\arraybackslash 65.2 \\

& SePPO
& \centering 66.2
& \centering 67.0
& \centering 62.0
& \centering \textbf{81.8}
& \centering 66.5
& \centering\arraybackslash \underline{68.7} \\

& SPO
& \centering 67.4
& \centering \underline{67.8}
& \centering 56.8
& \centering 72.2
& \centering \underline{67.6}
& \centering\arraybackslash 66.4 \\

& Diff.-DPO
& \centering 62.0
& \centering 57.0
& \centering 58.2
& \centering 61.0
& \centering 64.4
& \centering\arraybackslash 60.5 \\

\\[-1em] \cdashline{2-8}[0.4pt/1.5pt] \\[-1em]

& \textbf{Diffusion-DPO}
& & & & & & \\

& \hspace{5mm}w/ HPDv2
& \centering 55.2
& \centering 54.4
& \centering 54.8
& \centering 56.2
& \centering 52.4
& \centering\arraybackslash 54.6 \\

& \hspace{5mm}w/ Pick-a-Pic-v2
& \centering 52.2
& \centering 50.0
& \centering 53.2
& \centering 52.0
& \centering 52.6
& \centering\arraybackslash 52.0 \\

& \hspace{5mm}w/ OpenPref
& \centering 53.8
& \centering 53.4
& \centering 56.2
& \centering 52.0
& \centering 54.6
& \centering\arraybackslash 54.0 \\

& \hspace{5mm}w/ LNR
& \centering 61.6
& \centering 54.8
& \centering 58.6
& \centering 62.0
& \centering 52.0
& \centering\arraybackslash 57.8 \\

& \hspace{5mm}w/ LNR + RS
& \centering 63.8
& \centering 57.2
& \centering 58.4
& \centering 69.4
& \centering 58.8
& \centering\arraybackslash 61.5 \\

& \hspace{5mm}w/ AugMix
& \centering 44.2
& \centering 56.2
& \centering 46.2
& \centering 53.4
& \centering 33.2
& \centering\arraybackslash 46.6 \\

& \hspace{5mm}w/ AugMix-MN
& \centering 47.2
& \centering 55.6
& \centering 51.4
& \centering 59.8
& \centering 33.6
& \centering\arraybackslash 49.5 \\

& \hspace{5mm}w/ \textsc{PrefLens}
& \centering 67.0
& \centering \underline{67.8}
& \centering \underline{64.8}
& \centering 74.6
& \centering 65.0
& \centering\arraybackslash 67.8 \\

\\[-1em] \cdashline{2-8}[0.4pt/1.5pt] \\[-1em]

& \textbf{DSPO}
& & & & & & \\

& \hspace{5mm}w/ HPDv2
& \centering 62.4
& \centering 58.8
& \centering 61.6
& \centering 72.0
& \centering 59.2
& \centering\arraybackslash 62.8 \\

& \hspace{5mm}w/ Pick-a-Pic-v2
& \centering 59.2
& \centering 54.2
& \centering 60.2
& \centering 70.6
& \centering 58.2
& \centering\arraybackslash 60.5 \\

& \hspace{5mm}w/ OpenPref
& \centering 62.6
& \centering 59.4
& \centering 57.8
& \centering 73.4
& \centering 61.8
& \centering\arraybackslash 63.0 \\

& \hspace{5mm}w/ LNR
& \centering 65.4
& \centering 64.6
& \centering 63.0
& \centering 75.4
& \centering 65.8
& \centering\arraybackslash 66.8 \\

& \hspace{5mm}w/ LNR + RS
& \centering 65.6
& \centering 63.0
& \centering 62.6
& \centering 73.0
& \centering 64.6
& \centering\arraybackslash 65.8 \\

& \hspace{5mm}w/ AugMix
& \centering 61.6
& \centering 62.2
& \centering 60.8
& \centering 72.4
& \centering 61.8
& \centering\arraybackslash 63.8 \\

& \hspace{5mm}w/ AugMix-MN
& \centering 62.6
& \centering 64.6
& \centering 61.8
& \centering 72.4
& \centering 59.2
& \centering\arraybackslash 64.1 \\

& \hspace{5mm}w/ \textsc{PrefLens}
& \centering \textbf{70.2}
& \centering \textbf{69.8}
& \centering \textbf{65.0}
& \centering \underline{79.0}
& \centering \textbf{70.6}
& \centering\arraybackslash \textbf{70.9} \\

\midrule

% ============================================================
% GenAIBench
% ============================================================
\multirow{23}{*}{\textbf{GenAIBench}}

& SFT
& \centering 47.8
& \centering 52.0
& \centering 47.4
& \centering 51.4
& \centering 50.2
& \centering\arraybackslash 49.8 \\

& SPIN
& \centering 64.0
& \centering \underline{67.6}
& \centering 56.2
& \centering 63.8
& \centering 59.2
& \centering\arraybackslash 62.2 \\

& SePPO
& \centering 63.2
& \centering 60.8
& \centering 61.8
& \centering 70.4
& \centering 65.4
& \centering\arraybackslash 64.3 \\

& SPO
& \centering 64.0
& \centering 66.4
& \centering 56.0
& \centering 68.8
& \centering \underline{65.6}
& \centering\arraybackslash 64.2 \\

& Diff.-DPO
& \centering 55.0
& \centering 54.4
& \centering 59.2
& \centering 60.8
& \centering 64.2
& \centering\arraybackslash 58.7 \\

\\[-1em] \cdashline{2-8}[0.4pt/1.5pt] \\[-1em]

& \textbf{Diffusion-DPO}
& & & & & & \\

& \hspace{5mm}w/ HPDv2
& \centering 52.8
& \centering 51.0
& \centering 53.0
& \centering 57.0
& \centering 52.2
& \centering\arraybackslash 53.2 \\

& \hspace{5mm}w/ Pick-a-Pic-v2
& \centering 50.0
& \centering 53.0
& \centering 53.0
& \centering 50.6
& \centering 50.2
& \centering\arraybackslash 51.4 \\

& \hspace{5mm}w/ OpenPref
& \centering 50.0
& \centering 53.2
& \centering 51.8
& \centering 51.4
& \centering 52.6
& \centering\arraybackslash 51.8 \\

& \hspace{5mm}w/ LNR
& \centering 54.0
& \centering 54.0
& \centering 53.0
& \centering 53.6
& \centering 54.2
& \centering\arraybackslash 53.8 \\

& \hspace{5mm}w/ LNR + RS
& \centering 58.0
& \centering 53.8
& \centering 51.8
& \centering 64.2
& \centering 59.0
& \centering\arraybackslash 57.4 \\

& \hspace{5mm}w/ AugMix
& \centering 45.0
& \centering 56.6
& \centering 46.0
& \centering 56.6
& \centering 36.8
& \centering\arraybackslash 48.2 \\

& \hspace{5mm}w/ AugMix-MN
& \centering 44.8
& \centering 53.0
& \centering 48.2
& \centering 59.6
& \centering 38.2
& \centering\arraybackslash 48.8 \\

& \hspace{5mm}w/ \textsc{PrefLens}
& \centering \underline{66.8}
& \centering 62.8
& \centering 57.2
& \centering 68.2
& \centering 63.0
& \centering\arraybackslash 63.6 \\

\\[-1em] \cdashline{2-8}[0.4pt/1.5pt] \\[-1em]

& \textbf{DSPO}
& & & & & & \\

& \hspace{5mm}w/ HPDv2
& \centering 55.2
& \centering 60.0
& \centering 59.0
& \centering 69.4
& \centering 60.8
& \centering\arraybackslash 60.9 \\

& \hspace{5mm}w/ Pick-a-Pic-v2
& \centering 57.2
& \centering 59.2
& \centering 60.0
& \centering 72.0
& \centering 60.4
& \centering\arraybackslash 61.8 \\

& \hspace{5mm}w/ OpenPref
& \centering 62.4
& \centering 61.2
& \centering 57.0
& \centering 70.0
& \centering 61.0
& \centering\arraybackslash 62.3 \\

& \hspace{5mm}w/ LNR
& \centering 64.8
& \centering 64.8
& \centering \underline{62.4}
& \centering \underline{74.6}
& \centering 63.0
& \centering\arraybackslash \underline{65.9} \\

& \hspace{5mm}w/ LNR + RS
& \centering 62.4
& \centering 63.2
& \centering 59.6
& \centering 72.8
& \centering 61.8
& \centering\arraybackslash 64.0 \\

& \hspace{5mm}w/ AugMix
& \centering 61.8
& \centering 61.6
& \centering 57.2
& \centering 72.6
& \centering 59.0
& \centering\arraybackslash 62.4 \\

& \hspace{5mm}w/ AugMix-MN
& \centering 60.6
& \centering 66.6
& \centering 60.4
& \centering 73.4
& \centering 61.2
& \centering\arraybackslash 64.4 \\

& \hspace{5mm}w/ \textsc{PrefLens}
& \centering \textbf{68.0}
& \centering \textbf{70.2}
& \centering \textbf{65.0}
& \centering \textbf{75.8}
& \centering \textbf{68.8}
& \centering\arraybackslash \textbf{69.6} \\

\bottomrule
\end{tabular}
}

\caption{
Win-rate (\%) comparison on SD v1.5.
Best results are highlighted in \textbf{boldface},
and second-best results are \underline{underlined}.
The Avg. column reports the mean across the five evaluation metrics.
All values are rounded to one decimal place.
}
\label{tab:results_win_rate}
\end{table*}
\begin{table*}[p]
\centering
\resizebox{0.78\linewidth}{!}{
\setlength{\tabcolsep}{6pt}
\begin{tabular}{c l p{1.8cm} p{1.8cm} p{1.8cm} p{1.8cm} p{1.8cm} p{1.5cm}}
\toprule
\multirow{3}{*}{\textbf{Benchmark}} &
\multirow{3}{*}{\textbf{Method}} &
\multicolumn{2}{c}{\textbf{Image Quality}} &
\multicolumn{3}{c}{\textbf{Human Preference}} &
\multirow{2}{*}{\centering\arraybackslash\textbf{Average}} \\
\cmidrule(lr){3-4}
\cmidrule(lr){5-7}
& &
\centering\textbf{Aesthetics} &
\centering\textbf{Aesthetics} &
\centering\textbf{ImageReward} &
\centering\textbf{HPS v2} &
\centering\textbf{PickScore} & \\
& &
\centering\textbf{(ViT-14)} &
\centering\textbf{(ViT-16)} &
\centering\textbf{} &
\centering\textbf{} &
\centering\textbf{} & \\
\midrule

% ============================================================
% PICK-A-PIC-V2
% ============================================================
\multirow{23}{*}{\textbf{Pick-a-Pic-V2}}
& SFT
& \centering 44.8 & \centering 49.4 & \centering 50.4
& \centering 51.4 & \centering 50.4
& \centering\arraybackslash 49.3 \\

& Diff.-DPO
& \centering 49.6 & \centering 46.0 & \centering 62.8
& \centering 72.0 & \centering 66.4
& \centering\arraybackslash 59.4 \\

& SPO
& \centering 67.0 & \centering 65.8 & \centering 69.0
& \centering 84.6 & \centering \textbf{75.4}
& \centering\arraybackslash 72.4 \\

\\[-1em] \cdashline{2-8}[0.4pt/1.5pt] \\[-1em]

& \textbf{Diffusion-DPO} & & & & & & \\

& \hspace{5mm}w/ HPDv2
& \centering 50.8 & \centering 52.0 & \centering 52.4
& \centering 52.4 & \centering 52.6
& \centering\arraybackslash 52.0 \\

& \hspace{5mm}w/ Pick-a-Pic-v2
& \centering 50.0 & \centering 50.8 & \centering 55.4
& \centering 52.2 & \centering 53.4
& \centering\arraybackslash 52.4 \\

& \hspace{5mm}w/ OpenPref
& \centering 46.6 & \centering 47.2 & \centering 49.2
& \centering 46.4 & \centering 47.2
& \centering\arraybackslash 47.3 \\

& \hspace{5mm}w/ LNR
& \centering 66.4 & \centering 67.2 & \centering 68.4
& \centering 84.6 & \centering 65.2
& \centering\arraybackslash 70.4 \\

& \hspace{5mm}w/ LNR + RS
& \centering 59.4 & \centering 61.8 & \centering 66.8
& \centering 80.0 & \centering 61.2
& \centering\arraybackslash 65.8 \\

& \hspace{5mm}w/ AugMix
& \centering 27.0 & \centering 44.4 & \centering 16.4
& \centering 6.6 & \centering 5.6
& \centering\arraybackslash 20.0 \\

& \hspace{5mm}w/ AugMix-MN
& \centering 35.4 & \centering 53.0 & \centering 30.0
& \centering 24.2 & \centering 15.0
& \centering\arraybackslash 31.5 \\

& \hspace{5mm}w/ \textsc{PrefLens}
& \centering \textbf{69.0}
& \centering \textbf{69.0}
& \centering \textbf{73.2}
& \centering \underline{87.6}
& \centering 69.6
& \centering\arraybackslash \textbf{73.7} \\

\\[-1em] \cdashline{2-8}[0.4pt/1.5pt] \\[-1em]

& \textbf{DSPO} & & & & & & \\

& \hspace{5mm}w/ HPDv2
& \centering 57.2 & \centering 53.2 & \centering 61.4
& \centering 61.2 & \centering 58.4
& \centering\arraybackslash 58.3 \\

& \hspace{5mm}w/ Pick-a-Pic-v2
& \centering 49.8 & \centering 48.6 & \centering 48.0
& \centering 51.6 & \centering 53.0
& \centering\arraybackslash 50.2 \\

& \hspace{5mm}w/ OpenPref
& \centering 49.2 & \centering 52.2 & \centering 50.4
& \centering 46.6 & \centering 51.8
& \centering\arraybackslash 50.0 \\

& \hspace{5mm}w/ LNR
& \centering 64.2 & \centering 65.2 & \centering 65.0
& \centering 83.4 & \centering 68.2
& \centering\arraybackslash 69.2 \\

& \hspace{5mm}w/ LNR + RS
& \centering \underline{68.4}
& \centering 67.0
& \centering 67.2
& \centering 86.8
& \centering 66.0
& \centering\arraybackslash 71.1 \\

& \hspace{5mm}w/ AugMix
& \centering 31.2 & \centering 50.0 & \centering 23.6
& \centering 16.0 & \centering 12.4
& \centering\arraybackslash 26.6 \\

& \hspace{5mm}w/ AugMix-MN
& \centering 39.6 & \centering 55.8 & \centering 35.4
& \centering 36.8 & \centering 15.6
& \centering\arraybackslash 36.6 \\

& \hspace{5mm}w/ \textsc{PrefLens}
& \centering 67.2
& \centering \underline{68.4}
& \centering \underline{70.6}
& \centering \textbf{89.0}
& \centering \underline{72.4}
& \centering\arraybackslash \underline{73.5} \\

\midrule

% ============================================================
% PARTIPROMPT
% ============================================================
\multirow{23}{*}{\textbf{PartiPrompt}}
& SFT
& \centering 48.8 & \centering 47.6 & \centering 49.4
& \centering 49.6 & \centering 46.8
& \centering\arraybackslash 48.4 \\

& Diff.-DPO
& \centering 52.6 & \centering 48.6 & \centering 63.2
& \centering 70.2 & \centering 57.0
& \centering\arraybackslash 58.3 \\

& SPO
& \centering 66.6 & \centering 63.0 & \centering \underline{64.0}
& \centering 86.4 & \centering \underline{68.8}
& \centering\arraybackslash 69.8 \\

\\[-1em] \cdashline{2-8}[0.4pt/1.5pt] \\[-1em]

& \textbf{Diffusion-DPO} & & & & & & \\

& \hspace{5mm}w/ HPDv2
& \centering 54.0 & \centering 51.4 & \centering 52.2
& \centering 51.8 & \centering 49.2
& \centering\arraybackslash 51.7 \\

& \hspace{5mm}w/ Pick-a-Pic-v2
& \centering 54.2 & \centering 51.6 & \centering 51.8
& \centering 51.6 & \centering 52.2
& \centering\arraybackslash 52.3 \\

& \hspace{5mm}w/ OpenPref
& \centering 47.6 & \centering 52.0 & \centering 51.6
& \centering 45.4 & \centering 48.6
& \centering\arraybackslash 49.0 \\

& \hspace{5mm}w/ LNR
& \centering 66.2 & \centering 64.8 & \centering 62.6
& \centering \underline{87.4} & \centering 68.6
& \centering\arraybackslash \underline{69.9} \\

& \hspace{5mm}w/ LNR + RS
& \centering 61.0 & \centering 57.4 & \centering 59.8
& \centering 78.0 & \centering 57.2
& \centering\arraybackslash 62.7 \\

& \hspace{5mm}w/ AugMix
& \centering 30.8 & \centering 46.8 & \centering 13.6
& \centering 7.4 & \centering 6.2
& \centering\arraybackslash 21.0 \\

& \hspace{5mm}w/ AugMix-MN
& \centering 41.8 & \centering 52.8 & \centering 29.8
& \centering 32.8 & \centering 16.2
& \centering\arraybackslash 34.7 \\

& \hspace{5mm}w/ \textsc{PrefLens}
& \centering \textbf{69.6}
& \centering \textbf{67.4}
& \centering \underline{64.0}
& \centering \textbf{89.4}
& \centering \textbf{73.0}
& \centering\arraybackslash \textbf{72.7} \\

\\[-1em] \cdashline{2-8}[0.4pt/1.5pt] \\[-1em]

& \textbf{DSPO} & & & & & & \\

& \hspace{5mm}w/ HPDv2
& \centering 59.0 & \centering 54.8 & \centering 57.4
& \centering 61.0 & \centering 58.6
& \centering\arraybackslash 58.2 \\

& \hspace{5mm}w/ Pick-a-Pic-v2
& \centering 52.2 & \centering 50.0 & \centering 51.6
& \centering 48.4 & \centering 51.4
& \centering\arraybackslash 50.7 \\

& \hspace{5mm}w/ OpenPref
& \centering 50.0 & \centering 48.4 & \centering 52.0
& \centering 48.4 & \centering 51.4
& \centering\arraybackslash 50.0 \\

& \hspace{5mm}w/ LNR
& \centering 63.6 & \centering 60.2 & \centering 59.4
& \centering 83.2 & \centering 62.8
& \centering\arraybackslash 65.8 \\

& \hspace{5mm}w/ LNR + RS
& \centering 65.6 & \centering 60.4 & \centering 61.0
& \centering 84.8 & \centering 60.8
& \centering\arraybackslash 66.5 \\

& \hspace{5mm}w/ AugMix
& \centering 28.8 & \centering 45.8 & \centering 22.0
& \centering 17.6 & \centering 12.2
& \centering\arraybackslash 25.3 \\

& \hspace{5mm}w/ AugMix-MN
& \centering 40.6 & \centering 53.8 & \centering 31.4
& \centering 37.8 & \centering 15.6
& \centering\arraybackslash 35.8 \\

& \hspace{5mm}w/ \textsc{PrefLens}
& \centering \underline{68.2}
& \centering 65.6
& \centering \textbf{66.4}
& \centering 80.6
& \centering 68.6
& \centering\arraybackslash 69.9 \\

\midrule

% ============================================================
% GENAIBENCH
% ============================================================
\multirow{23}{*}{\textbf{GenAIBench}}
& SFT
& \centering 47.0 & \centering 43.8 & \centering 49.0
& \centering 51.8 & \centering 48.6
& \centering\arraybackslash 48.0 \\

& Diff.-DPO
& \centering 51.8 & \centering 47.2 & \centering 61.6
& \centering 72.6 & \centering 64.8
& \centering\arraybackslash 59.6 \\

& SPO
& \centering 67.6 & \centering 68.6 & \centering 66.2
& \centering 85.6 & \centering 70.8
& \centering\arraybackslash 71.8 \\

\\[-1em] \cdashline{2-8}[0.4pt/1.5pt] \\[-1em]

& \textbf{Diffusion-DPO} & & & & & & \\

& \hspace{5mm}w/ HPDv2
& \centering 53.2 & \centering 53.0 & \centering 56.8
& \centering 61.2 & \centering 55.0
& \centering\arraybackslash 55.8 \\

& \hspace{5mm}w/ Pick-a-Pic-v2
& \centering 55.8 & \centering 52.8 & \centering 54.2
& \centering 52.0 & \centering 55.6
& \centering\arraybackslash 54.1 \\

& \hspace{5mm}w/ OpenPref
& \centering 48.4 & \centering 48.6 & \centering 48.6
& \centering 52.0 & \centering 51.4
& \centering\arraybackslash 49.8 \\

& \hspace{5mm}w/ LNR
& \centering 66.6 & \centering 68.4 & \centering 65.4
& \centering 84.8 & \centering \underline{72.6}
& \centering\arraybackslash 71.6 \\

& \hspace{5mm}w/ LNR + RS
& \centering 62.2 & \centering 58.0 & \centering 56.8
& \centering 79.6 & \centering 58.4
& \centering\arraybackslash 63.0 \\

& \hspace{5mm}w/ AugMix
& \centering 24.4 & \centering 38.8 & \centering 11.4
& \centering 7.6 & \centering 5.6
& \centering\arraybackslash 17.6 \\

& \hspace{5mm}w/ AugMix-MN
& \centering 43.0 & \centering 54.0 & \centering 27.2
& \centering 31.2 & \centering 18.4
& \centering\arraybackslash 34.8 \\

& \hspace{5mm}w/ \textsc{PrefLens}
& \centering \underline{69.8}
& \centering \underline{70.6}
& \centering \textbf{69.4}
& \centering \textbf{88.2}
& \centering \textbf{74.2}
& \centering\arraybackslash \textbf{74.4} \\

\\[-1em] \cdashline{2-8}[0.4pt/1.5pt] \\[-1em]

& \textbf{DSPO} & & & & & & \\

& \hspace{5mm}w/ HPDv2
& \centering 57.6 & \centering 56.0 & \centering 57.6
& \centering 63.6 & \centering 59.8
& \centering\arraybackslash 58.9 \\

& \hspace{5mm}w/ Pick-a-Pic-v2
& \centering 53.6 & \centering 52.0 & \centering 52.4
& \centering 55.4 & \centering 56.2
& \centering\arraybackslash 53.9 \\

& \hspace{5mm}w/ OpenPref
& \centering 51.8 & \centering 51.2 & \centering 52.0
& \centering 51.0 & \centering 52.8
& \centering\arraybackslash 51.8 \\

& \hspace{5mm}w/ LNR
& \centering 63.6 & \centering 60.4 & \centering 60.6
& \centering 81.6 & \centering 66.4
& \centering\arraybackslash 66.5 \\

& \hspace{5mm}w/ LNR + RS
& \centering 68.0 & \centering 67.0 & \centering 64.2
& \centering 86.4 & \centering 66.0
& \centering\arraybackslash 70.3 \\

& \hspace{5mm}w/ AugMix
& \centering 36.4 & \centering 46.0 & \centering 9.1
& \centering 17.4 & \centering 12.0
& \centering\arraybackslash 24.2 \\

& \hspace{5mm}w/ AugMix-MN
& \centering 39.3 & \centering 50.8 & \centering 23.5
& \centering 37.6 & \centering 17.2
& \centering\arraybackslash 33.7 \\

& \hspace{5mm}w/ \textsc{PrefLens}
& \centering \textbf{71.4}
& \centering \textbf{72.2}
& \centering \underline{68.0}
& \centering \underline{87.2}
& \centering 69.2
& \centering\arraybackslash \underline{73.6} \\

\bottomrule
\end{tabular}
}

\caption{
Win-rate (\%) comparison on SDXL-Base.
Best results are highlighted in \textbf{boldface},
and second-best results are \underline{underlined}.
The Avg. column reports the mean across the available metrics.
All values are rounded to one decimal place.
}
\label{tab:results_win_rate}
\end{table*}
% \clearpage

\subsection{Characterization of the reward-defined preference space.}
\label{sec:preference_space_analysis}
We further analyze the reward-defined preference space used for
neighborhood construction. Specifically, we consider three
representations: (i) a one-dimensional representation using only
Aesthetic Score, (ii) a one-dimensional representation using only
ImageReward, and (iii) the joint two-dimensional representation
$(\text{ImageReward},\text{Aesthetic})$.

To quantify the relationship between the two reward dimensions, we compute
Pearson~\cite{pearson1894iii} and Spearman~\cite{spearman1961proof}
correlations. At the image level, ImageReward and Aesthetic Score exhibit
moderate correlation (Pearson $0.68$, Spearman $0.68$). We additionally
compute the win--loss reward margin for each preference pair and obtain
Pearson and Spearman correlations of $0.65$ and $0.66$, respectively.
Thus, while the two dimensions are related, they are not redundant and
provide complementary preference information.

\begin{figure}[H]
    \centering
    \begin{subfigure}[t]{0.48\linewidth}
        \centering
        \includegraphics[width=\linewidth]{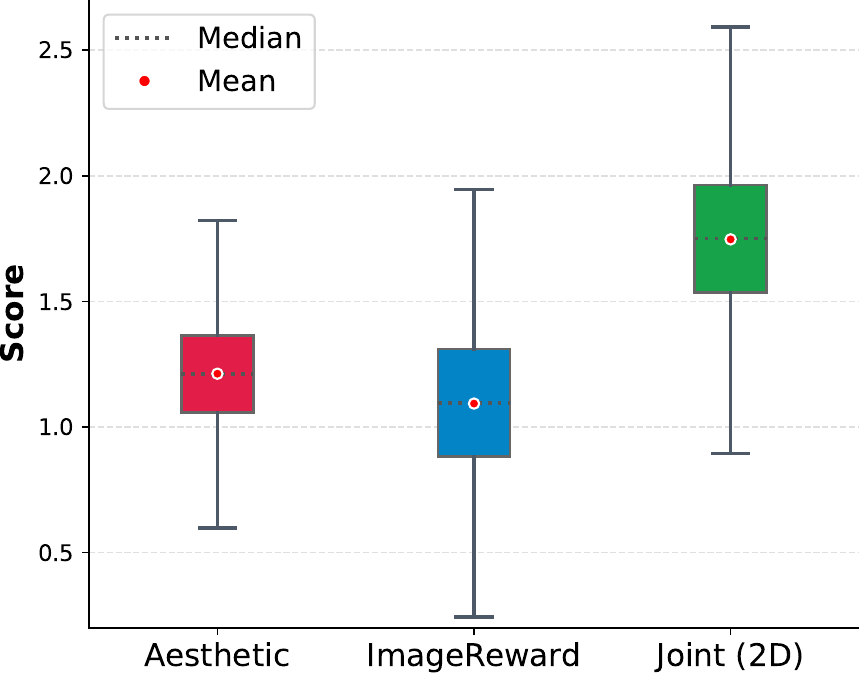}
        \caption{}
        \label{fig:pairwise_distance}
    \end{subfigure}
    \hfill
    \begin{subfigure}[t]{0.48\linewidth}
        \centering
        \includegraphics[width=\linewidth]{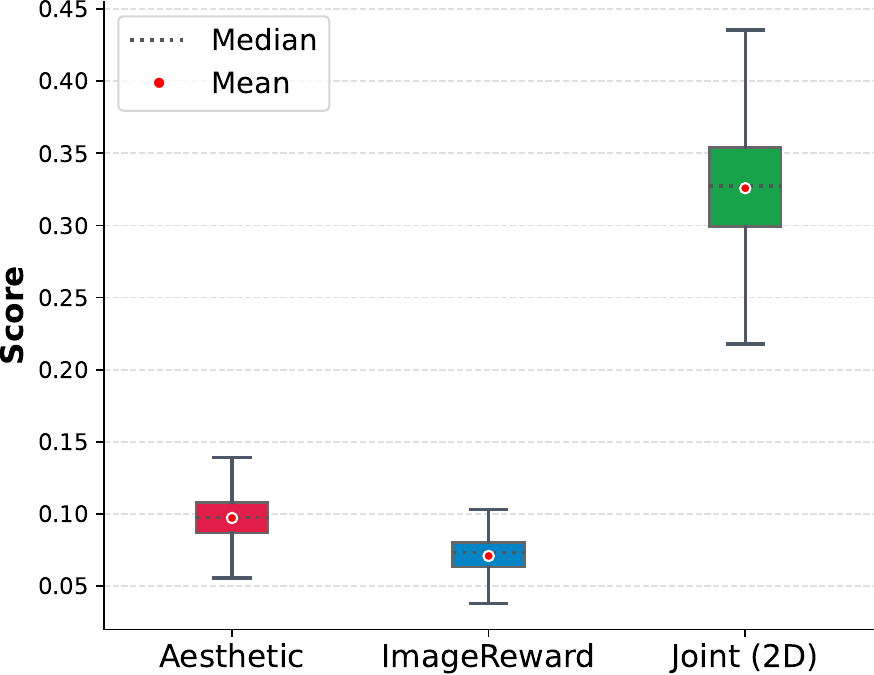}
        \caption{}
        \label{fig:knn_radius}
    \end{subfigure}
    
    \caption{Distribution of group-level diversity metrics under different scoring functions. (a) Average pairwise distance and (b) Average $k$-nearest neighbor (k-NN) radius. 
    % computed for images selected using \textit{Aesthetic}, \textit{ImageReward}, and their joint ($\ell_2$) scoring.
    }
    \label{fig:diversity_boxplots}
\end{figure}

We further examine the geometric structure induced by the different
representations (Figure~\ref{fig:diversity_boxplots}). For each anchor and its associated recovered neighbors,
we compute intra-neighborhood pairwise distances and $k$-nearest-neighbor
(kNN)~\cite{cover1967nearest} distance statistics. Compared with either
one-dimensional representation, the joint two-dimensional space exhibits
fewer near-zero-distance pairs. This indicates that candidates that appear
nearly indistinguishable along a single reward dimension can remain
separable when both preference signals are considered, providing a more
discriminative representation of the local neighborhood.

\section{Optimization Robustness and Convergence Analysis}
\label{sec:optimization_analysis}

\subsection{Training Convergence}

\begin{figure*}[p]
    \centering

    \resizebox{0.90\textwidth}{!}{%
    \begin{minipage}{\textwidth}

        \begin{subfigure}[t]{\linewidth}
            \centering
            \includegraphics[width=\linewidth]{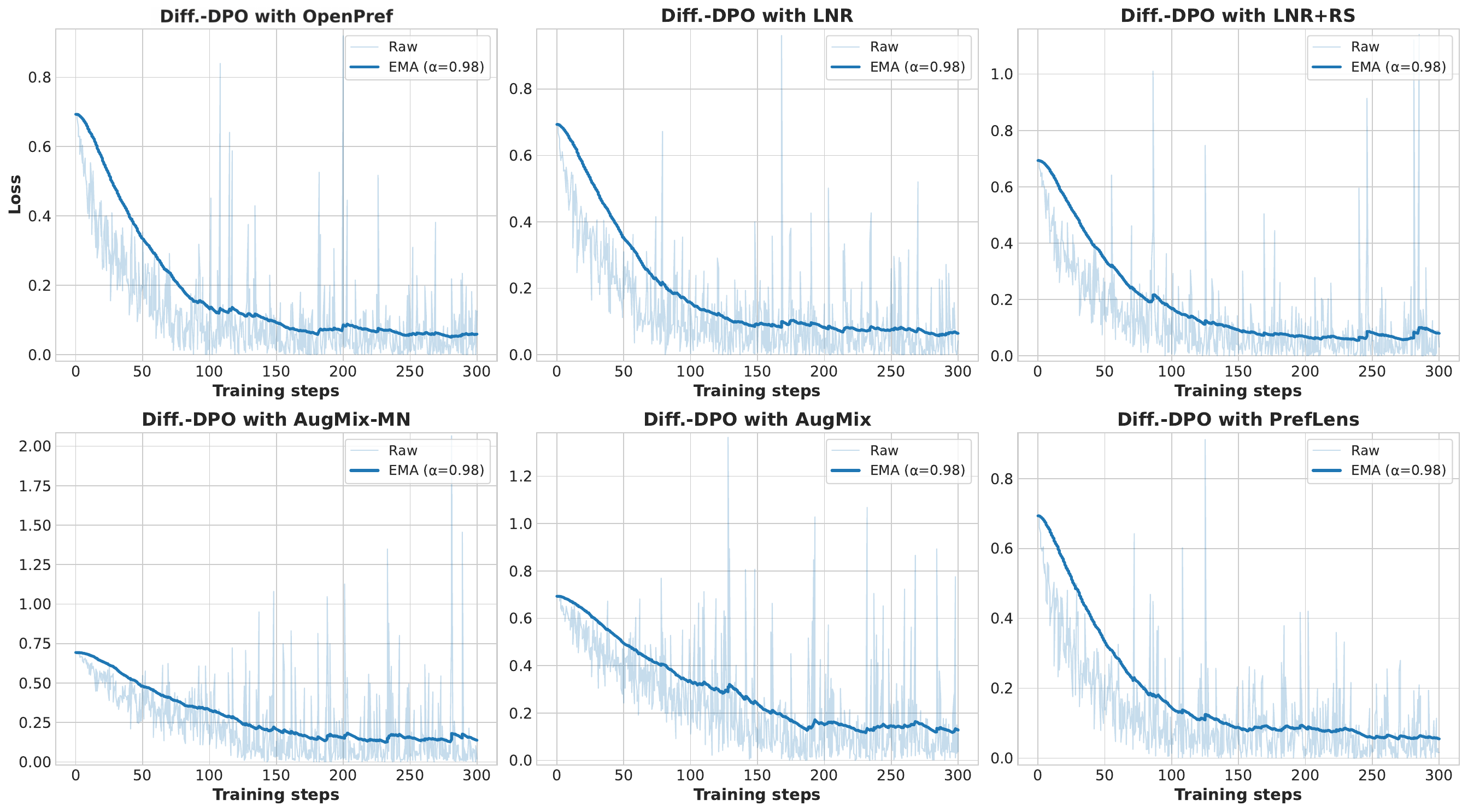}
            \caption{SD v1.5}
            \label{fig:tifa_blip}
        \end{subfigure}

        \vspace{56pt}

        \begin{subfigure}[t]{\linewidth}
            \centering
            \includegraphics[width=\linewidth]{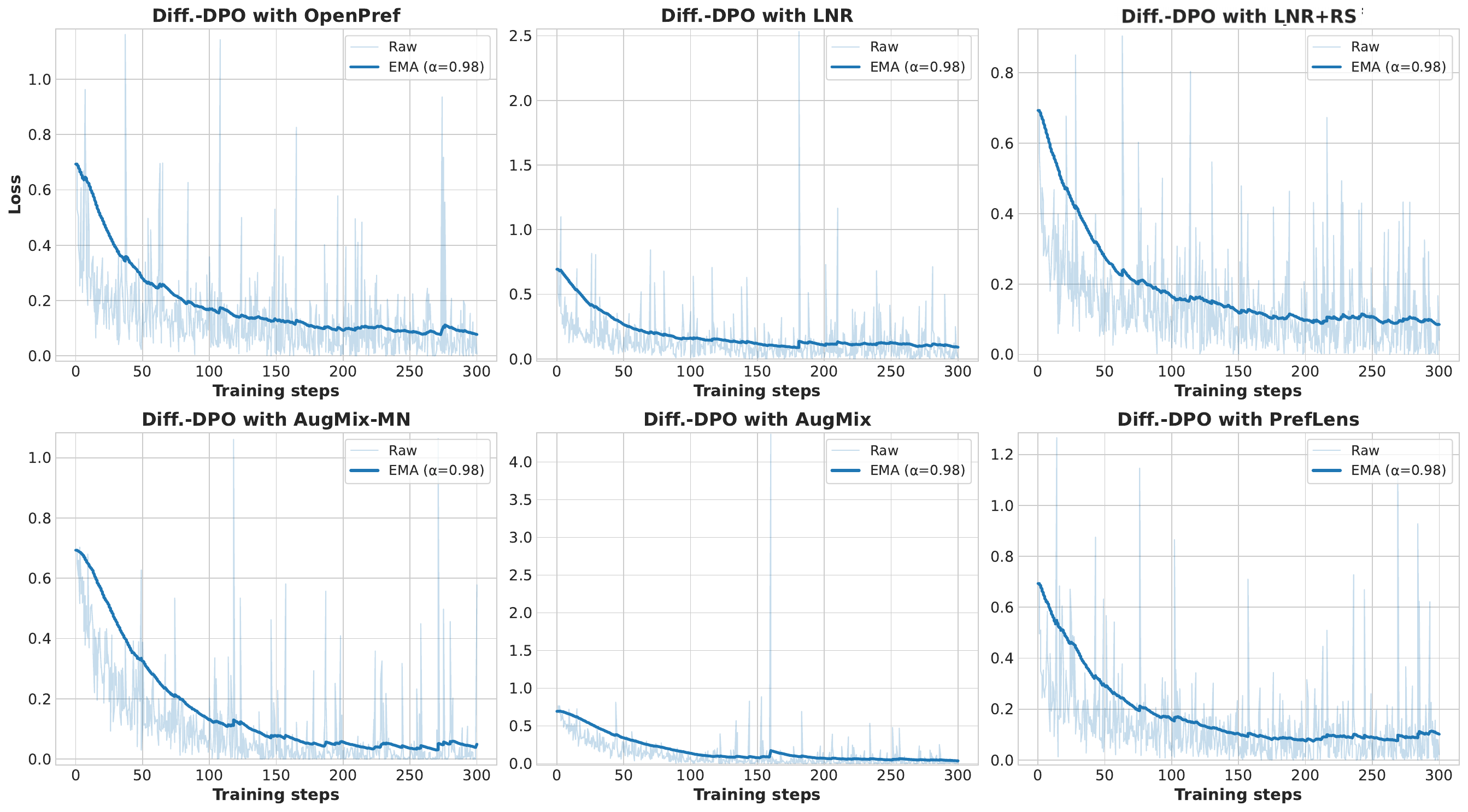}
            \caption{SDXL-Base}
            \label{fig:tifa_git}
        \end{subfigure}

    \end{minipage}%
    }

   \caption{
\textbf{Training-loss convergence for the compared configurations on (a) SD v1.5 and (b) SDXL-Base.}
We report the raw Diffusion-DPO training loss together with its exponential
moving average (EMA, $\alpha=0.98$). Across both backbones, the compared
configurations exhibit a clear decreasing trend and converge within the
training budget. \textsc{PrefLens} shows stable convergence comparable to
the alternative neighborhood-construction strategies, indicating that its
improved downstream alignment does not come at the cost of poorer
optimization convergence.
}
    \label{fig:loss_plot}
\end{figure*}

We further examine the optimization behavior of the compared training
configurations by analyzing their Diffusion-DPO loss trajectories.
Figure~\ref{fig:loss_plot} reports both the raw training loss and its
exponential moving average (EMA) for SD v1.5 and SDXL-Base. Across both
backbones, the compared configurations exhibit a clear decreasing trend and
converge within the training budget.

\subsection{Hyperparameter Sensitivity}

We analyze the sensitivity of preference optimization to two key training
hyperparameters: the learning rate and LoRA rank. The analysis is conducted
for both SD v1.5 and SDXL-Base while keeping the remaining training settings
fixed. These experiments characterize how optimization choices affect
downstream preference-alignment performance across the two diffusion
backbones.

\begin{table*}[p]
    \centering
    \begin{minipage}{\linewidth}
\captionsetup{type=table}
\centering
\scriptsize
\setlength{\tabcolsep}{2.5pt}
\renewcommand{\arraystretch}{1.12}

% ============================================================
% (a) SD v1.5 -- PrefLens
% ============================================================
\begin{subtable}[t]{0.45\textwidth}
\centering

\resizebox{0.9\linewidth}{!}{%
\begin{tabular}{lccccccc}
\toprule
\multirow{2}{*}{\textbf{Dataset}} &
\multirow{2}{*}{\textbf{Rank}} &
\multirow{2}{*}{\textbf{LR}} &
\multicolumn{2}{c}{\textbf{Image Quality}} &
\multicolumn{3}{c}{\textbf{Human Preference}} \\
\cmidrule(lr){4-5}
\cmidrule(lr){6-8}
& & &
\textbf{Aest.14} &
\textbf{Aest.16} &
\textbf{IR} &
\textbf{HPS} &
\textbf{PS} \\
\midrule

% ---------------- Pick-a-Pic-v2 ----------------
\multirow{4}{*}{\textbf{Pick-a-Pic-v2}}
& 8 & $1{\times}10^{-8}$
& 47.8 & 48.0 & 49.6 & 55.6 & 51.2 \\

& 128 & $1{\times}10^{-8}$
& 48.2 & 49.8 & 50.2 & 51.6 & 50.0 \\

& 8 & $5{\times}10^{-5}$
& 63.4 & 61.4 & 65.6 & 69.0 & 63.4 \\

& \cellcolor{black!5}128
& \cellcolor{black!5}$5{\times}10^{-5}$
& \cellcolor{black!5}\textbf{68.8}
& \cellcolor{black!5}\textbf{67.0}
& \cellcolor{black!5}\textbf{66.6}
& \cellcolor{black!5}\textbf{72.4}
& \cellcolor{black!5}\textbf{71.0} \\

\midrule

% ---------------- PartiPrompts ----------------
\multirow{4}{*}{\textbf{PartiPrompts}}
& 8 & $1{\times}10^{-8}$
& 53.6 & 51.8 & 51.6 & 49.8 & 49.6 \\

& 128 & $1{\times}10^{-8}$
& 55.8 & 50.2 & 53.8 & 51.0 & 51.0 \\

& 8 & $5{\times}10^{-5}$
& 64.6 & 56.4 & 60.8 & 65.4 & 61.6 \\

& \cellcolor{black!5}128
& \cellcolor{black!5}$5{\times}10^{-5}$
& \cellcolor{black!5}\textbf{67.0}
& \cellcolor{black!5}\textbf{67.8}
& \cellcolor{black!5}\textbf{64.8}
& \cellcolor{black!5}\textbf{74.6}
& \cellcolor{black!5}\textbf{65.0} \\

\midrule

% ---------------- GenAIBench ----------------
\multirow{4}{*}{\textbf{GenAIBench}}
& 8 & $1{\times}10^{-8}$
& 50.0 & 48.4 & 51.4 & 52.2 & 47.8 \\

& 128 & $1{\times}10^{-8}$
& 51.2 & 51.8 & 49.2 & 53.0 & 49.2 \\

& 8 & $5{\times}10^{-5}$
& 57.8 & 51.0 & 51.0 & 58.6 & 60.8 \\

& \cellcolor{black!5}128
& \cellcolor{black!5}$5{\times}10^{-5}$
& \cellcolor{black!5}\textbf{66.8}
& \cellcolor{black!5}\textbf{62.8}
& \cellcolor{black!5}\textbf{57.2}
& \cellcolor{black!5}\textbf{68.2}
& \cellcolor{black!5}\textbf{63.0} \\

\bottomrule
\end{tabular}%
}

\caption{SD v1.5 -- \textsc{PrefLens}}
\label{tab:hyperparam_preflens}
\end{subtable}
\hfill
% ============================================================
% (b) SD v1.5 -- OpenPref
% ============================================================
\begin{subtable}[t]{0.45\textwidth}
\centering

\resizebox{0.9\linewidth}{!}{%
\begin{tabular}{lccccccc}
\toprule
\multirow{2}{*}{\textbf{Dataset}} &
\multirow{2}{*}{\textbf{Rank}} &
\multirow{2}{*}{\textbf{LR}} &
\multicolumn{2}{c}{\textbf{Image Quality}} &
\multicolumn{3}{c}{\textbf{Human Preference}} \\
\cmidrule(lr){4-5}
\cmidrule(lr){6-8}
& & &
\textbf{Aest.14} &
\textbf{Aest.16} &
\textbf{IR} &
\textbf{HPS} &
\textbf{PS} \\
\midrule

% ---------------- Pick-a-Pic-v2 ----------------
\multirow{4}{*}{\textbf{Pick-a-Pic-v2}}
& 8 & $1{\times}10^{-8}$
& 47.0 & 47.4 & 51.8 & 52.8 & 48.2 \\

& 128 & $1{\times}10^{-8}$
& \textbf{49.6} & 49.2 & 50.2 & 52.0 & 49.6 \\

& 8 & $5{\times}10^{-5}$
& 47.6 & 48.0 & \textbf{54.4} & 53.4 & \textbf{50.8} \\

& \cellcolor{black!5}128
& \cellcolor{black!5}$5{\times}10^{-5}$
& \cellcolor{black!5}47.2
& \cellcolor{black!5}\textbf{49.4}
& \cellcolor{black!5}49.8
& \cellcolor{black!5}\textbf{53.6}
& \cellcolor{black!5}50.4 \\

\midrule

% ---------------- PartiPrompts ----------------
\multirow{4}{*}{\textbf{PartiPrompts}}
& 8 & $1{\times}10^{-8}$
& 54.8 & 51.0 & 50.0 & 51.0 & 47.8 \\

& 128 & $1{\times}10^{-8}$
& 54.0 & 52.0 & 49.6 & 49.0 & 48.8 \\

& 8 & $5{\times}10^{-5}$
& \textbf{57.8} & 52.6 & 49.4 & \textbf{56.0} & 54.0 \\

& \cellcolor{black!5}128
& \cellcolor{black!5}$5{\times}10^{-5}$
& \cellcolor{black!5}53.8
& \cellcolor{black!5}\textbf{53.4}
& \cellcolor{black!5}\textbf{56.2}
& \cellcolor{black!5}52.0
& \cellcolor{black!5}\textbf{54.6} \\

\midrule

% ---------------- GenAIBench ----------------
\multirow{4}{*}{\textbf{GenAIBench}}
& 8 & $1{\times}10^{-8}$
& \textbf{51.6} & 52.0 & 51.2 & 50.6 & 50.0 \\

& 128 & $1{\times}10^{-8}$
& 50.8 & 49.6 & 48.6 & 51.6 & 52.0 \\

& 8 & $5{\times}10^{-5}$
& 51.4 & 52.2 & 50.8 & \textbf{53.6} & 50.2 \\

& \cellcolor{black!5}128
& \cellcolor{black!5}$5{\times}10^{-5}$
& \cellcolor{black!5}50.0
& \cellcolor{black!5}\textbf{53.2}
& \cellcolor{black!5}\textbf{51.8}
& \cellcolor{black!5}51.4
& \cellcolor{black!5}\textbf{52.6} \\

\bottomrule
\end{tabular}%
}

\caption{SD v1.5 -- OpenPref}
\label{tab:hyperparam_openpref}
\end{subtable}

\vspace{8pt}

% ============================================================
% (c) SDXL-Base -- PrefLens
% ============================================================
\begin{subtable}[t]{0.45\textwidth}
\centering

\resizebox{0.9\linewidth}{!}{%
\begin{tabular}{lccccccc}
\toprule
\multirow{2}{*}{\textbf{Dataset}} &
\multirow{2}{*}{\textbf{Rank}} &
\multirow{2}{*}{\textbf{LR}} &
\multicolumn{2}{c}{\textbf{Image Quality}} &
\multicolumn{3}{c}{\textbf{Human Preference}} \\
\cmidrule(lr){4-5}
\cmidrule(lr){6-8}
& & &
\textbf{Aest.14} &
\textbf{Aest.16} &
\textbf{IR} &
\textbf{HPS} &
\textbf{PS} \\
\midrule

% ---------------- Pick-a-Pic-v2 ----------------
\multirow{4}{*}{\textbf{Pick-a-Pic-v2}}
& 8 & $1{\times}10^{-8}$
& 56.6 & 53.4 & 54.4 & 55.4 & \textbf{71.0} \\

& 128 & $1{\times}10^{-8}$
& 50.2 & 48.6 & 50.8 & 50.2 & 53.2 \\

& 8 & $5{\times}10^{-5}$
& 54.2 & 58.6 & 49.2 & 44.0 & 27.8 \\

& \cellcolor{black!5}128
& \cellcolor{black!5}$5{\times}10^{-5}$
& \cellcolor{black!5}\textbf{69.0}
& \cellcolor{black!5}\textbf{69.0}
& \cellcolor{black!5}\textbf{73.2}
& \cellcolor{black!5}\textbf{87.6}
& \cellcolor{black!5}69.6 \\

\midrule

% ---------------- PartiPrompts ----------------
\multirow{4}{*}{\textbf{PartiPrompts}}
& 8 & $1{\times}10^{-8}$
& 59.6 & 47.4 & \textbf{64.8} & 63.4 & \textbf{79.4} \\

& 128 & $1{\times}10^{-8}$
& 51.6 & 46.6 & 51.6 & 47.8 & 52.0 \\

& 8 & $5{\times}10^{-5}$
& 49.8 & 46.2 & 41.4 & 43.4 & 25.2 \\

& \cellcolor{black!5}128
& \cellcolor{black!5}$5{\times}10^{-5}$
& \cellcolor{black!5}\textbf{69.6}
& \cellcolor{black!5}\textbf{67.4}
& \cellcolor{black!5}64.0
& \cellcolor{black!5}\textbf{89.4}
& \cellcolor{black!5}73.0 \\

\midrule

% ---------------- GenAIBench ----------------
\multirow{4}{*}{\textbf{GenAIBench}}
& 8 & $1{\times}10^{-8}$
& 57.0 & 53.6 & 64.6 & 60.8 & 72.4 \\

& 128 & $1{\times}10^{-8}$
& 50.8 & 50.8 & 52.8 & 52.0 & 53.8 \\

& 8 & $5{\times}10^{-5}$
& 51.8 & 52.0 & 38.6 & 43.4 & 27.0 \\

& \cellcolor{black!5}128
& \cellcolor{black!5}$5{\times}10^{-5}$
& \cellcolor{black!5}\textbf{69.8}
& \cellcolor{black!5}\textbf{70.6}
& \cellcolor{black!5}\textbf{69.4}
& \cellcolor{black!5}\textbf{88.2}
& \cellcolor{black!5}\textbf{74.2} \\

\bottomrule
\end{tabular}%
}

\caption{SDXL-Base -- \textsc{PrefLens}}
\label{tab:hyperparam_preflens_sdxl}
\end{subtable}
\hfill
% ============================================================
% (d) SDXL-Base -- OpenPref
% ============================================================
\begin{subtable}[t]{0.45\textwidth}
\centering

\resizebox{0.9\linewidth}{!}{%
\begin{tabular}{lccccccc}
\toprule
\multirow{2}{*}{\textbf{Dataset}} &
\multirow{2}{*}{\textbf{Rank}} &
\multirow{2}{*}{\textbf{LR}} &
\multicolumn{2}{c}{\textbf{Image Quality}} &
\multicolumn{3}{c}{\textbf{Human Preference}} \\
\cmidrule(lr){4-5}
\cmidrule(lr){6-8}
& & &
\textbf{Aest.14} &
\textbf{Aest.16} &
\textbf{IR} &
\textbf{HPS} &
\textbf{PS} \\
\midrule

% ---------------- Pick-a-Pic-v2 ----------------
\multirow{4}{*}{\textbf{Pick-a-Pic-v2}}
& 8 & $1{\times}10^{-8}$
& \textbf{54.4} & 46.4 & 48.6 & \textbf{55.0} & 50.2 \\

& 128 & $1{\times}10^{-8}$
& 46.6 & 49.0 & 47.8 & 47.6 & 50.0 \\

& 8 & $5{\times}10^{-5}$
& 46.6 & \textbf{49.4} & 47.0 & 52.0 & \textbf{50.8} \\

& \cellcolor{black!5}128
& \cellcolor{black!5}$5{\times}10^{-5}$
& \cellcolor{black!5}46.6
& \cellcolor{black!5}47.2
& \cellcolor{black!5}\textbf{49.2}
& \cellcolor{black!5}46.4
& \cellcolor{black!5}47.2 \\

\midrule

% ---------------- PartiPrompts ----------------
\multirow{4}{*}{\textbf{PartiPrompts}}
& 8 & $1{\times}10^{-8}$
& 51.4 & 49.4 & 49.8 & \textbf{50.8} & 46.6 \\

& 128 & $1{\times}10^{-8}$
& 49.0 & 44.4 & 49.8 & 48.4 & 47.6 \\

& 8 & $5{\times}10^{-5}$
& \textbf{54.4} & 49.8 & 51.2 & 50.4 & 47.4 \\

& \cellcolor{black!5}128
& \cellcolor{black!5}$5{\times}10^{-5}$
& \cellcolor{black!5}47.6
& \cellcolor{black!5}\textbf{52.0}
& \cellcolor{black!5}\textbf{51.6}
& \cellcolor{black!5}45.4
& \cellcolor{black!5}\textbf{48.6} \\

\midrule

% ---------------- GenAIBench ----------------
\multirow{4}{*}{\textbf{GenAIBench}}
& 8 & $1{\times}10^{-8}$
& 50.2 & 47.6 & 48.6 & 49.0 & 50.0 \\

& 128 & $1{\times}10^{-8}$
& 51.2 & 47.4 & 50.6 & 51.2 & \textbf{51.4} \\

& 8 & $5{\times}10^{-5}$
& \textbf{55.0} & 48.0 & \textbf{54.4} & 51.4 & 51.0 \\

& \cellcolor{black!5}128
& \cellcolor{black!5}$5{\times}10^{-5}$
& \cellcolor{black!5}48.4
& \cellcolor{black!5}\textbf{48.6}
& \cellcolor{black!5}48.6
& \cellcolor{black!5}\textbf{52.0}
& \cellcolor{black!5}\textbf{51.4} \\

\bottomrule
\end{tabular}%
}

\caption{SDXL-Base -- OpenPref}
\label{tab:hyperparam_openpref_sdxl}
\end{subtable}

% ============================================================
% Overall caption
% ============================================================
\caption{
\textbf{Sensitivity to training hyperparameters for SD v1.5 and SDXL-Base.}
We vary the LoRA rank and learning rate for OpenPref and \textsc{PrefLens}
while keeping the remaining training configuration fixed. We select a
single configuration based on its overall performance across the evaluated
datasets and metrics, rather than tuning separately for individual
dataset--metric combinations. Here, Aest.14 and Aest.16 denote Aesthetic Score using ViT-L/14 and ViT-B/16, respectively; IR denotes ImageReward, HPS denotes HPSv2, and PS denotes PickScore. Rank $128$ with learning rate
$5\times10^{-5}$ performs best in the largest number of evaluated settings
and is therefore used as the default configuration; it is highlighted in
gray. The substantial variation across rank--learning-rate combinations
also indicates that preference-based diffusion fine-tuning is sensitive to
optimization hyperparameters. The maximum value for each metric within each
benchmark is shown in \textbf{bold}; ties are bolded jointly.
}
\label{tab:hyperparameter_search}

\end{minipage}
    \par\vspace{10pt}
    \begin{minipage}{\linewidth}
\captionsetup{type=table}
\centering
\scriptsize
\setlength{\tabcolsep}{4pt}
\renewcommand{\arraystretch}{1.12}

% ============================================================
% (a) SD v1.5
% ============================================================
\begin{subtable}[t]{0.42\textwidth}
\centering

\resizebox{\linewidth}{!}{%
\begin{tabular}{llcccccc}
\toprule
\multirow{2}{*}{\textbf{Dataset}} &
\multirow{2}{*}{\textbf{Preference Space}} &
\multicolumn{2}{c}{\textbf{Image Quality}} &
\multicolumn{3}{c}{\textbf{Human Preference}} &
\multirow{2}{*}{\textbf{Avg.}} \\
\cmidrule(lr){3-4}
\cmidrule(lr){5-7}

& &
\textbf{Aest.14} &
\textbf{Aest.16} &
\textbf{IR} &
\textbf{HPS} &
\textbf{PS} & \\
\midrule

% ---------------- Pick-a-Pic ----------------
\multirow{2}{*}{\textbf{Pick-a-Pic}}
& Normalized
& 67.0
& \textbf{68.4}
& 63.8
& \textbf{75.6}
& 69.4
& 68.8 \\

& Raw
& \textbf{68.8}
& 67.0
& \textbf{66.6}
& 72.4
& \textbf{71.0}
& \textbf{69.2} \\

\midrule

% ---------------- PartiPrompt ----------------
\multirow{2}{*}{\textbf{PartiPrompt}}
& Normalized
& 66.8
& 64.4
& \textbf{65.4}
& 73.6
& \textbf{66.0}
& 67.2 \\

& Raw
& \textbf{67.0}
& \textbf{67.8}
& 64.8
& \textbf{74.6}
& 65.0
& \textbf{67.8} \\

\midrule

% ---------------- GenAI ----------------
\multirow{2}{*}{\textbf{GenAI}}
& Normalized
& \textbf{68.8}
& 62.4
& 55.8
& 67.6
& 61.8
& 63.3 \\

& Raw
& 66.8
& \textbf{62.8}
& \textbf{57.2}
& \textbf{68.2}
& \textbf{63.0}
& \textbf{63.6} \\

\bottomrule
\end{tabular}%
}

\caption{SD v1.5}
\label{tab:normalization_sd15}
\end{subtable}
\hfill
% ============================================================
% (b) SDXL-Base
% ============================================================
\begin{subtable}[t]{0.42\textwidth}
\centering

\resizebox{\linewidth}{!}{%
\begin{tabular}{llcccccc}
\toprule
\multirow{2}{*}{\textbf{Dataset}} &
\multirow{2}{*}{\textbf{Preference Space}} &
\multicolumn{2}{c}{\textbf{Image Quality}} &
\multicolumn{3}{c}{\textbf{Human Preference}} &
\multirow{2}{*}{\textbf{Avg.}} \\
\cmidrule(lr){3-4}
\cmidrule(lr){5-7}

& &
\textbf{Aest.14} &
\textbf{Aest.16} &
\textbf{IR} &
\textbf{HPS} &
\textbf{PS} & \\
\midrule

% ---------------- Pick-a-Pic ----------------
\multirow{2}{*}{\textbf{Pick-a-Pic}}
& Normalized
& \textbf{71.4}
& \textbf{71.8}
& \textbf{74.8}
& 85.4
& 68.4
& \textbf{74.4} \\

& Raw
& 69.0
& 69.0
& 73.2
& \textbf{87.6}
& \textbf{69.6}
& 73.7 \\

\midrule

% ---------------- PartiPrompt ----------------
\multirow{2}{*}{\textbf{PartiPrompt}}
& Normalized
& 67.6
& 67.0
& \textbf{65.4}
& 88.4
& 71.4
& 72.0 \\

& Raw
& \textbf{69.6}
& \textbf{67.4}
& 64.0
& \textbf{89.4}
& \textbf{73.0}
& \textbf{72.7} \\

\midrule

% ---------------- GenAI ----------------
\multirow{2}{*}{\textbf{GenAI}}
& Normalized
& 66.6
& 67.6
& 67.2
& 87.2
& 73.2
& 72.4 \\

& Raw
& \textbf{69.8}
& \textbf{70.6}
& \textbf{69.4}
& \textbf{88.2}
& \textbf{74.2}
& \textbf{74.4} \\

\bottomrule
\end{tabular}%
}

\caption{SDXL-Base}
\label{tab:normalization_sdxl}
\end{subtable}

\caption{
\textbf{Sensitivity to preference-space normalization.}
We compare the default raw reward-defined preference space with a
$z$-score-normalized variant for SD v1.5 and SDXL-Base across
Pick-a-Pic-v2, PartiPrompt, and GenAIBench. Here, Aest.14 and Aest.16 denote Aesthetic Score using ViT-L/14 and ViT-B/16, respectively; IR denotes ImageReward, HPS denotes HPSv2, and PS denotes PickScore. In most cases, the two
constructions yield closely matched performance, with neither variant
consistently dominating across metrics or backbones. This behavior is
consistent with the similar variances of the two reward dimensions
($1.32$ and $1.52$) and the high overlap in selected neighbors, with
$94.4\%$ overlap in selected samples. Together, these results
suggest that the reported gains are not driven by a substantial imbalance
in reward scales for OpenPref. However, this behavior may be
dataset-dependent; for datasets with more heterogeneous reward
distributions, normalization may become more important. We therefore view
$z$-score normalization as a preferable default when transferring the
construction to other preference datasets.
}
\label{tab:normalization_ablation}
\end{minipage}
\end{table*}

\section{Ablations}
\label{sec:additional_ablations}
\textbf{\textit{Perturbation families.}}
We separately evaluate the three perturbation families---\emph{Photometric}, \emph{Geometric}, and \emph{Fidelity Reduction}---as well as their combined use. Combining all three families performs best in most settings, indicating that variations in appearance, spatial structure, and visual fidelity provide complementary preference supervision.

\begin{figure*}[p]
    \centering
    
    \begin{subfigure}{\linewidth}
        \centering
        \includegraphics[width=0.8\linewidth]{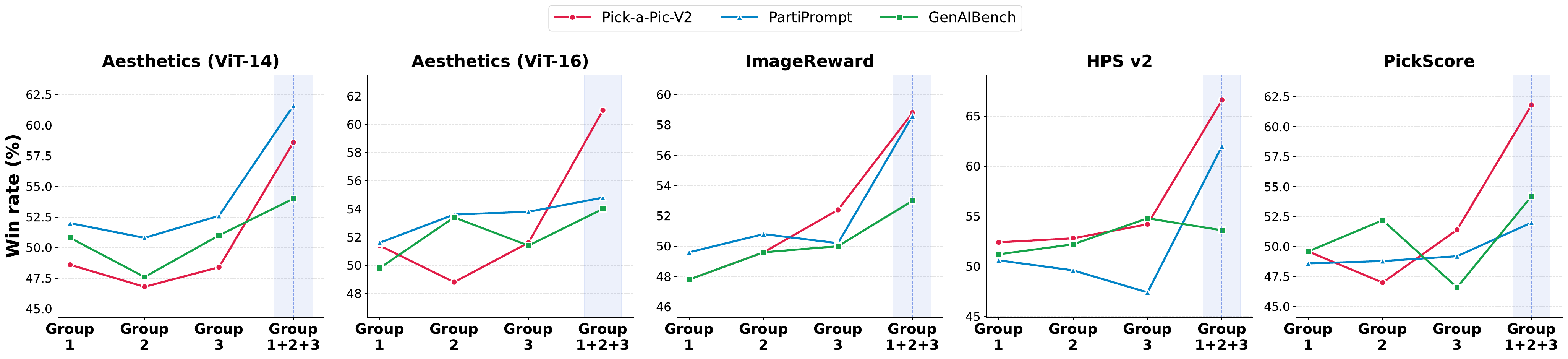}
        \caption{SD v1.5}
        \label{fig:ablation-sd15}
    \end{subfigure}
    
    \begin{subfigure}{\linewidth}
        \centering
        \includegraphics[width=0.8\linewidth]{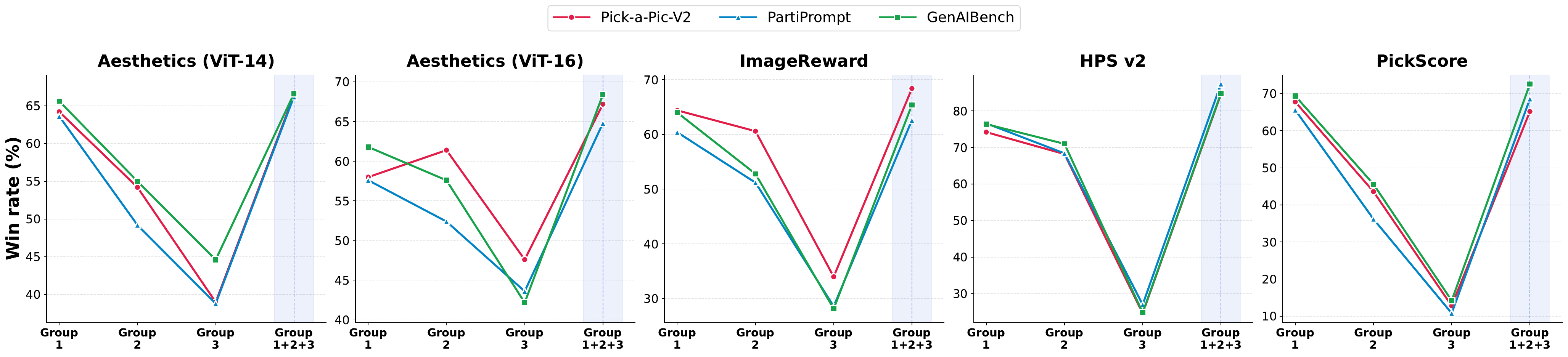}
        \caption{SDXL-Base}
        \label{fig:ablation-sdxl}
    \end{subfigure}

\caption{
\textbf{Ablation on perturbation families.}
We group the controlled perceptual perturbations into three families:
Photometric (Group~1), Geometric (Group~2), and Fidelity Reduction
(Group~3), and compare each family independently with their combined use
(Group~1+2+3) in \textsc{PrefLens}. The individual families exhibit
different behavior across datasets, metrics, and backbones, with no single
family consistently dominating the others. Notably, for SDXL-Base,
Group~3 alone yields substantially lower performance on several metrics,
indicating that the effectiveness of an isolated perturbation family can
be strongly model-dependent. This also suggests that relying on any single
family may provide an incomplete or poorly matched neighborhood for a given
backbone. In contrast, combining all three families generally yields the
strongest and most consistent performance, suggesting that their
complementary variations in appearance, spatial structure, and visual
fidelity provide a more robust basis for local neighborhood construction.
}
\label{fig:perturbation_ablation}
\end{figure*}

\textbf{\textit{Perturbation strength.}}
For perturbations with configurable severity, the default configuration
samples strength from the corresponding predefined range
$[\lambda_{\min},\lambda_{\max}]$. We compare this range-based strategy against fixed minimum-strength and fixed maximum-strength variants. Sampling across the full range generally performs better than either fixed-strength alternative, suggesting that exposing the model to varying perturbation intensities provides more effective supervision.

\begin{figure*}[p]
    \centering
    
    \begin{subfigure}{\linewidth}
        \centering
        \includegraphics[width=0.8\linewidth]{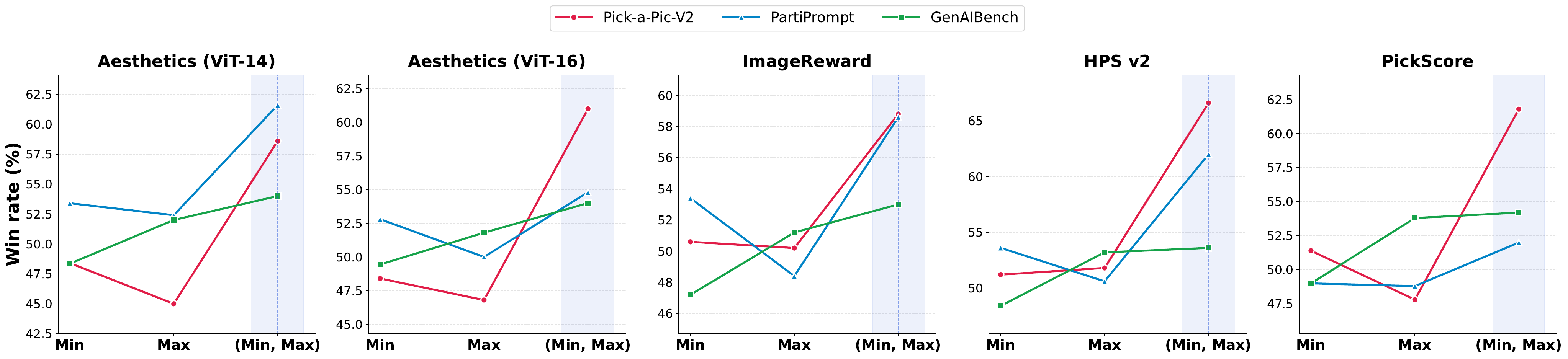}
        \caption{SD v1.5}
        \label{fig:ablation-sd15}
    \end{subfigure}
    
    \begin{subfigure}{\linewidth}
        \centering
        \includegraphics[width=0.8\linewidth]{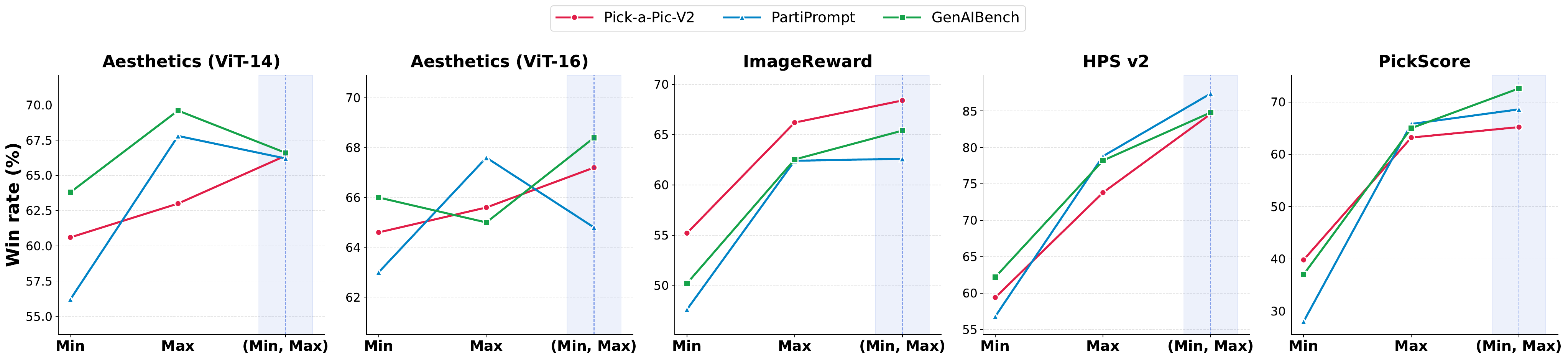}
        \caption{SDXL-Base}
        \label{fig:ablation-sdxl}
    \end{subfigure}

\caption{
\textbf{Ablation on perturbation-strength sampling.}
We compare the default \textsc{PrefLens} strategy, which samples perturbation
strengths from $[\lambda_{\min},\lambda_{\max}]$, with fixed-strength
variants using only $\lambda_{\min}$ (Min) or $\lambda_{\max}$ (Max).
Across SD v1.5 and SDXL-Base, sampling over the full strength range generally
provides the strongest and most consistent performance across benchmarks and
metrics. Using only the minimum strength can produce variations that are too
subtle to provide sufficiently informative preference comparisons, whereas
using only the maximum strength restricts neighborhood construction to
consistently severe perturbations. Sampling from the full range instead
exposes the model to variations of different perceptual strengths, yielding
a more diverse local neighborhood. All other components of
the pipeline are kept unchanged.
}
\label{fig:strength_ablation}
\end{figure*}

\textbf{\textit{Margin threshold.}}
We evaluate the sensitivity of Margin-Based Filtering (MBF) to
$\epsilon\in\{0.1,0.5,1.0\}$. The results support the default choice
$\epsilon=0.5$, which provides the strongest overall performance among the evaluated thresholds.

\begin{figure*}[p]
    \centering
    
    \begin{subfigure}{\linewidth}
        \centering
        \includegraphics[width=0.8\linewidth]{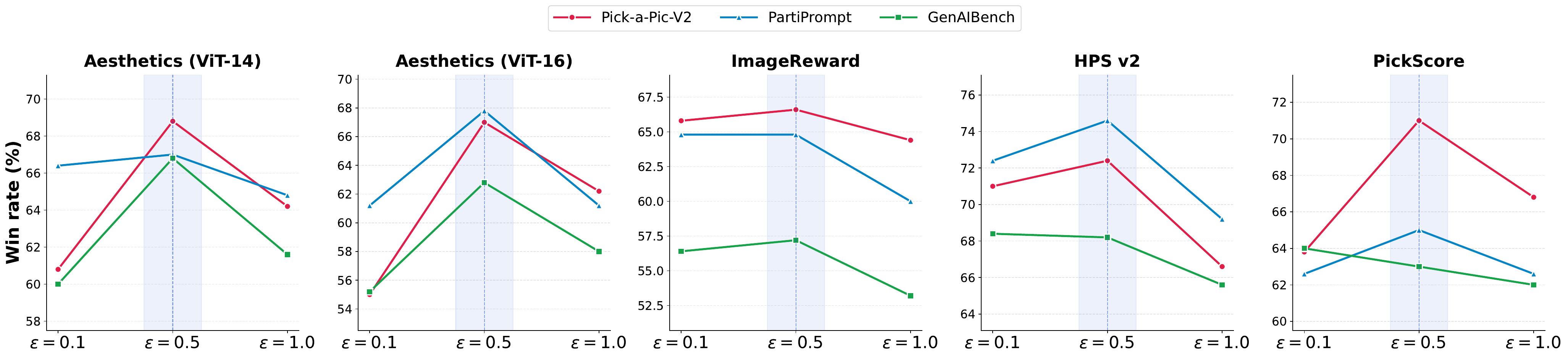}
        \caption{SD v1.5}
        \label{fig:ablation-sd15}
    \end{subfigure}
    
    \begin{subfigure}{\linewidth}
        \centering
        \includegraphics[width=0.8\linewidth]{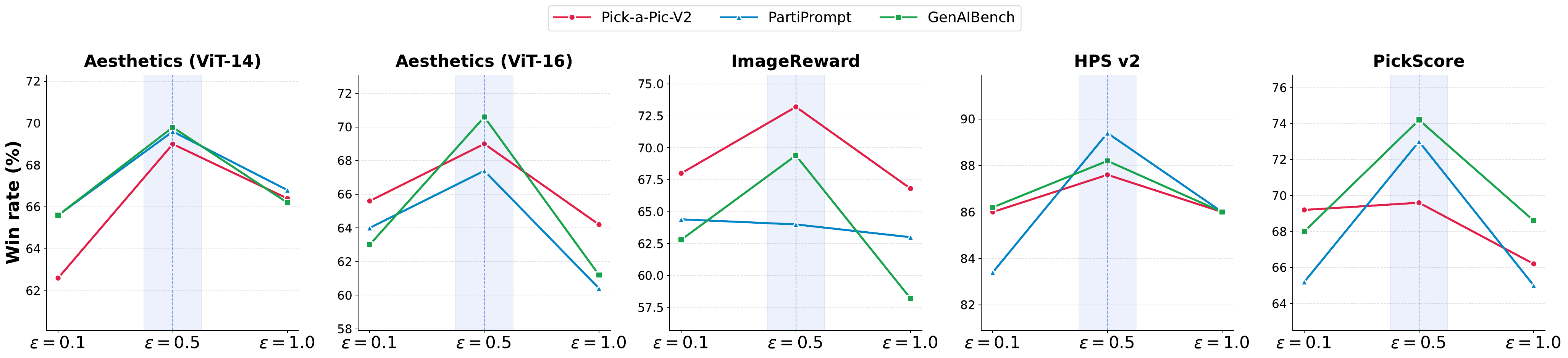}
        \caption{SDXL-Base}
        \label{fig:ablation-sdxl}
    \end{subfigure}

\caption{
\textbf{Effect of the MBF margin threshold $\epsilon$.}
The margin $\epsilon$ controls how far a candidate must be from the preferred
anchor in the reward-defined preference space to be retained by MBF.
When $\epsilon$ is too small, many retained samples remain nearly identical
to the anchor, providing only weak perceptual differences and potentially
falling below a meaningful just-noticeable-difference regime
\cite{wu2019survey}. In contrast, an excessively large $\epsilon$ removes
too many candidates from the recovered neighborhood, reducing the diversity
and coverage of available preference comparisons. Intermediate values
therefore provide a better balance between perceptual distinctiveness and
neighborhood coverage for both SD v1.5 and SDXL-Base.
}
    \label{fig:epsilon_abalation}
\end{figure*}

\textbf{\textit{Neighborhood-selection strategy and sampling budget.}}
We compare the proposed density-aware informative selection against matched-budget random sampling (\textsc{PrefLens}-RS) and uniform cluster sampling (\textsc{PrefLens}-US). We additionally vary the sampling ratio $K\in\{25\%,50\%,75\%\}$ to study the trade-off between neighborhood size and informativeness. These experiments show that the choice of neighbors, rather than neighborhood expansion alone, plays an important role in downstream preference alignment.

\begin{figure*}[p]
    \centering
    
    \begin{subfigure}{\linewidth}
        \centering
        \includegraphics[width=0.8\linewidth]{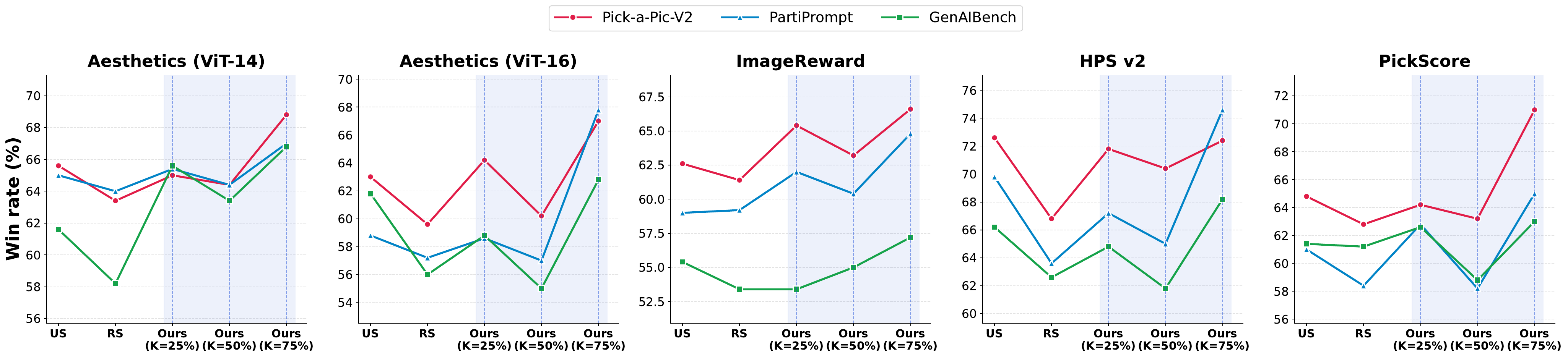}
        \caption{SD v1.5}
        \label{fig:ablation-sd15}
    \end{subfigure}
    
    \begin{subfigure}{\linewidth}
        \centering
        \includegraphics[width=0.8\linewidth]{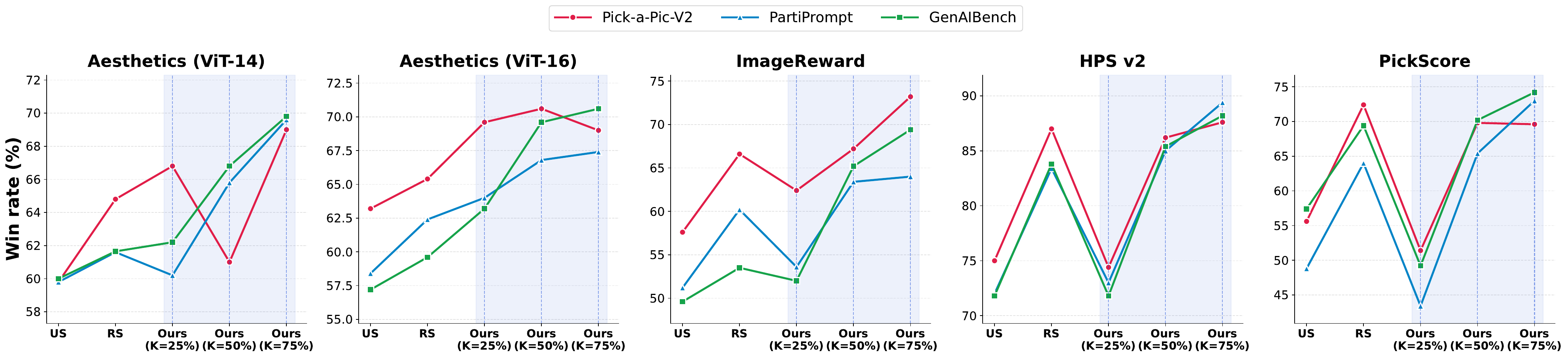}
        \caption{SDXL-Base}
        \label{fig:ablation-sdxl}
    \end{subfigure}

\caption{
\textbf{Ablation of neighborhood sampling strategy and sampling ratio $K$.}
We compare uniform sampling (US), random sampling (RS), and the proposed
\textsc{PrefLens} selection with sampling ratios
$K\in\{25\%,50\%,75\%\}$ across Pick-a-Pic-v2, PartiPrompt, and
GenAIBench. US and RS provide weaker or less consistent performance,
showing that simply drawing samples from the recovered neighborhood is
insufficient to fully exploit the available comparisons. In contrast,
the proposed informative selection generally yields stronger performance,
with $K=75\%$ providing the best overall results across metrics and
backbones. Performance is nevertheless not strictly monotonic with $K$;
in several cases, $K=50\%$ performs below $K=25\%$, indicating that
alignment quality depends not only on the number of selected pairs but
also on their composition and diversity. The shaded region denotes the
\textsc{PrefLens} variants.
}
    \label{fig:ablation}
\end{figure*}

\textbf{\textit{Preference-space normalization.}}
Our default construction uses the raw ImageReward and Aesthetic Score values. We additionally construct the preference space after $z$-score normalization using training-set statistics. The raw reward dimensions have similar variances ($1.32$ and $1.52$), and the raw and normalized pipelines exhibit a 94.4\% overlap in selected samples. We report downstream results for both constructions, showing limited sensitivity to reward-scale normalization.

\clearpage
\clearpage

\begin{figure*}[p]
\centering

\begin{minipage}[t]{0.48\textwidth}
\small
\section{Additional Qualitative Results}
\label{sec:additional_qualitative}

We provide additional qualitative comparisons in Figures~\ref{fig:sd15_picapic}, \ref{fig:sd15_parti}, \ref{fig:sd15_genai}, \ref{fig:sdxl_picapic}, \ref{fig:sdxl_partiprompt}, and \ref{fig:sdxl_genai}. These figures cover both SD v1.5 and SDXL-Base across Pick-a-Pic-v2, PartiPrompts, and GenAIBench. Each example compares generations from the corresponding base model, Diffusion-DPO trained with OpenPref, and Diffusion-DPO trained with \textsc{PrefLens}.

The examples span diverse prompts involving multiple objects, spatial relations, counting, text rendering, attributes, and complex visual compositions. Across these settings, \textsc{PrefLens} generally produces generations with improved prompt adherence, stronger compositional fidelity, and better overall visual quality.
\end{minipage}\hfill%
\begin{minipage}[t]{0.48\textwidth}
\small
\section{Evaluation Prompts and Annotation Instructions}
\label{sec:evaluation_instructions}

In this section, we include instructions for
(i) human ranking of generated images, (ii) human validation of recovered
preference pairs together with Easy/Medium/Hard perceptual-difficulty
annotation, (iii) VLM-as-a-Judge evaluation of overall preference alignment,
and (iv) VLM-based evaluation of compositional prompt adherence. These
instructions standardize the comparison criteria used across the different
evaluation settings and clarify how preference quality, compositional
correctness, and perceptual difficulty are assessed. The corresponding
evaluation prompts and annotation protocols are provided in
Figures~\ref{fig:prompt_adherence},
\ref{fig:human_pref_prompt},
\ref{fig:comprehensive_eval}, and
\ref{fig:pair_validation_difficulty}.
\end{minipage}
\par\vspace{30pt}
\begin{minipage}{\linewidth}
\captionsetup{type=figure}
    \centering
\scalebox{0.68}{
\begin{tabular}{>{\raggedright\arraybackslash}m{3cm} *{4}{>{\centering\arraybackslash}m{3.2cm}}}

& \makecell{
\textbf{\fontsize{10}{8}\selectfont Cyberpunk}\\[-2pt]
\textbf{\fontsize{10}{8}\selectfont Cat}
}
& \makecell{
\textbf{\fontsize{10}{8}\selectfont Logo of blue}\\[-2pt]
\textbf{\fontsize{10}{8}\selectfont elephant,}\\[-2pt]
\textbf{\fontsize{10}{8}\selectfont flat modern}\\[-2pt]
\textbf{\fontsize{10}{8}\selectfont vector icon}
}
& \makecell{
\textbf{\fontsize{10}{8}\selectfont A dog riding}\\[-2pt]
\textbf{\fontsize{10}{8}\selectfont bicycle}
}
& \makecell{
\textbf{\fontsize{10}{8}\selectfont Sticker}\\[-2pt]
\textbf{\fontsize{10}{8}\selectfont for a}\\[-2pt]
\textbf{\fontsize{10}{8}\selectfont hacker}
}
 \\[-1pt]

% Row 1 – Base (SD v1.5)
\makecell[c]{\textbf{Base} \\ \textbf{(SD v1.5)}}

& \includegraphics[width=\linewidth]{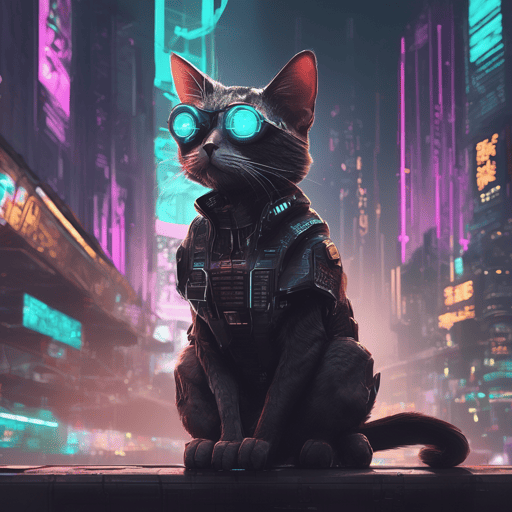}
& \includegraphics[width=\linewidth]{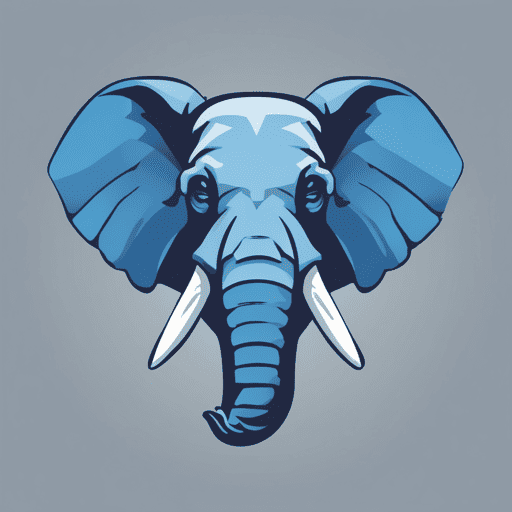}
& \includegraphics[width=\linewidth]{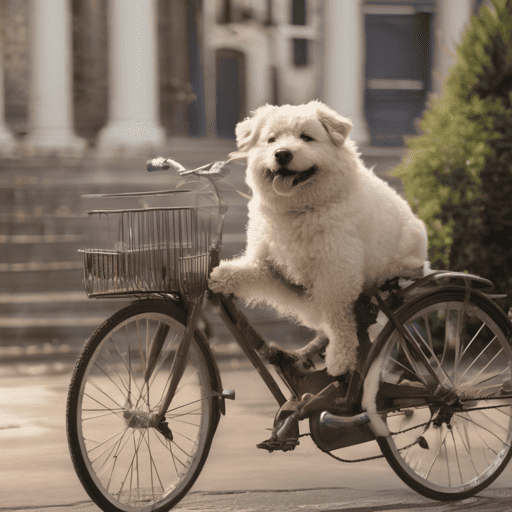}
& \includegraphics[width=\linewidth]{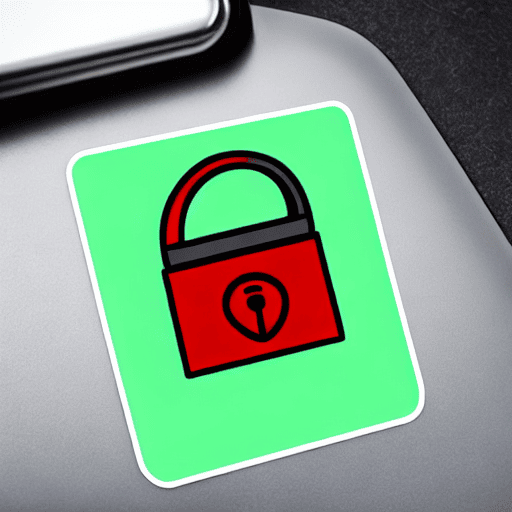} \\

% Row 2 – Diff-DPO
\makecell[c]{\textbf{Diff-DPO}\\\textbf{with}\\\textbf{\textit{OpenPref}}} 
& \includegraphics[width=\linewidth]{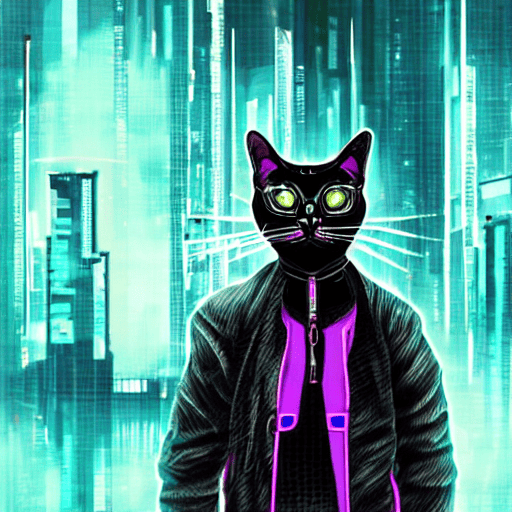}
& \includegraphics[width=\linewidth]{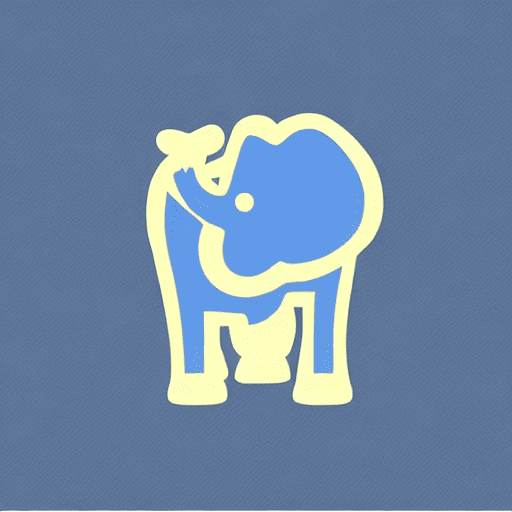}
& \includegraphics[width=\linewidth]{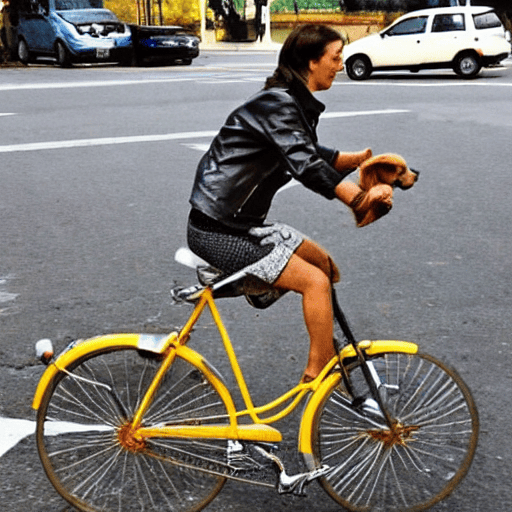}
& \includegraphics[width=\linewidth]{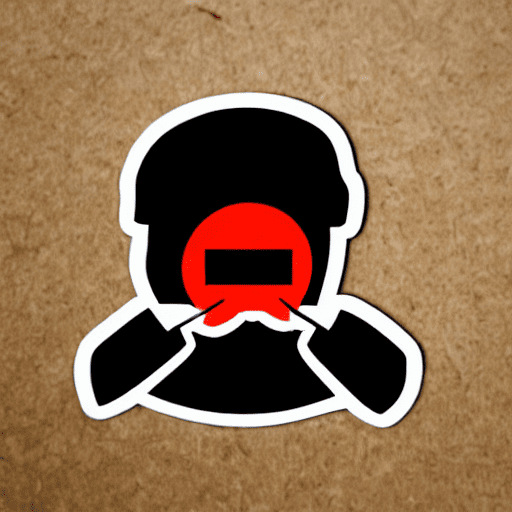} \\

% Row 3 – Ours with Diff-DPO
\makecell[c]{\textbf{Diff-DPO}\\\textbf{with}\\\textbf{\textsc{PrefLens}}} 
& \includegraphics[width=\linewidth]{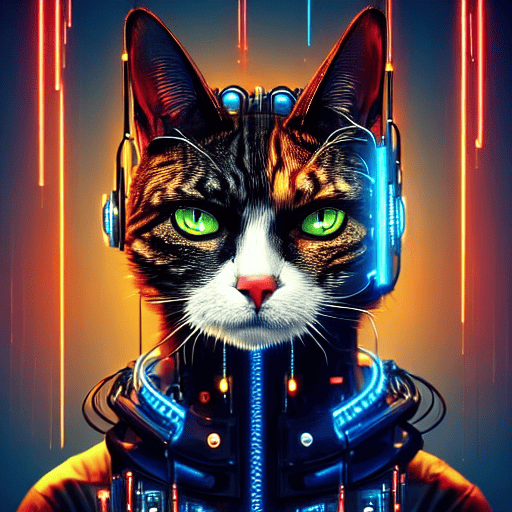}
& \includegraphics[width=\linewidth]{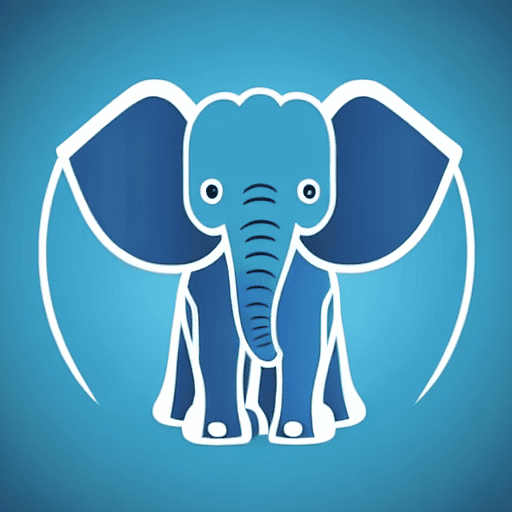}
& \includegraphics[width=\linewidth]{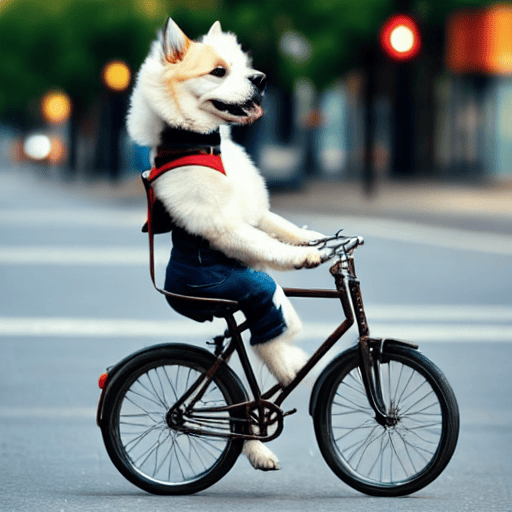}
& \includegraphics[width=\linewidth]{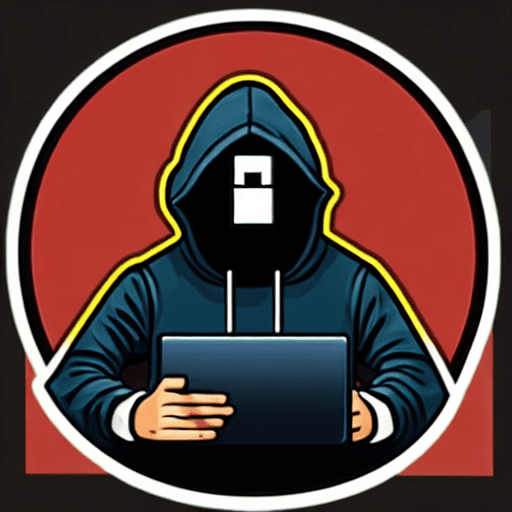} \\

\end{tabular}
}

\caption{
\textbf{Qualitative comparison on Pick-a-Pic-v2 using SD v1.5.} For
``Cyberpunk Cat,'' \textsc{PrefLens} preserves a clearly recognizable cat
while retaining the requested cyberpunk appearance, whereas Diffusion-DPO
produces a more humanoid depiction. For the ``blue elephant'' prompt,
\textsc{PrefLens} better matches the requested flat vector-logo style.
For ``a dog riding a bicycle,'' the OpenPref-trained model depicts a human
riding while holding a dog, whereas \textsc{PrefLens} correctly places the
dog itself on the bicycle. For ``sticker for a hacker,'' \textsc{PrefLens}
renders a more explicit hacker-themed sticker with a hooded figure and
laptop, rather than generic security iconography.
}
\label{fig:sd15_picapic}
\end{minipage}

\end{figure*}

\clearpage
\begin{figure*}[p]
    \centering
\scalebox{0.68}{
\begin{tabular}{>{\raggedright\arraybackslash}m{3cm} *{4}{>{\centering\arraybackslash}m{3.2cm}}}

& \makecell{
\textbf{\fontsize{10}{8}\selectfont Red and yellow}\\[-2pt]
\textbf{\fontsize{10}{8}\selectfont balloons hanging}\\[-2pt]
\textbf{\fontsize{10}{8}\selectfont from a ceiling}\\[-2pt]
\textbf{\fontsize{10}{8}\selectfont fan}
}

& \makecell{
\textbf{\fontsize{10}{8}\selectfont A motorcycle}\\[-2pt]
\textbf{\fontsize{10}{8}\selectfont in front of}\\[-2pt]
\textbf{\fontsize{10}{8}\selectfont a rhinoceros}
}

& \makecell{
\textbf{\fontsize{10}{8}\selectfont A giant cobra}\\[-2pt]
\textbf{\fontsize{10}{8}\selectfont snake made}\\[-2pt]
\textbf{\fontsize{10}{8}\selectfont from corn}
}

& \makecell{
\textbf{\fontsize{10}{8}\selectfont A group of}\\[-2pt]
\textbf{\fontsize{10}{8}\selectfont not more than}\\[-2pt]
\textbf{\fontsize{10}{8}\selectfont five meerkats}\\[-2pt]
\textbf{\fontsize{10}{8}\selectfont with sunset}
}
 \\[16pt]

% Row 1 – Base (SD v1.5)
\makecell[c]{\textbf{Base} \\ \textbf{(SD v1.5)}}

& \includegraphics[width=\linewidth]{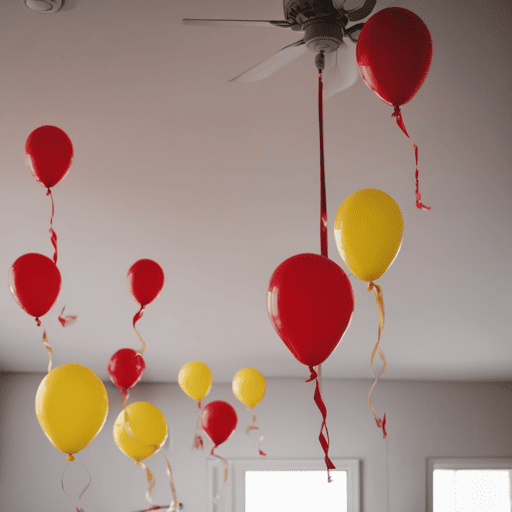}
& \includegraphics[width=\linewidth]{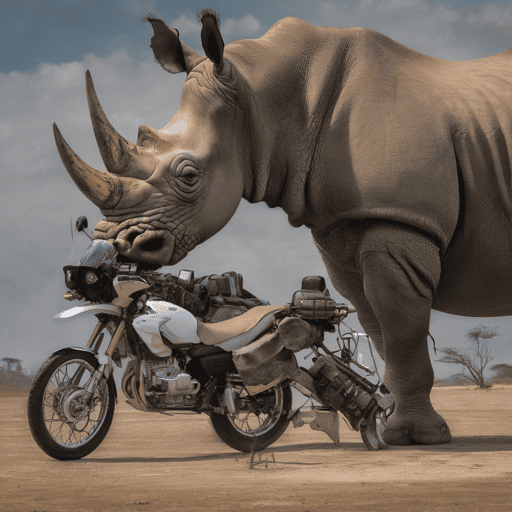}
& \includegraphics[width=\linewidth]{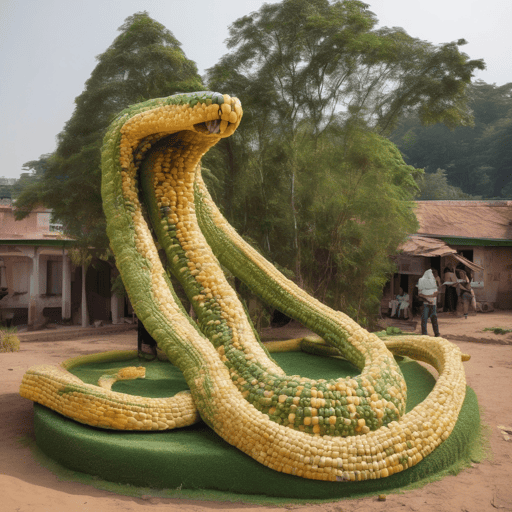}
& \includegraphics[width=\linewidth]{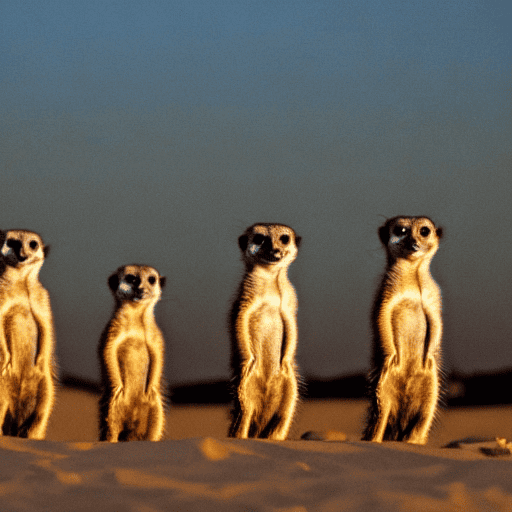} \\

% Row 2 – Diff-DPO
\makecell[c]{\textbf{Diff-DPO}\\\textbf{with}\\\textbf{\textit{OpenPref}}} 
& \includegraphics[width=\linewidth]{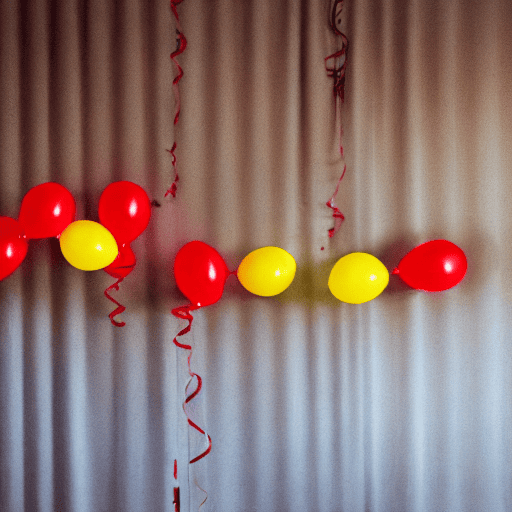}
& \includegraphics[width=\linewidth]{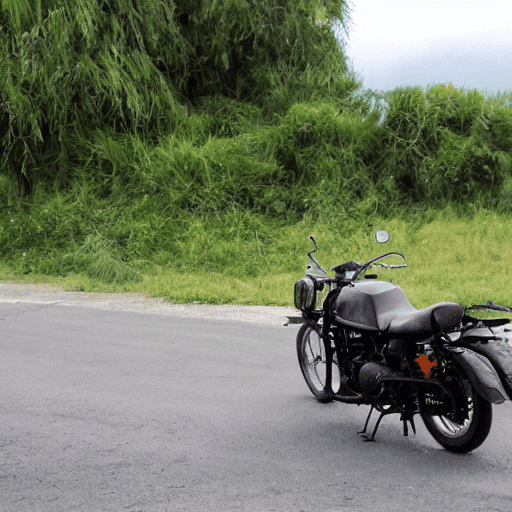}
& \includegraphics[width=\linewidth]{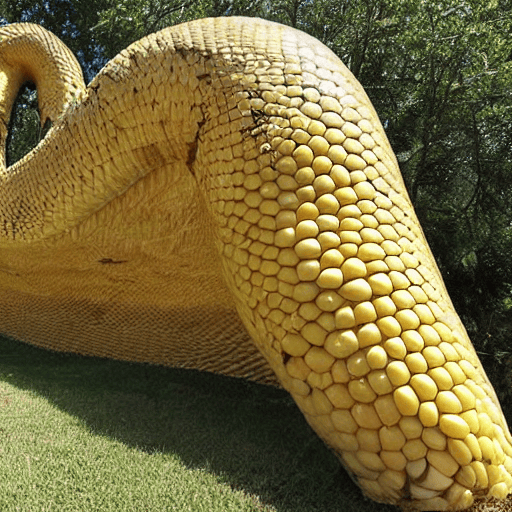}
& \includegraphics[width=\linewidth]{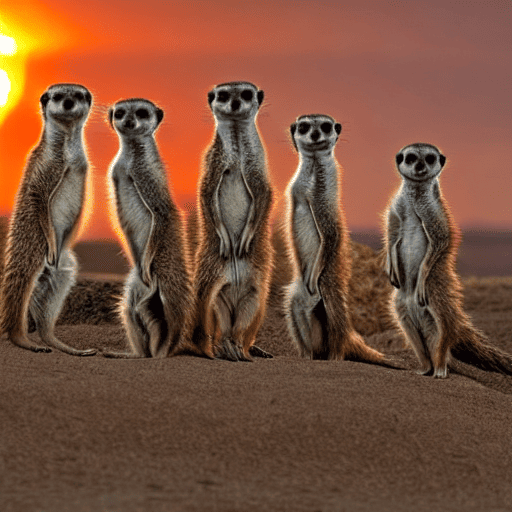} \\

% Row 3 – Ours with Diff-DPO
\makecell[c]{\textbf{Diff-DPO}\\\textbf{with}\\\textbf{\textsc{PrefLens}}} 
& \includegraphics[width=\linewidth]{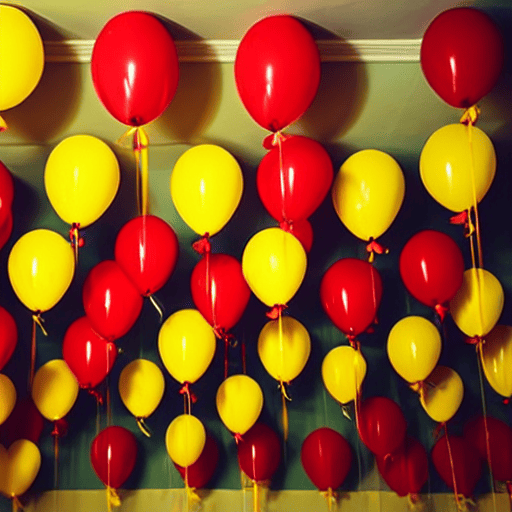}
& \includegraphics[width=\linewidth]{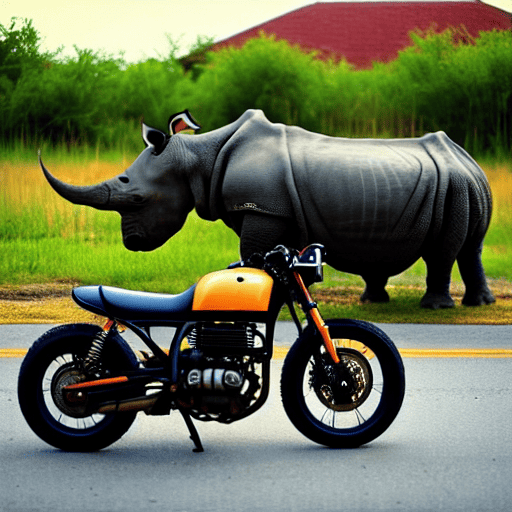}
& \includegraphics[width=\linewidth]{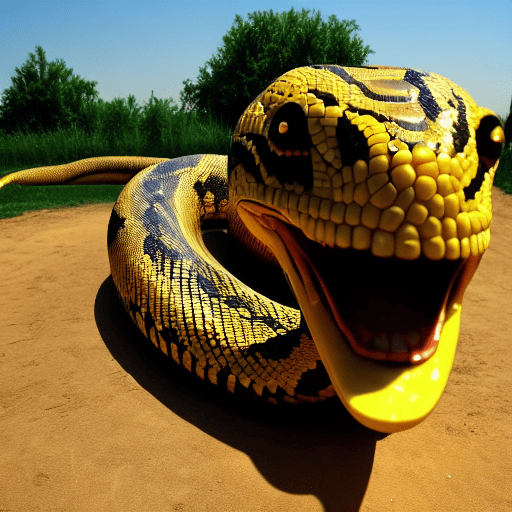}
& \includegraphics[width=\linewidth]{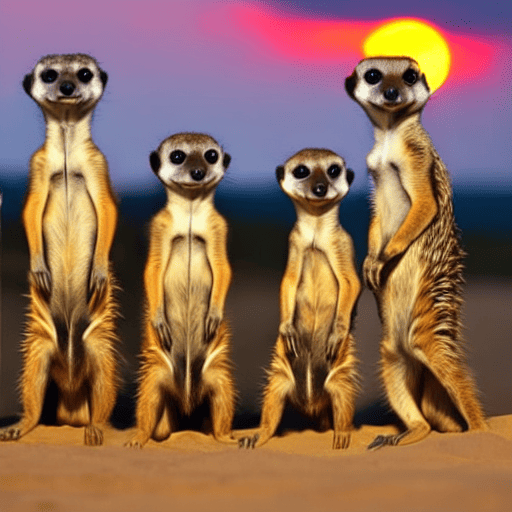} \\

\end{tabular}
}
\caption{
\textbf{Qualitative comparison on PartiPrompt using SD v1.5.} For
``red and yellow balloons hanging from a ceiling fan,''
\textsc{PrefLens} better captures both the requested colors and the
dense hanging arrangement, whereas the OpenPref-trained model misses the
ceiling-fan structure. For ``a motorcycle in front of a rhinoceros,''
\textsc{PrefLens} more clearly preserves both objects and their spatial
relation, while the OpenPref-trained model omits the rhinoceros. For
``a giant cobra snake made from corn,'' \textsc{PrefLens} produces a
clear cobra-shaped figure with recognizable corn texture, whereas the
OpenPref-trained model yields a less faithful close-up view. For
``a group of not more than five meerkats with sunset,''
\textsc{PrefLens} better satisfies both the counting constraint and the
sunset appearance.
}
\label{fig:sd15_parti}
\end{figure*}

\begin{figure*}[p]
    \centering
\scalebox{0.68}{
\begin{tabular}{>{\raggedright\arraybackslash}m{3cm} *{4}{>{\centering\arraybackslash}m{3.2cm}}}

& \makecell{
\textbf{\fontsize{10}{8}\selectfont A garden path lined}\\[-2pt]
\textbf{\fontsize{10}{8}\selectfont with glowing stones}\\[-2pt]
\textbf{\fontsize{10}{8}\selectfont under a twilight}\\[-2pt]
\textbf{\fontsize{10}{8}\selectfont sky.}
}

& \makecell{
\textbf{\fontsize{10}{8}\selectfont A couple, with the}\\[-2pt]
\textbf{\fontsize{10}{8}\selectfont taller one hugging}\\[-2pt]
\textbf{\fontsize{10}{8}\selectfont the shorter one}\\[-2pt]
\textbf{\fontsize{10}{8}\selectfont from behind.}
}

& \makecell{
\textbf{\fontsize{10}{8}\selectfont A vase with water,}\\[-2pt]
\textbf{\fontsize{10}{8}\selectfont but no flowers}\\[-2pt]
\textbf{\fontsize{10}{8}\selectfont to nourish.}
}

& \makecell{
\textbf{\fontsize{10}{8}\selectfont A green chalkboard}\\[-2pt]
\textbf{\fontsize{10}{8}\selectfont with a detailed}\\[-2pt]
\textbf{\fontsize{10}{8}\selectfont chalk drawing}\\[-2pt]
\textbf{\fontsize{10}{8}\selectfont of a cityscape.}
}
\\[6pt]

% Row 1 – Base (SD v1.5)
\makecell[c]{\textbf{Base} \\ \textbf{(SD v1.5)}}

& \includegraphics[width=\linewidth]{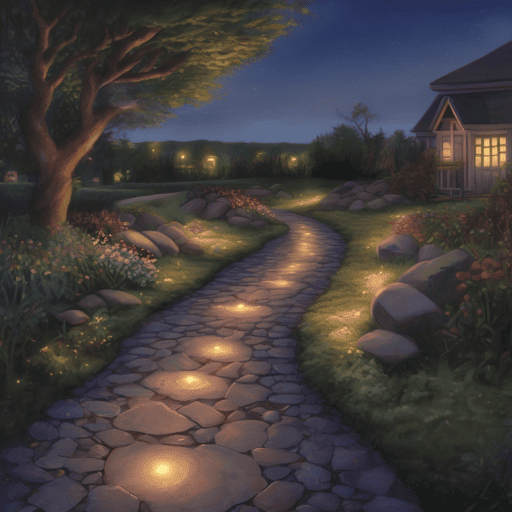}
& \includegraphics[width=\linewidth]{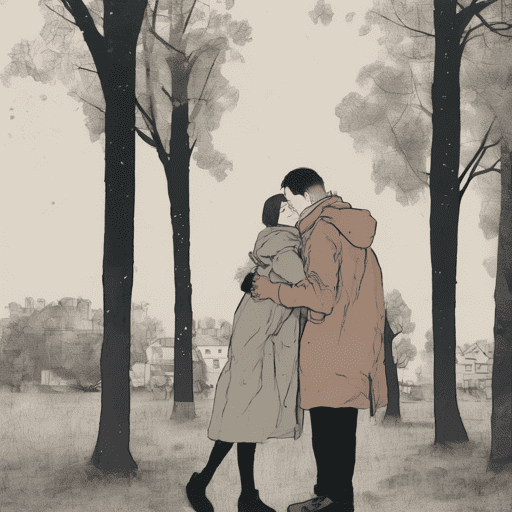}
& \includegraphics[width=\linewidth]{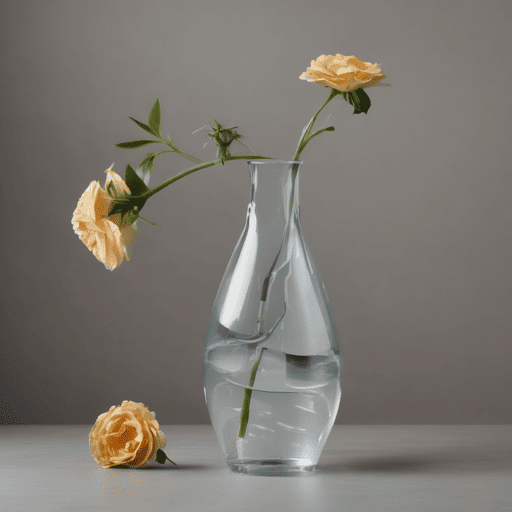}
& \includegraphics[width=\linewidth]{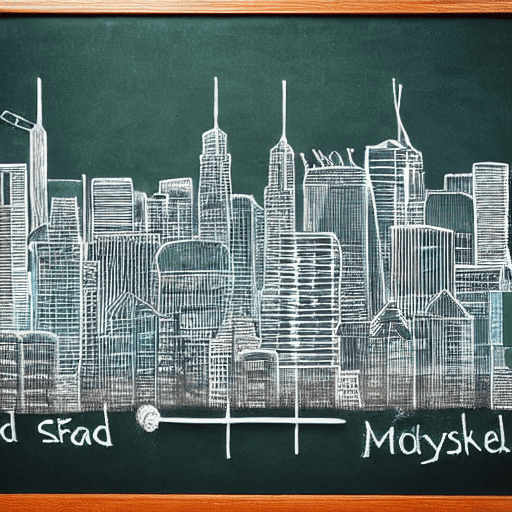} \\

% Row 2 – Diff-DPO
\makecell[c]{\textbf{Diff-DPO}\\\textbf{with}\\\textbf{\textit{OpenPref}}}
& \includegraphics[width=\linewidth]{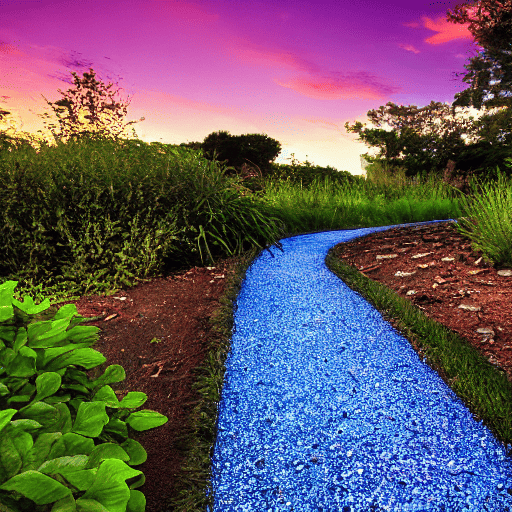}
& \includegraphics[width=\linewidth]{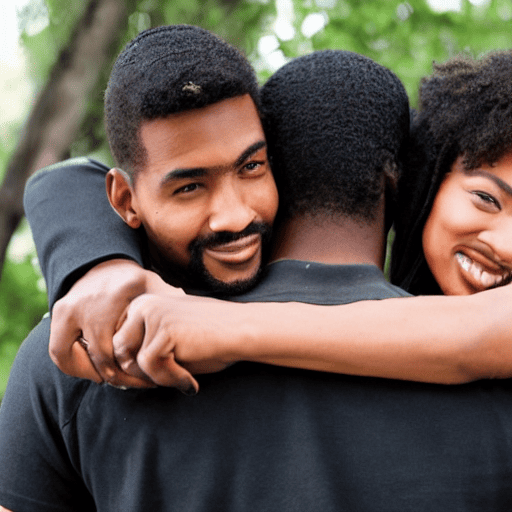}
& \includegraphics[width=\linewidth]{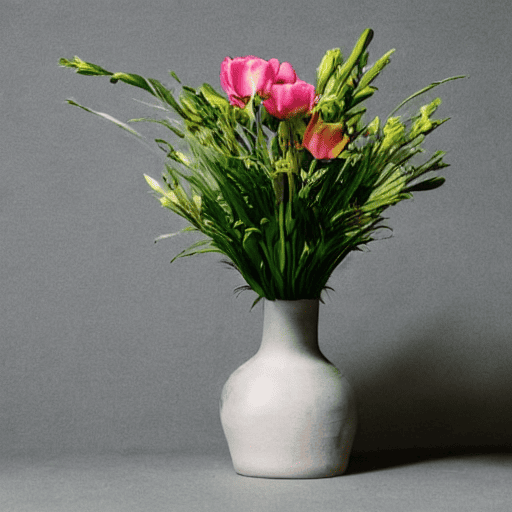}
& \includegraphics[width=\linewidth]{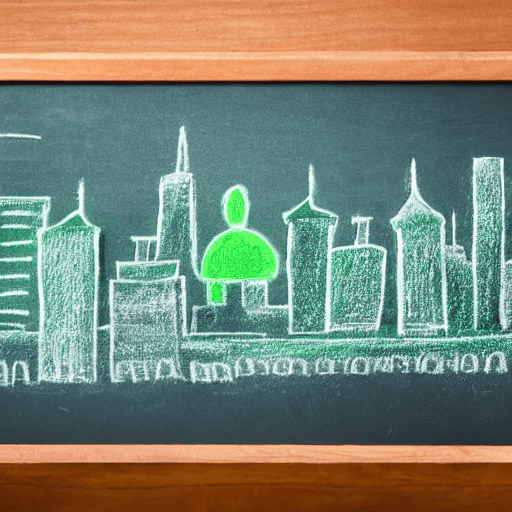} \\

% Row 3 – Ours with Diff-DPO
\makecell[c]{\textbf{Diff-DPO}\\\textbf{with}\\\textbf{\textsc{PrefLens}}} 
& \includegraphics[width=\linewidth]{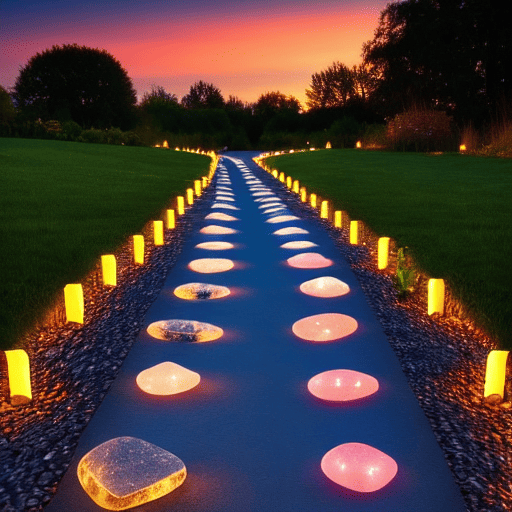}
& \includegraphics[width=\linewidth]{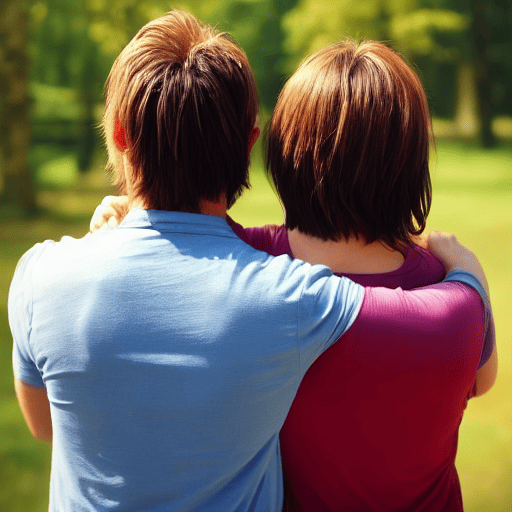}
& \includegraphics[width=\linewidth]{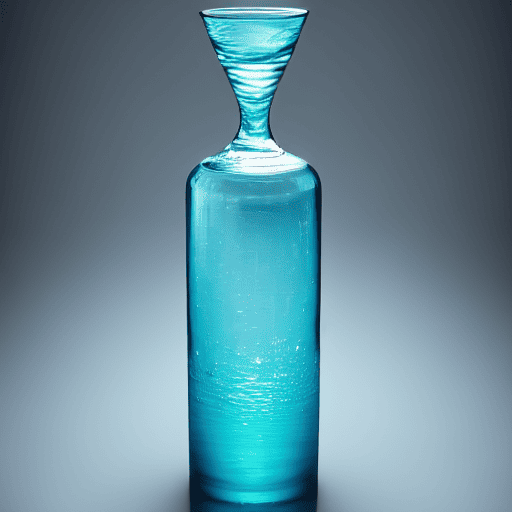}
& \includegraphics[width=\linewidth]{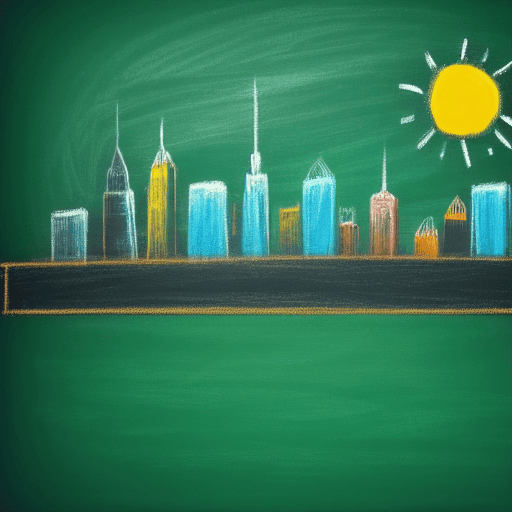} \\

\end{tabular}
}

\caption{
\textbf{Qualitative comparison on GenAIBench using SD v1.5.} For the ``garden path'' prompt,
\textsc{PrefLens} renders a path explicitly lined with glowing stones under
a twilight setting, while Diffusion-DPO with OpenPref mainly produces a
brightly colored path. For the hugging prompt, \textsc{PrefLens} better
preserves the requested two-person composition and rear embrace, whereas
the OpenPref model introduces an additional person. The strongest
difference appears for ``a vase with water, but no flowers,'' where both
the Base and OpenPref models add flowers, while \textsc{PrefLens} correctly
renders a water-filled vase without them. For the chalkboard example,
\textsc{PrefLens} retains the requested green-board cityscape composition,
although with a more stylized rendering.
}

\label{fig:sd15_genai}
\end{figure*}

\begin{figure*}[p]
\centering
\begin{minipage}{\linewidth}
\captionsetup{type=figure,font=footnotesize}
    \centering

\scalebox{0.68}{
\begin{tabular}{
>{\raggedright\arraybackslash}m{3cm}
*{4}{>{\centering\arraybackslash}m{3.2cm}}
}

% ============================================================
% Prompt headers
% ============================================================

&
\makecell{
\textbf{\fontsize{10}{9}\selectfont Three demons}\\[-2pt]
\textbf{\fontsize{10}{9}\selectfont together holding}\\[-2pt]
\textbf{\fontsize{10}{9}\selectfont swords around hell}
}

&
\makecell{
\textbf{\fontsize{10}{9}\selectfont A guinea pig}\\[-2pt]
\textbf{\fontsize{10}{9}\selectfont riding a}\\[-2pt]
\textbf{\fontsize{10}{9}\selectfont motorcycle}
}

&
\makecell{
\textbf{\fontsize{10}{9}\selectfont 3 guys standing}\\[-2pt]
\textbf{\fontsize{10}{9}\selectfont in front of}\\[-2pt]
\textbf{\fontsize{10}{9}\selectfont Overland Trucks}
}

&
\makecell{
\textbf{\fontsize{10}{9}\selectfont A cube in the}\\[-2pt]
\textbf{\fontsize{10}{9}\selectfont style of}\\[-2pt]
\textbf{\fontsize{10}{9}\selectfont Pastel Colors}
}

\\

% ============================================================
% Row 1 -- Base SDXL
% ============================================================

\makecell[c]{
\textbf{Base}\\
\textbf{(SDXL)}
}

&
\includegraphics[width=\linewidth]
{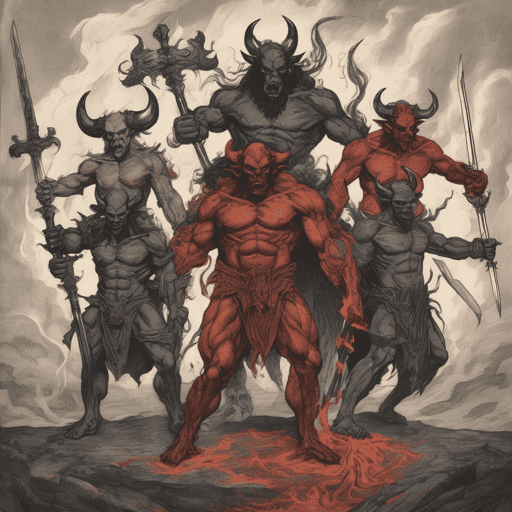}

&
\includegraphics[width=\linewidth]
{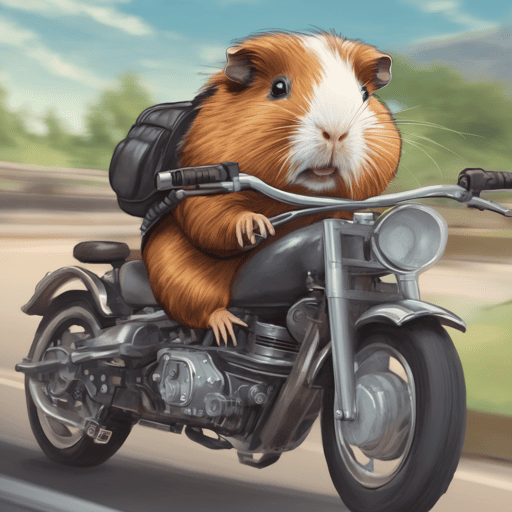}

&
\includegraphics[width=\linewidth]
{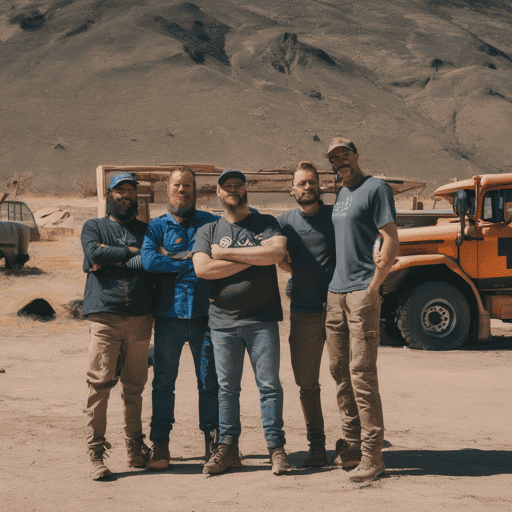}

&
\includegraphics[width=\linewidth]
{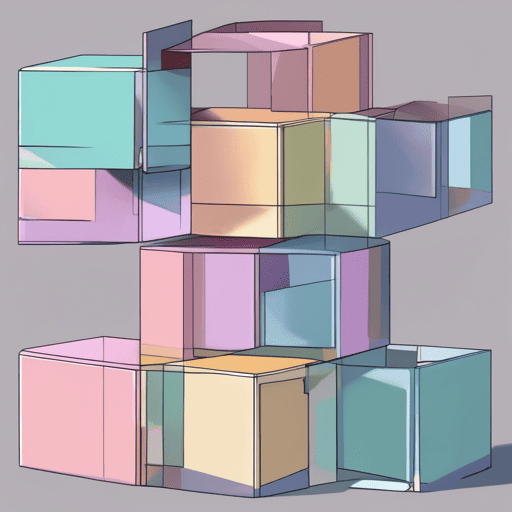}

\\

% ============================================================
% Row 2 -- Diffusion-DPO with OpenPref
% ============================================================

\makecell[c]{
\textbf{Diff-DPO}\\
\textbf{with}\\
\textbf{\textit{OpenPref}}
}

&
\includegraphics[width=\linewidth]
{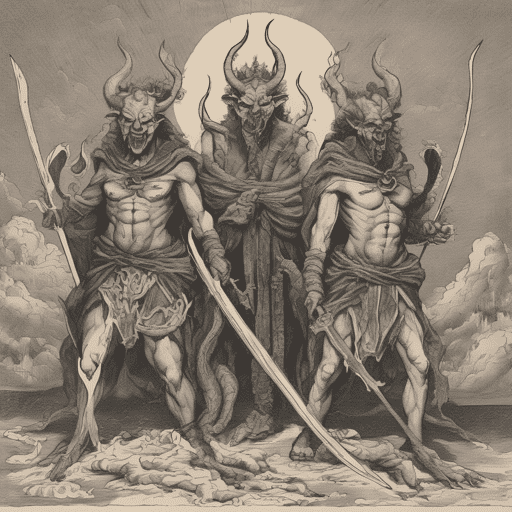}

&
\includegraphics[width=\linewidth]
{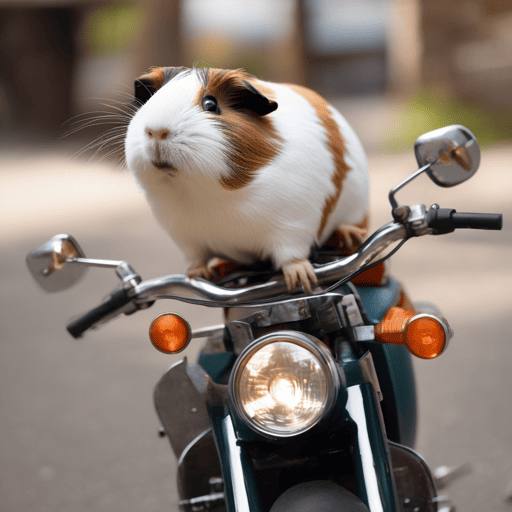}

&
\includegraphics[width=\linewidth]
{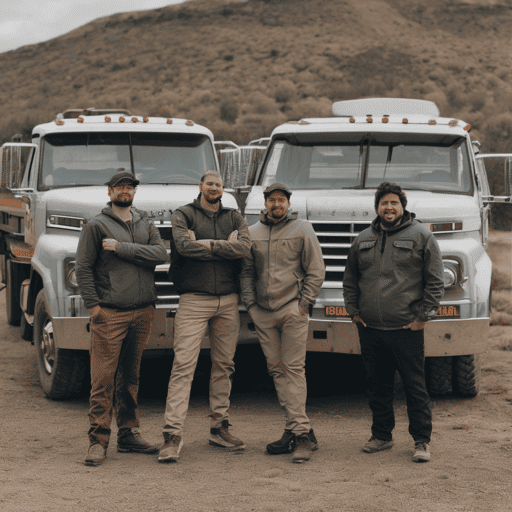}

&
\includegraphics[width=\linewidth]
{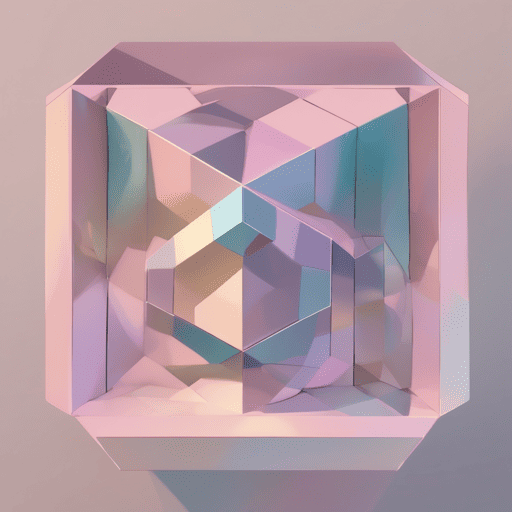}

\\

% ============================================================
% Row 3 -- Diffusion-DPO with PrefLens
% ============================================================

\makecell[c]{
\textbf{Diff-DPO}\\
\textbf{with}\\
\textbf{\textsc{PrefLens}}
}

&
\includegraphics[width=\linewidth]
{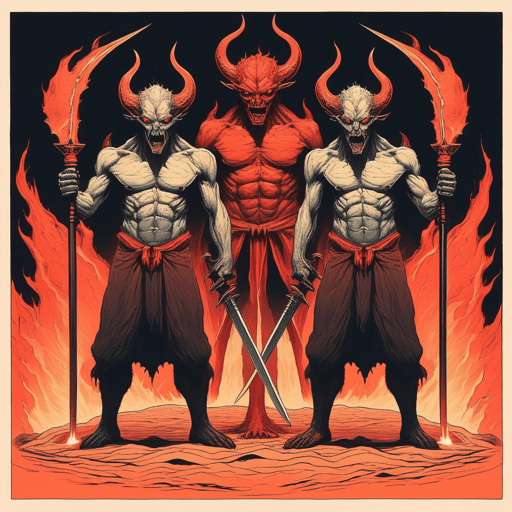}

&
\includegraphics[width=\linewidth]
{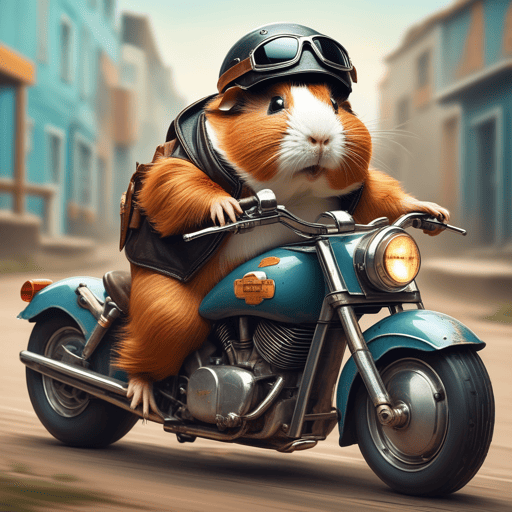}

&
\includegraphics[width=\linewidth]
{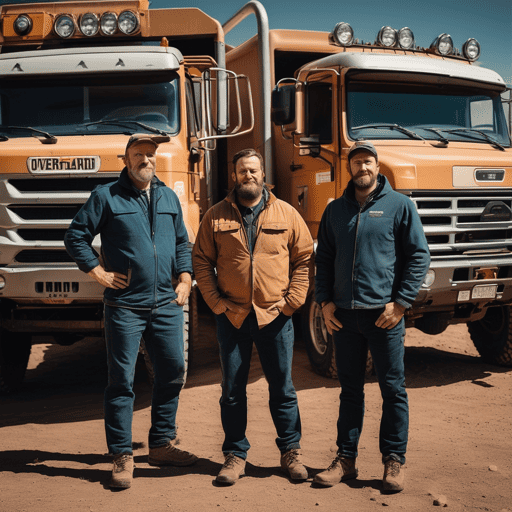}

&
\includegraphics[width=\linewidth]
{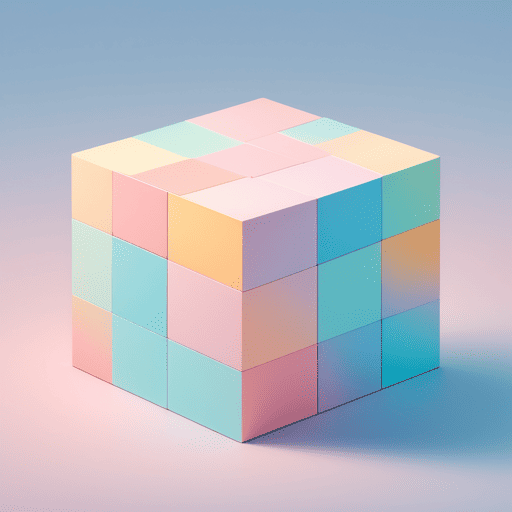}

\\

\end{tabular}
}

\caption{
\textbf{Qualitative comparison on Pick-a-Pic-v2 using SDXL-Base.} The demon example preserves the requested three-character
composition together with the swords and infernal visual theme. For the guinea-pig example,
\textsc{PrefLens} more clearly depicts the animal riding the motorcycle
rather than merely being positioned on it. For ``3 guys standing in front
of Overland Trucks,'' it better satisfies both the requested object count
and the spatial relation, rendering three people directly in front of the
trucks. In the pastel-cube example, the Base and OpenPref variants produce
more complex multi-cube or abstract structures, whereas \textsc{PrefLens}
renders a single, clearly defined cube with the requested pastel appearance.
}
\label{fig:sdxl_picapic}

\end{minipage}
\par\vspace{4pt}
\begin{minipage}{\linewidth}
\captionsetup{type=figure,font=footnotesize}
    \centering
\scalebox{0.68}{
\begin{tabular}{>{\raggedright\arraybackslash}m{3cm} *{4}{>{\centering\arraybackslash}m{3.2cm}}}

& \makecell{
\textbf{\fontsize{10}{8}\selectfont An orange pickup}\\[-2pt]
\textbf{\fontsize{10}{8}\selectfont truck next to a}\\[-2pt]
\textbf{\fontsize{10}{8}\selectfont yellow Porsche 911}
}

& \makecell{
\textbf{\fontsize{10}{8}\selectfont The Statue of Liberty}\\[-2pt]
\textbf{\fontsize{10}{8}\selectfont with the face}\\[-2pt]
\textbf{\fontsize{10}{8}\selectfont of an owl}
}

& \makecell{
\textbf{\fontsize{10}{8}\selectfont A group of not more}\\[-2pt]
\textbf{\fontsize{10}{8}\selectfont than five meerkats}\\[-2pt]
\textbf{\fontsize{10}{8}\selectfont standing with the sun}\\[-2pt]
\textbf{\fontsize{10}{8}\selectfont setting behind them}
}

& \makecell{
\textbf{\fontsize{10}{8}\selectfont A cartoon of a boy}\\[-2pt]
\textbf{\fontsize{10}{8}\selectfont playing with}\\[-2pt]
\textbf{\fontsize{10}{8}\selectfont a tiger}
}
\\

% Row 1 – Base
\makecell[c]{\textbf{Base}\\\textbf{(SDXL)}}
& \includegraphics[width=\linewidth]{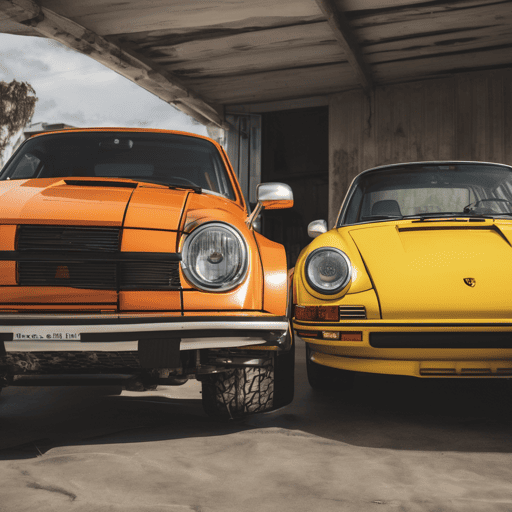}
& \includegraphics[width=\linewidth]{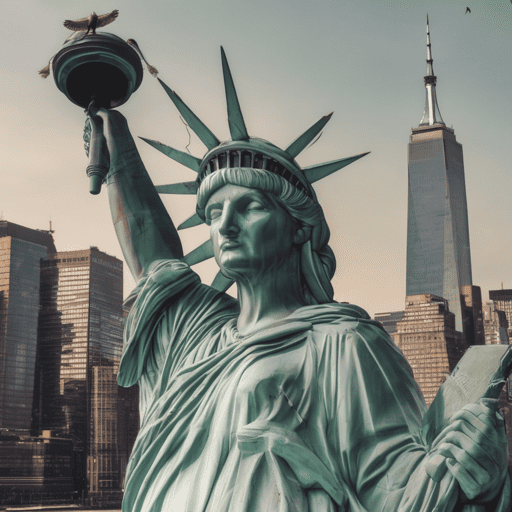}
& \includegraphics[width=\linewidth]{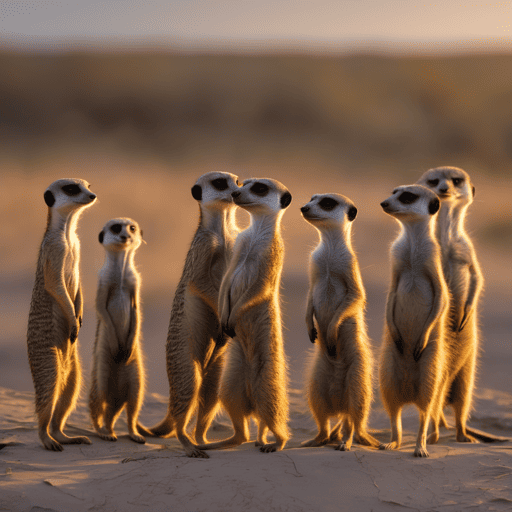}
& \includegraphics[width=\linewidth]{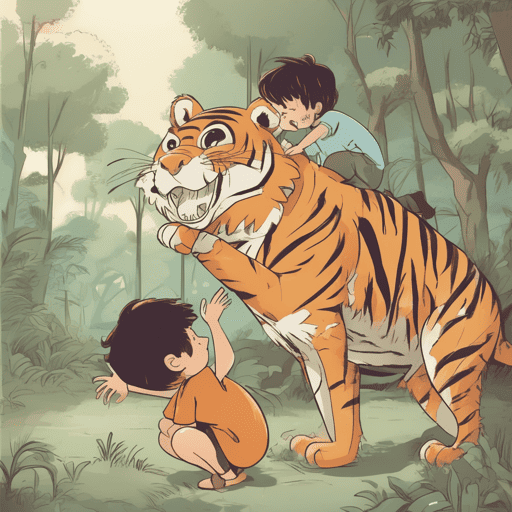} \\

% Row 2 – Diff-DPO with OpenPref
\makecell[c]{\textbf{Diff-DPO}\\\textbf{with}\\\textbf{\textit{OpenPref}}}
& \includegraphics[width=\linewidth]{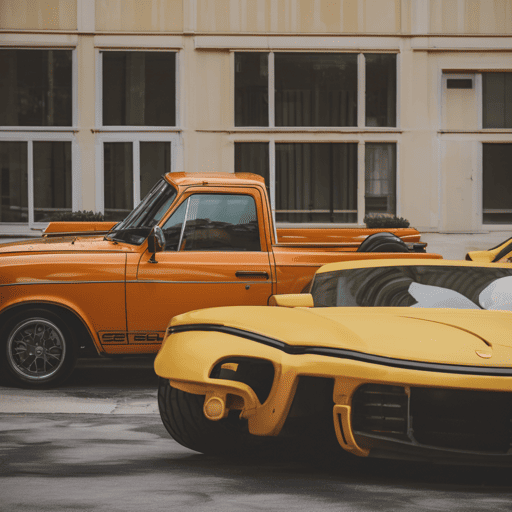}
& \includegraphics[width=\linewidth]{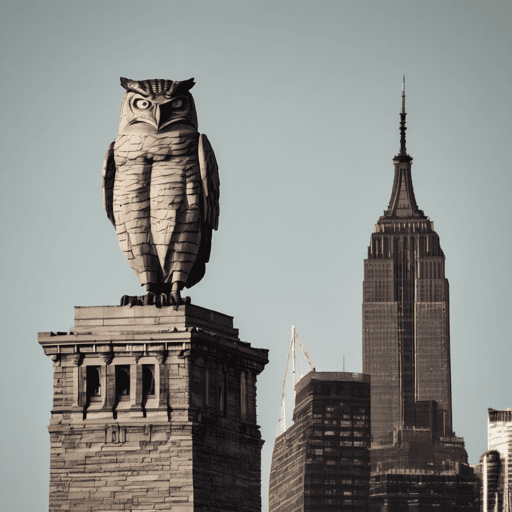}
& \includegraphics[width=\linewidth]{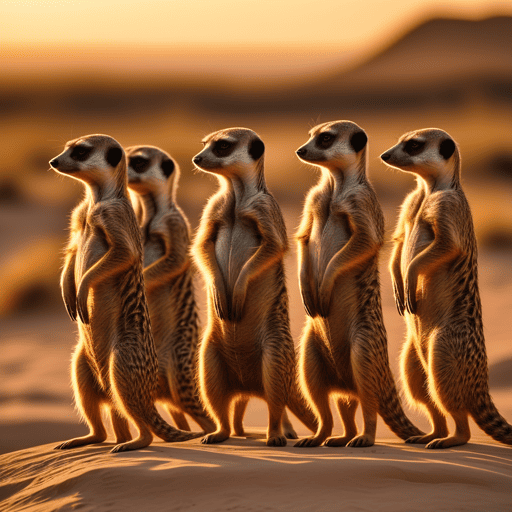}
& \includegraphics[width=\linewidth]{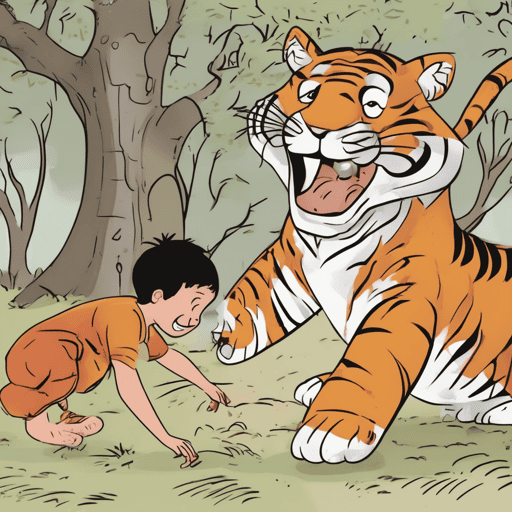} \\

% Row 3 – Diff-DPO with PrefLens
\makecell[c]{\textbf{Diff-DPO}\\\textbf{with}\\\textbf{\textsc{PrefLens}}}
& \includegraphics[width=\linewidth]{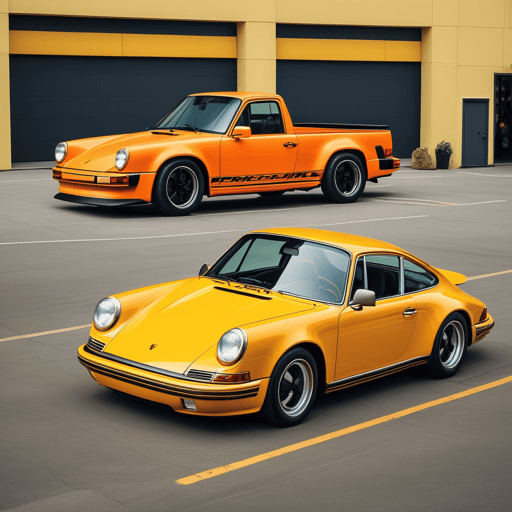}
& \includegraphics[width=\linewidth]{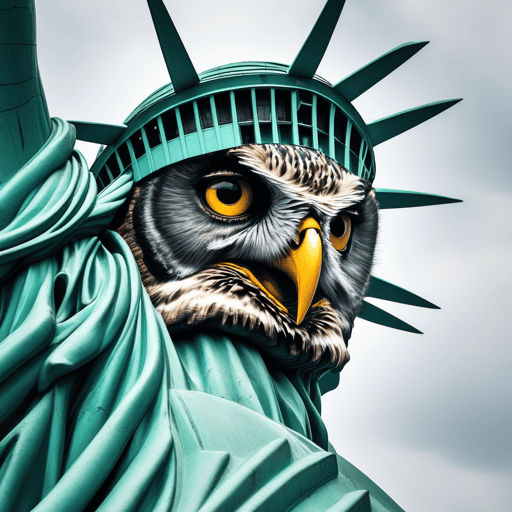}
& \includegraphics[width=\linewidth]{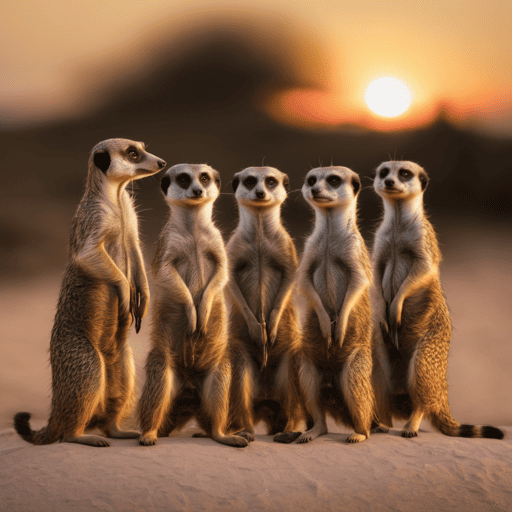}
& \includegraphics[width=\linewidth]{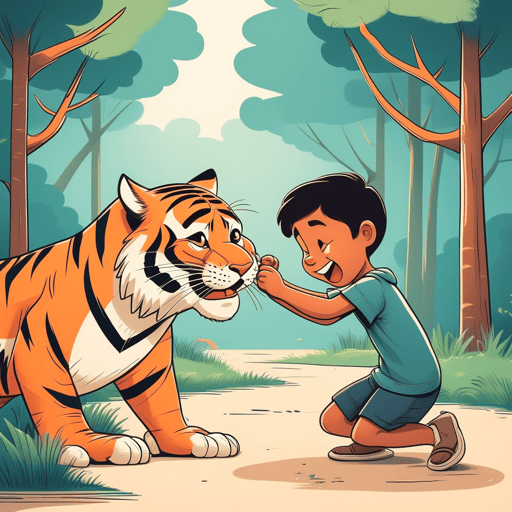} \\

\end{tabular}
}

\caption{
\textbf{Qualitative comparison on PartiPrompt using SDXL-Base.} In the first example, \textsc{PrefLens}
clearly renders an orange pickup truck next to a yellow Porsche 911,
preserving both the requested object identities and their relation. In the
second, it integrates an owl face into the Statue of Liberty, whereas the
other variants either retain a conventional statue or depict an owl without
the requested statue composition. In the third, \textsc{PrefLens} respects
the ``not more than five meerkats'' constraint while retaining the sunset
setting. Finally, for the boy-and-tiger prompt, it produces a clear
cartoon-style interaction between a single boy and a tiger, more closely
matching the requested scene.
}
\label{fig:sdxl_partiprompt}
\end{minipage}
\end{figure*}
\newtcolorbox{evalprompt}[2][]{
    enhanced,
    sharp corners,
    colback=gray!4,
    colframe=black!65,
    boxrule=0.6pt,
    left=7pt,
    right=7pt,
    top=13pt,
    bottom=7pt,
    before skip=8pt,
    after skip=4pt,
    title=\textbf{#2},
    fonttitle=\bfseries\small,
    coltitle=white,
    attach boxed title to top left={
        xshift=8pt,
        yshift=-7pt
    },
    boxed title style={
        sharp corners,
        colback=black,
        colframe=black,
        boxrule=0pt,
        left=7pt,
        right=7pt,
        top=3.5pt,
        bottom=3.5pt
    },
    #1
}

% ============================================================
% Composition Evaluation Prompt
% ============================================================

\begin{figure*}[p]
\centering

\begin{minipage}{\linewidth}
\captionsetup{type=figure}
    \centering
\scalebox{0.68}{
\begin{tabular}{>{\raggedright\arraybackslash}m{3cm} *{4}{>{\centering\arraybackslash}m{3.2cm}}}

% ============================================================
% Prompt headers
% ============================================================

& \makecell{
\textbf{\fontsize{10}{8}\selectfont A small cat}\\[-2pt]
\textbf{\fontsize{10}{8}\selectfont waves a wand}
}

& \makecell{
\textbf{\fontsize{10}{8}\selectfont A pilot with}\\[-2pt]
\textbf{\fontsize{10}{8}\selectfont aviator sunglasses}
}

& \makecell{
\textbf{\fontsize{10}{8}\selectfont The dog with a leash}\\[-2pt]
\textbf{\fontsize{10}{8}\selectfont sits quietly, and the}\\[-2pt]
\textbf{\fontsize{10}{8}\selectfont other without a leash}\\[-2pt]
\textbf{\fontsize{10}{8}\selectfont runs wildly}
}

& \makecell{
\textbf{\fontsize{10}{8}\selectfont A musician with}\\[-2pt]
\textbf{\fontsize{10}{8}\selectfont a guitar, but there}\\[-2pt]
\textbf{\fontsize{10}{8}\selectfont is no audience}\\[-2pt]
\textbf{\fontsize{10}{8}\selectfont to listen}
}
\\[19pt]

% ============================================================
% Row 1 -- Base
% ============================================================

\makecell[c]{\textbf{Base}\\\textbf{(SDXL)}}

& \includegraphics[width=\linewidth]
{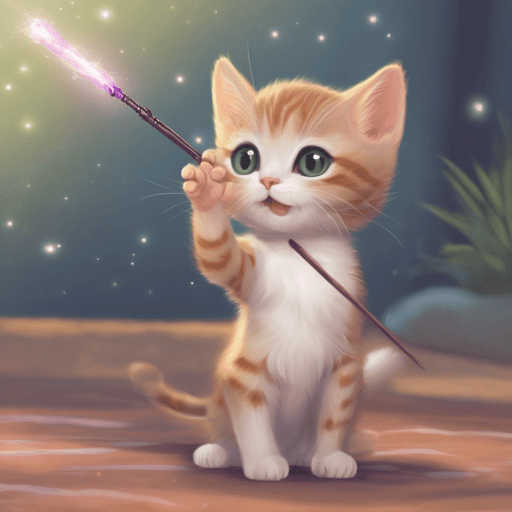}

& \includegraphics[width=\linewidth]
{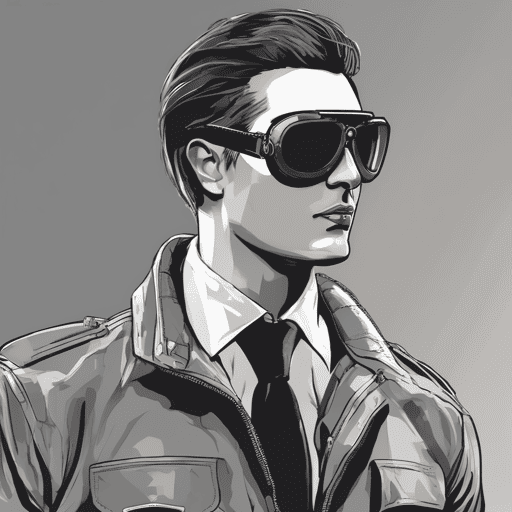}

& \includegraphics[width=\linewidth]
{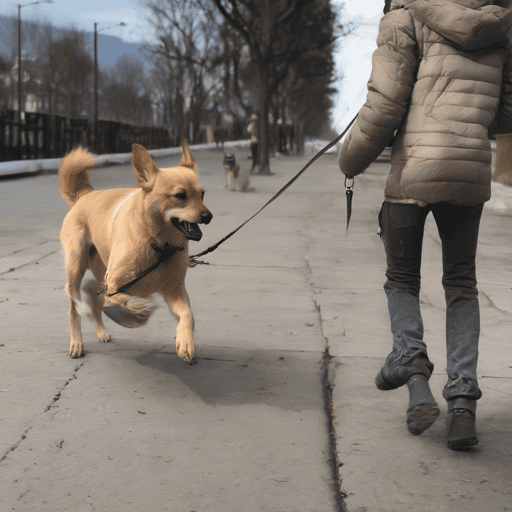}

& \includegraphics[width=\linewidth]
{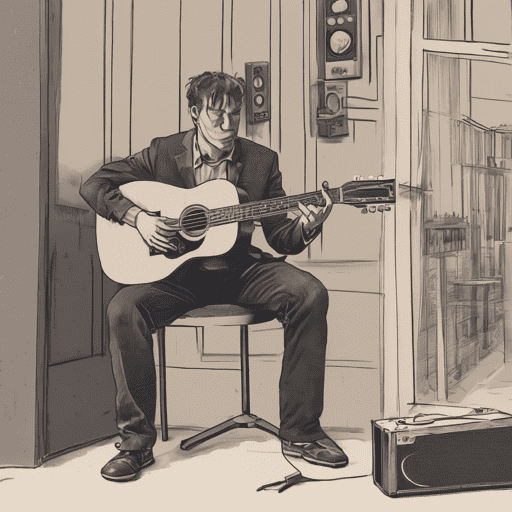}

\\

% ============================================================
% Row 2 -- Diffusion-DPO with OpenPref
% ============================================================

\makecell[c]{
\textbf{Diff-DPO}\\
\textbf{with}\\
\textbf{\textit{OpenPref}}
}

& \includegraphics[width=\linewidth]
{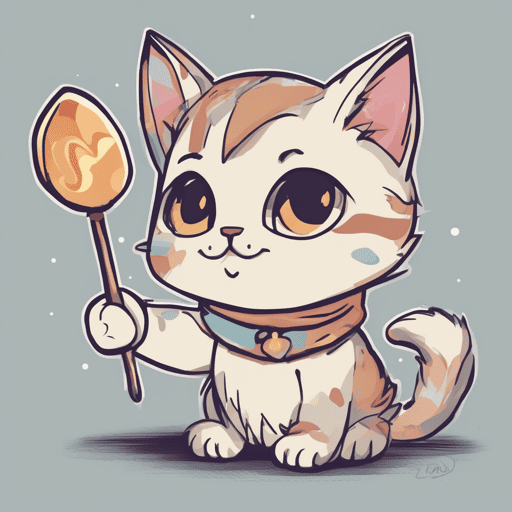}

& \includegraphics[width=\linewidth]
{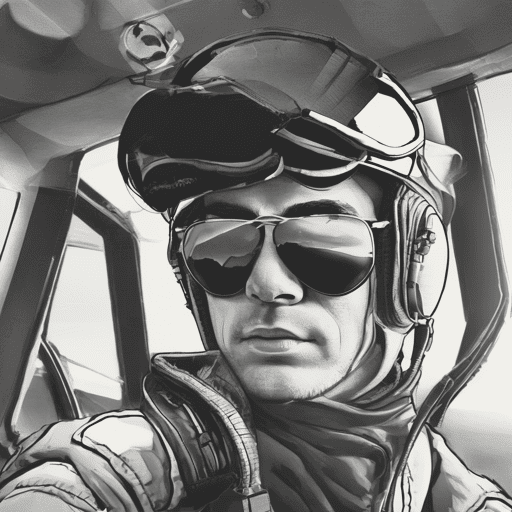}

& \includegraphics[width=\linewidth]
{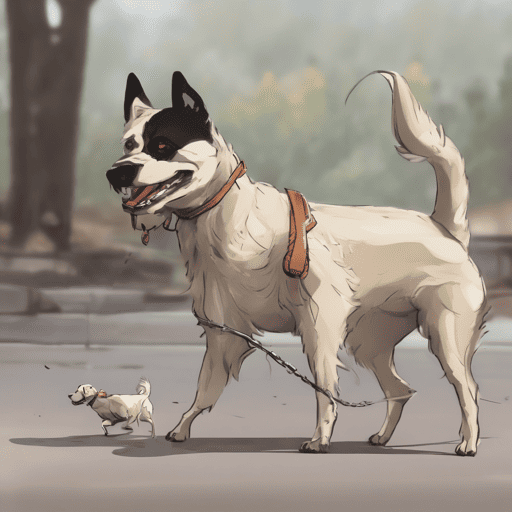}

& \includegraphics[width=\linewidth]
{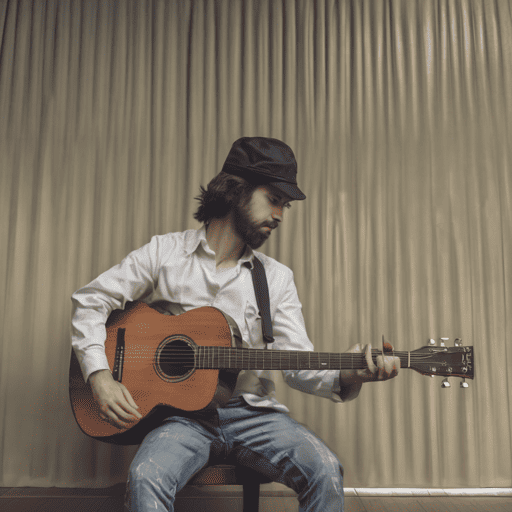}

\\

% ============================================================
% Row 3 -- Diffusion-DPO with PrefLens
% ============================================================

\makecell[c]{
\textbf{Diff-DPO}\\
\textbf{with}\\
\textbf{\textsc{PrefLens}}
}

& \includegraphics[width=\linewidth]
{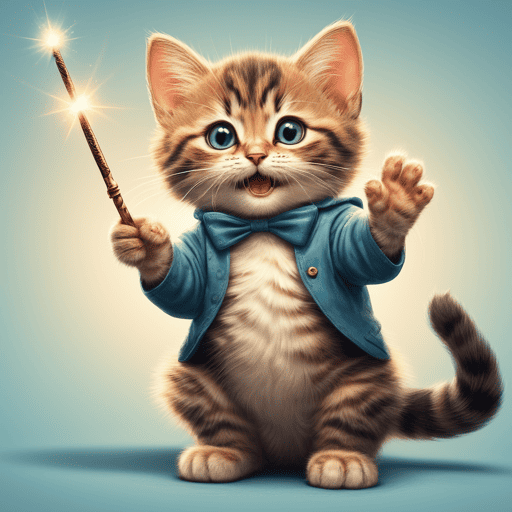}

& \includegraphics[width=\linewidth]
{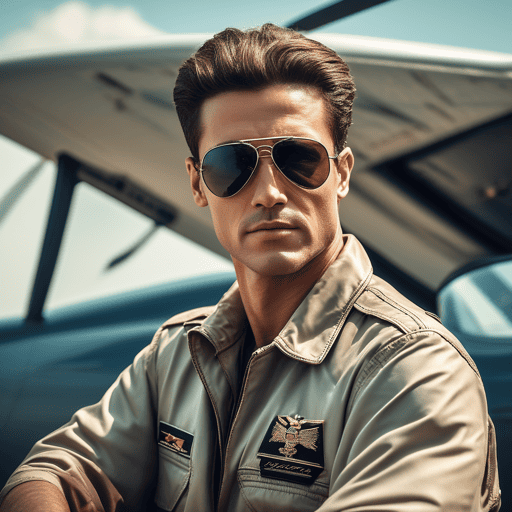}

& \includegraphics[width=\linewidth]
{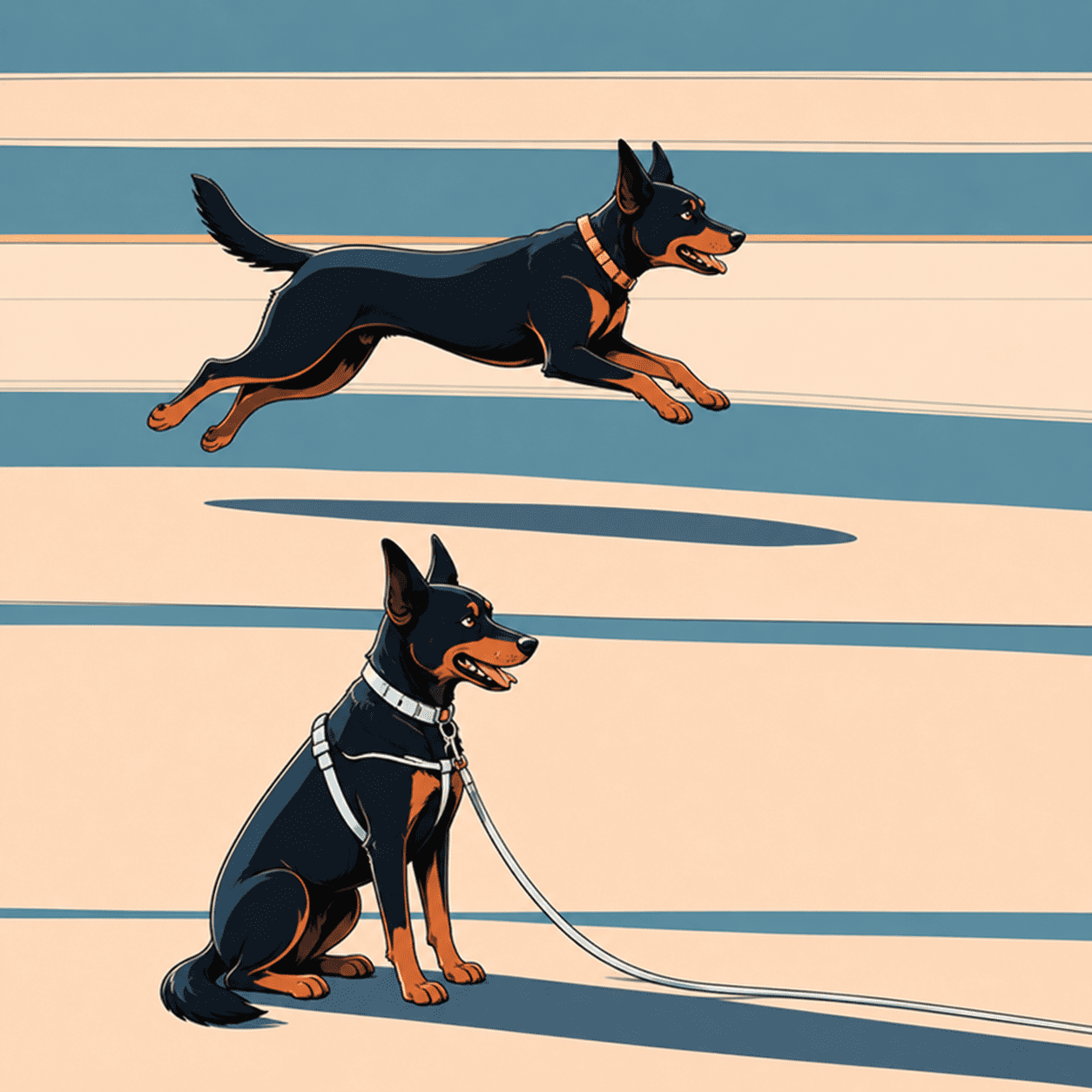}

& \includegraphics[width=\linewidth]
{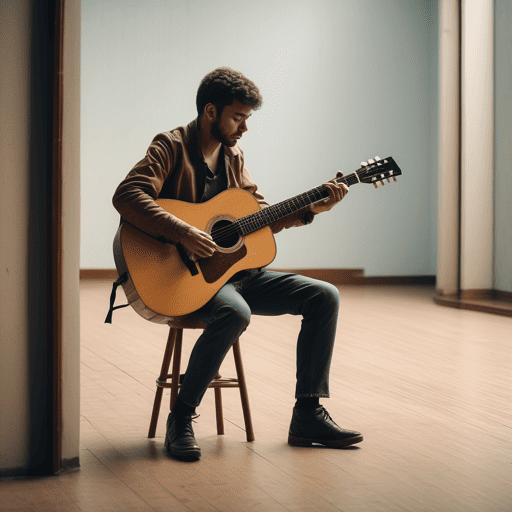}

\\

\end{tabular}
}

\caption{
\textbf{Qualitative comparison on GenAIBench using SDXL-Base.}
Rows show the Base model, Diffusion-DPO with OpenPref, and Diffusion-DPO
with \textsc{PrefLens}, respectively. In the first example,
\textsc{PrefLens} clearly renders a small cat actively holding and waving
a wand. In the second, it preserves both the pilot appearance and the
requested aviator sunglasses. The third example requires two contrasting
actions: one dog with a leash sitting quietly and another unleashed dog
running; \textsc{PrefLens} captures both the leash condition and the
distinct actions more explicitly. Finally, for the musician prompt,
\textsc{PrefLens} renders a guitarist in an isolated setting without an
audience, preserving the requested absence constraint.
}
\label{fig:sdxl_genai}

\end{minipage}
\par\vspace{4pt}

\begin{evalprompt}[top=7pt,bottom=5pt]{Composition Evaluation Prompt}

{\ttfamily\footnotesize

You are given a prompt and three images.

Prompt: \{prompt\_text\}

Image A is the first image.\\
Image B is the second image.\\
Image C is the third image.

Rank the images from best to worst based on how well each image satisfies the prompt.

Evaluate prompt adherence using factors such as:
\begin{itemize}[nosep]
\item correct object count
\item required object presence
\item absence of unprompted objects
\item object color
\item object shape and geometry
\item object scale
\item object location
\item relative spatial positioning
\item object interactions
\item overall visual clarity
\item background/context consistency
\end{itemize}

Use the prompt as the only reference.

Output strictly one ranking using only the letters A, B, and C separated by greater-than signs.

Valid examples:

\hspace*{1em}A $>$ C $>$ B\\
\hspace*{1em}A $>$ B $>$ C\\
\hspace*{1em}C $>$ A $>$ B

Do not output explanations, scores, bullets, or any additional text.
}

\end{evalprompt}

\caption{Prompt used for VLM-as-a-Judge evaluation of compositional fidelity.}
\label{fig:prompt_adherence}

\end{figure*}

% ============================================================
% Human Evaluation Instructions
% ============================================================

\begin{figure*}[p]
\centering

\begin{evalprompt}{Human Evaluation Instructions}

{\ttfamily\footnotesize

You are given a prompt and three images.

Prompt: \{prompt\_text\}

Image A is the first image.\\
Image B is the second image.\\
Image C is the third image.

Your task is to rank the three images from best to worst as a human evaluator would.

Carefully read the prompt and evaluate all images based on the following factors:

\textbf{1. Prompt Alignment}
\begin{itemize}
\item correctness of objects and attributes
\item adherence to the required scene structure
\end{itemize}

\textbf{2. Aesthetic Quality}
\begin{itemize}
\item realism
\item image quality
\item artistic appeal
\item overall attractiveness
\end{itemize}

\textbf{3. Composition and Visual Coherence}
\begin{itemize}
\item colors and orientation
\item object positioning
\item spatial relationships
\item interaction consistency
\end{itemize}

Use all factors jointly to determine the final ranking.

Output exactly one ranking using only the letters A, B, and C.

Valid examples:

\hspace*{1em}A $>$ B $>$ C\\
\hspace*{1em}B $>$ C $>$ A\\
\hspace*{1em}C $>$ A $>$ B

Do not output explanations, scores, bullets, or any additional text.
}

\end{evalprompt}

\caption{
Instructions provided for human evaluation. Annotators were asked to rank
images by jointly considering prompt alignment, aesthetic quality, and
compositional coherence.
}
\label{fig:comprehensive_eval}

\end{figure*}

% ============================================================
% Overall Quality Evaluation Prompt
% ============================================================

\begin{figure*}[p]
\centering

\begin{evalprompt}{Overall Quality Evaluation Prompt}

{\ttfamily\footnotesize

You are given a prompt and three images.

Prompt: \{prompt\_text\}

Image A is the first image.\\
Image B is the second image.\\
Image C is the third image.

Rank the images from most preferred to least preferred as a human evaluator would.

Use the prompt as the main reference and consider:
\begin{itemize}
\item overall human preference
\item prompt faithfulness
\item visual quality
\item realism
\item coherence
\item naturalness
\end{itemize}

Output exactly one ranking in the format:

\hspace*{1em}A $>$ B $>$ C

Do not output explanations, scores, bullets, or any additional text.
}

\end{evalprompt}

\caption{Prompt used for VLM-as-a-Judge evaluation of overall image quality.}
\label{fig:human_pref_prompt}

\end{figure*}

% ============================================================
% Recovered-Pair Human Validation + Difficulty
% ============================================================

\begin{figure*}[p]
\centering

\begin{evalprompt}{Recovered-Pair Human Validation}

{\ttfamily\footnotesize

You are given a text prompt and two images.

Prompt: \{prompt\_text\}

Image A is the first image.\\
Image B is the second image.

\medskip
\textbf{Task : Preference Validation}

Select the image that you prefer overall with respect to the given prompt.

When making your decision, jointly consider:

\begin{itemize}
    \item prompt alignment and correctness of depicted content
    \item object presence, attributes, and spatial relationships
    \item composition and visual coherence
    \item structural fidelity and absence of visible artifacts
    \item image quality, sharpness, and visual clarity
    \item color, lighting, and overall aesthetic quality
\end{itemize}

Choose the image that provides the stronger overall realization of the
prompt. Do not base the decision on a single criterion alone.

Valid example:

\hspace*{1em}Preference: A\\

Do not provide additional explanations or scores.
}

\end{evalprompt}

\caption{
\textbf{Instructions for human validation of recovered preference pairs.}
Annotators select the preferred image by jointly considering prompt
alignment, composition, structural fidelity, visual quality, and aesthetics.
}
\label{fig:pair_validation_difficulty}

\end{figure*}   
\end{document}